\documentclass[runningheads]{llncs}
\usepackage{graphicx}
\usepackage{comment}
\usepackage{amsmath,amssymb} % define this before the line numbering.
\usepackage{color}
\usepackage{booktabs}
\usepackage{enumitem}
\usepackage{float}
\usepackage{multirow}
\usepackage{xcolor}
\usepackage{pgfplots}
\usepackage{wrapfig}
\usepackage{lipsum}
\usepackage{hyperref}  % Custom import

\newenvironment{customlegend}[1][]{%
    \begingroup
    \csname pgfplots@init@cleared@structures\endcsname
    \pgfplotsset{#1}%
}{%
    \csname pgfplots@createlegend\endcsname
    \endgroup
}%
\def\addlegendimage{\csname pgfplots@addlegendimage\endcsname}

\newcommand{\addlegendimageintext}[1]{%
    \tikz {
        \begin{customlegend}[
            legend entries={\empty},
            legend style={
                draw=none,
                inner sep=0pt,
                column sep=0pt,
                nodes={inner sep=0pt}}]
        \addlegendimage{#1}
        \end{customlegend}
    }%
}

\definecolor{others}{RGB}{26, 83, 92}
\definecolor{residential}{RGB}{78, 205, 196}
\definecolor{commercial}{RGB}{255, 107, 107}
\definecolor{industrial}{RGB}{255, 230, 109}
\definecolor{zurichgrey}{RGB}{100, 100, 100}
\definecolor{zurichdarkgreen}{RGB}{0, 125, 0}
\definecolor{zurichbrightgreen}{RGB}{0, 255, 0}
\definecolor{zurichbrown}{RGB}{150, 80, 0}
\definecolor{zurichyellow}{RGB}{255, 255, 0}
\definecolor{zurichdarkblue}{RGB}{0, 0, 150}
\definecolor{zurichlightblue}{RGB}{150, 150, 255}

\begin{document}
% \linenumbers
\pagestyle{headings}
\mainmatter
\def\ECCVSubNumber{4524}  % Insert your submission number here

\title{SideInfNet: A Deep Neural Network for Semi-Automatic Semantic Segmentation with Side Information} % Replace with your title

\begin{comment}
\titlerunning{ECCV-20 submission ID \ECCVSubNumber} 
\authorrunning{ECCV-20 submission ID \ECCVSubNumber} 
\author{Anonymous ECCV submission}
\institute{Paper ID \ECCVSubNumber}
\end{comment}
%******************

\titlerunning{SideInfNet: Semi-Automatic Semantic Segmentation with Side Information}
\author{
Jing Yu Koh\inst{1}\thanks{Currently an AI Resident at Google.} %\orcidID{0000-0002-4402-7489}
\and
Duc Thanh Nguyen\inst{2}%\orcidID{0000-0002-2285-2066}
\and
Quang-Trung Truong\inst{1}%\orcidID{0000-0002-6242-2191}
\and
Sai-Kit Yeung\inst{3}%\orcidID{0000-0001-7974-0607}
\and
Alexander Binder\inst{1}%\orcidID{0000-0001-9605-6209}
}
\index{Koh, Jing Yu} % Jing Yu Koh, first name "Jing Yu", last name "Koh"

\authorrunning{J.Y. Koh et al.}
\institute{Singapore University of Technology and Design, Singapore \and
Deakin University, Australia \and
Hong Kong University of Science and Technology, Hong Kong \\
}
%******************
\maketitle

\begin{abstract}
% max 150 words
Fully-automatic execution is the ultimate goal for many Computer Vision applications. However, this objective is not always realistic in tasks associated with high failure costs, such as medical applications. For these tasks, semi-automatic methods allowing minimal effort from users to guide computer algorithms are often preferred due to desirable accuracy and performance. Inspired by the practicality and applicability of the semi-automatic approach, this paper proposes a novel deep neural network architecture, namely SideInfNet that effectively integrates features learnt from images with side information extracted from user annotations. To evaluate our method, we applied the proposed network to three semantic segmentation tasks and conducted extensive experiments on benchmark datasets. Experimental results and comparison with prior work have verified the superiority of our model, suggesting the generality and effectiveness of the model in semi-automatic semantic segmentation. 

\end{abstract}

\section{Introduction}
Most studies in Computer Vision tackle fully-automatic inference tasks which, ideally, perform automatically without human intervention. To achieve this, machine learning models are often well trained on rich datasets. However, these models may still fail in reality when dealing with unseen samples. A possible solution for this challenge is using assistive information provided by users, e.g., user-provided brush strokes and bounding boxes \cite{rother2004grabcut}. Human input is also critical for tasks with high costs of failure. Examples include medical applications where predictions generated by computer algorithms have to be verified by human experts before they can be used in treatment plans. In such cases, a semi-automatic approach that allows incorporation of easy-and-fast side information provided from human annotations may prove more reliable and preferable.

Semantic segmentation is an important Computer Vision problem aiming to associate each pixel in an image with a semantic class label. Recent semantic segmentation methods have been built upon deep neural networks \cite{long2015fully,girshick2014rich,chen2014semantic,chen2018deeplab}. However, these methods are not flexible to be extended with additional information from various sources, such as human annotations or multi-modal data. In addition, human interactions are not allowed seamlessly and conveniently.

In this paper, we propose SideInfNet, a general model that is capable of integrating domain knowledge learnt from domain data (e.g., images) with side information from user annotations or other modalities in an end-to-end fashion. SideInfNet is built upon a combination of advanced deep learning techniques. In particular, the backbone of SideInfNet is constructed from state-of-the-art convolutional neural network (CNN) based semantic segmentation models. To effectively calibrate the dense domain-dependent information against the spatially sparse side information, fractionally strided convolutions are added to the model. To speed up the inference process and reduce the computational cost while maintaining the quality of segmentation, adaptive inference gates are proposed to make the network's topology flexible and optimal. To the best of our knowledge, this combination presents a novel architecture for semi-automatic segmentation.

A key challenge in designing such a model is in making it generalize to different sparsity and modalities of side information. Existing work focuses on sparse pixel-wise side information, such as user-defined keypoints \cite{tripathi2017pose2instance,papandreou2018personlab}, and geotagged photos~\cite{workman2017unified,feng2018urban}. However, these methods may not perform optimally when the side information is non-uniformly distributed and/or poorly provided, e.g., brush strokes which can be drawn dense and intertwined. In~\cite{workman2017unified}, street-level panorama information is used as a source of side information. However, such knowledge is not available in tasks other than remote sensing, e.g., in tasks where the side information is provided as brush strokes. Furthermore, expensive nearest neighbor search is used for the kernel regression in~\cite{workman2017unified}, which we replace by efficient trainable fractionally strided convolutions. The Higher-Order Markov Random Field model proposed in~\cite{feng2018urban} can be adapted to various side information types but is not end-to-end trainable. Compared with these works, SideInfNet provides superior performance in various tasks and on different datasets. Importantly, our model provides a principled compromise between fully-automatic and manual segmentation. The benefit gained by the model is well shown in tasks where there exists a mismatch between training and test distribution. A few brush strokes can drastically improve the performance on these tasks. We show the versatility of our proposed model on three tasks:
\begin{itemize}
\item \textbf{Zone segmentation} of low-resolution satellite imagery \cite{feng2018urban}. Geotagged street-level photographs from social media are used as side information.
\item \textbf{BreAst Cancer Histology (BACH) segmentation} \cite{aresta2019bach}. Whole-slide images are augmented with expert-created brush strokes to segment the slides into \textit{normal}, \textit{benign}, \textit{in situ carcinoma} and \textit{invasive carcinoma} regions.
\item \textbf{Urban segmentation} of very high-resolution (VHR) overhead crops taken of the city of Zurich \cite{volpi2015semantic}. Brush annotations indicate geographic features and are augmented with imagery features to identify eight different urban and peri-urban classes from the Zurich Summer dataset \cite{volpi2015semantic}.
\end{itemize}

\section{Related Work}

\subsection{Interactive Segmentation}
GrabCut~\cite{rother2004grabcut} is a seminal work of interactive segmentation that operates in an unsupervised manner. The method allows users to provide interactions in the form of brush strokes and bounding boxes demarcating objects. Several methods have extended the GrabCut framework for both semantic segmentation and instance segmentation, e.g., \cite{goring2012semantic,xu2017deep}. However, these methods only support bounding box annotations and thus cannot be used in datasets containing irregular object shapes, e.g., non-rectangular zones in the Zurich Summer dataset \cite{volpi2015semantic}.

Users can also provide prior and reliable cues to guide the segmentation process on-the-fly~\cite{perazzi2017learning,caelles2017one,shankar2015video,li2018instance}. For instance, Perazzi et al. \cite{perazzi2017learning} proposed a CNN-based guidance method for segmenting user-defined objects from video data. In this work, users provide object bounding boxes or regions. It is also shown that increasing the number of user annotations led to improved segmentation quality. In a similar manner, Nagaraja et al. \cite{shankar2015video} tackled the task of object segmentation from video by combining motion cues and user annotations. In their work, users make scribbles to delineate the objects of interests. Experimental results verified the cooperation of sparse user annotations and motion cues, filling the gap between fully automatic and manual object segmentation. However, in the above methods, user annotations play a role as auxiliary cues but are not effectively incorporated (as features) into the segmentation process. 

\subsection{Semantic Segmentation with Side Information}

One form of side information used in several segmentation problems is keypoint annotations. The effectiveness of oracle keypoints in human segmentation is illustrated in~\cite{tripathi2017pose2instance}. Similarly, in~\cite{papandreou2018personlab}, a method for automatically learning keypoints was proposed. The keypoints are grouped into pose instances and used for instance segmentation of human subjects. The spatial layout of keypoints is important to represent meaningful human structures, but such constraint are not always held for other object types, such as cell masses in histopathology.

Literature has also demonstrated the advantages of using ground-level imagery as side information in remote sensing. For instance, in~\cite{mattyus2016hd}, multi-view imagery data, including aerial and ground images, were fused into a Markov Random Field (MRF) model to enhance the quality of fine-grained road segmentation. In~\cite{feng2018urban}, domain-dependent features from satellite images were learnt using CNNs while street-level photos were classified and considered as higher-orders in a Higher-Order MRF model. These methods are flexible to various CNN architectures but are not trainable in an end-to-end fashion.

Workman et al. \cite{workman2017unified} proposed a model for fusing multi-view imagery data into a deep neural network for estimating geospatial functions land cover and land use. While this model is end-to-end, it has heavy computational requirements for its operation, e.g., for calculating and storing $k$ nearest annotations, and thus may not be tractable for tasks with high density annotations. In addition, the model requires panorama knowledge to infer street-view photography.

\section{SideInfNet}
\begin{figure*}[!t] % This is placed here so it'll appear in the right location
\centering
\includegraphics[width=1.0\textwidth]{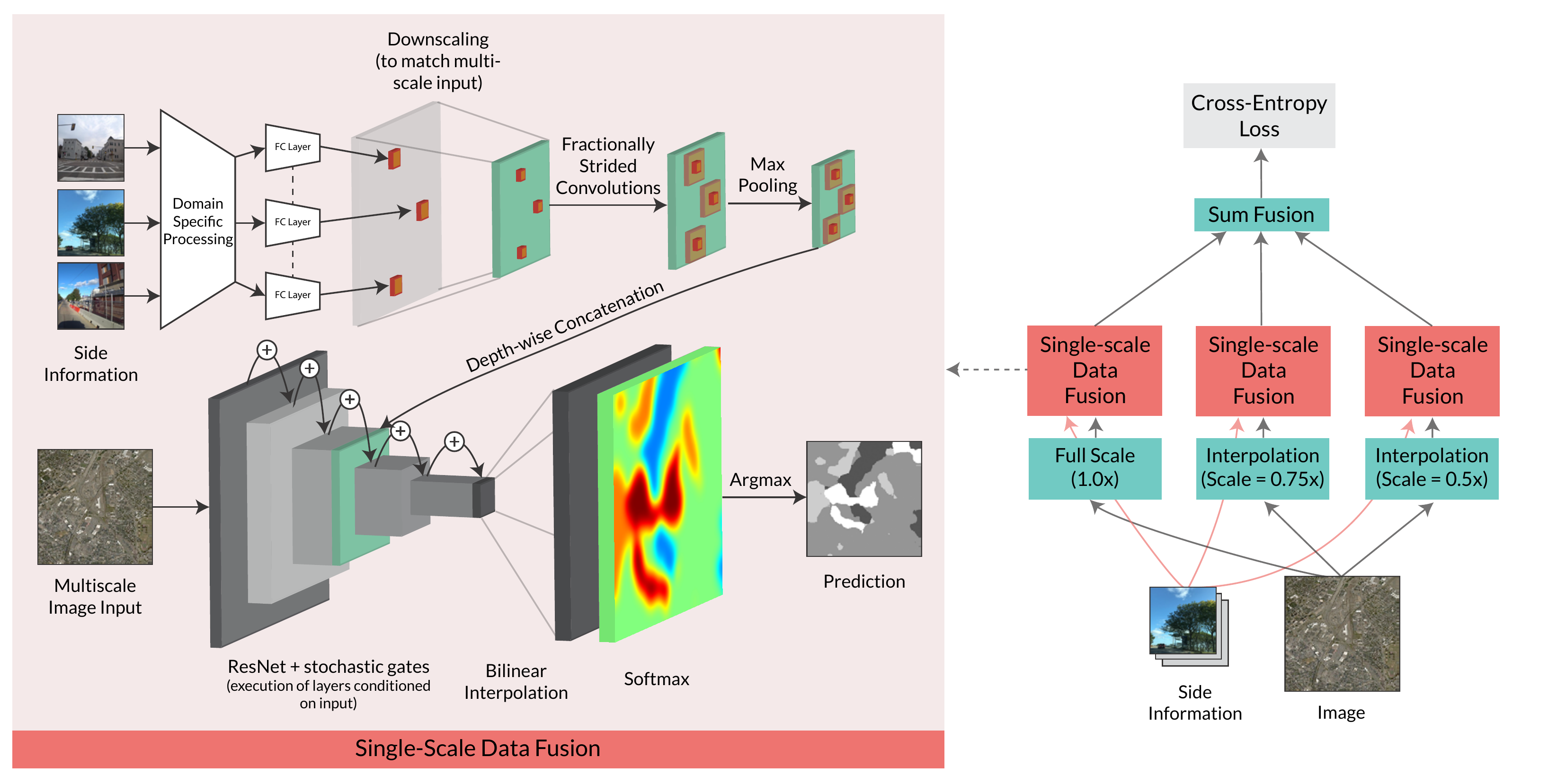}
\caption{Our proposed network architecture. A feature map of annotations is constructed based on the task. Our architecture for semantic segmentation is built on top of Deeplab-ResNet \cite{chen2018deeplab}.}
\label{fig:network}
\end{figure*}

We propose SideInfNet, a novel neural network that fuses domain knowledge and user-provided side information in an end-to-end trainable architecture. SideInfNet allows the incorporation of multi-modal data, and is flexible with different annotation types and adaptive to various segmentation models. SideInfNet is built upon state-of-the-art semantic segmentation \cite{long2015fully,chen2018deeplab,ren2015faster} and recent advances in adaptive neural networks \cite{veit2018convolutional,shazeer2017outrageously}. This combination makes our model optimal while maintaining high quality segmentation results. For the sake of ease in presentation, we describe our method in the view of zone segmentation, a case study. However, our method is general and can be applied in different scenarios.

Zone segmentation aims to provide a zoning map for an aerial image, i.e., to identify the zone type for every pixel on the aerial image. Side information in this case includes street-level photos. These photos are captured by users and associated with geocodes that refer to their locations on the aerial image. Domain-dependent features are extracted from the input aerial image using some CNN-based semantic segmentation model (see Section~\ref{sec:cnnsegment}). Side information features are then constructed from user-provided street-level photos (see Section~\ref{sec:sideinfoconstruction} and Section~\ref{sec:learning}). Associated geocodes in the street-level photos help to identify their locations in the receptive fields in the SideInfNet architecture where both domain-dependent and side information features are fused. To reduce the computational cost of the model while not sacrificing the quality of segmentation, adaptive inference gates are proposed to skip layers conditioned on input (see Section~\ref{sec:adaptive}). Fig.~\ref{fig:network} illustrates the workflow of SideInfNet whose components are described in detail in the following subsections.

\subsection{CNN-based Semantic Segmentation} \label{sec:cnnsegment}
To extract domain-dependent features, we adopt the  Deeplab-ResNet \cite{chen2018deeplab}, a state-of-the-art CNN-based semantic segmentation. Deeplab-ResNet makes use of a series of dilated convolutional layers, with increasing rates to aggregate multi-scale features. To adapt Deeplab-ResNet into our framework, we retain the same architecture but extend the \textit{conv2\_3} layer with side information (see Section~\ref{sec:sideinfoconstruction}). 

Specifically, the side information feature map is concatenated to the output of the \textit{conv2\_3} layer (see  Fig.~\ref{fig:network}). As the original \textit{conv2\_3} layer outputs a feature map with 256 channels, concatenating the side information feature map results in a $\frac{H}{4}\times\frac{W}{4}\times (256+d)$ dimensional feature map where $H$ and $W$ are the height and width of the input image, and $d$ is the number of channels of the side information feature map. This extended feature map is the input to the next convolutional layer, \textit{conv3\_1}. We provide an ablation study on varying the dimension $d$ in our supplementary material.

\subsection{Side Information Feature Map Construction} \label{sec:sideinfoconstruction}

Depending on applications, domain specific preprocessing may need to be applied to the side information. For instance, in the zone segmentation problem, we use the Places365-CNN in \cite{zhou2014learning} to create vector representations for street-level photos (see details in Section~\ref{sec:zoningexpsetup}). These vectors are then passed through a fully-connected layer returning $d$-dimensional vectors. Suppose that the input aerial image is of size $H \times W$. A side information feature map $\boldsymbol{x}^{l}$ of size $H \times W \times d$ can be created by initializing the $d$-dimensional vector at every location in $H \times W$ with the feature vector of the corresponding street-level photo, if one exists there. The feature vectors at locations that are not associated to any street-level photos are padded with zeros. Mapping image locations to street-level photos can be done using the associated geocodes of the street-level photos. Nearest neighbor interpolation is applied on the side information feature map to create multi-scale features. Features that fall in the same image locations (on the aerial image) due to downscaling are averaged. To make feature vectors consistent across scales and data samples, all feature vectors are normalized to the unit length. 

There may exist misalignment in associating the side information features with their corresponding locations on the side information feature map. For instance, a brush stroke provided by a user may not well align with a true region. In the application of zoning, a street-level photo may not record the scene at the exact location where the photo is captured. Therefore, a direct reference of a street-level photo to a location on the feature map via the photo's geocode may not be a perfect association. However, one could expect that the side information could be propagated from nearby locations. To address this issue, we apply a series of fractionally-strided convolutions to the normalized feature map $\boldsymbol{x}^{l}$ to distribute the side information spatially. In our implementation, we use $3\times3$ kernels of ones, with stride length of 1 and padding of 1. After a single fractionally-strided convolution, side information features are distributed onto neighbouring $3\times3$ regions. We repeat this operation (denoted as $f_c$) $n$ times and sum up all the feature maps to create the features for the next layer as follows,
\begin{align}
\label{eq:fractionallystridedconv}
\boldsymbol{x}^{l+1} = F(\boldsymbol{x}^{l})=\sum_{i=1}^n w_i f_c^i(\boldsymbol{x}^{l})
\end{align}
where $w_i$ are learnable parameters and $f_c^i$ is the $i$-th functional power of $f_c$, i.e., 
\begin{align}
f_c^i(\boldsymbol{x}^{l}) = \begin{cases}
    f_c(\boldsymbol{x}^{l}), & i=1 \\ 
    f_c(f_c^{i-1}(\boldsymbol{x}^{l})), & \text{otherwise}
    \end{cases}
\end{align}

The parameters $w_i$ in (\ref{eq:fractionallystridedconv}) allow our model to learn the importance of spatial extent. We observe a decreasing pattern in $w_i$ (i.e., $w_1>w_2>\cdots$) after training. This matches our intuition that information is likely to become less relevant with increased distances. The resulting feature map $\boldsymbol{x}^{l+1}$ represents a weighted sum of nearby feature vectors. We also normalize the feature vector at each location in the feature map by the number of the fractionally-strided convolutions used at that location. This has the effect of averaging overlapping features.

Lastly, we perform maxpooling to further downsample the side information feature map to fit with the counterpart domain-dependent feature map for feature fusion. We choose to perform feature fusion before the second convolutional block of Deeplab-ResNet, with the output of the \textit{conv2\_3} layer. We empirically found that this provided a good balance between computational complexity and segmentation quality. The output of the maxpooling layer is concatenated in the channels dimension to the output of the original layer (see Fig.~\ref{fig:network}). It is important to note that our proposed side information feature map construction method is general and can be applied alongside any CNN-based semantic segmentation architectures. 

\subsection{Fusion Weight Learning}
\label{sec:learning}
As defined in (\ref{eq:fractionallystridedconv}), the output for each pixel $(p,q)$ in the feature map $f_c^{i+1}(x^l)$ (after applying $3\times3$ fractionally-strided convolution of 1s) can be described as:
\begin{align}
f_c^{i+1}(x^l)_{p,q}=\sum_{j=1}^3\sum_{k=1}^3w_ix^l_{p-2+j,q-2+k}.
\end{align}
Gradient of the fusion weight $w_i$ for each layer can be computed as,
\begin{align} 
\label{eq:fusionbp}
\frac{\partial L}{\partial w_i}=\frac{\partial L}{\partial f_c^{i+1}(x^l)}\frac{\partial f_c^{i+1}(x^l)}{\partial w_i}=\sum_p\sum_q\frac{\partial L}{\partial f_c^{i+1}(x^l)_{p,q}}\sum_{j=1}^3\sum_{k=1}^3 x^l_{p-2+n, q-2+n}
\end{align}
where $\frac{\partial L}{\partial f_c^{i+1}(x^l)}$ is back-propagated from the \textit{conv2\_3} layer.

For the fully-connected layers used for domain-specific processing (see Fig.~\ref{fig:network}), the layers are shared for each side-information instance. The shared weights $w_{\text{fc}}$ can be learnt through standard back-propagation of a fully-connected layer:
\begin{align}
\frac{\partial L}{\partial w_{\text{fc}}}=\frac{\partial L}{\partial f_c^1}\frac{\partial f_c^1}{\partial w_{\text{fc}}}
\end{align}
where $\frac{\partial L}{\partial f_c^1}$ is back-propagated from the first fusion layer (see~(\ref{eq:fusionbp})).

\subsection{Adaptive Architecture} \label{sec:adaptive}
\noindent Inspired by advances in adaptive neural networks \cite{veit2018convolutional,shazeer2017outrageously}, we adopt adaptive inference graphs in SideInfNet. Adaptive inference graphs decide skip-connections in the network architecture using adaptive gates $\boldsymbol{z}^l$. Specifically, we define,
\begin{align}
\label{eq:adaptivemodel}
\boldsymbol{x}^{l+1} = \boldsymbol{x}^{l} + \boldsymbol{z}^l(h(\boldsymbol{x}^{l})) \cdot F(\boldsymbol{x}^{l})
\end{align}
where $\boldsymbol{z}^l(h(\boldsymbol{x}^{l})) \in \{0,1\}$ and $h$ is some function that maps $\boldsymbol{x}^{l} \in H \times W \times d$ into a lower-dimensional space of $1 \times 1 \times d$. The gate $\boldsymbol{z}^l$ is conditioned on $\boldsymbol{x}^{l}$ and takes a binary decision (1 for ``on'' and 0 for ``off'').

Like \cite{veit2018convolutional}, we set the early layers and the final classification layer of our model to always be executed, as these layers are critical for maintaining the accuracy. The gates are included in every other layer. We define the function $h$ as,
\begin{align}
\label{eq:h}
h(\boldsymbol{x}^l) = \frac{1}{H \times W} \sum_{i=1}^H \sum_{j=1}^W \boldsymbol{x}^{l}_{i,j}
\end{align}

The feature map $h(\boldsymbol{x}^l)$ is passed into a multi-layer perceptron (MLP), which computes a relevance score to determine whether the layer $l$ is executed. We also use a gate target rate $t$, that determines what fraction of layers should be activated. This is implemented as a mean squared error (MSE) loss and jointly optimized with the cross entropy loss. Each separate MLP determines whether its corresponding layer should be executed (contributing 1 to the total count), or not (contributing 0). Thus, the MSE loss encourages the overall learnt execution rate to be close to $t$. This is dynamic, i.e., more important layers would be executed more frequently and vice versa. For instance, a target rate $t=0.8$ imposes a penalty on the loss function when the proportion of layers executed is greater or less than 80\%. Our experimental results on this adaptive model are presented in Section \ref{sec:results}, where we find that allowing a proportion of layers to be skipped helps improve segmentation quality.

\section{Experiments and Results} \label{sec:results}
\noindent In this section, we extensively evaluate our proposed SideInfNet in three different case studies. In each case study, we compare our method with its baseline and other existing works. We also evaluate our method under various levels of side information usage and with another CNN backbone. 

\subsection{Zone Segmentation} \label{sec:urban}
\subsubsection{Experimental Setup}
\label{sec:zoningexpsetup}
Like~\cite{feng2018urban}, we conducted experiments on three US cities: Boston (BOS), New York City (NYC), and San Francisco (SFO). Freely available satellite images hosted on Microsoft Bing Maps~\cite{bingmaps} were used. Ground-truth maps were retrieved at a service level of 12, which corresponds to a resolution of 38.2185 meters per pixel. An example of the satellite imagery is shown in Fig.~\ref{fig:satellitesfo}. We retrieved street-level photos from Mapillary \cite{mapillary}, a service for sharing crowdsourced geotagged photos. There were four zone types: \textit{Residential}, \textit{Commercial}, \textit{Industrial} and \textit{Others}. Table~\ref{table:mapillary} summarizes the dataset used in this case study.

% \vspace{-0.2in}
\noindent
\begin{minipage}{\textwidth}
   \begin{minipage}{0.55\textwidth}
     \begin{figure}[H]
      \centering
      \includegraphics[width=\linewidth]{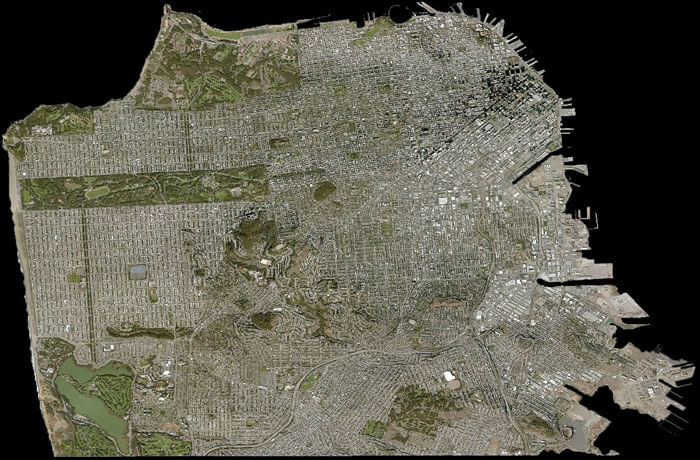}
    %   \vspace{-0.1in}
      \caption{Satellite image of San Francisco.}
      \label{fig:satellitesfo}  
     \end{figure}
   \end{minipage}
%   \hspace{0.05\linewidth}
   \begin{minipage}{0.4\textwidth}
     \centering
     \begin{table}[H]
      \caption{Proportion of street-level photos (\#photos).}
      \begin{tabular}{|c|c|c|c|}
      \hline
        \multirow{2}{*}{\textbf{Zone Type}} &\multicolumn{3}{|c|}{\textbf{City}} \\
        \cline{2-4} 
        & \textbf{BOS}& \textbf{NYC} & \textbf{SFO} \\
        \hline
        Residential & 25,607 & 16,395 & 50,116 \\
        \hline
        Commercial  & 13,412 & 5,556  & 19,641 \\
        \hline
        Industrial  & 2,876  & 9,327  & 15,219 \\
        \hline
        Others      & 25,402 & 15,281 & 50,214 \\
        \hline
     \end{tabular}
     \label{table:mapillary}
    \end{table}
   \end{minipage}
\end{minipage}
% \vspace{0.1in}

To extract side information features, we utilized the pre-trained model of Places365-CNN~\cite{zhou2014learning}, which was designed for scene recognition. We fine-tuned the model on our data. During training the model, we froze the weights of the Places365-CNN and used this fine-tuned model to generate side information feature maps. We also applied a series of $n=5$ fractionally-strided convolutions on feature maps generated from Places365-CNN. This acts as to distribute the side information from each geotagged photo 5 pixels in each cardinal direction. 

%We empirically found that $n=5$ gave a good balance between segmentation quality and computational requirements.

\subsubsection{Results}

\begin{figure*}[t]
% \centering
% \scalebox{0.95}{
\begin{center}
\begin{tabular}{l@{\ }l@{\ }c@{\ }c@{\ }c@{\ }c}
% \hline
 \toprule
\multicolumn{1}{c}{}
& \multicolumn{1}{c}{HO-MRF \cite{feng2018urban}}
& \multicolumn{1}{c}{Unified \cite{workman2017unified}}
& \multicolumn{1}{c}{SideInfNet}
& \multicolumn{1}{c}{Groundtruth}
\\\midrule
\rotatebox{90}{\textbf{BOS}} & \multicolumn{1}{c}{
\includegraphics[width=0.23\linewidth]{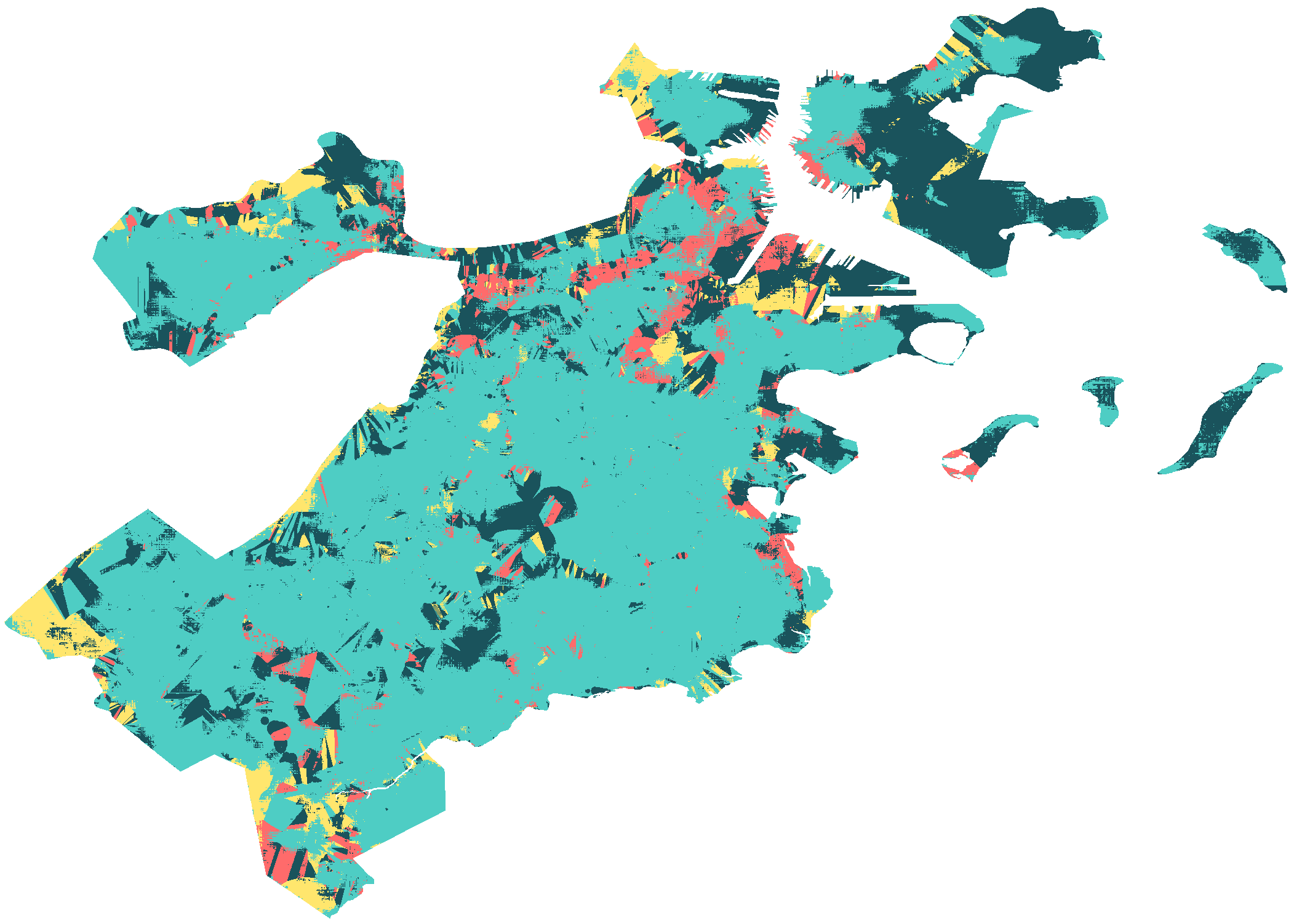}} &
\includegraphics[width=0.23\linewidth]{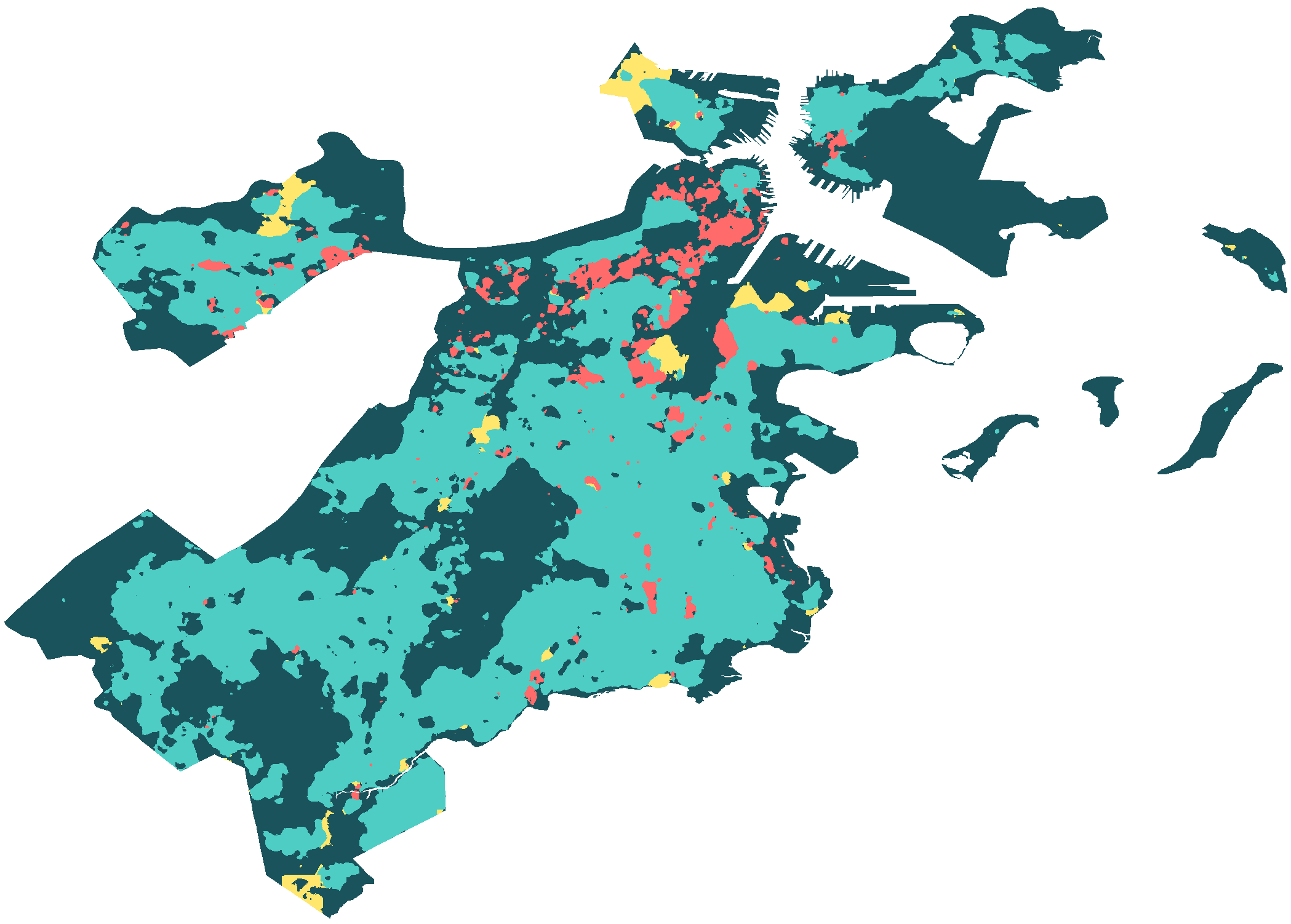} &
\includegraphics[width=0.23\linewidth]{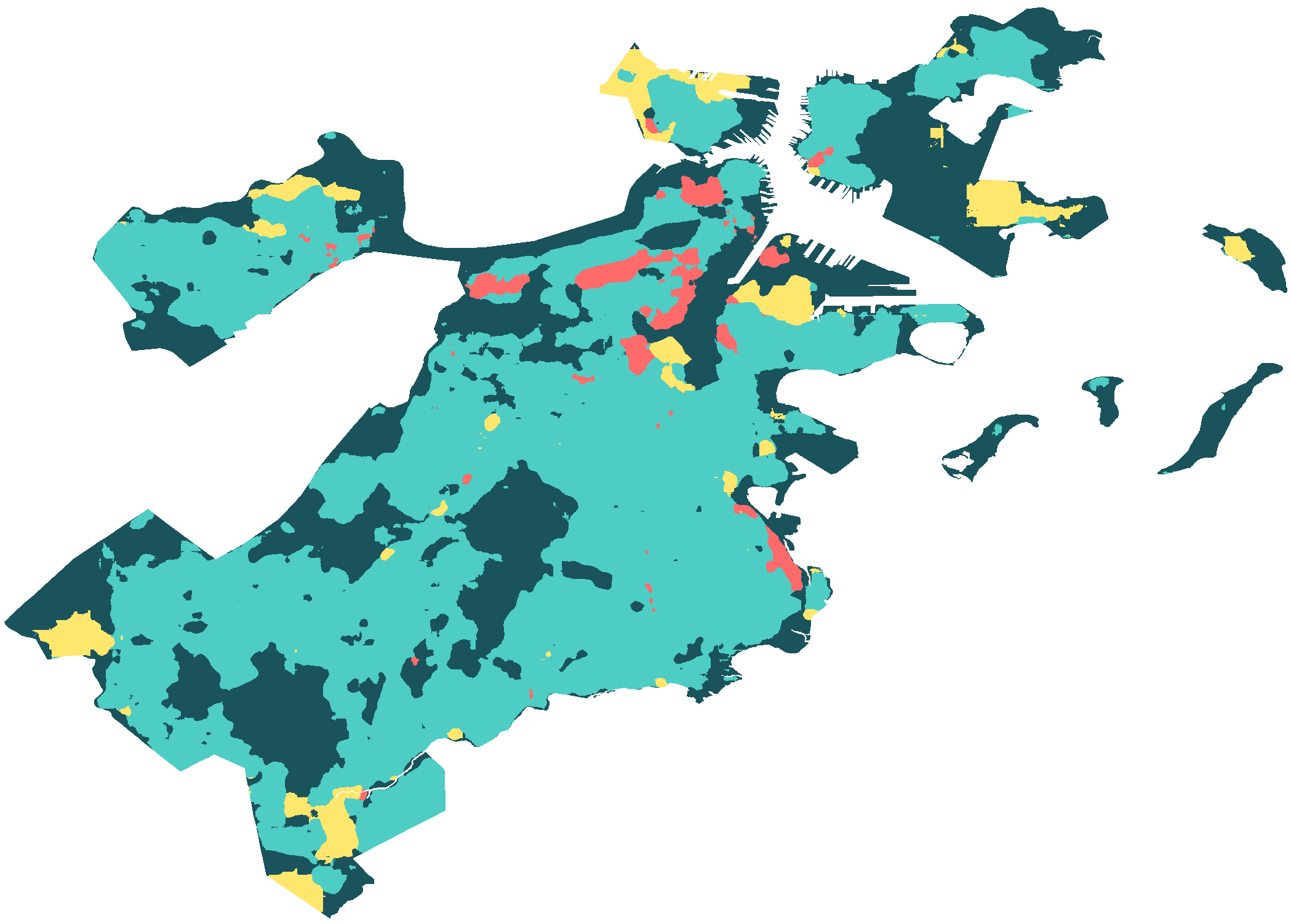} &
\includegraphics[width=0.23\linewidth]{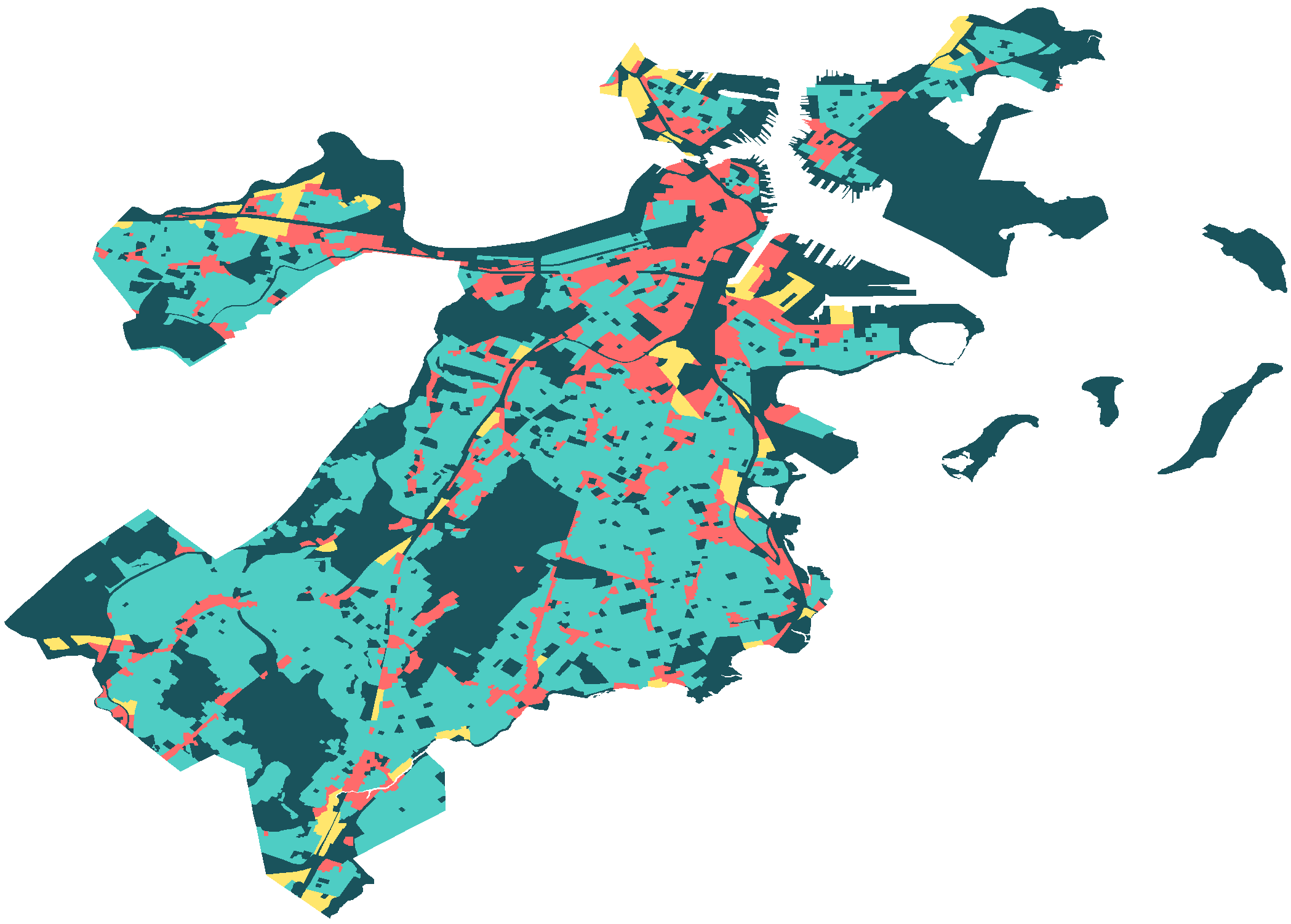}
\\\midrule
\rotatebox{90}{\textbf{NYC}} & \multicolumn{1}{c}{
\includegraphics[width=0.23\linewidth]{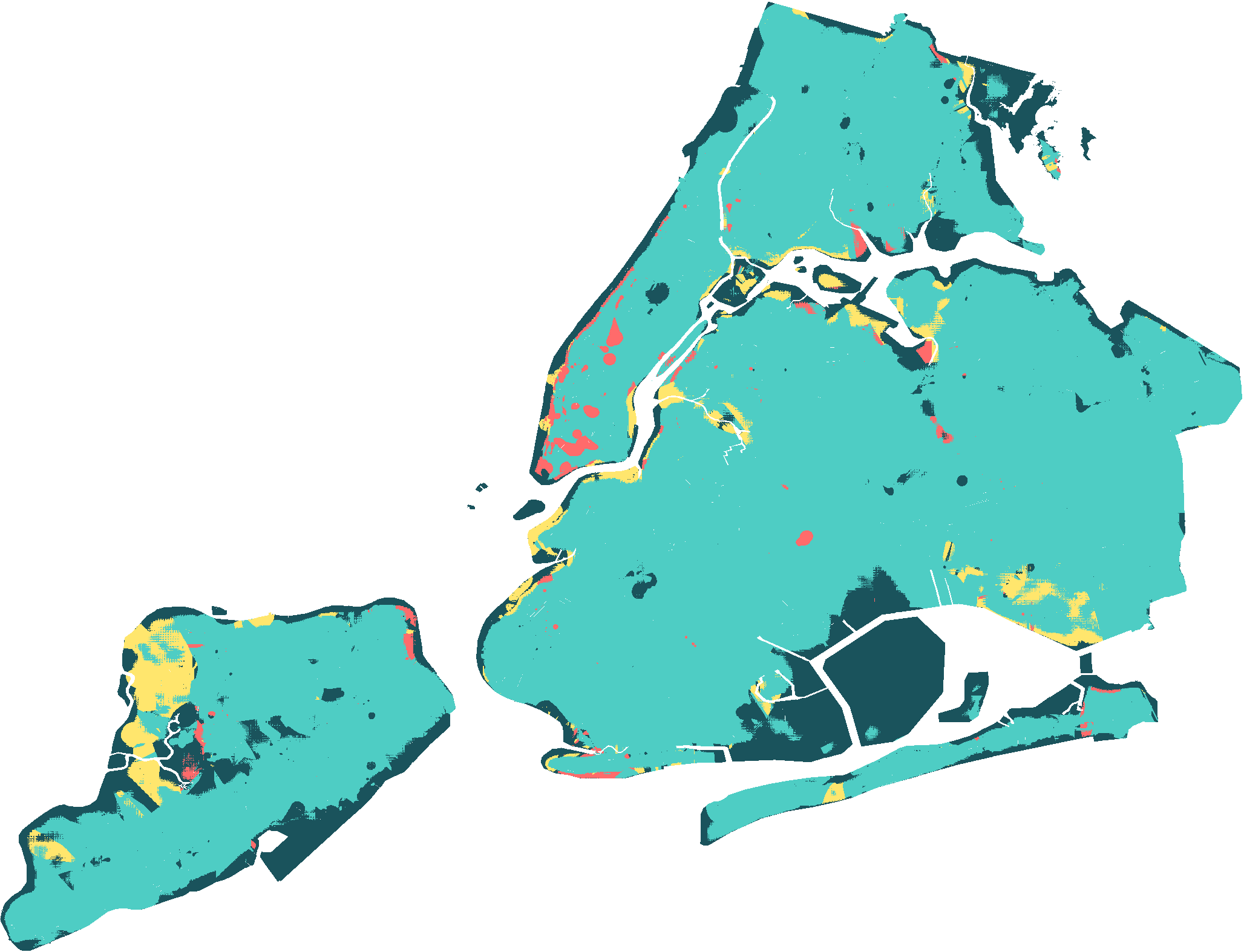}} &
\includegraphics[width=0.23\linewidth]{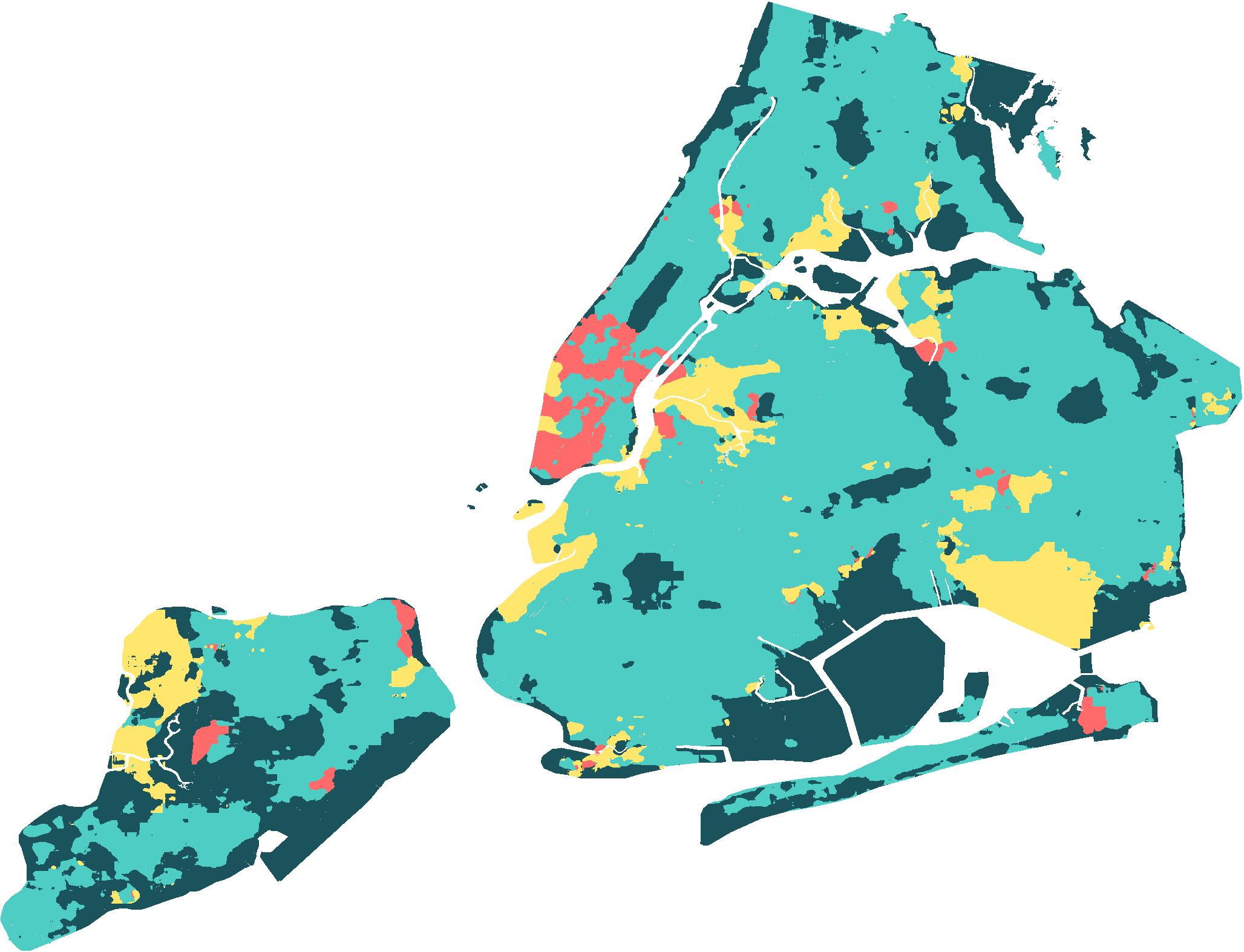} &
\includegraphics[width=0.23\linewidth]{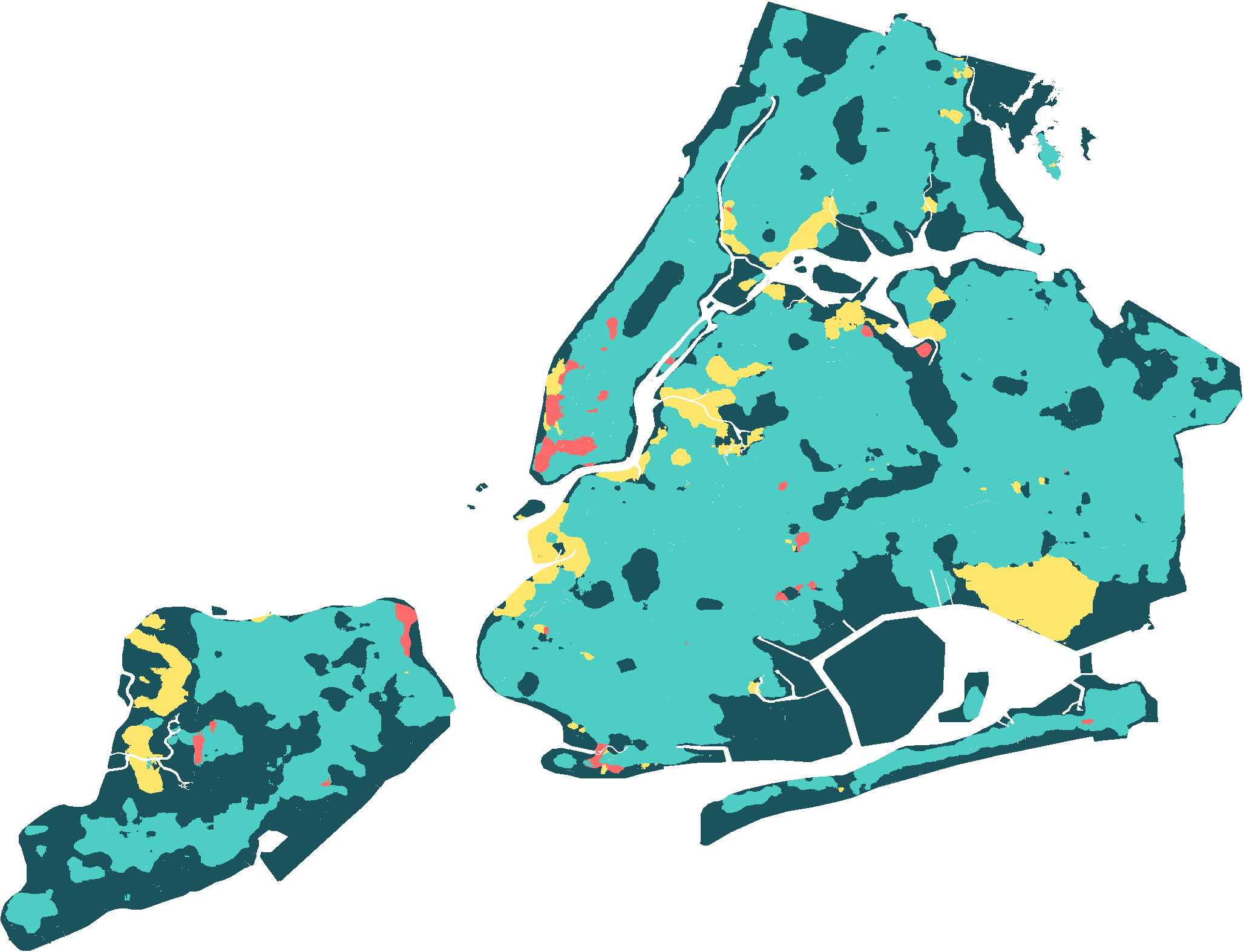} &
\includegraphics[width=0.23\linewidth]{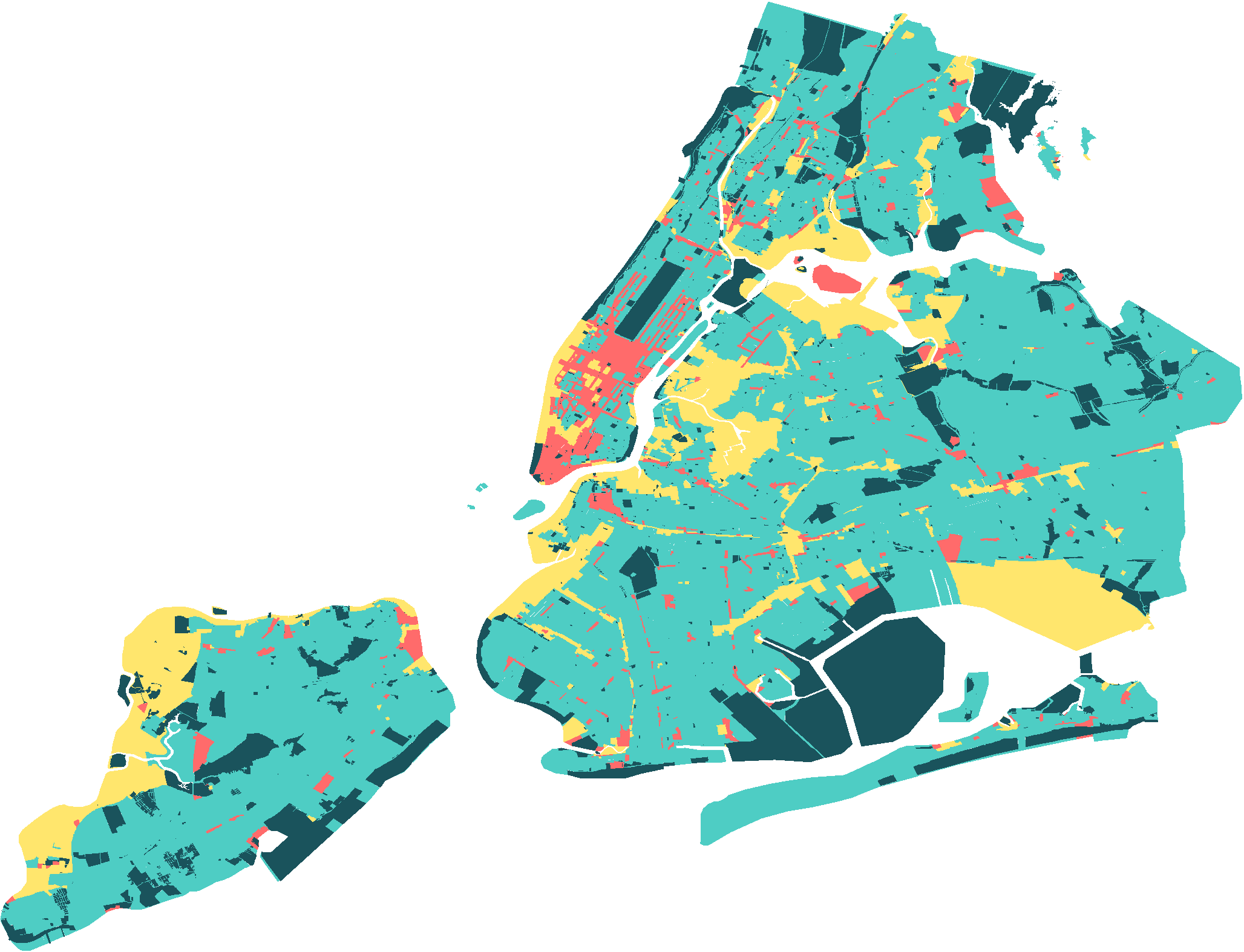}
\\\midrule
\rotatebox{90}{\textbf{SFO}} & \multicolumn{1}{c}{
\includegraphics[width=0.23\linewidth]{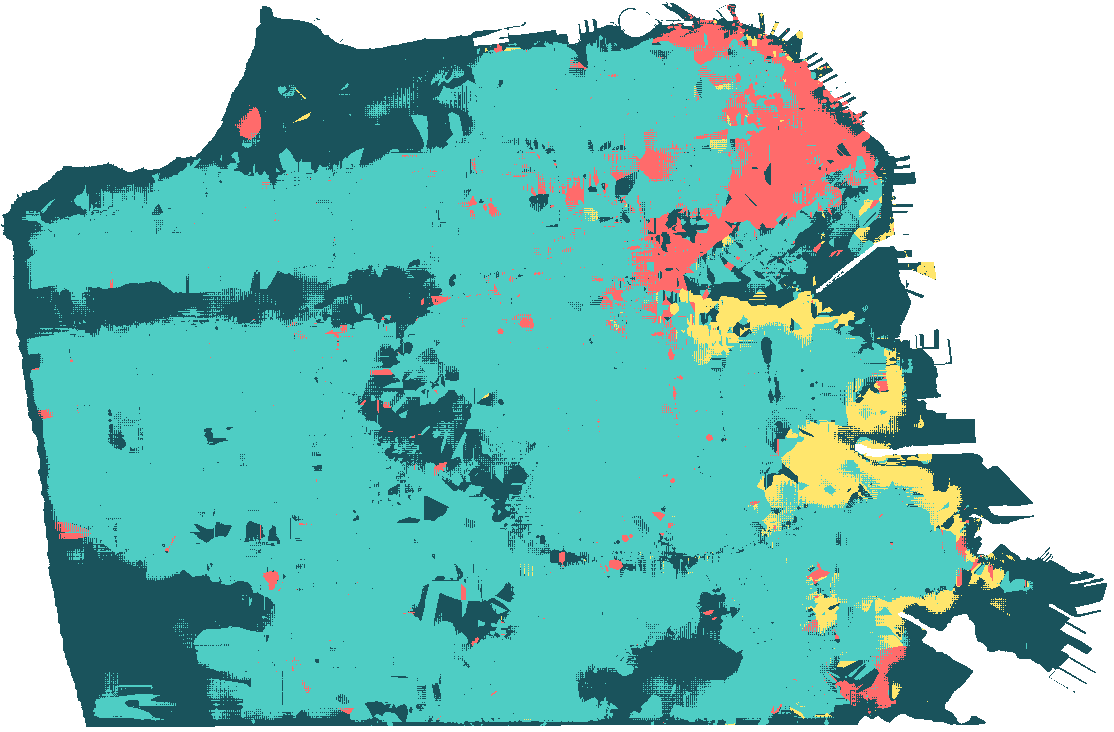}} &
\includegraphics[width=0.23\linewidth]{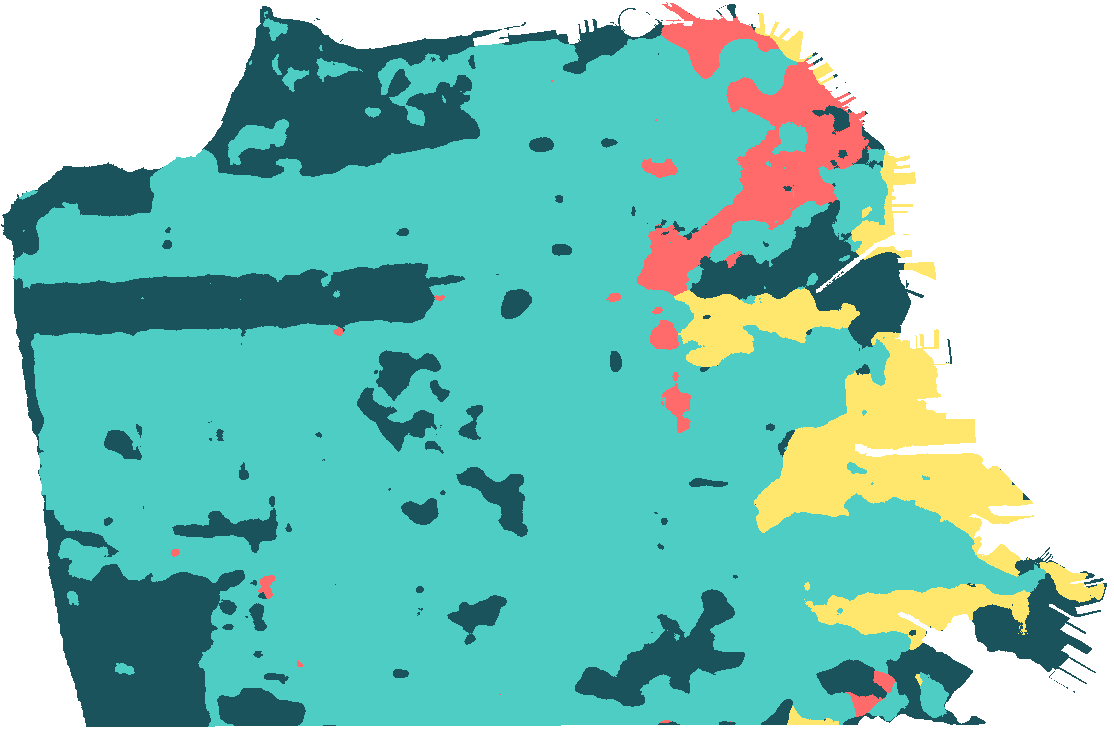}&
\includegraphics[width=0.23\linewidth]{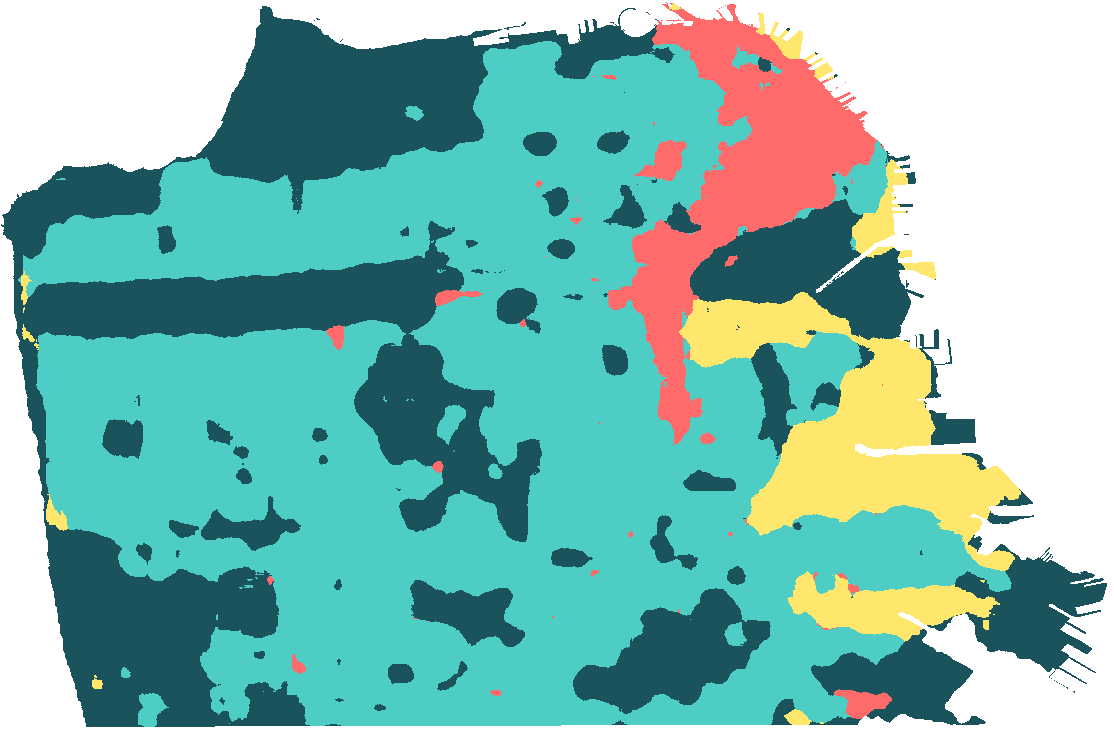} &
\includegraphics[width=0.23\linewidth]{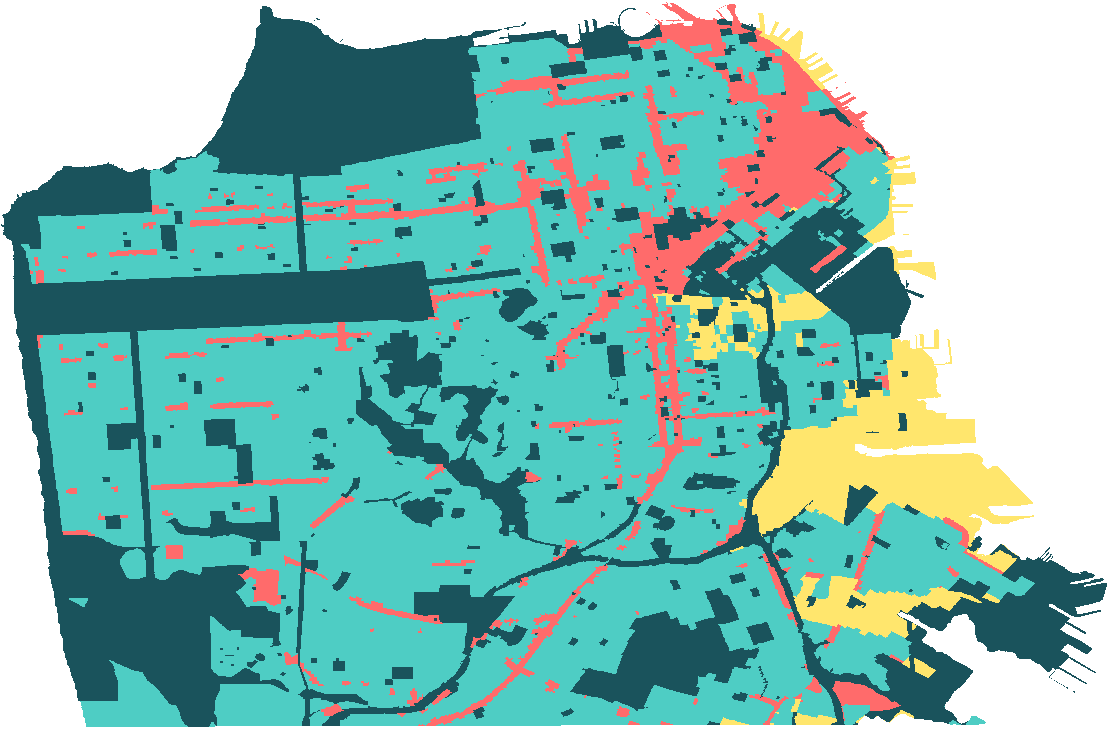}
% \\\hline
\\\bottomrule
% \vspace{0.01pt}
\end{tabular}
\addlegendimageintext{fill=others, area legend} Others
% \hspace{3pt}
\addlegendimageintext{fill=residential, area legend} Residential
% \hspace{3pt}
\addlegendimageintext{fill=commercial, area legend} Commercial
% \hspace{3pt}
\addlegendimageintext{fill=industrial, area legend} Industrial
\end{center}
\caption{
Comparison of our method and previous works. Best viewed in color.}

\label{fig:qualitative}
\end{figure*}

\begin{table*}[t] % Placed here so it will appear on the right page
\caption{Segmentation performance on zoning. Best performances are highlighted.}
\begin{center}
\begin{tabular}{@{}ccccccccc@{}}
\toprule
\multirow{2}{*}{\textbf{Approach}} & \multicolumn{4}{c}{\textbf{Accuracy}} & \multicolumn{4}{c}{\textbf{mIOU}} \\ \cmidrule{2-9}
 & \textbf{BOS} & \textbf{NYC} & \textbf{SFO} & \textbf{Mean} & \textbf{BOS} & \textbf{NYC} & \textbf{SFO} & \textbf{Mean} \\ \cmidrule{1-9}
Deeplab-ResNet \cite{chen2018deeplab} & 60.79\% & 59.58\% & 72.21\% & 64.19\% & 28.85\% & 23.77\% & 38.40\% & 30.34\% \\
HO-MRF$^{\mathrm{*}}$ \cite{feng2018urban} & 59.52\%  & \underline{72.25\%} & 73.93\% & 68.57\% & 31.92\% & 34.99\% & 46.53\% & 37.81\% \\
Unified$^{\mathrm{*}}$ \cite{workman2017unified} & 67.91\% & 70.92\% & 75.92\% & 71.58\% & 40.51\% & 39.27\% & 55.36\% & 45.05\% \\
SideInfNet & \underline{71.33\%} & 71.08\% & \underline{79.59\%} & \underline{74.00\%} & \underline{41.96\%} & \underline{39.59\%} & \underline{60.31\%} & \underline{47.29\%} \\
\bottomrule
\multicolumn{5}{l}{$^{\mathrm{*}}$ Our implementation.}
\end{tabular}
\label{table:results}
\end{center}
\end{table*}

We evaluate our method and compare it with two recent works: Higher-Order Markov Random Field (HO-MRF) \cite{feng2018urban} and Unified model \cite{workman2017unified} using 3-fold cross validation, i.e., two cities are used for training and the other one is used for testing. To have a fair comparison, the same Places365-CNN model is used to extract side information in all methods. We also compare our method against the baseline Deeplab-ResNet, which directly performs semantic segmentation of satellite imagery without the use of geotagged photos.

Our results on both pixel accuracy and mean intersection over union (mIOU) are reported in Table~\ref{table:results}. As shown in the table, our method significantly improves over its baseline, Deeplab-ResNet, proving the importance of side information. SideInfNet also outperforms prior work, with a relative improvement in pixel accuracy from the Unified model by 3.38\% and from the HO-MRF by 7.92\%. Improvement on mIOU scores is also significant, e.g., by 4.97\% relative to the Unified model, and 25.07\% relative to the HO-MRF model.

In addition to improved accuracy, our method offers several advantages over the previous works. First, compared with the HO-MRF~\cite{feng2018urban}, our method is trained end-to-end, allowing it to jointly learn optimal parameters for both semantic segmentation and side information feature extraction. Second, our method is efficient in computation. It simply performs a single forward pass through the network to produce segmentation results, opposed to iterative inference in the HO-MRF. Third, by using fractionally-strided convolutions, the complexity of our method is invariant to the side information density. This allows optimal performance on regions with high density of side information. In contrast, the Unified model \cite{workman2017unified} requires exhaustive searches to determine nearest street-level photos for every pixel on satellite image and thus depend on the density of the street-level photos and the size of the satellite image.  

We qualitatively show the segmentation results of our method and other works in Fig.~\ref{fig:qualitative}. A clear drawback of the HO-MRF is that the results tend to be grainy, likely due to the sparsity of street-level imagery. In contrast, our method generally provides smoother results that form contiguous regions. Moreover, our method better captures fine grained details from street-level imagery.

\subsection{BreAst Cancer Histology Segmentation} 
\label{sec:databach}

\begin{figure*}[t]
    \centering
    \includegraphics[width=0.32\textwidth]{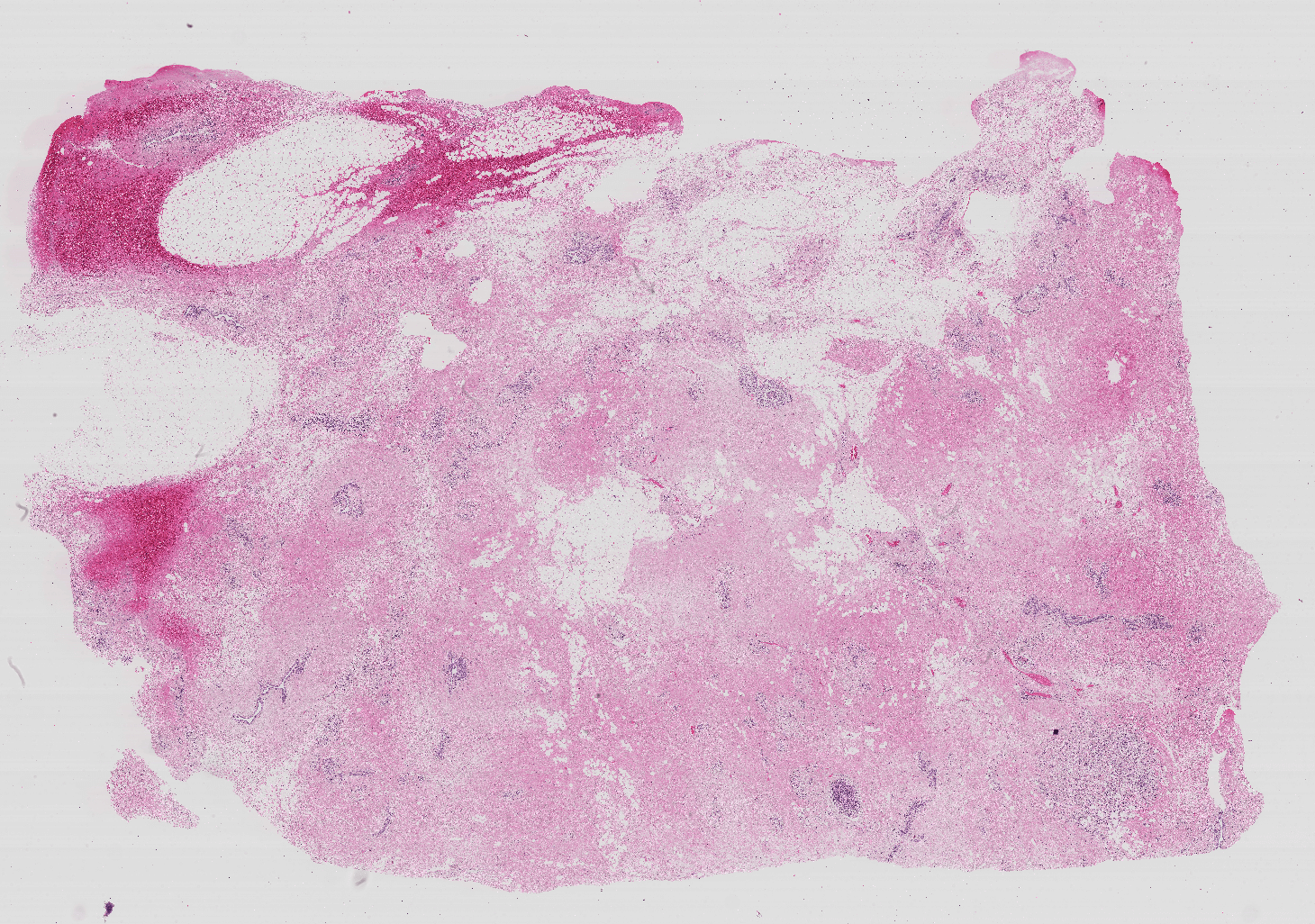}
    \includegraphics[width=0.32\textwidth]{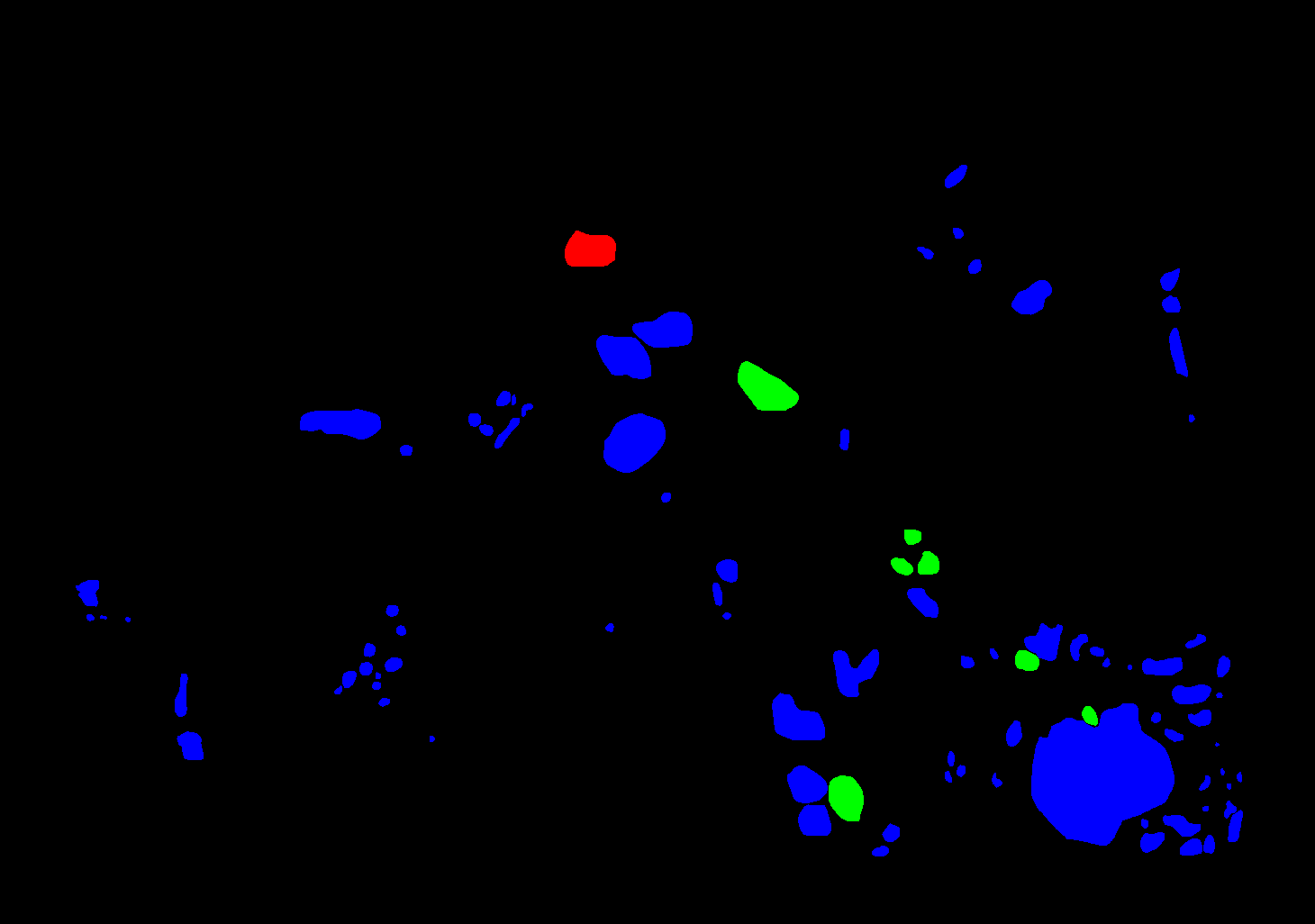}
    \includegraphics[width=0.32\textwidth]{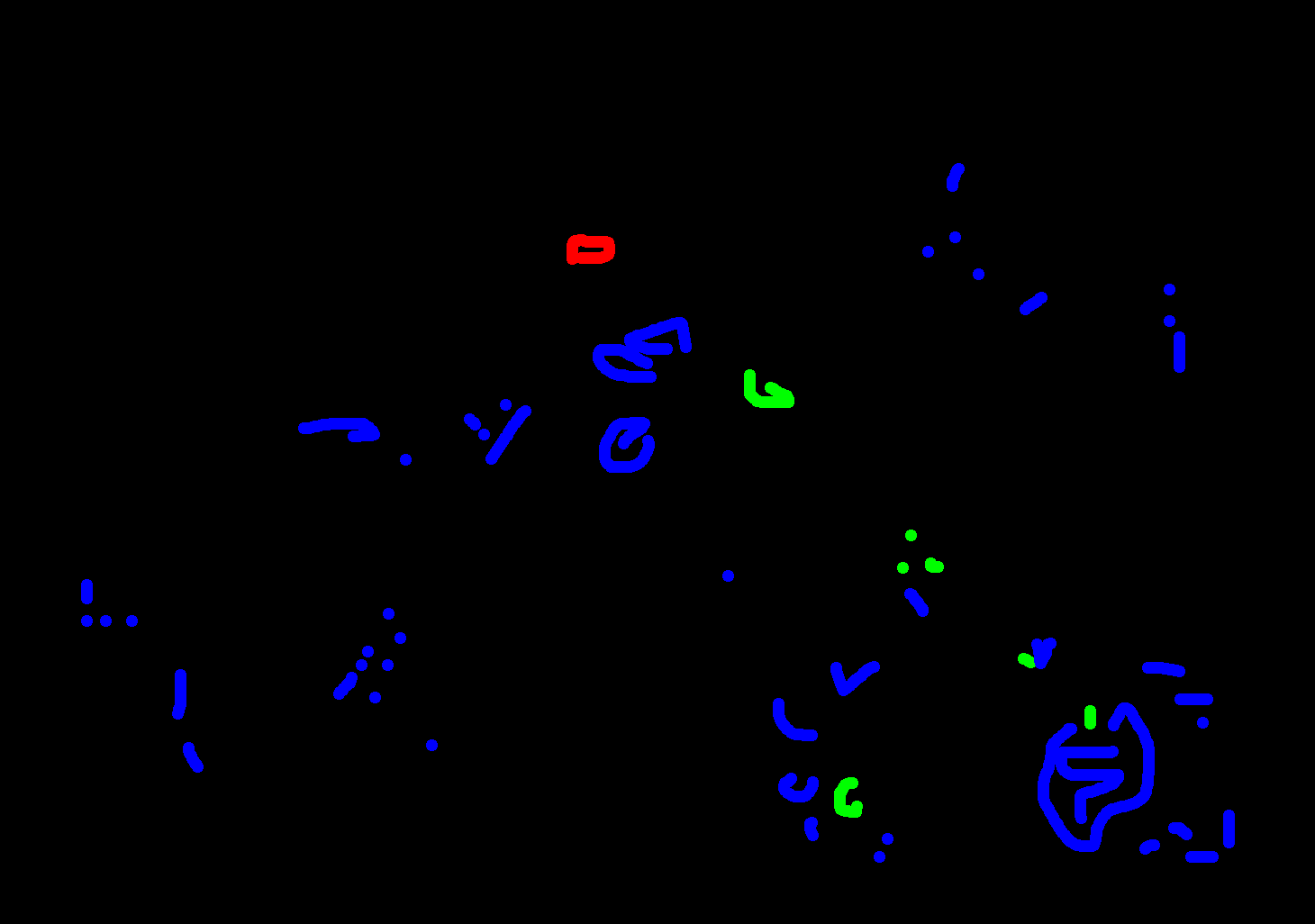}
    \caption{Left: Whole-slide image. Middle: True labels from the ground-truth. Right: Simulated brush strokes. Best viewed in color.}
    \label{fig:bachexample}
\end{figure*}

\begin{figure*}[t]
% \centering
% \scalebox{0.95}{
\begin{center}
\begin{tabular}{l@{\ }c@{\ }c@{\ }c@{\ }c}
% \hline
% \toprule
\multicolumn{1}{c}{Deeplab-ResNet \cite{chen2018deeplab}}
& \multicolumn{1}{c}{Unified \cite{workman2017unified}}
& \multicolumn{1}{c}{SideInfNet}
& \multicolumn{1}{c}{Groundtruth}\\
%\\\midrule
\includegraphics[width=.24\textwidth]{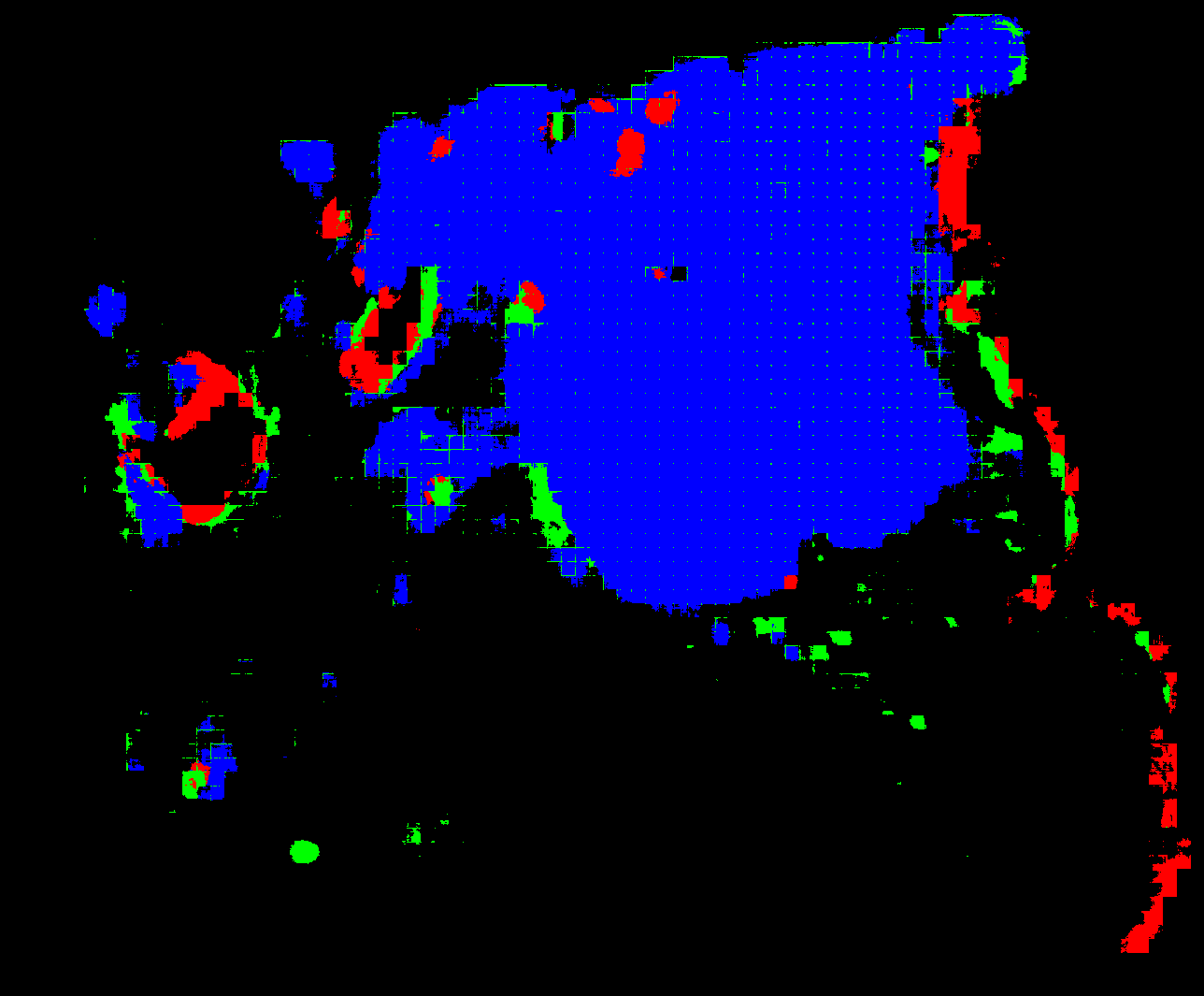} &
\includegraphics[width=.24\textwidth]{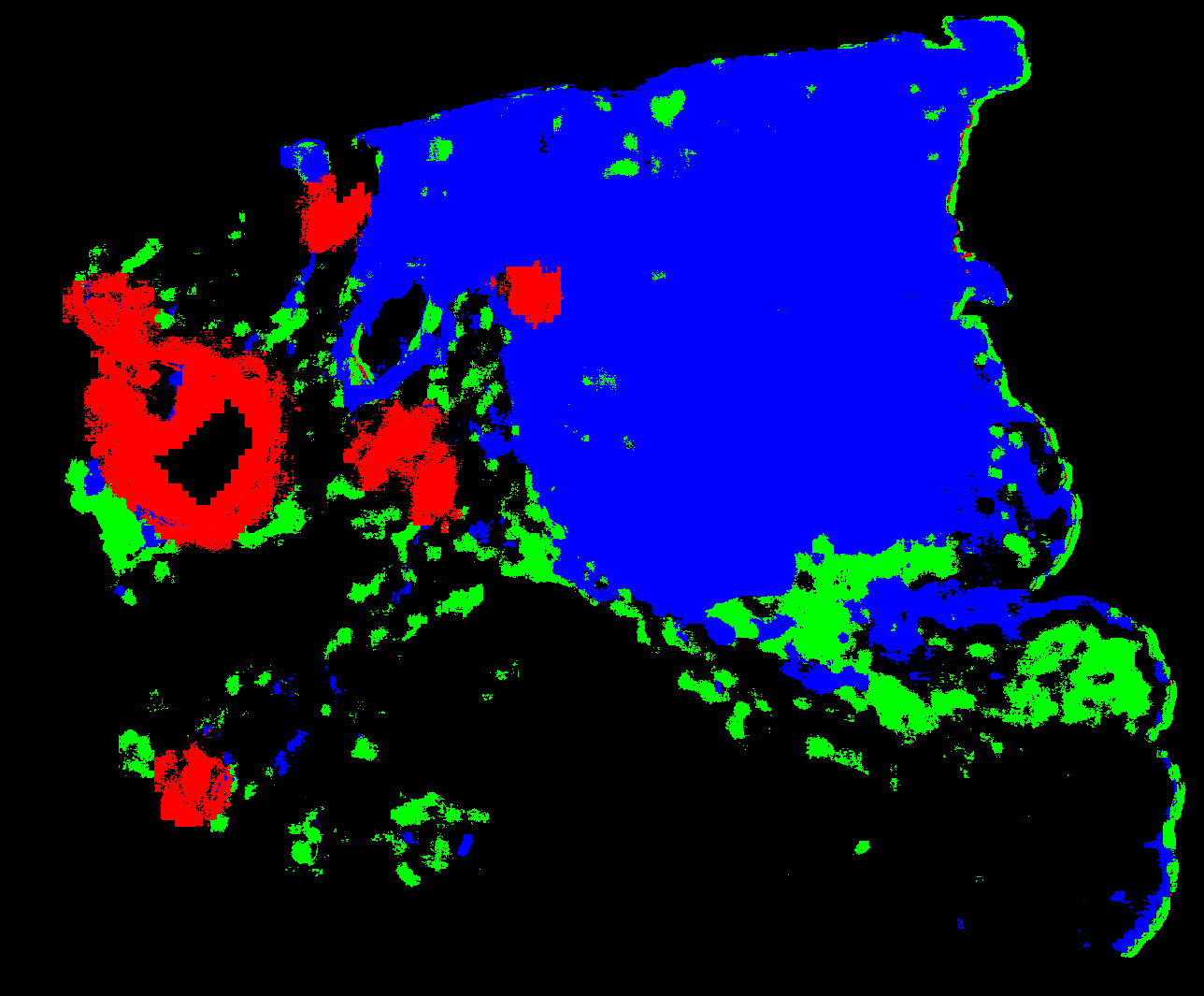} &
\includegraphics[width=.24\textwidth]{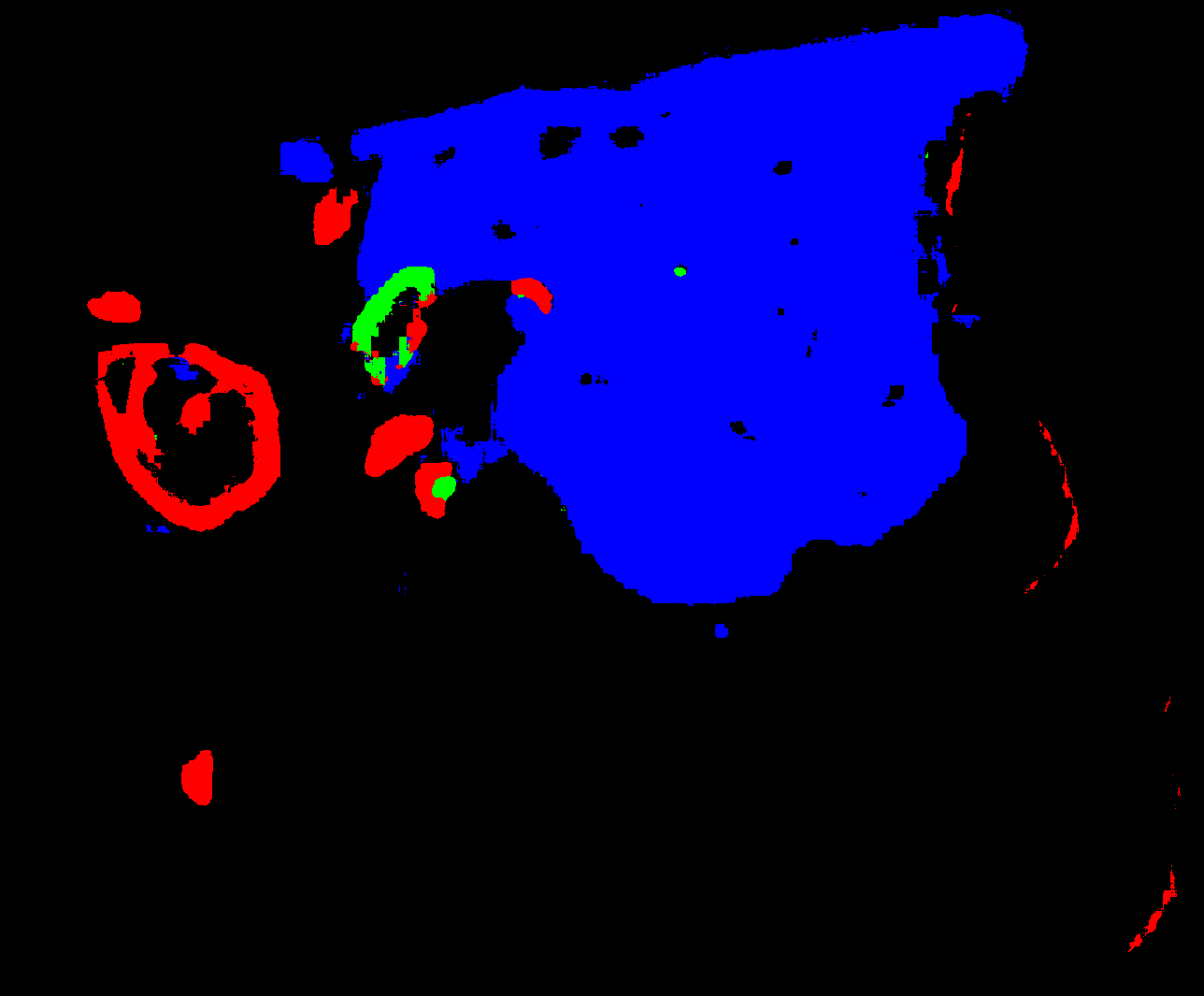} &
\includegraphics[width=.24\textwidth]{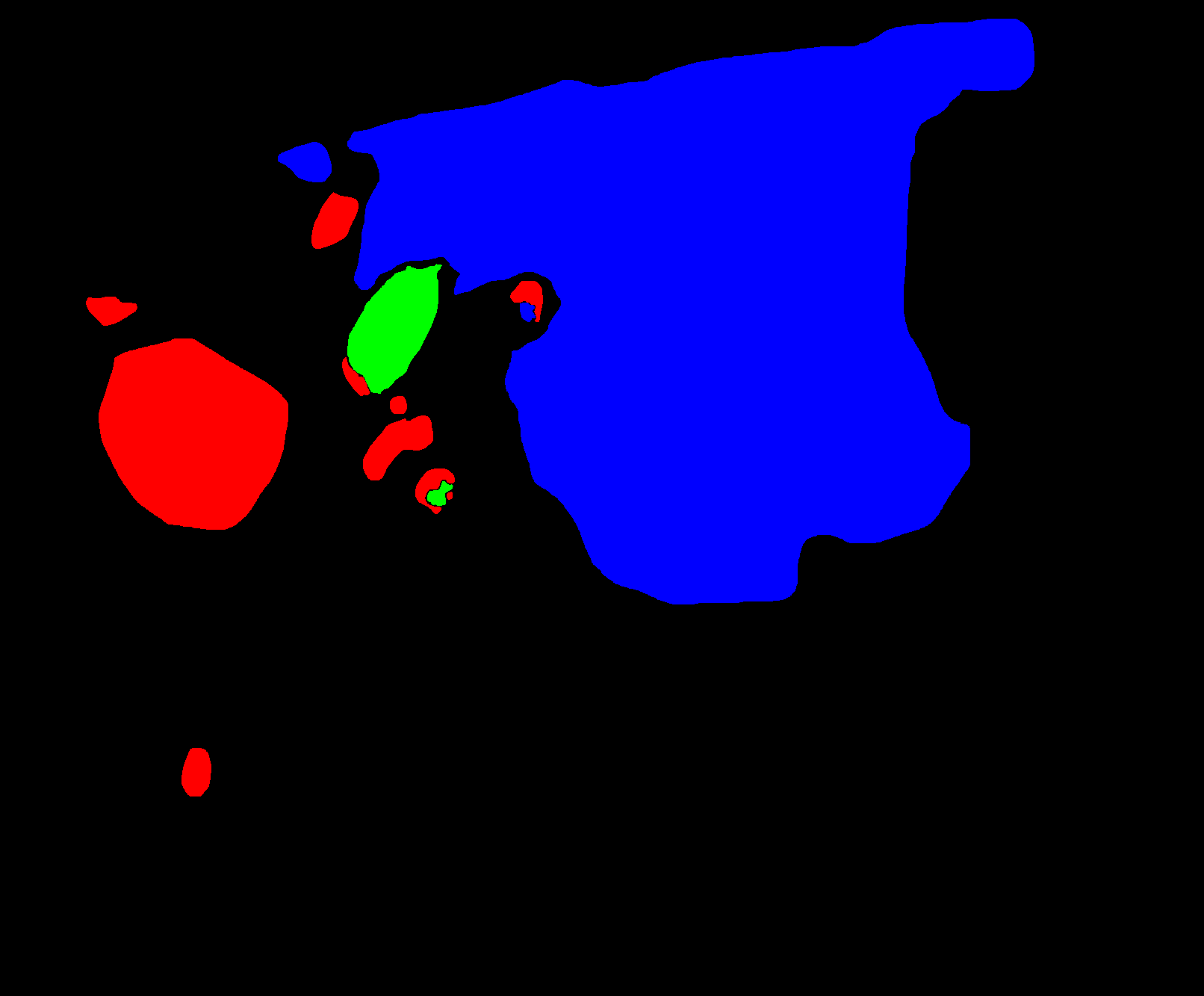}\\
%\vspace{0.01pt}
\end{tabular}
\end{center}
\caption{Comparison of our method and other works on image A05 in the BACH dataset \cite{aresta2019bach}. Best viewed in color.}
\label{fig:bachresults}
\end{figure*}

\subsubsection{Experimental Setup}
BACH (BreAst Cancer Histology)~\cite{aresta2019bach} is a dataset for breast cancer histology microscopy segmentation\footnote{Data can be found at \url{https://iciar2018-challenge.grand-challenge.org/}. Due to the unavailability of the actual test set, we used slides A05 and A10 for testing, slide A02 for validation, and all other slides for training. This provides a fair class distribution, as not all slides contained all semantic classes.}. This dataset consists of high resolution whole-slide images that contain an entire sampled tissue. The whole-slide images were annotated by two medical experts, and images with disagreements were discarded. There are four classes: \textit{normal}, \textit{benign}, \textit{in situ carcinoma} and \textit{invasive carcinoma}. An example of a whole-slide image and its labels is shown in Fig.~\ref{fig:bachexample}. As the \textit{normal} class is considered background, it is not evaluated. Side information for BACH consists of expert brush stroke annotations, indicating the potential presence of each class. In this case study, we use four different brush stroke colors to annotate the four classes.

BACH dataset does not include actual expert-annotated brush strokes. Therefore, to evaluate our method, we simulated expert annotations by using ground-truth labels in the dataset. Since the ground-truth was created by two experts, our brush strokes can be viewed as simulated rough expert input. To simulate situations where users have limited annotation time, we skipped annotating small regions that are likely to be omitted under time constraints. Fig.~\ref{fig:bachexample} shows an example of our simulated brush strokes. In our experiments, we used slides A05 and A10 for testing, slide A02 for validation, and all other slides for training. 

% \vspace{-0.2in}
\noindent
\begin{minipage}{\textwidth}
   \begin{minipage}{0.5\textwidth}
     \centering
     \begin{table}[H]
        \caption{Segmentation performance (mIOU) on BACH dataset. Best performances are highlighted.}
        \begin{tabular}{cccc}
        \toprule
        \textbf{Approach} & \textbf{A05} & \textbf{A10} & \textbf{Mean}\\
        \cmidrule{1-4}
        Deeplab-ResNet~\cite{chen2018deeplab} & 34.08\% & 21.64\% & 27.86\% \\ 
        GrabCut~\cite{rother2004grabcut} & 30.20\%  & 25.21\% & 27.70\% \\
        Unified$^{\mathrm{*}}$~\cite{workman2017unified} & 41.50\% & 17.23\% & 29.37\% \\
        SideInfNet & \underline{59.03\%} & \underline{35.45\%} & \underline{47.24\%} \\
        \bottomrule
        %\multicolumn{4}{l}{$^{\dagger}$ Averaged over binary segmentation scores.} \\
        \multicolumn{4}{l}{$^{\mathrm{*}}$ Our implementation.}
        \end{tabular}
        \label{table:bachresults}
    \end{table}
   \end{minipage}
%   \hspace{0.03\linewidth}
   \begin{minipage}{0.45\textwidth}
     \centering
     \begin{table}[H]
        \caption{Segmentation performance on Zurich Summer dataset. Best performances are highlighted.}
        \begin{tabular}{ccc}
        \toprule
        \textbf{Approach} & \textbf{Accuracy} & \textbf{mIOU} \\
        \cmidrule{1-3}
        Deeplab-ResNet~\cite{chen2018deeplab} & 73.20\% & 42.95\% \\
        GrabCut~\cite{rother2004grabcut} & 60.53\% & 26.89\% \\
        Unified$^{\mathrm{*}}$~\cite{workman2017unified} & 68.20\% & 42.09\% \\
        SideInfNet & \underline{78.97\%} & \underline{58.31\%} \\
        \bottomrule
        %\multicolumn{3}{l}{$^{\dagger}$ Averaged over binary segmentation scores.} \\
        \multicolumn{3}{l}{$^{\mathrm{*}}$ Our implementation.}
        %\multicolumn{3}{l}{$^{\dagger}$ Binary segmentation. Cannot be directly compared.}
        \end{tabular}
        \label{table:zurichresults}
     \end{table}
   \end{minipage}
\end{minipage}
%\vspace{0.1in}

%\begin{wraptable}{R}{0.52\textwidth}
%\small
%    \centering
%    \caption{Experimental results on BACH.}
%    %\vspace{-20mm}
%    \begin{tabular}{@{}cccc@{}}
%        \toprule
%        \multirow{2}{*}{\textbf{Approach}} & %\multicolumn{3}{c}{\textbf{mIOU}} \\ \cmidrule{2-4} 
%        & \textbf{A05} & \textbf{A10} & \textbf{Average} \\ %\cmidrule{1-4}
%        Deeplab-ResNet \cite{chen2018deeplab} & 34.08\% & %21.64\% & 27.86\% \\ 
%        GrabCut$^{\dagger}$ \cite{rother2004grabcut} & 30.20\%  %& 25.21\% & 27.70\% \\
%        Unified$^{\mathrm{*}}$ \cite{workman2017unified} & %41.50\% & 17.23\% & 29.37\% \\
%        SideInfNet & \underline{59.03\%} & \underline{35.45\%} & %\underline{47.24\%} \\
%        \bottomrule
%        \multicolumn{4}{l}{$^{\dagger}$ Averaged over binary %segmentation scores.} \\
%        \multicolumn{4}{l}{$^{\mathrm{*}}$ Our implementation.}
%    \end{tabular}
%    \label{table:bachresults}
%    %\vspace{-20mm}
%\end{wraptable}

\subsubsection{Results}
We evaluate three different methods: our proposed SideInfNet, Unified model \cite{workman2017unified}, and GrabCut \cite{rother2004grabcut}. We were unable to run the HO-MRF model \cite{feng2018urban} on the BACH dataset due to the large size of the whole-slide images (note that the HO-MRF makes use of fully-connected MRF and thus is not computationally feasible under this context). In addition, since GrabCut is a binary segmentation method, to adapt this work to our case study, we ran the GrabCut model independently for each class. We report the performance of all the methods in Table~\ref{table:bachresults}. We also provide some qualitative results in Fig.~\ref{fig:bachresults}.

Experimental results show that our method greatly outperforms previous works on BACH dataset. Furthermore, the Unified model~\cite{workman2017unified} even performs worse than the baseline Deeplab-ResNet that used only whole-slide imagery. This suggests the limitation of the Unified model~\cite{workman2017unified} in learning from dense annotations. Table~\ref{table:bachresults} also confirms the role played by the side information (i.e., the Deeplab-ResNet vs SideInfNet). This aligns with our intuition, as we would expect that brush strokes provide stronger cues to guide the segmentation. 

\subsection{Urban Segmentation}
\subsubsection{Experimental Setup}
\label{sec:zurich}
\begin{figure*}[t]
\centering
\includegraphics[width=.30\textwidth]{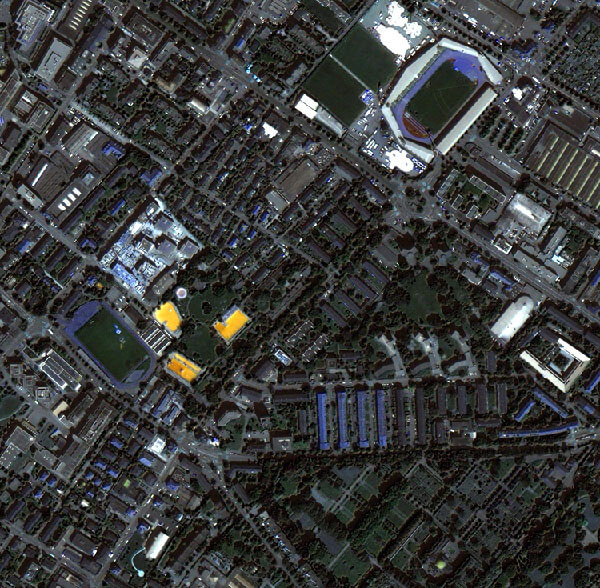}\hfill
\includegraphics[width=.30\textwidth]{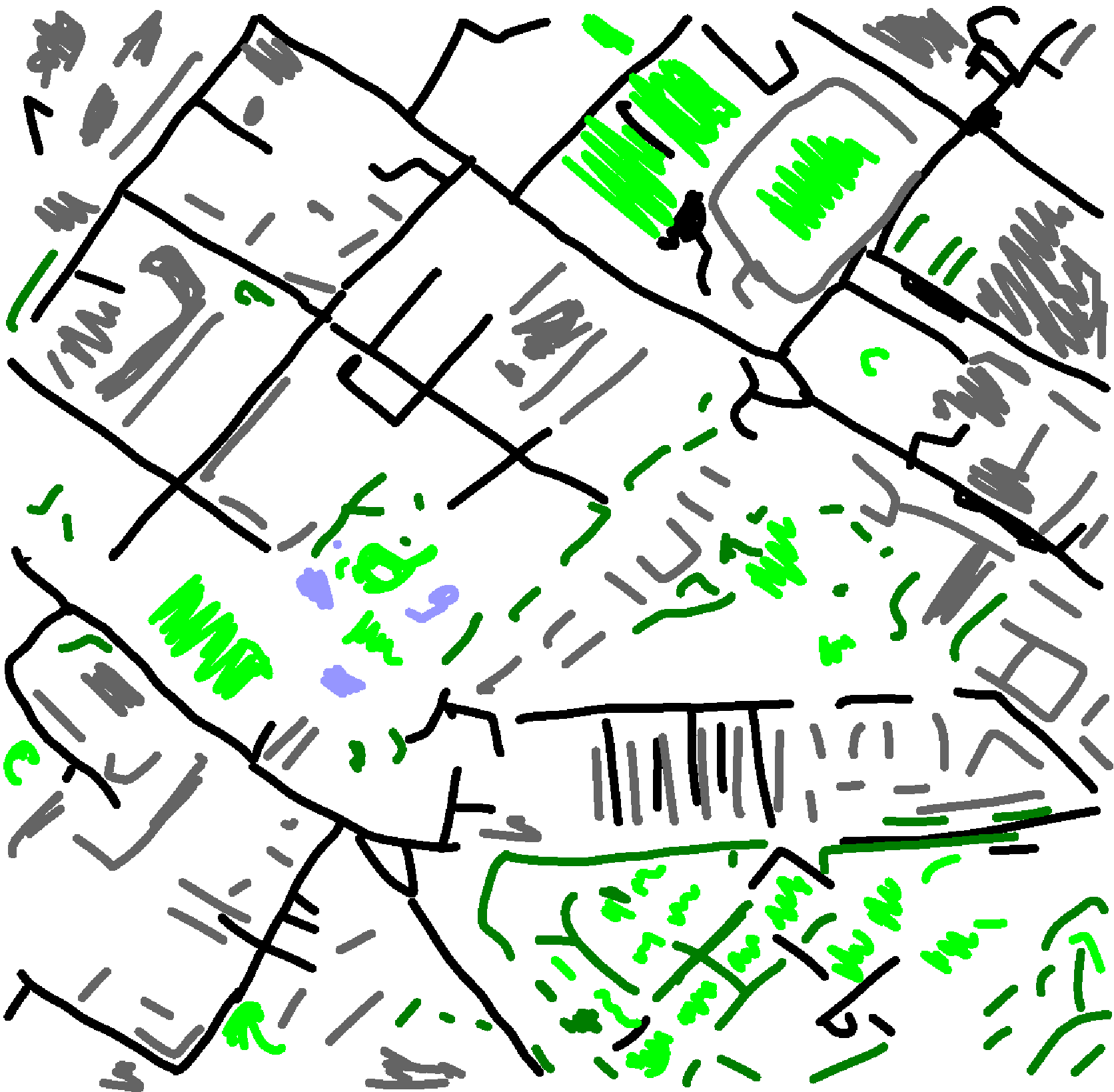}\hfill
\includegraphics[width=.30\textwidth]{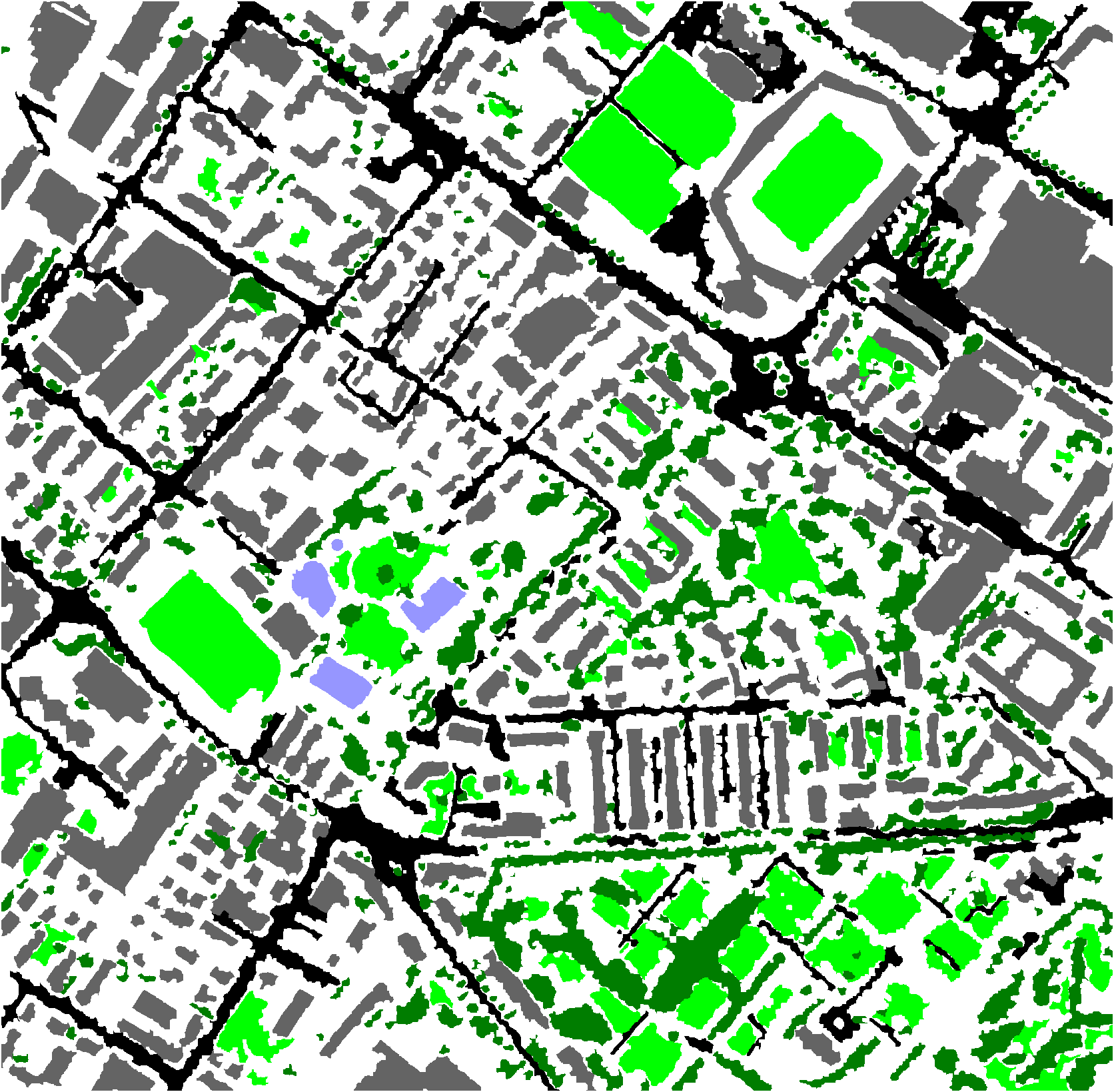}
\caption{Example satellite image, brush annotations, and ground-truth map from the Zurich Summer dataset \cite{volpi2015semantic}. Best viewed in color.}
\label{fig:zurichexample}
\end{figure*}

The Zurich Summer v1.0 dataset~\cite{volpi2015semantic} includes 20 very high resolution (VHR) overview crops taken from the city of Zurich, pansharpened to a PAN resolution of about 0.62 centimeters ground sampling distance (GSD). This is a much higher resolution compared to the low-resolution satellite imagery used in the zoning dataset. The Zurich Summer dataset contains eight different urban and periurban classes: Roads, Buildings, Trees, Grass, Bare Soil, Water, Railways and Swimming pools. Examples of satellite imagery, ground-truth labels, and brush annotations are shown in Fig.~\ref{fig:zurichexample}. Preprocessing steps and feature map construction are performed similarly to that of BACH. We also used rough brush strokes demarcating potential urban classes as side information.

\begin{figure*}[t]
% \centering
% \scalebox{0.95}{
\begin{center}
\begin{tabular}{l@{\ }c@{\ }c@{\ }c@{\ }c}
% \hline

\multicolumn{1}{c}{Deeplab-ResNet \cite{chen2018deeplab}}
& \multicolumn{1}{c}{Unified \cite{workman2017unified}}
& \multicolumn{1}{c}{SideInfNet}
& \multicolumn{1}{c}{Groundtruth} \\
\includegraphics[width=.245\textwidth]{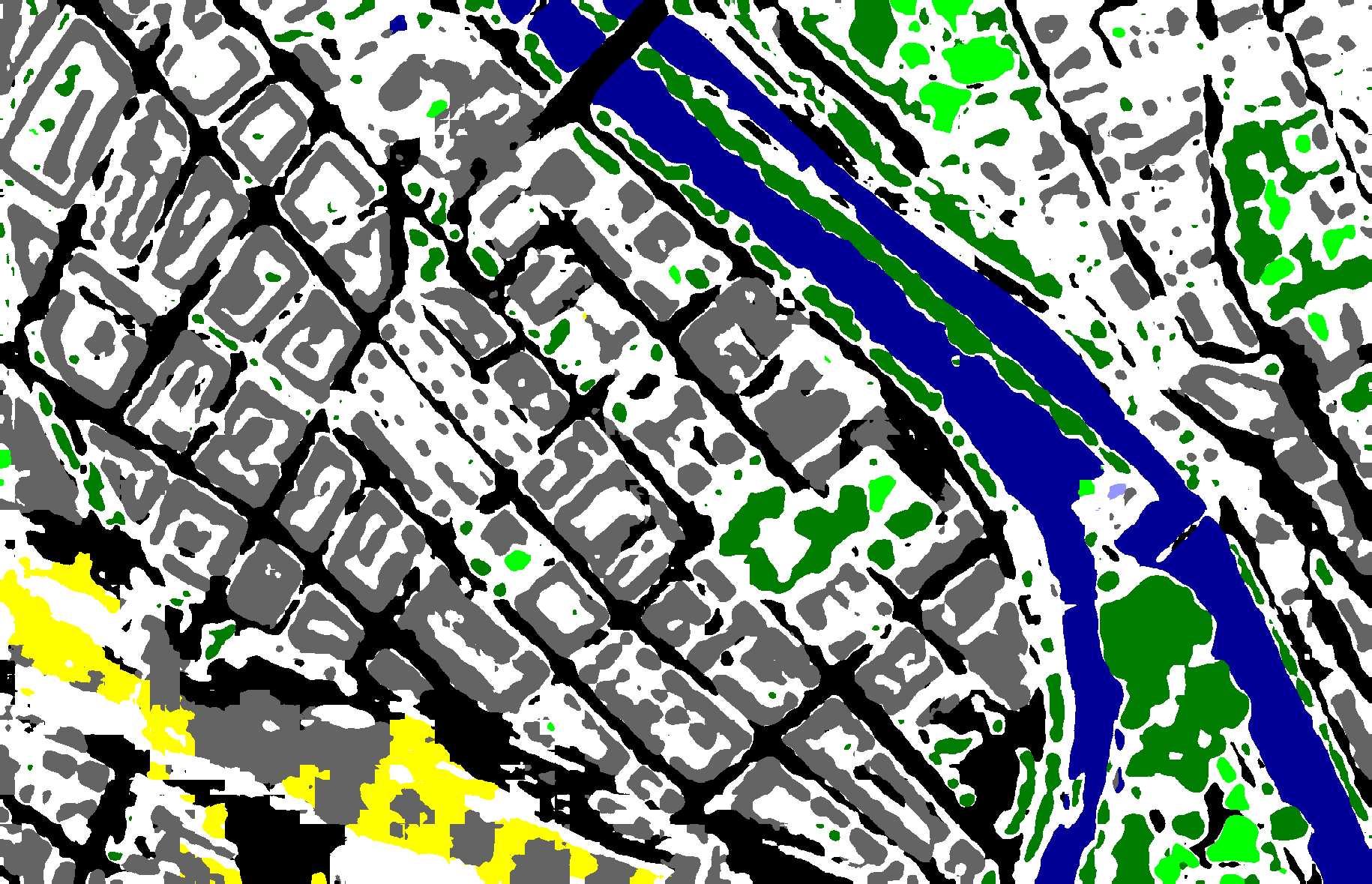} &
\includegraphics[width=.245\textwidth]{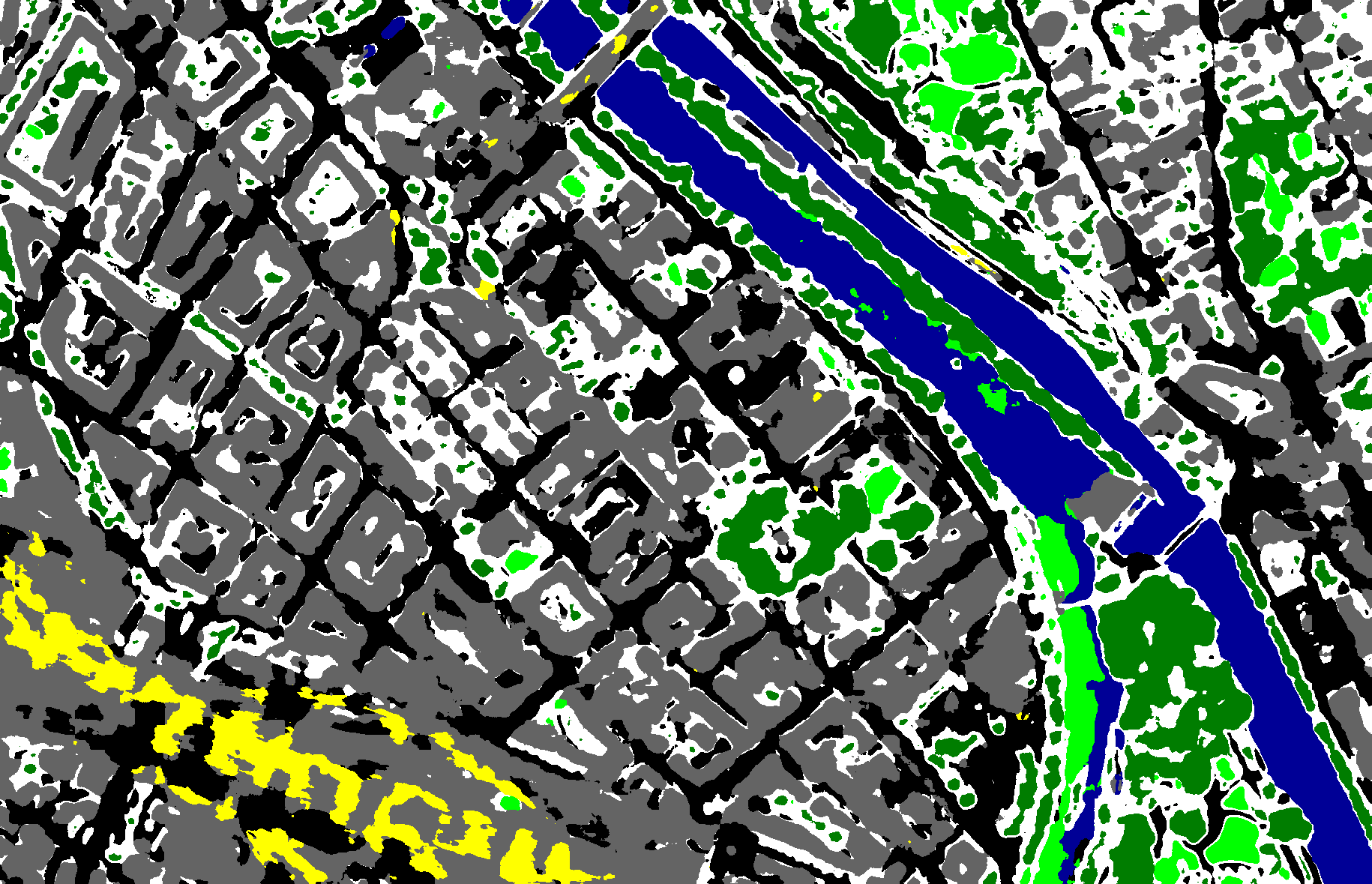} &
\includegraphics[width=.245\textwidth]{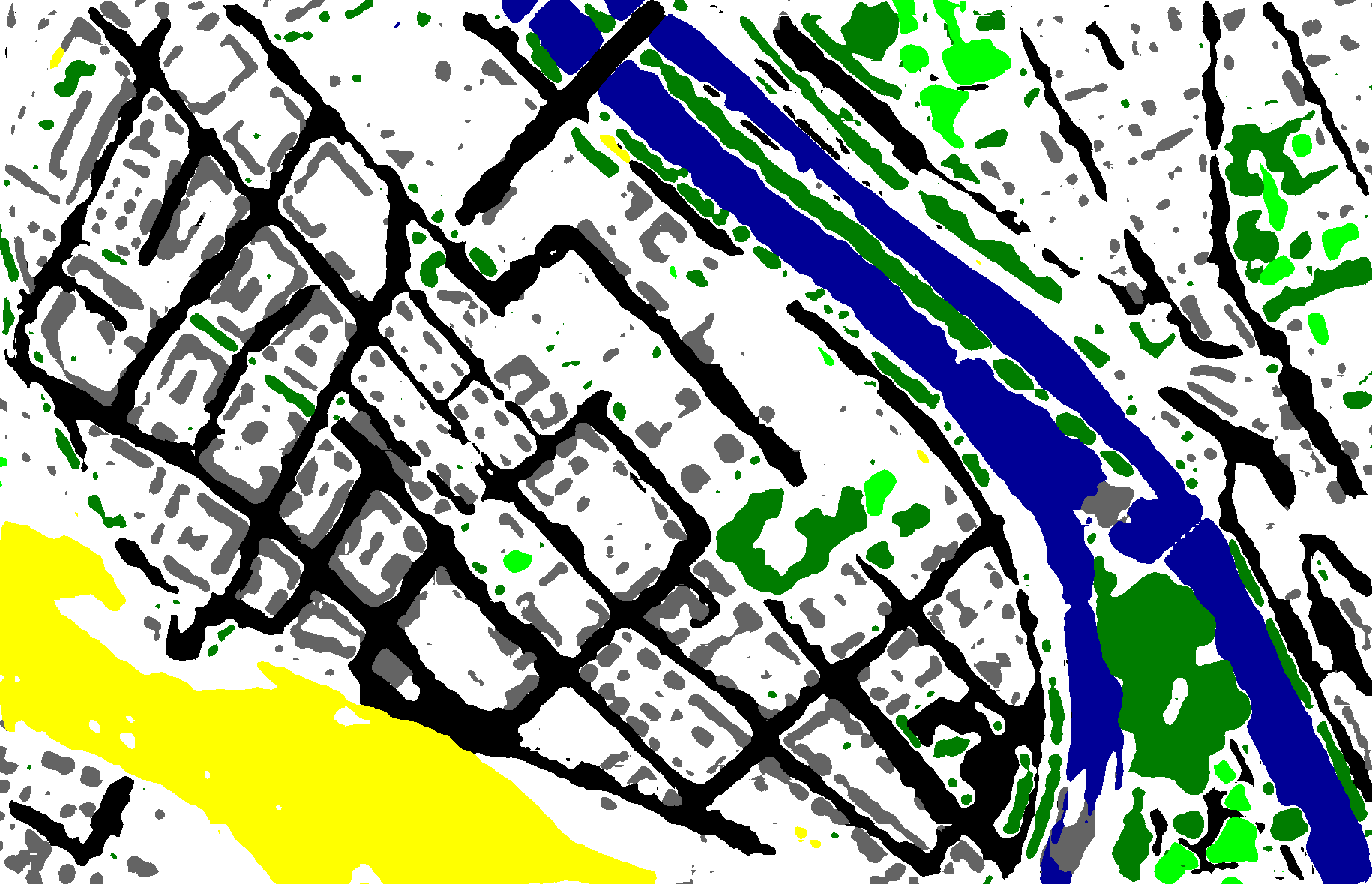} &
\includegraphics[width=.245\textwidth]{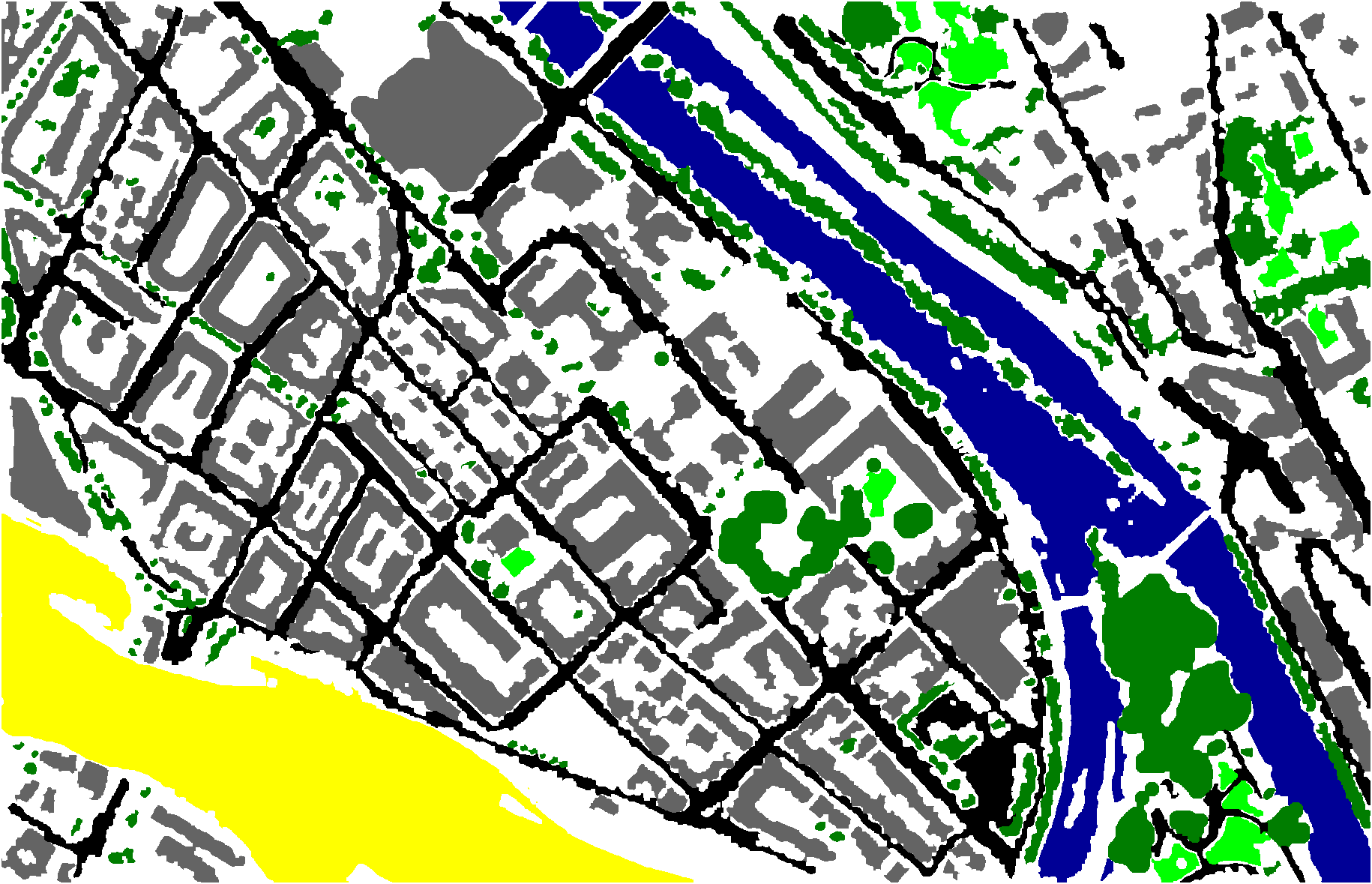}\\
%\vspace{0.01pt}
\end{tabular}
\end{center}
\caption{Qualitative comparison of our method and other works on the Zurich Summer dataset \cite{volpi2015semantic}. Best viewed in color.}
\label{fig:zurichresults}
\end{figure*}

\subsubsection{Results}
Our experimental results on the Zurich Summer dataset are summarized in Table~\ref{table:zurichresults}. In general, similar trends with the BACH dataset are found, and our proposed method outperforms all prior works. Specifically, by using brush strokes, we are able to gain a relative improvement of 7.88\% on accuracy and 35.76\% on mIOU over the baseline Deeplab-ResNet. The Zurich dataset contains high-resolution satellite imagery, which suggests the usefulness of including brush annotations even with high fidelity image data. SideInfNet also outperforms the Unified model~\cite{workman2017unified} with a relative improvement of 15.79\% on accuracy and 38.53\% on mIOU. This result proves the robustness of our method in dealing with dense annotations, which challenge the Unified model. GrabCut also under-performs due to its limitations as an unsupervised binary segmentation method. A qualitative comparison of our method with other works is also shown in Fig.~\ref{fig:zurichresults}.

\subsection{Varying Levels of Side Information}

In this experiment, we investigate the performance of our method when varying the availability of side information. To simulate various densities of brush strokes for an input image, we sample the original brush strokes (e.g., from 0\% to 100\% of the total number) and evaluate the segmentation performance of our method accordingly. The brush strokes could be randomly sampled. However, this approach may bias the spatial distribution of the brush strokes. To maintain the spatial distribution of the brush strokes for every sampling case, we perform $k$-means clustering on the original set of the brush strokes. For instance, if we wish to utilize a percentage $p$ of the total brush strokes, and $n$ brush strokes are present in total, we apply $k$-means algorithm with $k = \text{ceil}(np)$ on the centers of the brush strokes to spatially cluster the brush strokes into $k$ groups. For each group, we select the brush stroke whose center is closest to the group's centroid. This step results in $k$ brush strokes. We note that a similar procedure can be applied to sample street-level photos for zone segmentation.

\begin{table*}[t]
\caption{Performance of SideInfNet with varying side information.}
\begin{center}
\begin{tabular}{ccccccc}
\toprule
\multirow{2}{*}{\multirow{2}{*}{\begin{tabular}[c]{@{}c@{}}\textbf{Side Information}\\ \textbf{Used}\end{tabular}}} & \multicolumn{3}{c}{\textbf{mIOU}} & \multicolumn{3}{c}{\textbf{Mean Accuracy}} \\
\cmidrule{2-7}
& Zoning~\cite{feng2018urban} & BACH~\cite{aresta2019bach} & Zurich~\cite{volpi2015semantic} & Zoning~\cite{feng2018urban} & BACH~\cite{aresta2019bach} & Zurich~\cite{volpi2015semantic} \\
\cmidrule{1-7}
100\% & 47.29\% & 47.24\% & 58.31\% & 74.00\% & 71.99\% & 78.97\% \\
% 90\%  & & 45.15\% & & & & \\
80\%  & 40.27\% & 40.53\% & 52.32\% & 72.46\% & 68.60\% & 77.58\% \\
% 70\%  & & 37.35\% & & & \\
60\%  & 39.56\% & 34.16\% & 52.14\% & 72.39\% & 68.56\% & 76.33\% \\
% 50\%  & & 30.60\% & & & \\
40\%  & 37.70\% & 29.56\% & 49.49\% & 71.01\% & 64.87\% & 75.83\% \\
% 30\%  & & 27.27\% & & & & \\
20\%  & 34.04\% & 26.15\% & 47.72\% & 68.11\% & 56.86\% & 74.29\% \\
% 10\%  & & 25.46\% & & & & \\
0\%   & 28.11\% & 23.86\% & 45.98\% & 58.63\% & 60.48\% & 73.36\% \\
\bottomrule
\end{tabular}
\label{table:reducedresults}
\end{center}
\end{table*}

We report the quantitative results of our method w.r.t varying side information in Table~\ref{table:reducedresults}. In general, we observe a decreasing trend over the accuracy and mIOU as the proportion of side information decreases. This supports our hypothesis that side information is a key signal for improving segmentation accuracy. We also observe a trade off between human effort and segmentation accuracy. For instance, on the zone segmentation dataset~\cite{feng2018urban}, improvement over the baseline Deeplab-ResNet is achieved with as little as 20\% of the original number of geotagged photos. This suggests that our proposed method can provide significant performance gains even with minimal human effort.

\subsection{SideInfNet with another CNN Backbone}

\begin{table}[t]
\caption{Performance (mIOU) of SideInfNet with VGG.}
\begin{center}
\begin{tabular}{cccc}
\toprule
\textbf{Model} & Zoning \cite{feng2018urban} & BACH \cite{aresta2019bach} & Zurich \cite{volpi2015semantic} \\ % & Zoning & BACH & Zurich \\
\cmidrule{1-4}
SideInfNet-VGG & \underline{46.12\%} & \underline{49.53\%} & \underline{49.73\%} \\ % & 70.76\% & \% & \underline{77.74\%} \\
Unified \cite{workman2017unified} & 45.05\% & 29.37\% & 42.09\% \\ % & \underline{71.58\%} & \% & 68.20\% \\
\bottomrule
\end{tabular}
\label{table:compare}
\end{center}
% \vspace{-0.5cm}
\end{table}

To show the adaptability of SideInfNet, we experimented SideInfNet built with another CNN backbone. In particular, we adopted the VGG-19 as the backbone in our architecture. Note that VGG was also used in the Unified model~\cite{workman2017unified}. To provide a fair comparison, we re-implemented both SideInfNet and Unified model with the same VGG architecture and evaluated both models using the same training/test split. We also utilized the original hyperparameters proposed in~\cite{workman2017unified} in our implementation. We report the results of this experiment in Table~\ref{table:compare}. 

Experimental results show that SideInfNet outperforms the Unified model~\cite{workman2017unified} on all segmentation tasks when the same VGG backbone is used. These results confirm again the advantages of our method in feature construction and fusion.

%Experimental results show that, in general SideInfNet with Deeplab-ResNet backbone achieves state-of-the-art performance on many tasks. Moreover, Deeplab-ResNet outperforms VGG in both our architecture and that in~\cite{workman2017unified}. Specifically, a slight improvement (about 2\% mIOU) is observed by Deeplab-ResNet on the zone and BACH segmentation tasks. In contrast, on the urban segmentation task, Deeplab-ResNet greatly outperforms VGG (about 9\% mIOU increase with our architecture and 4\% mIOU with ~\cite{workman2017unified}). These resutls also confirm the superiority of our architecture over \cite{workman2017unified} on both Deeplab-ResNet and VGG backbone.

\section{Conclusion}
\noindent This paper proposes SideInfNet, a novel end-to-end neural network for semi-automatic semantic segmentation with additional side information. Through extensive experiments on various datasets and modalities, we have shown the advantages of our method across a wide range of applications, including but not limited to remote sensing and medical image segmentation. In addition to being general, our method boasts improved accuracy and computational advantages over prior models. Lastly, our architecture is easily adapted to various semantic segmentation models and side information feature extractors.

The method proposed in this paper acts as a compromise between fully-automatic and manual segmentation. This is essential for many applications with high cost of failure, in which fully-automatic methods may not be widely accepted as of yet. Our model works well with dense brush stroke information, providing a quick and intuitive way for human experts to refine the model's outputs. In addition, our model also outperforms prior work on sparse pixel-wise annotations. By including side information to shape predictions, we are able to achieve an effective ensemble of human expertise and machine efficiency, producing both fast and accurate segmentation results.

\section{Acknowledgement}

\begin{itemize}
\item Duc Thanh Nguyen was partially supported by an internal SEBE 2019 RGS grant from Deakin University.
\item Sai-Kit Yeung was partially supported by an internal grant from HKUST (R9429) and HKUST-WeBank Joint Lab.
\item Alexander Binder was supported by the MoE Tier2 Grant MOE2016-T2-2-154, Tier1 grant TDMD 2016-2, SUTD grant SGPAIRS1811, TL grant RTDST1907012.
\end{itemize}

\pagebreak
\appendix

\section{Supplementary Material}
In this supplementary material, we provide implementation details of our SideInfNet in Section~\ref{sec:experimentalsetup}. We present ablation experiments conducted to ascertain the effectiveness of our method in Section~\ref{sec:ablationstudies}, in which we compare against existing fusion methods by implementing our model with the same baseline segmentation network. Computational analysis of our method is performed in Section~\ref{sec:computationalanalysis}. We present additional qualitative evaluations of our method and prior works in three case studies in Section~\ref{sec:qualitativeevalution}.

\section{Implementation Details}
\label{sec:experimentalsetup}
\noindent
In this section, we detail the settings used to train our proposed SideInfNet in various case studies. All models of our SideInfNet were implemented in PyTorch v1.2 \cite{paszke2017automatic}.

\subsection{General Settings}
The domain-dependent feature extractor of our proposed SideInfNet is built based on the Deeplab-ResNet \cite{chen2018deeplab} model. In our implementation, we optimized the Deeplab-ResNet using stochastic gradient descent (SGD) with a momentum of 0.9. The Deeplab-ResNet receives input an image of size $H\times W$ pixels and produces a \textit{conv2\_3} layer output of approximately size $\frac{H}{4}\times \frac{W}{4}$. Therefore, our maxpool layer uses a kernel size of 6 and a stride of 4 to achieve the desired size.

%Given an image input of size $H\times W$ pixels as input, the original Deeplab-ResNet \textit{conv2\_3} layer output is of approximately $\frac{H}{4}\times \frac{W}{4}$ size. Hence, our maxpool layer uses a kernel size of 6 and a stride of 4 to achieve the desired size.

For processing side information, we used a single fully-connected layer that mapped input vectors to a 64-dimensional space. As shown in Section~\ref{sec:ablationstudies}, this setting (i.e., 64 dimensions for side information) balanced both the accuracy and computational cost and worked well in all case studies. We also experimented with deeper multi-layer perceptrons (MLP) and non-linear activation functions, but found no improvement from these settings.

%For the Unified model \cite{workman2017unified}, we followed the original implementation of the work for hyperparameters and optimization choices.

% We follow the implementation of \cite{veit2018convolutional} in the stochastic gates.

%\subsubsection{Maxpooling for Fusion}

%\subsection{Transfer Learning}
%\noindent

For all tasks, transfer learning was applied. We initialized our SideInfNet models with the weights of the Deeplab-ResNet trained for semantic segmentation on the Microsoft COCO (MS-COCO) dataset \cite{lin2014microsoft}. Due to concatenation of data in our models, the weights of the layers at the concatenation point (i.e., the \textit{conv2\_3} layer in the Deeplab-ResNet architecture) cannot be directly restored. Instead, we randomly initialized additional channels that are required for the concatenation. Specifically, we restored the first 256 channels of the \textit{conv2\_3} layer from the MS-COCO pretrained weights, and randomly initialized the additional 64 channels.

\subsection{Zone Segmentation}

\subsubsection{Training}
\label{sec:zoneseg_training}
For training, we used $80 \times 80$ pixels crops of each city from the zone segmentation dataset \cite{feng2018urban}. Patches containing more than 60\% of masked data were discarded. Each patch was saved along with its coordinate information for retrieval of the geotagged photo data. We normalized images by performing mean subtraction from the RGB channels using the training set mean.

To augment the training data, we performed random horizontal and vertical flips of image patches. We also experimented with scaling the patches, but did not observe any improvement in performance.

%\subsubsection{Hyperparameters} \label{sec:urbanhyperparams}
Training was performed with a mini-batch size of 16 and over 20 epochs. We used a base learning rate of 0.00025 with a polynomial learning rate decay with power of 0.9. In addition, we set the learning rate for the MLP to 0.025, and for the data fusion \textit{conv2\_3} to 0.0005 respectively. The reason for this setting is that these layers are not restored through transfer learning, and benefit from higher learning rates. To make the training stable, we used a learning rate warmup of 20 data epochs, in which the learning rate linearly increased from epoch 1 to epoch 20.

In our models, each fractionally-strided convolution was multiplied by a learnable scalar. We initialized all the scalars to 1, which we found to be helpful in diffusing geotagged photo data. Intuitively, this initialization could result in maximum diffusion by default, which we found essential to aid in learning meaningful representations for sparse side information.

\subsubsection{Evaluation}
We tested our model using 3-fold cross validation, in which two cities were used for training, and the other city was used for testing. In each validation, we scanned the test satellite image by a window of size $80 \times 80$ pixels and a spatial stride of $21 \times 21$ pixels for NYC and BOS, and $23 \times 23$ pixels for SFO (due to the different scales of the input data).

Inference was performed individually on the windows to retrieve the softmax class probabilities. The resulting softmax patches were then merged, and overlapping regions were averaged. The final inference result was achieved by taking an argmax over the averaged softmax result.

\subsection{BreAst Cancer Histology Segmentation}
\subsubsection{Training}
Due to the large size of whole-slide images in the BreAst Cancer Histology (BACH) dataset \cite{aresta2019bach}, we downscaled the whole-slide images for computational efficiency. We first resized the whole-slide images to $\frac{1}{4}$ of their original size. We then cropped patches of $299 \times 299$ pixels with a stride length of $99 \times 99$ pixels. We discarded all patches that contained less than 5\% of non-normal classes. Each patch was saved along with its coordinate information for retrieval of the brush stroke annotations. Lastly, we normalized images through mean subtraction where the mean was derived from training dataset. We also performed random horizontal and vertical flips for data augmentation.

%\subsubsection{Hyperparameters} \label{sec:bachhyperparams}
%Training was performed using a mini-batch size of 4 with gradients accumulated over 4 iterations for an effective mini-batch size of 16.

Training was performed using a mini-batch size of 4 with gradients accumulated over 4 iterations. We used a base learning rate of 0.0001 with a polynomial learning rate decay with power of 0.9. In addition, we set the learning rate for the MLP to 0.01, and of the fusion layer \textit{conv2\_3} to 0.0002 respectively. The learning rate for the classification layer was set to 0.001.

The model was trained for 20 epochs, with early stopping imposed if accuracy on the validation set did not increase for 3 epochs. We used a learning rate warmup of 20 data epochs for stability in training; the learning rate linearly increased from epoch 1 to epoch 20.

The learnable scalar for the first fractionally-strided convolution was set to 1, and all others were set to 0, resulting in no diffusion by default. We found this essential to aid in learning good representations for dense side information, such as brush stroke annotations.

\subsubsection{Evaluation}
We performed inference using patches processed as in the training procedure. We averaged the softmax probabilities of any overlapping regions. Similarly to the zone segmentation case study, the final results were achieved by taking an argmax over the averaged softmax result.

\subsection{Urban Segmentation}

For urban segmentation on the Zurich Summer Dataset \cite{volpi2015semantic}, we cropped training images to patches of size $80 \times 80$ pixels with a stride length of $20 \times 20$ pixels. Images were saved with associated coordinate information for retrieval of brush stroke annotations. We normalized images by performing mean subtraction from the RGB channels using the training set mean. Data augmentation was also done using random horizontal and vertical flips.

Hyperparameter setting was similar to that of the zone segmentation case study. We initialized the learnable scalar for the first fractionally-strided convolution to 1, and all others to 0, similarly to the BACH dataset.

In our experiments, we used the images \textit{zh5}, \textit{zh7}, \textit{zh8}, \textit{zh11}, and \textit{zh18} for testing. All other images were used for training. This split ensures that all classes are present in both training and testing.

\section{Ablation Studies}
\label{sec:ablationstudies}

\begin{table*}[t] % Placed here so it appears in the right page
\caption{Performance of variants of SideInfNet in zone segmentation \cite{feng2018urban}. Best performances are highlighted.}
\centering

\scriptsize

\begin{tabular}{@{}cccccccc@{}}

\toprule
\multirow{2}{*}{\textbf{Approach}} & \multicolumn{1}{c}{\multirow{2}{*}{\begin{tabular}[c]{@{}c@{}}\textbf{Street}\\ \textbf{Photo}\end{tabular}}} & \multicolumn{1}{c}{\multirow{2}{*}{\begin{tabular}[c]{@{}c@{}}\textbf{Fractionally}\\ \textbf{Strided Convolutions}\end{tabular}}} & \multicolumn{1}{c}{\multirow{2}{*}{\begin{tabular}[c]{@{}c@{}}\textbf{Gate}\\ \textbf{Rate} ($t$)\end{tabular}}} & \multicolumn{4}{c}{\textbf{Pixel Accuracy}} \\ \cmidrule(l){5-8}
 & \multicolumn{1}{c}{} & \multicolumn{1}{c}{} & \multicolumn{1}{c}{} & \textbf{BOS} & \textbf{NYC} & \textbf{SFO} & \textbf{Average} \\
\cmidrule{1-8}
Deeplab-ResNet \cite{chen2018deeplab} & - & - & - & 60.79\% & 59.58\% & 72.21\% & 64.19\% \\
Geotagged & \checkmark & - & - & 60.19\% & 58.87\% & 74.18\% & 64.41\% \\
Diffused & \checkmark & \checkmark & - & 69.08\% & \underline{71.95\%} & 79.49\% & 73.51\% \\
SideInfNet & \checkmark & \checkmark & 0.8 & 70.10\% & 70.67\% & 79.38\% & 73.38\% \\
SideInfNet & \checkmark & \checkmark & 0.6 & \underline{71.33\%} & 71.08\% & \underline{79.59\%} & \underline{74.00\%} \\
SideInfNet & \checkmark & \checkmark & 0.4 & 70.45\% & 70.58\% & 79.51\% & 73.51\% \\
\bottomrule
\label{table:ablation}
\end{tabular}
\end{table*}

\subsection{Components of SideInfNet}

In order to validate the benefits of our various technical novelties, we performed several ablation experiments on the main components of our proposed SideInfNet. In this supplementary material we present experimental results from the zone segmentation application, although similar trends are observed from the other case studies as well.

The results are summarized in Table~\ref{table:ablation}. It is shown that the inclusion of side information in the form of street-level photos is essential in improving the segmentation accuracy. In particular, our best performing model (SideInfNet), fusing both domain-dependent features from satellite data and side information, achieved a relative gain of 15.28\% over the baseline Deeplab-ResNet \cite{chen2018deeplab} that uses only satellite imagery. In addition, the results prove that side information diffusion using fractionally-strided convolutions (\textit{Diffused} model) was important for performance improvements. This method of diffusion gained a relative improvement of 14.89\% over the \textit{Geotagged} model, which simply diffused the side information upon spatial distance (via nearest neighbor interpolation).

%concatenated the nearest neighbor outputs without spatial diffusion.

The SideInfNet model with adaptive inference gates also slightly improved over the \textit{Diffused} model. An additional benefit of the adaptive inference gates is reduced computational complexity and model parameters, as not all the layers in the network architecture are executed for each run.

\subsection{Varying Feature Dimension of Side Information}

As presented in the implementation details of SideInfNet in Section~\ref{sec:experimentalsetup}, the side information is fed through a single fully-connected layer to produce a 64-dimensional feature vector. We experimented our SideInfNet with various out sizes of the fully-connected layer including 64, 128, 256, and 512, and report the results on the zoning~\cite{feng2018urban}, BACH~\cite{aresta2019bach}, and Zurich Summer dataset~\cite{volpi2015semantic} in Table~\ref{table:dimSideInformation}.

\begin{table*}[t] % Placed here so it appears in the right page
\caption{Performance of SideInfNet when varying the dimension of side information. Note that ``-'' in the BACH dataset indicates that the model is unable to learn. Best performances are highlighted.}
\centering
\centerline{(a) Zoning~\cite{feng2018urban}}
\begin{tabular}{@{}ccccccccc@{}}
\toprule
\multirow{2}{*}{\textbf{Dimension}} & \multicolumn{4}{c}{\textbf{mIOU}} & \multicolumn{4}{c}{\textbf{Pixel Accuracy}} \\ \cmidrule(l){2-9}
 & \textbf{BOS} & \textbf{NYC} & \textbf{SFO} & \textbf{Average} & \textbf{BOS} & \textbf{NYC} & \textbf{SFO} & \textbf{Average} \\
\cmidrule{1-9}
64 & \underline{41.96\%} & 39.59\% & \underline{60.31\%} & \underline{47.29\%} & \underline{71.33\%} & 71.08\% & \underline{79.59\%} & \underline{74.00\%} \\
128 & 40.63\% & \underline{40.71\%} & 44.98\% & 42.11\% & 70.79\% & \underline{72.00\%} & 72.32\% & 71.70\% \\
256 & 39.52\% & 39.10\% & 57.67\% & 45.43\% & 70.29\% & 70.10\% & 78.31\% & 72.90\% \\
512 & 38.78\% & 40.10\% & 57.00\% & 45.30\% & 69.18\% & 71.15\% & 77.07\% & 72.74\% \\
\bottomrule \\
\end{tabular}

\centerline{(b) BACH~\cite{aresta2019bach}}
\begin{tabular}{@{}ccccccc@{}}
\toprule
\multirow{2}{*}{\textbf{Dimension}} & \multicolumn{3}{c}{\textbf{mIOU}} & \multicolumn{3}{c}{\textbf{Pixel Accuracy}} \\ \cmidrule(l){2-7}
 & \textbf{A05} & \textbf{A10} & \textbf{Average} & \textbf{A05} & \textbf{A10} & \textbf{Average} \\
\cmidrule{1-7}
64 & 59.03\% & 35.45\% & 47.24\% & 89.68\% & 54.29\% & 71.99\% \\
128/256/512 & - & - & - & - & - & - \\
\bottomrule \\
\end{tabular}

\centerline{(c) Zurich~\cite{volpi2015semantic}}
\begin{tabular}{@{}ccc@{}}
\toprule
\textbf{Dimension} & \textbf{mIOU} & \textbf{Pixel Accuracy} \\
\cmidrule{1-3}
64 & \underline{58.31\%} & \underline{78.97\%} \\
128 & 51.69\% & 74.71\% \\
256 & 45.89\% & 73.09\% \\
512 & 41.37\% & 69.94\% \\
\bottomrule
\end{tabular}
\label{table:dimSideInformation}
\end{table*}

Experimental results show that increasing the feature dimensionality of side information (i.e., the output size of the fully-connected layer) on the zoning dataset has a negligible effect on the performance, e.g., <2\% of deviation in mIOU, as the high dimensional side information vectors can be mapped meaningfully. In contrast, on the BACH and Zurich Summer datasets, worse performance is observed when increasing the feature dimensionality of side information. This is likely due to the simplicity of the side information in these datasets, e.g., brush strokes can be represented simply by scalars corresponding to different semantic classes. We also observe that, on the BACH dataset, when the size of the side information exceeds 64, SideInfNet is unable to learn any meaningful features, leading to either random or biased predictions. On the Zurich Summer dataset, side information of brush strokes has dimensionality of 8 and increasing the side information's dimensionality leads to overfitting. Therefore, to make a balance between the performance and computational complexity, we recommend 64-dimensional side information vectors for all the case studies and datasets.

\subsection{Varying Levels of Side Information}

In our main paper, we present an experiment on varying the availability of side information (see Section 4.4 in the main paper). In this experiment, we used all the side information available in training datasets to train the SideInfNet model and tested the model by varying the level of side information in test sets. Recall that, to simulate various levels of side information while keeping the same spatial distribution, we sample the side information, e.g., brush strokes, using $k$-means algorithm applied on the centers of the brush strokes.

\begin{figure*}[t]
\centering
\includegraphics[width=.48\textwidth]{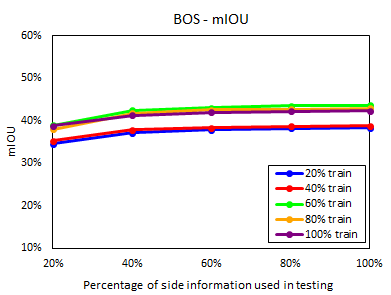}
\includegraphics[width=.48\textwidth]{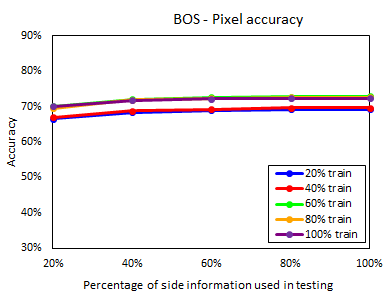}
\includegraphics[width=.48\textwidth]{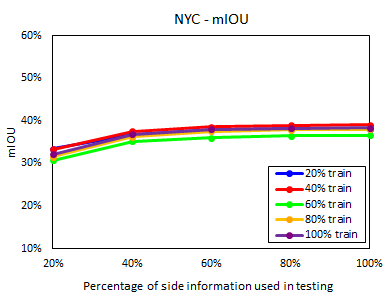}
\includegraphics[width=.48\textwidth]{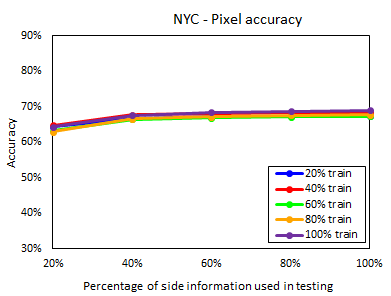}
\includegraphics[width=.48\textwidth]{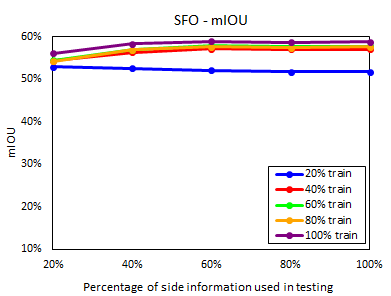}
\includegraphics[width=.48\textwidth]{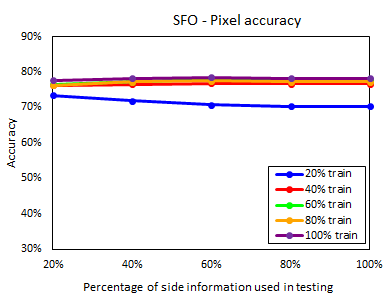}
\includegraphics[width=.48\textwidth]{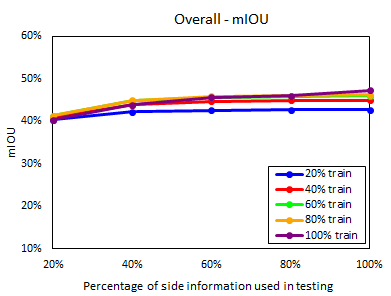}
\includegraphics[width=.48\textwidth]{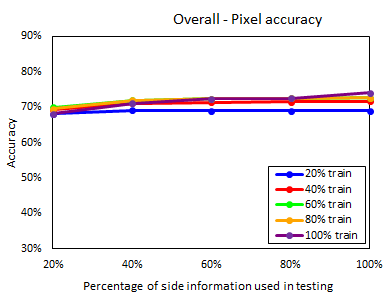}
\caption{Performance of SideInfNet when varying the availability of side information in both training and testing on zoning dataset~\cite{feng2018urban}.}
\label{fig:varyingSideInfZoning}
\end{figure*}

\begin{figure*}[t]
\centering
\includegraphics[width=.48\textwidth]{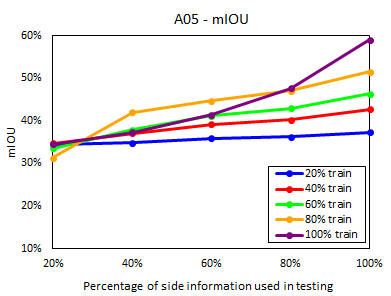}
\includegraphics[width=.48\textwidth]{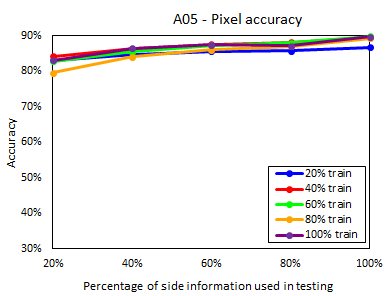}
\includegraphics[width=.48\textwidth]{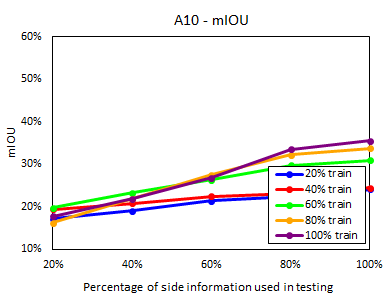}
\includegraphics[width=.48\textwidth]{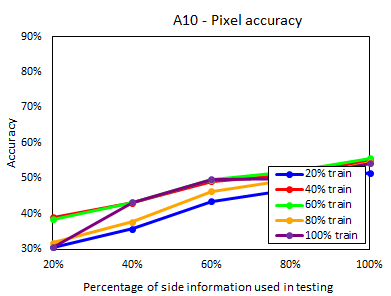}
\includegraphics[width=.48\textwidth]{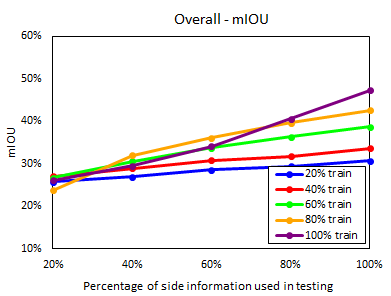}
\includegraphics[width=.48\textwidth]{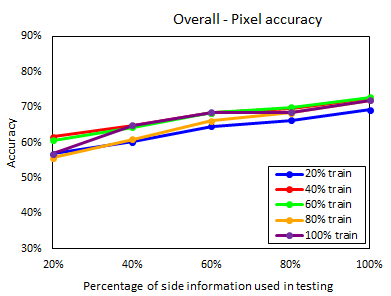}
\caption{Performance of SideInfNet when varying the availability of side information in both training and testing on BACH dataset~\cite{aresta2019bach}.}
\label{fig:varyingSideInfBACH}
\end{figure*}

\begin{figure*}[t]
\centering
\includegraphics[width=.48\textwidth]{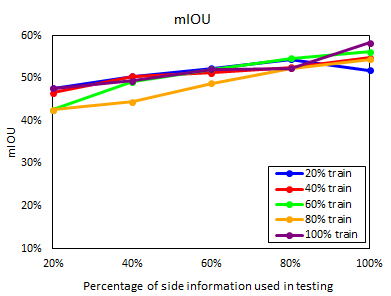}
\includegraphics[width=.48\textwidth]{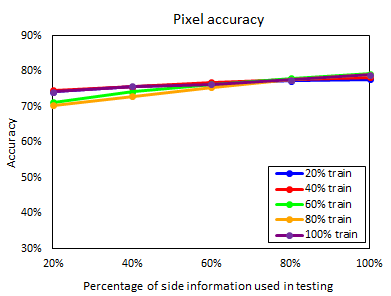}
\caption{Performance of SideInfNet when varying the availability of side information in both training and testing on Zurich Summer dataset~\cite{volpi2015semantic}.}
\label{fig:varyingSideInfZurich}
\end{figure*}

In this supplementary material, we provide more detailed results and in-depth analysis on the results. Specifically, we varied the availability of side information in both the training and test sets, e.g., $x$\% of available side information is used in training vs $y$\% of available side information is used in testing, where $x$ and $y$ vary in 20\%, 40\%, ..., 100\%. We show the detailed performance of SideInfNet (in both mIOU and pixel accuracy) on the zoning~\cite{feng2018urban}, BACH~\cite{aresta2019bach}, and Zurich Summer dataset~\cite{volpi2015semantic} in Fig.~\ref{fig:varyingSideInfZoning}, Fig.~\ref{fig:varyingSideInfBACH}, and Fig.~\ref{fig:varyingSideInfZurich} respectively. From experimental results, we observe that, to achieve the best overall performance, SideInfNet should be trained with 100\% side information available in the training data but can work well at inference time even with fewer side information. This confirms the practicality and applicability of our model in situations where a few annotations from users can significantly improve the segmentation quality.

\begin{figure*}[t]
\centering
\begin{center}
\begin{tabular}{l@{\ }c@{\ }c@{\ }c@{\ }c@{\ }c@{\ }}
 \toprule
\multicolumn{1}{c}{}
& \multicolumn{1}{c}{0\%}
& \multicolumn{1}{c}{20\%}
& \multicolumn{1}{c}{40\%}
& \multicolumn{1}{c}{60\%}
& \multicolumn{1}{c}{80\%}
\\\midrule
\rotatebox{90}{SFO Photos} & 
 &
\includegraphics[width=.19\textwidth]{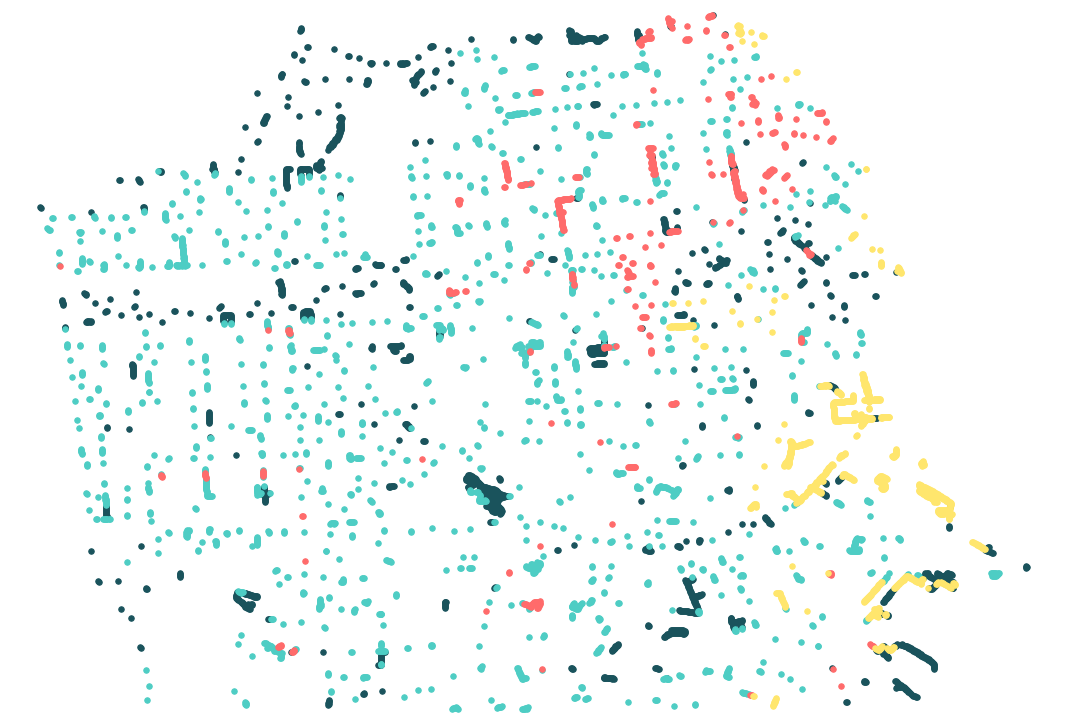} &
\includegraphics[width=.19\textwidth]{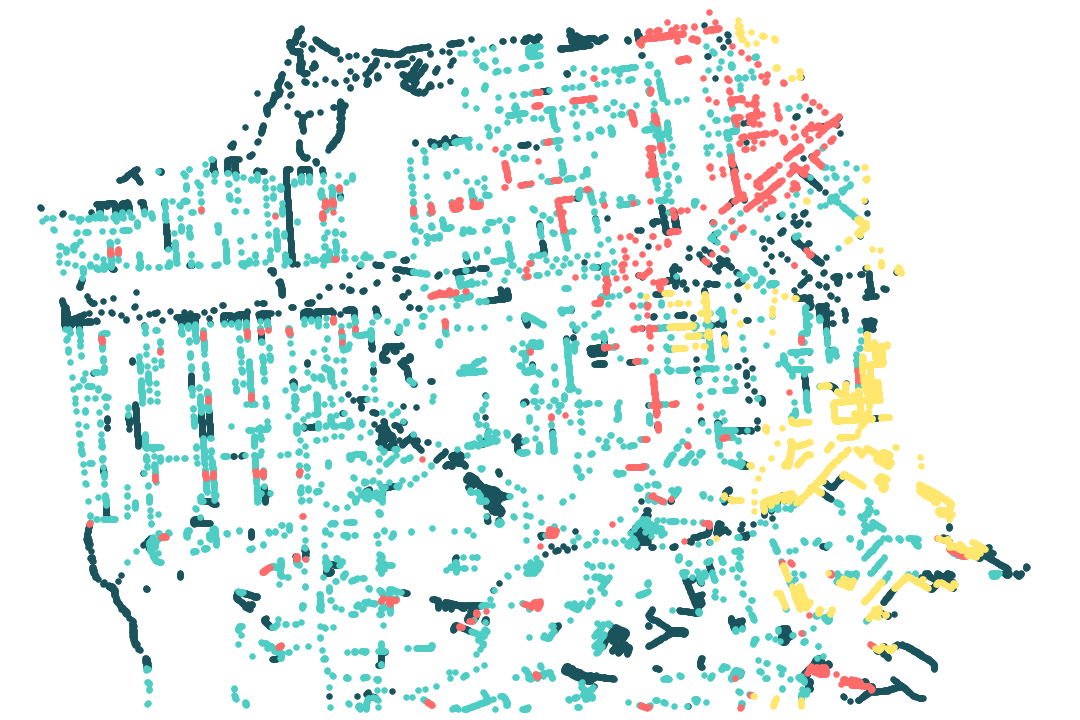} &
\includegraphics[width=.19\textwidth]{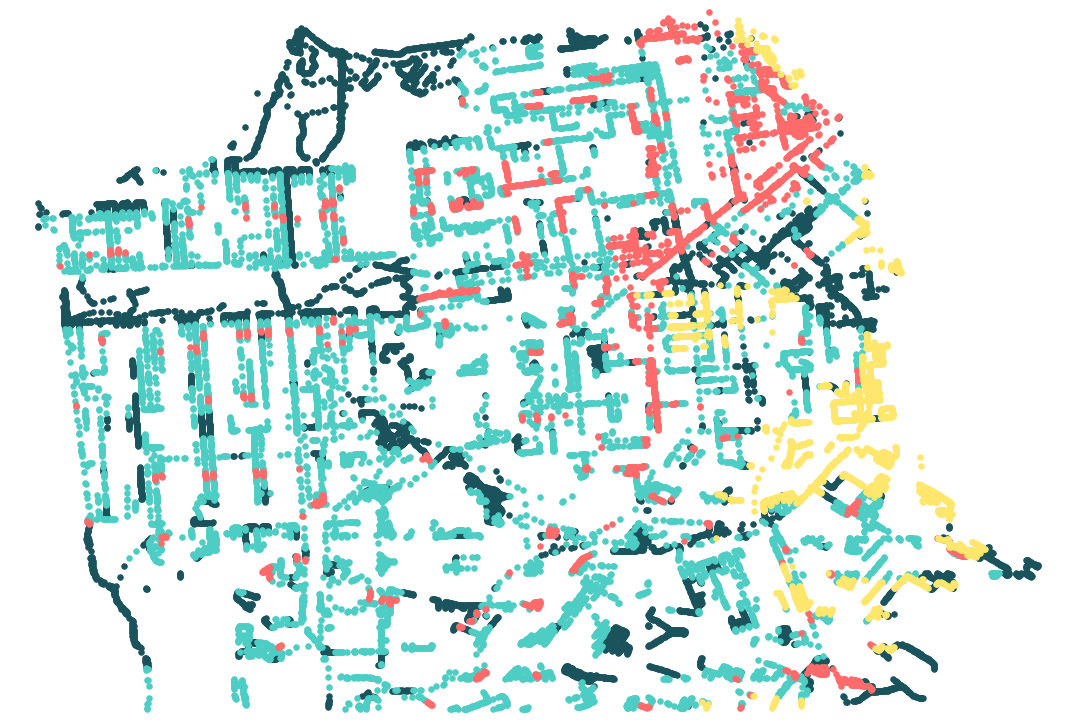} &
\includegraphics[width=.19\textwidth]{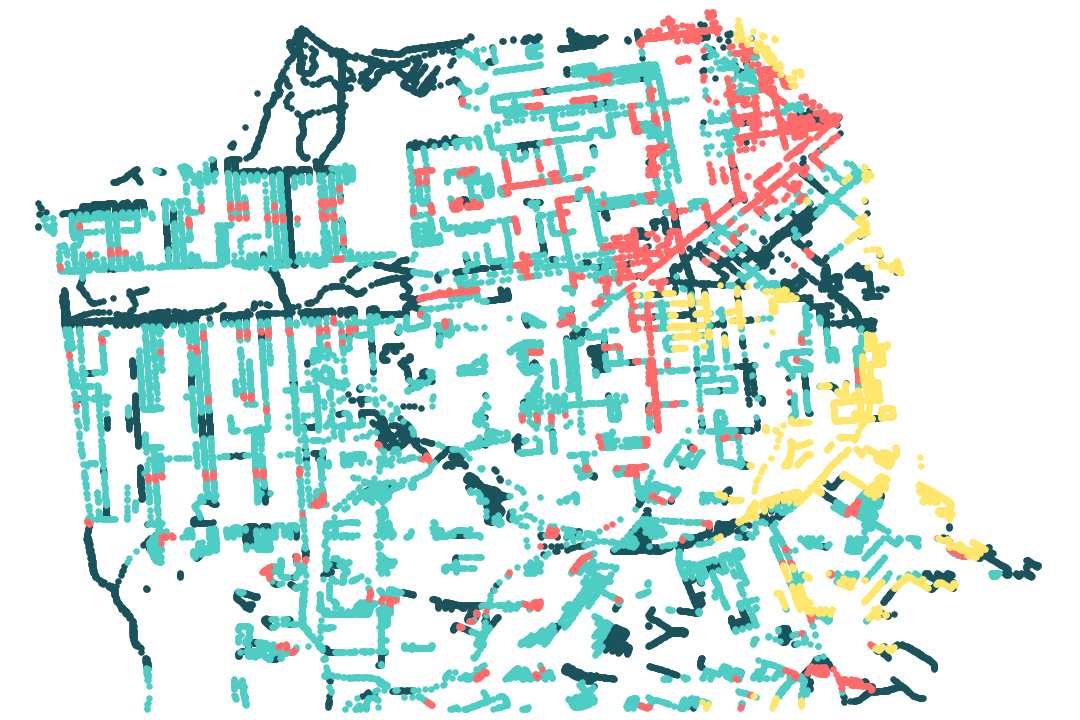} 
\\\midrule
\rotatebox{90}{SFO Results} & 
\includegraphics[width=.19\textwidth]{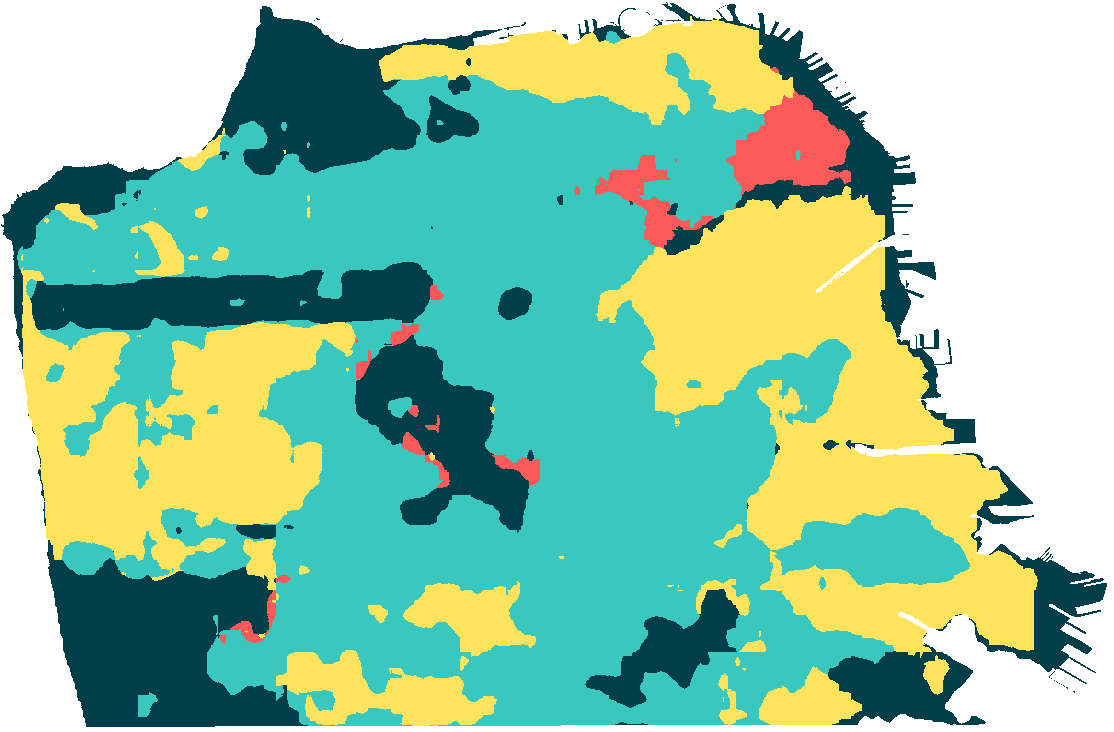} &
\includegraphics[width=.19\textwidth]{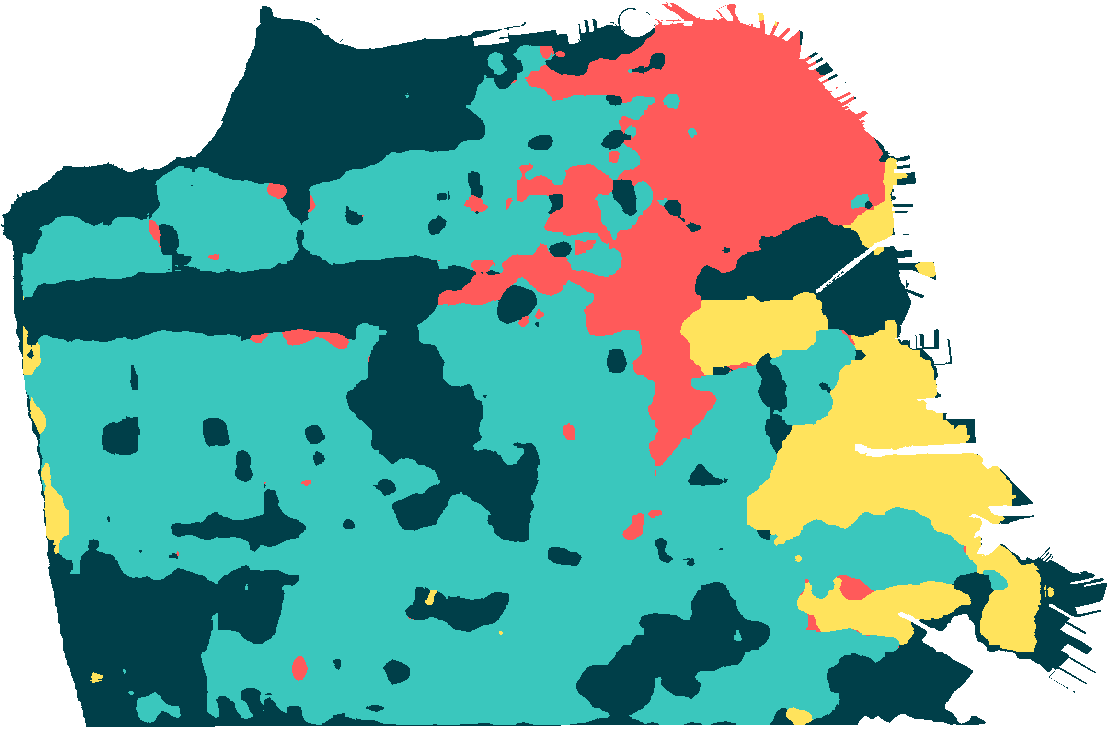} &
\includegraphics[width=.19\textwidth]{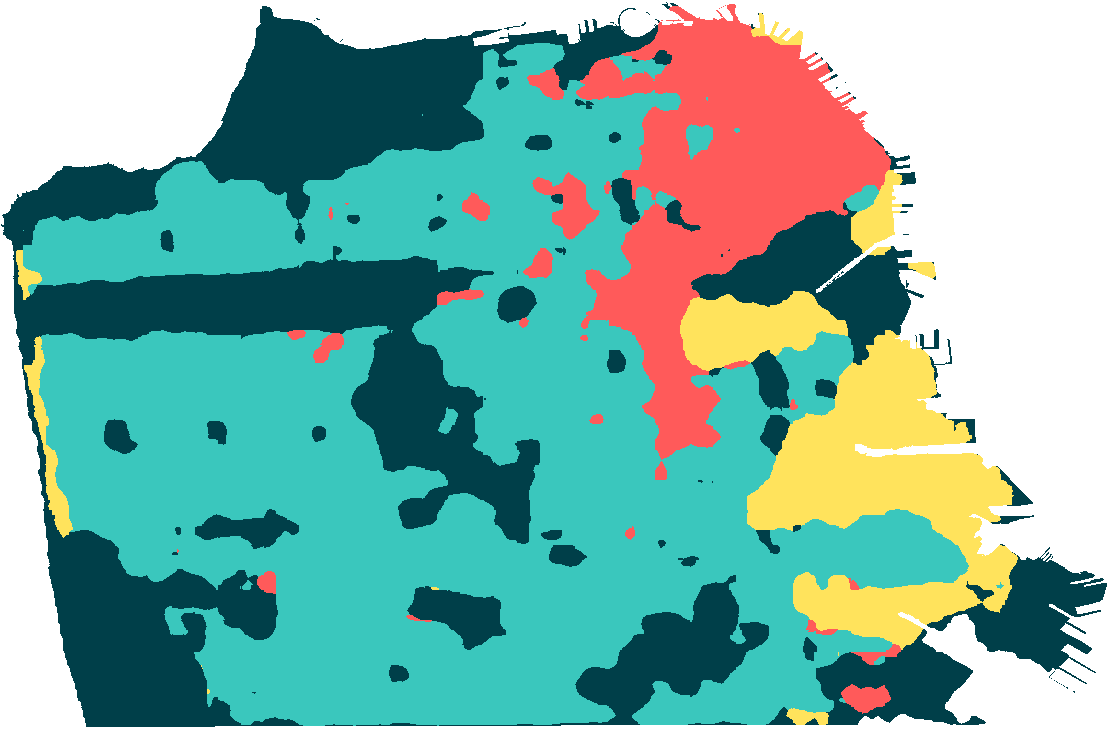} &
\includegraphics[width=.19\textwidth]{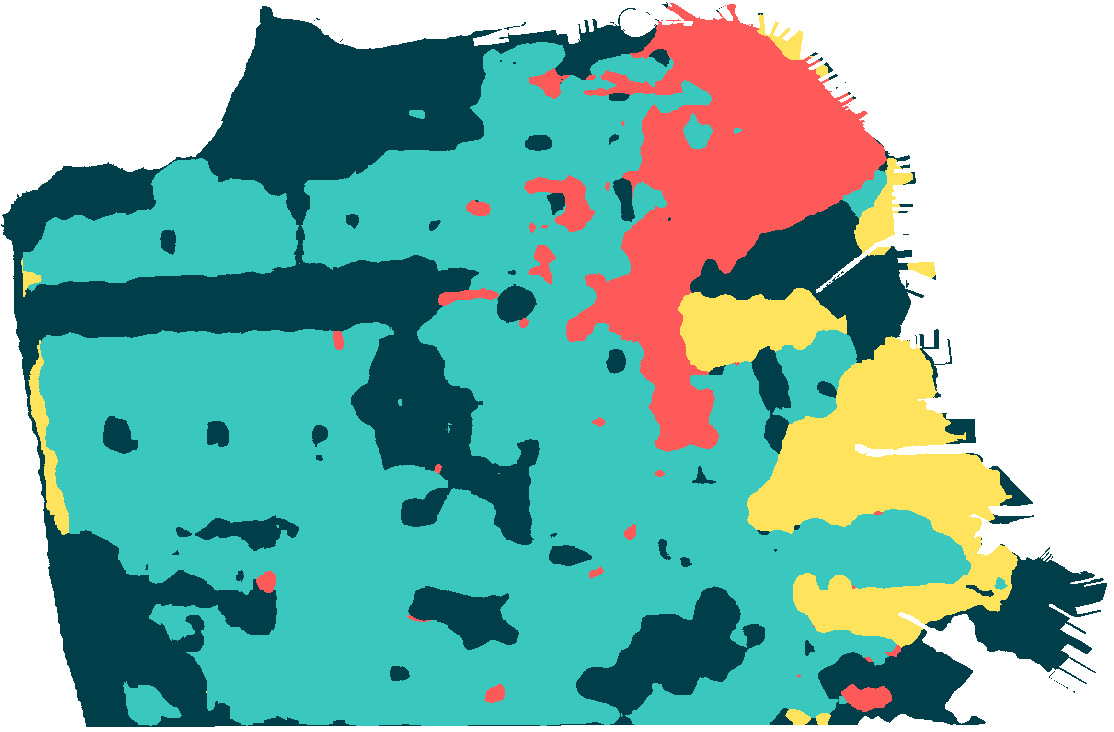} &
\includegraphics[width=.19\textwidth]{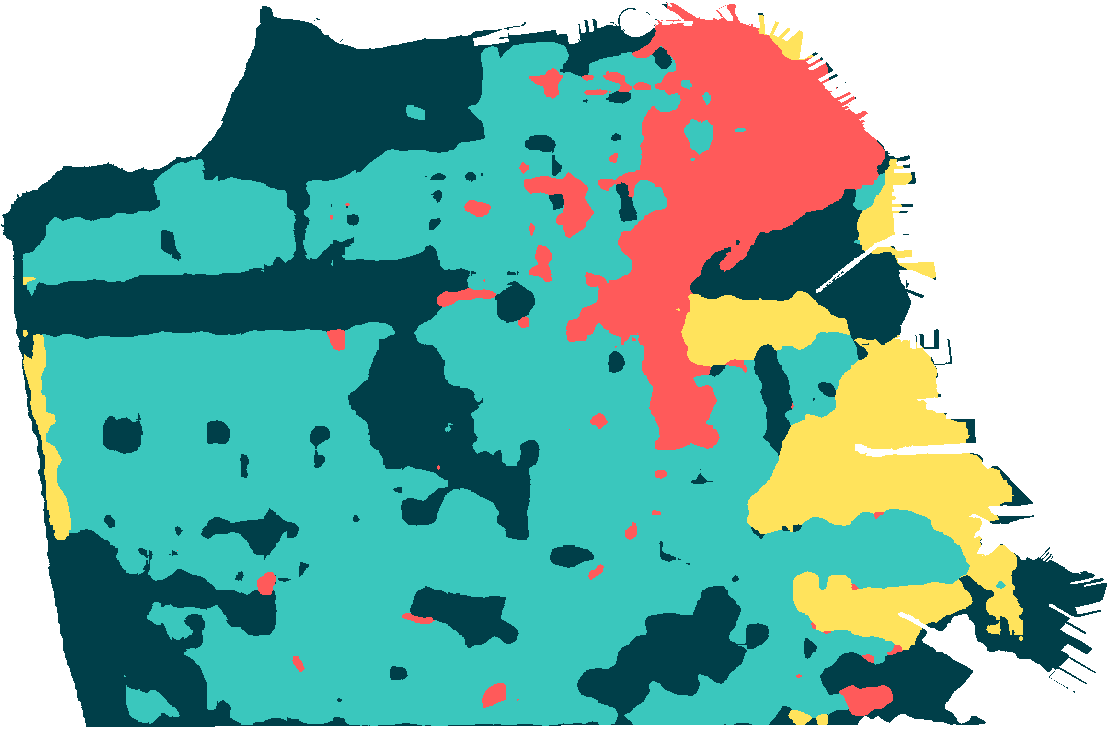} 
\\\midrule
\rotatebox{90}{NYC Photos} & 
 &
\includegraphics[width=.19\textwidth]{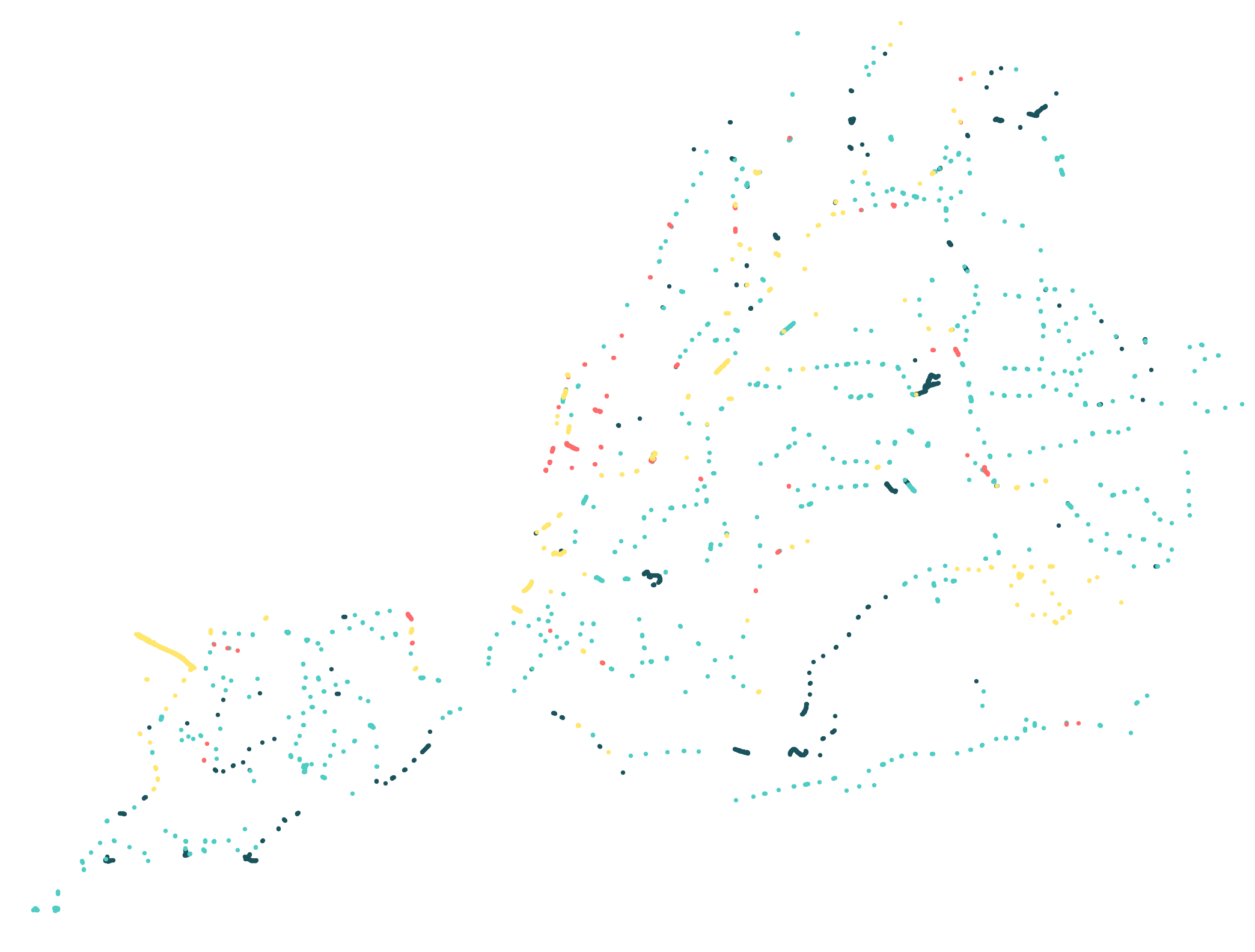} &
\includegraphics[width=.19\textwidth]{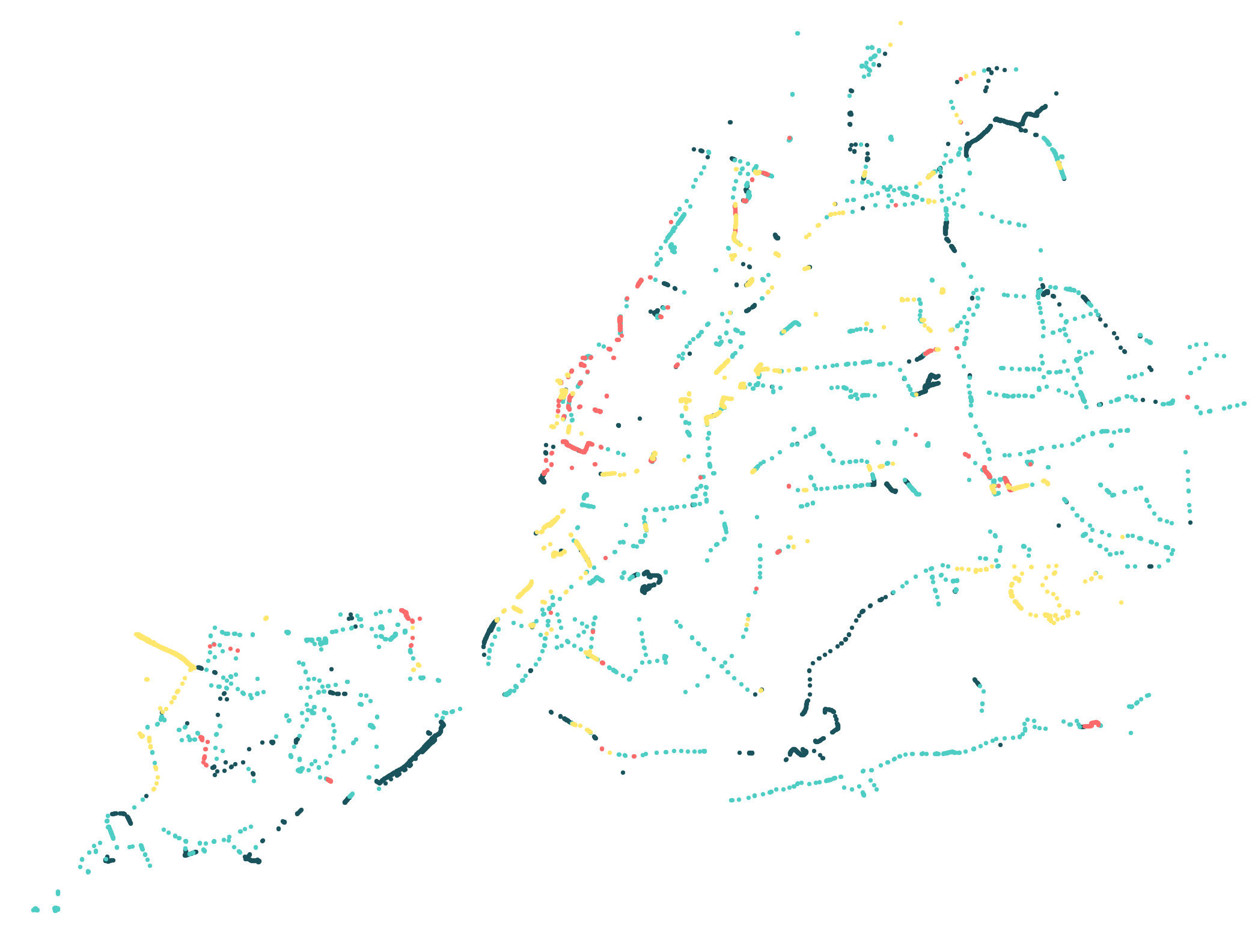} &
\includegraphics[width=.19\textwidth]{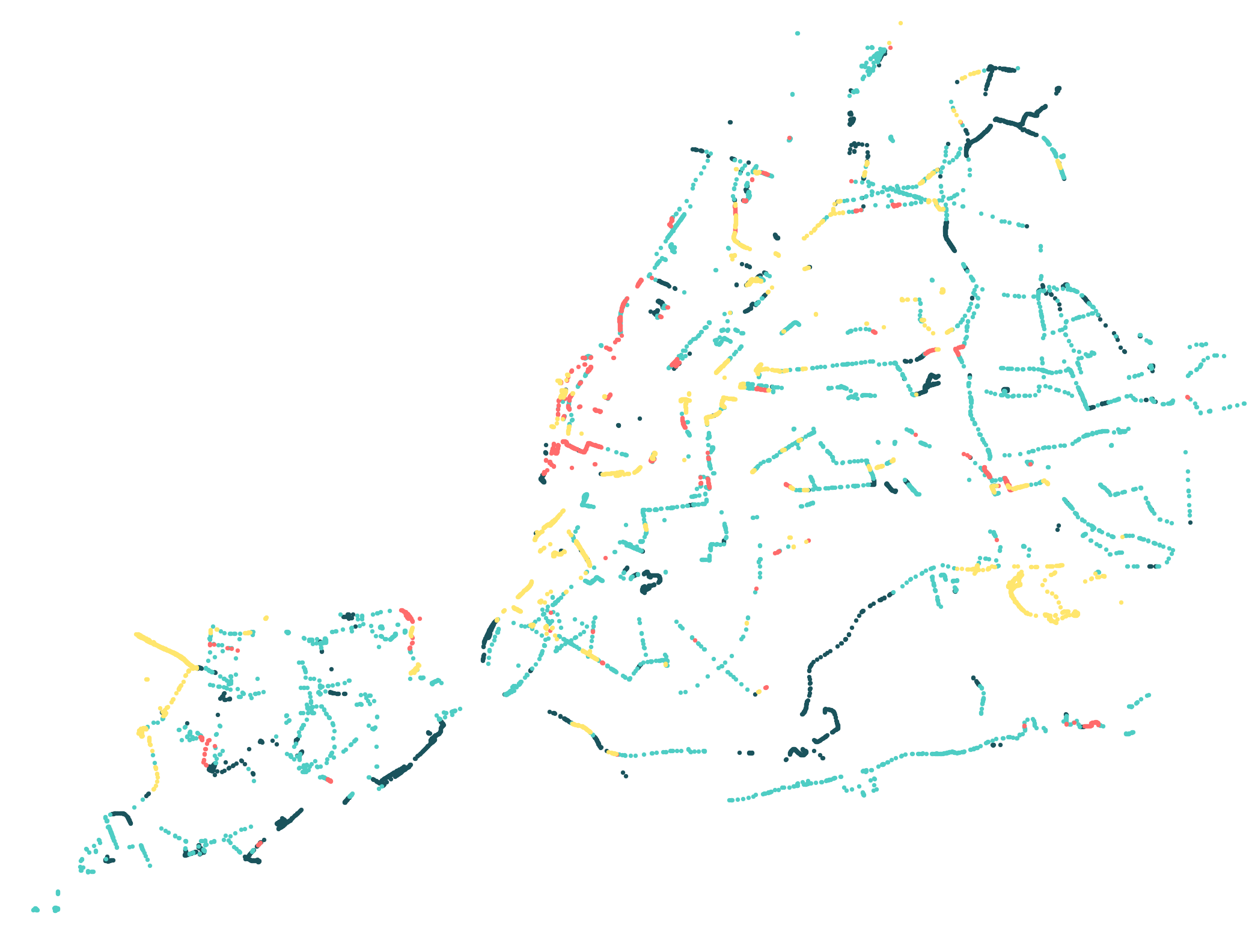} &
\includegraphics[width=.19\textwidth]{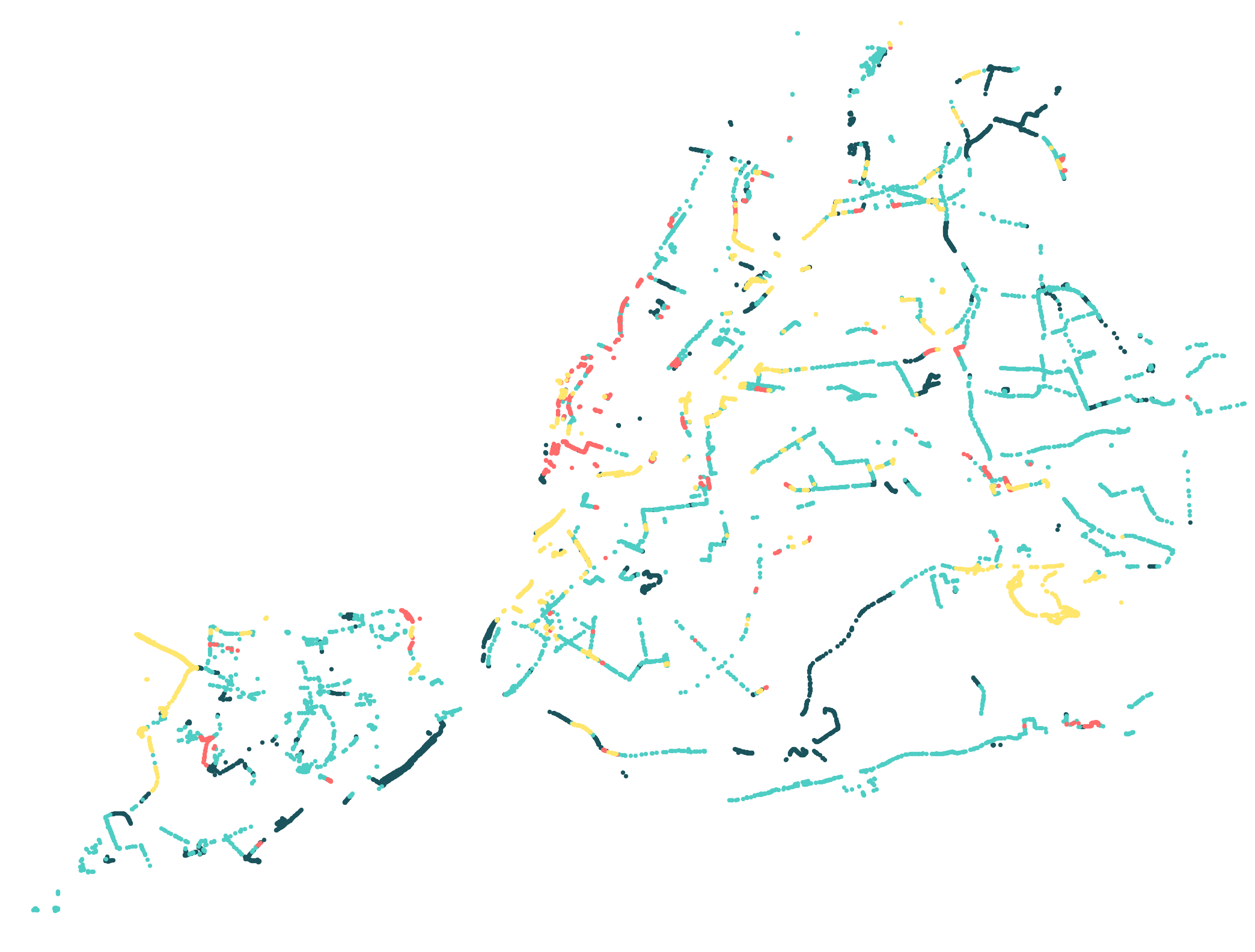} 
\\\midrule
\rotatebox{90}{NYC Results} & 
\includegraphics[width=.19\textwidth]{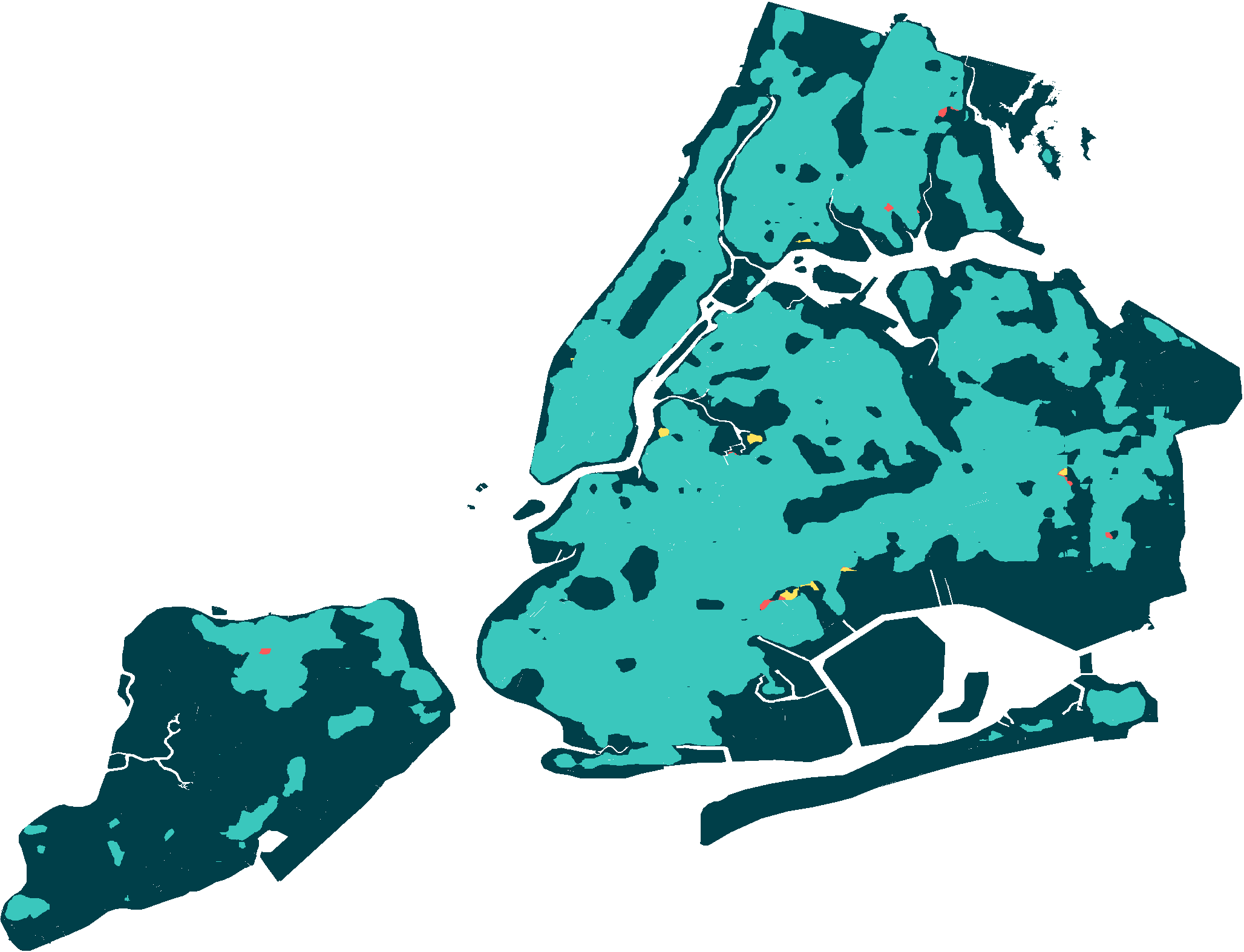} &
\includegraphics[width=.19\textwidth]{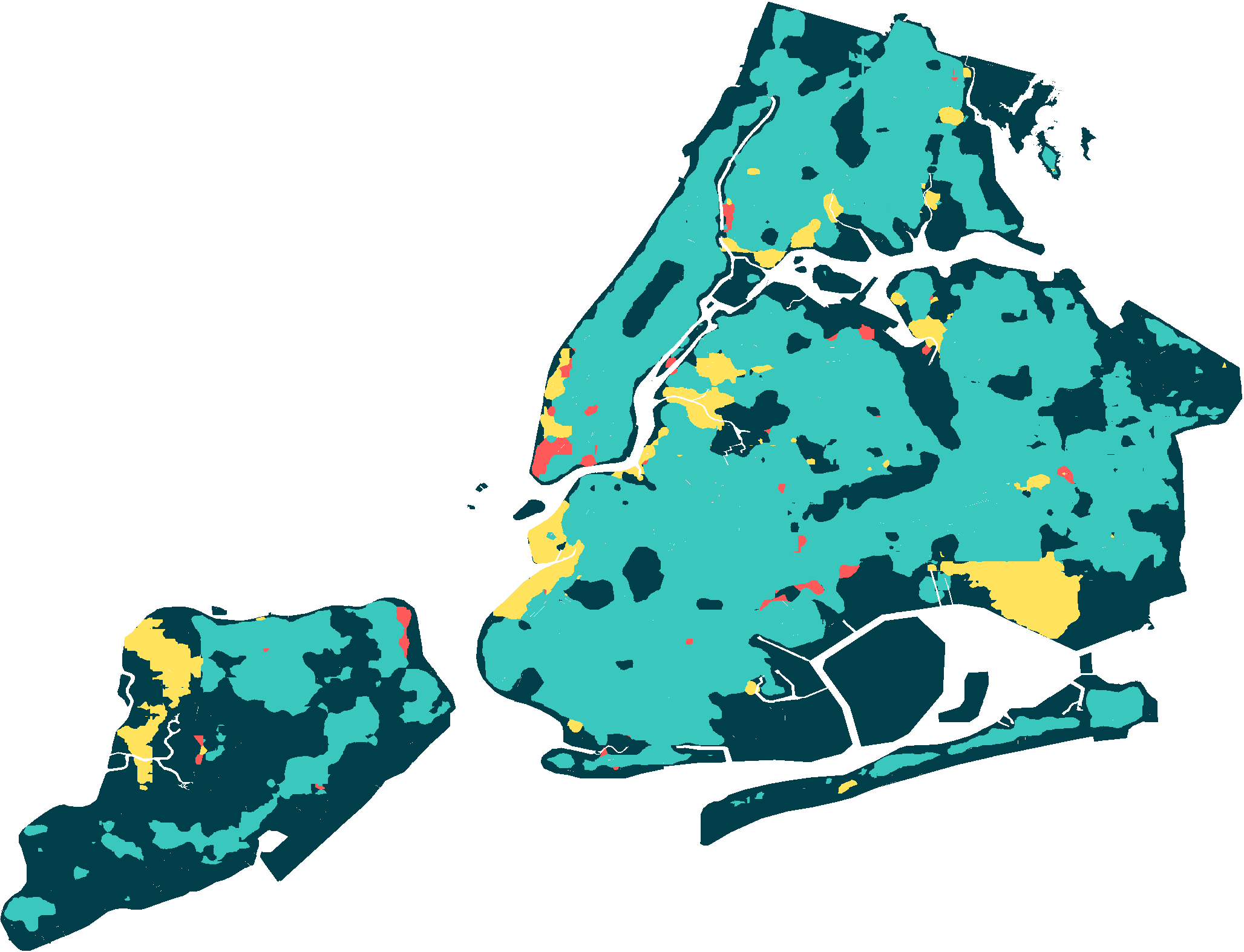} &
\includegraphics[width=.19\textwidth]{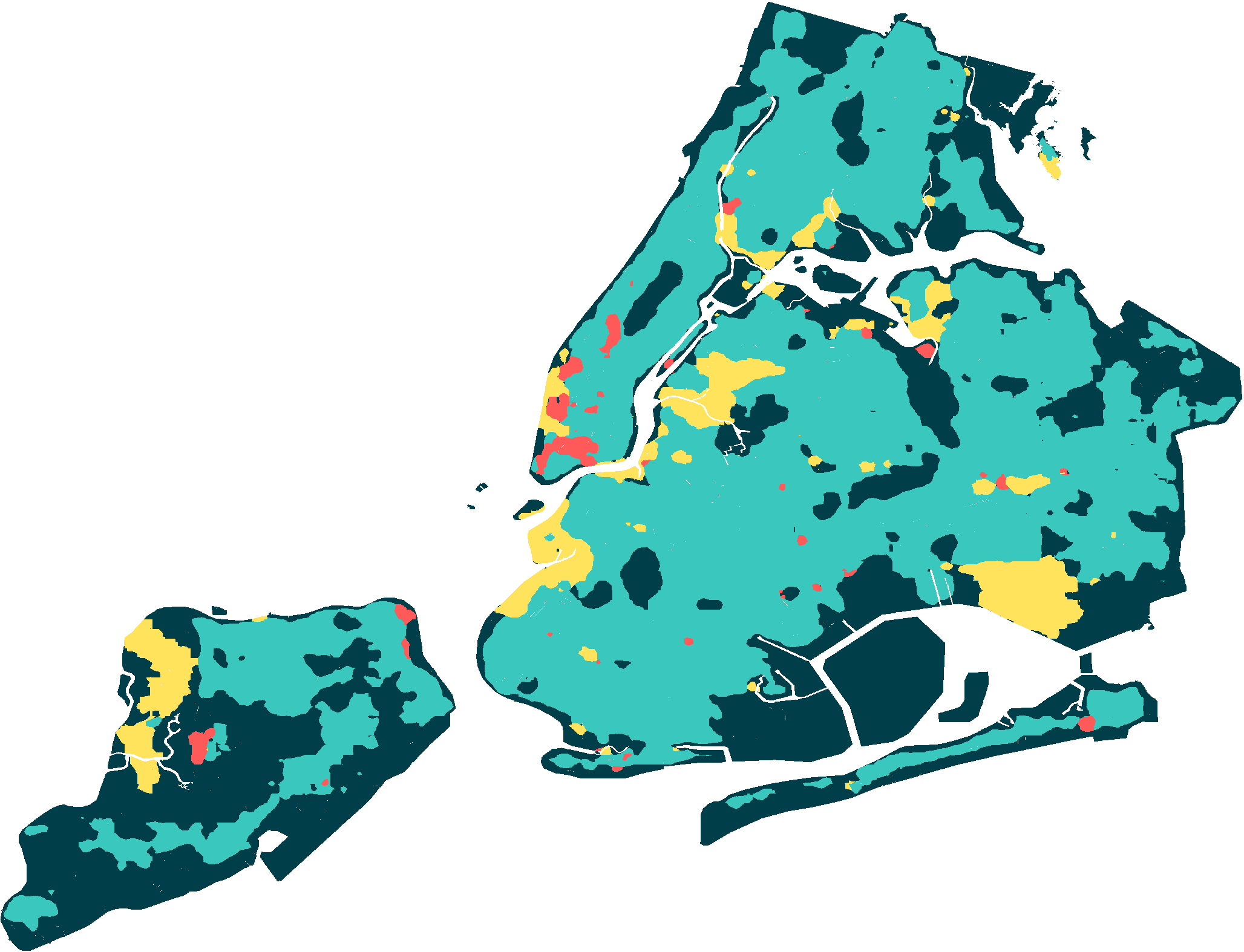} &
\includegraphics[width=.19\textwidth]{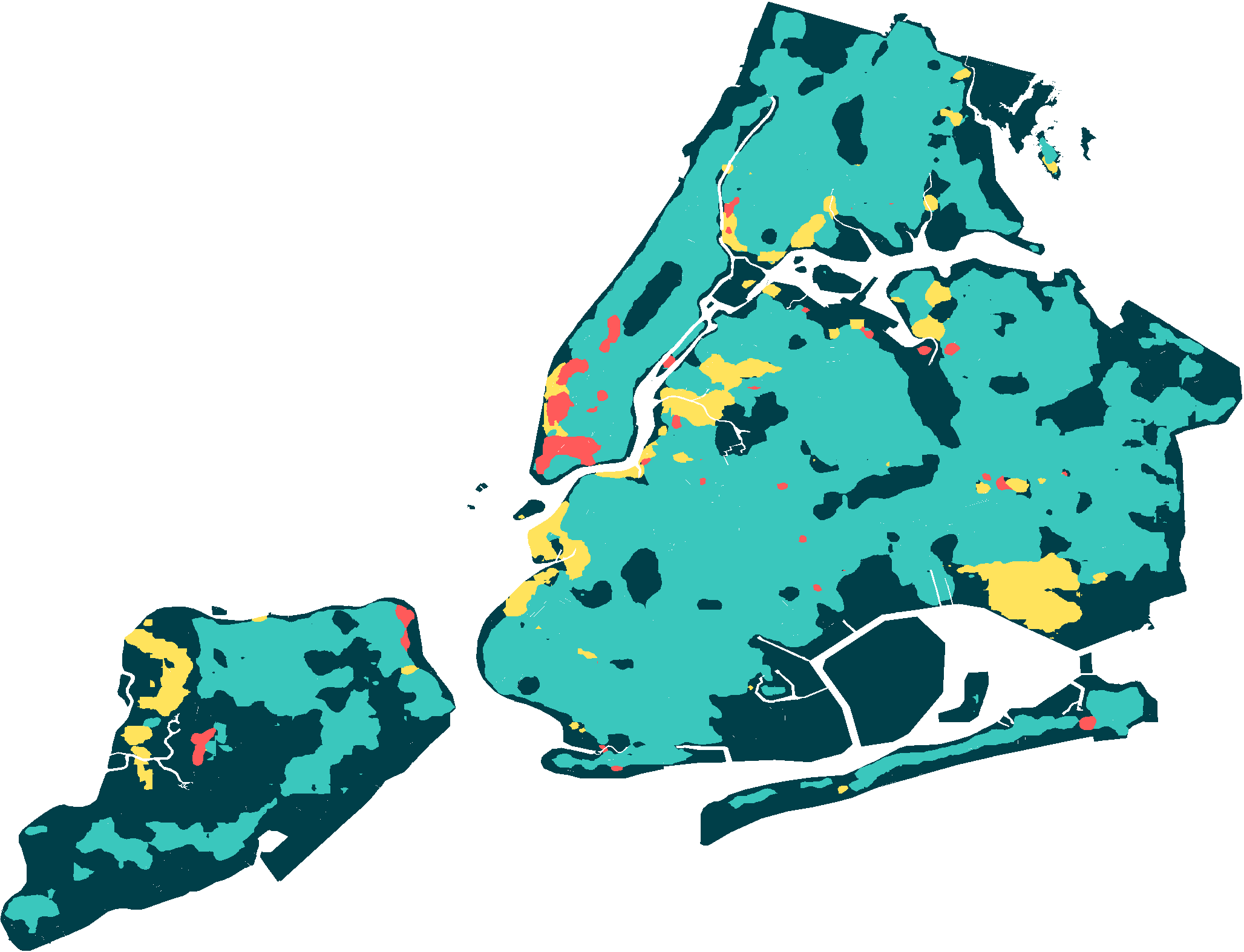} &
\includegraphics[width=.19\textwidth]{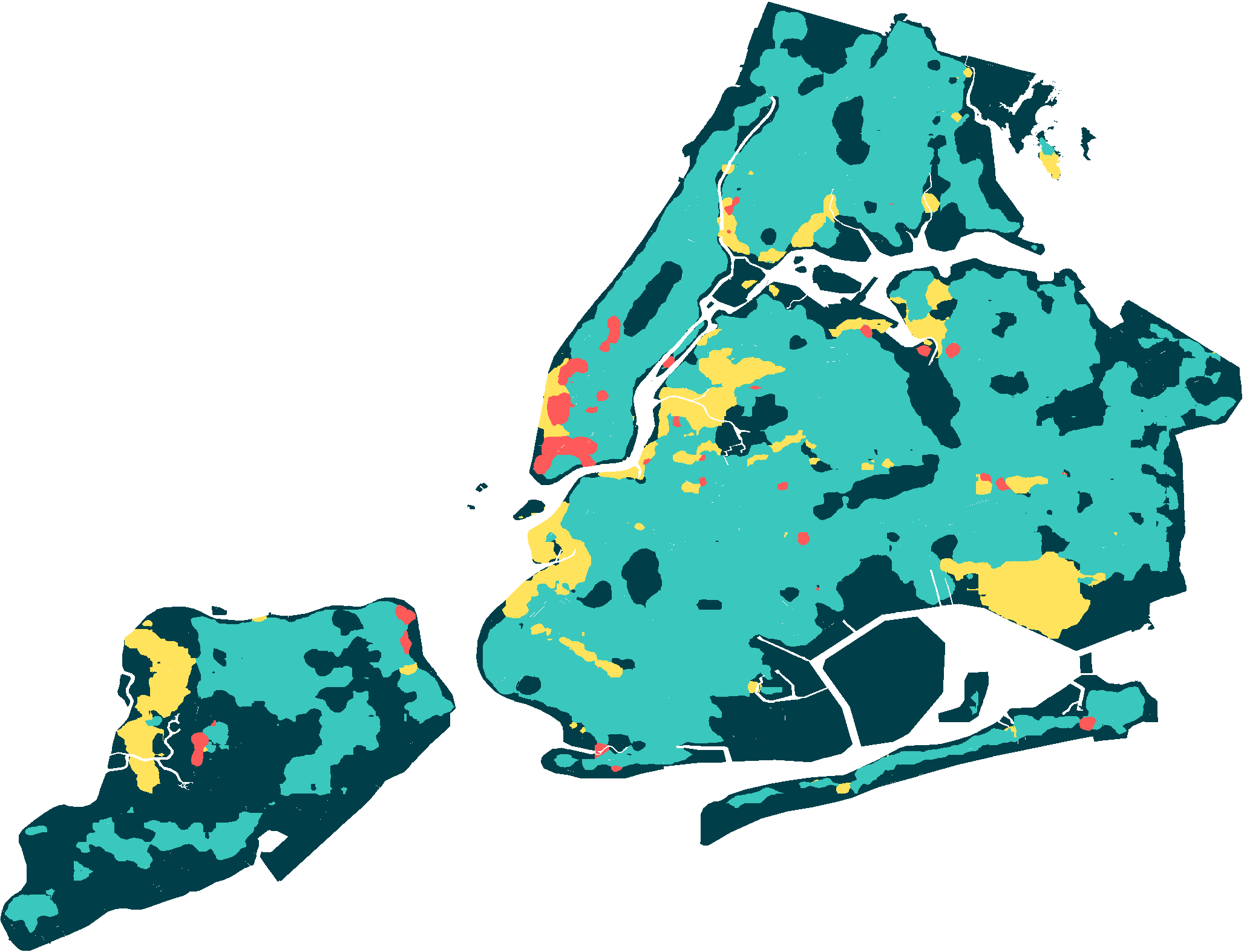} 
\\\midrule
\rotatebox{90}{BOS Photos} & 
 &
\includegraphics[width=.19\textwidth]{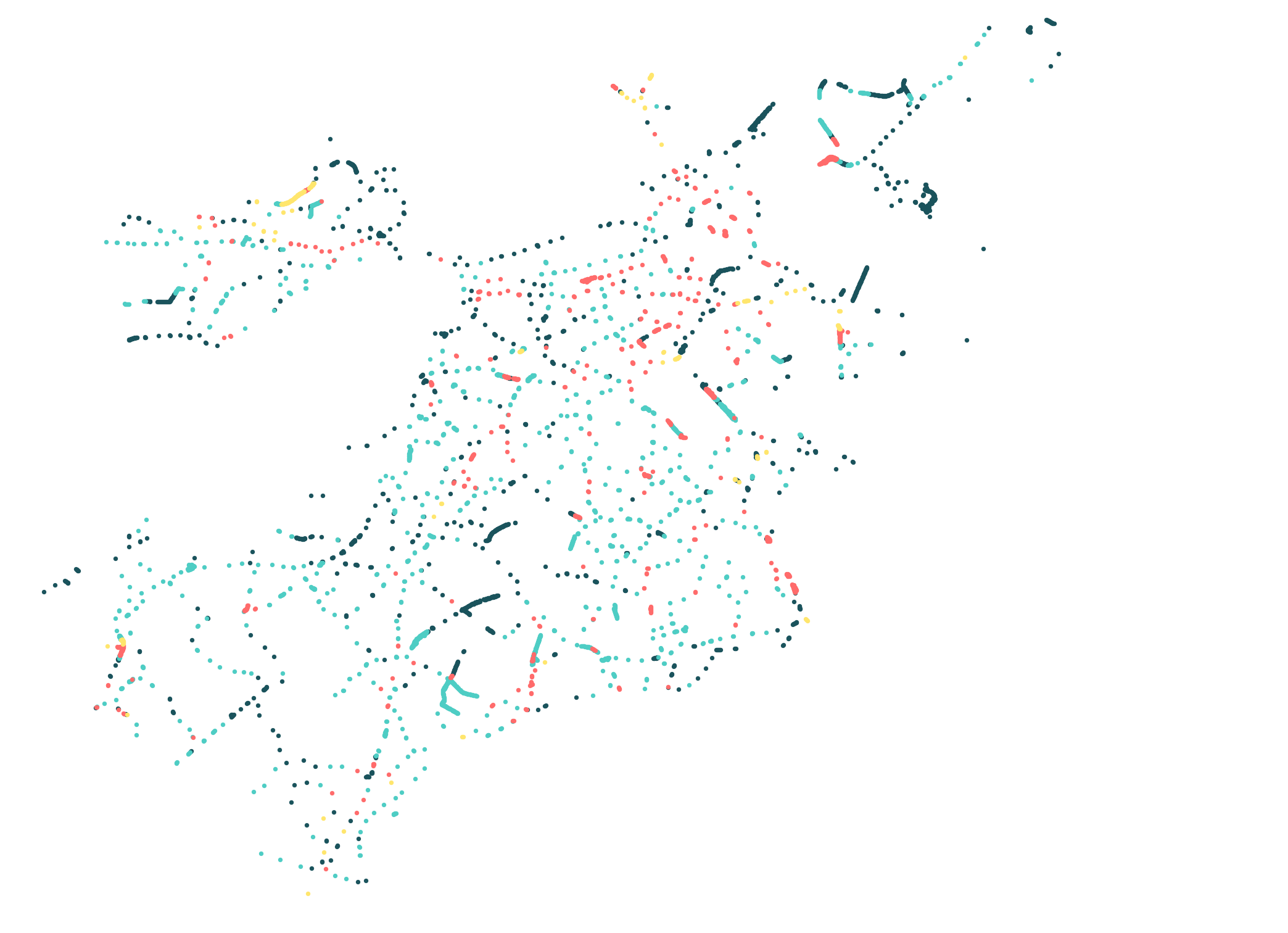} &
\includegraphics[width=.19\textwidth]{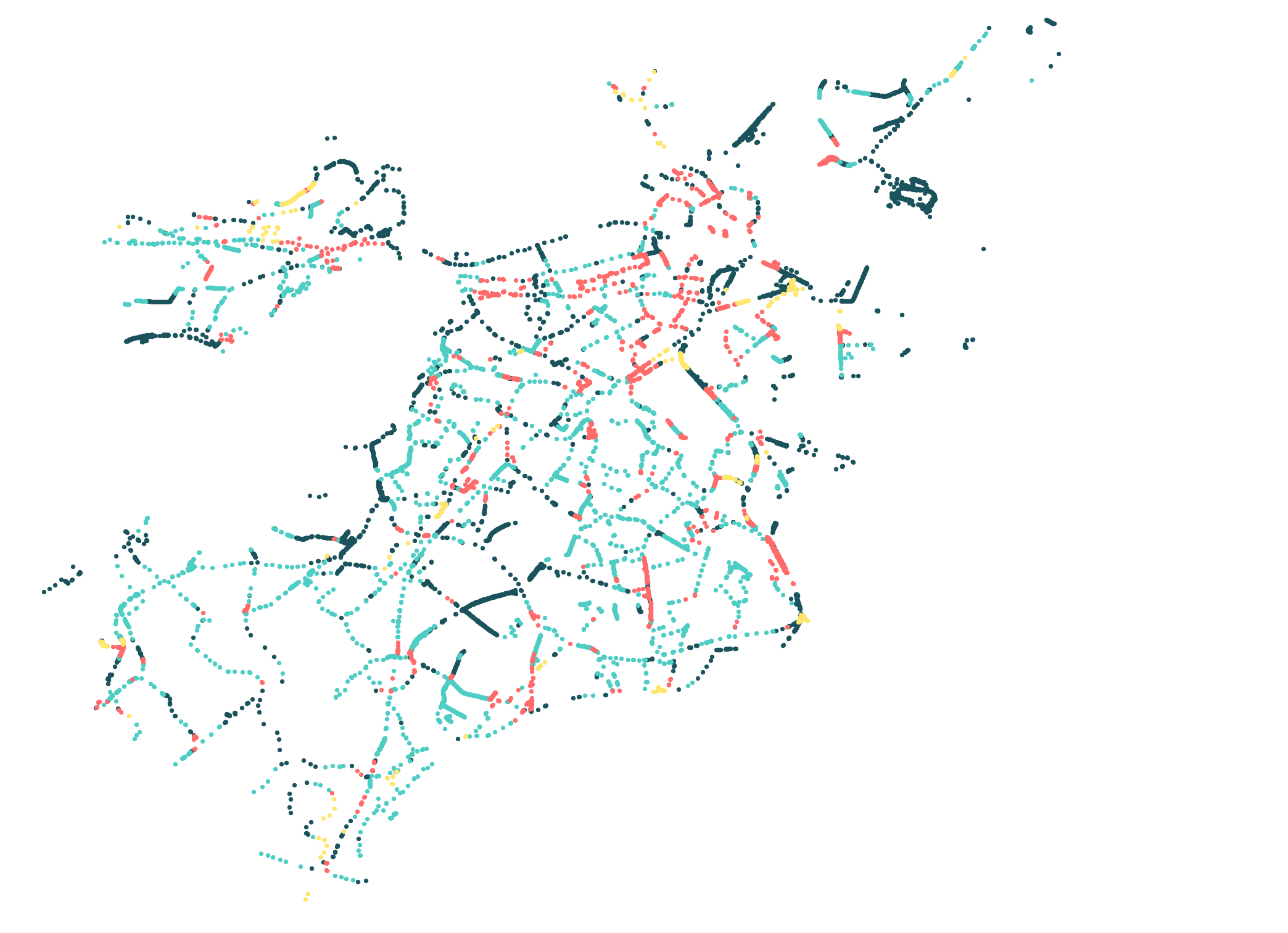} &
\includegraphics[width=.19\textwidth]{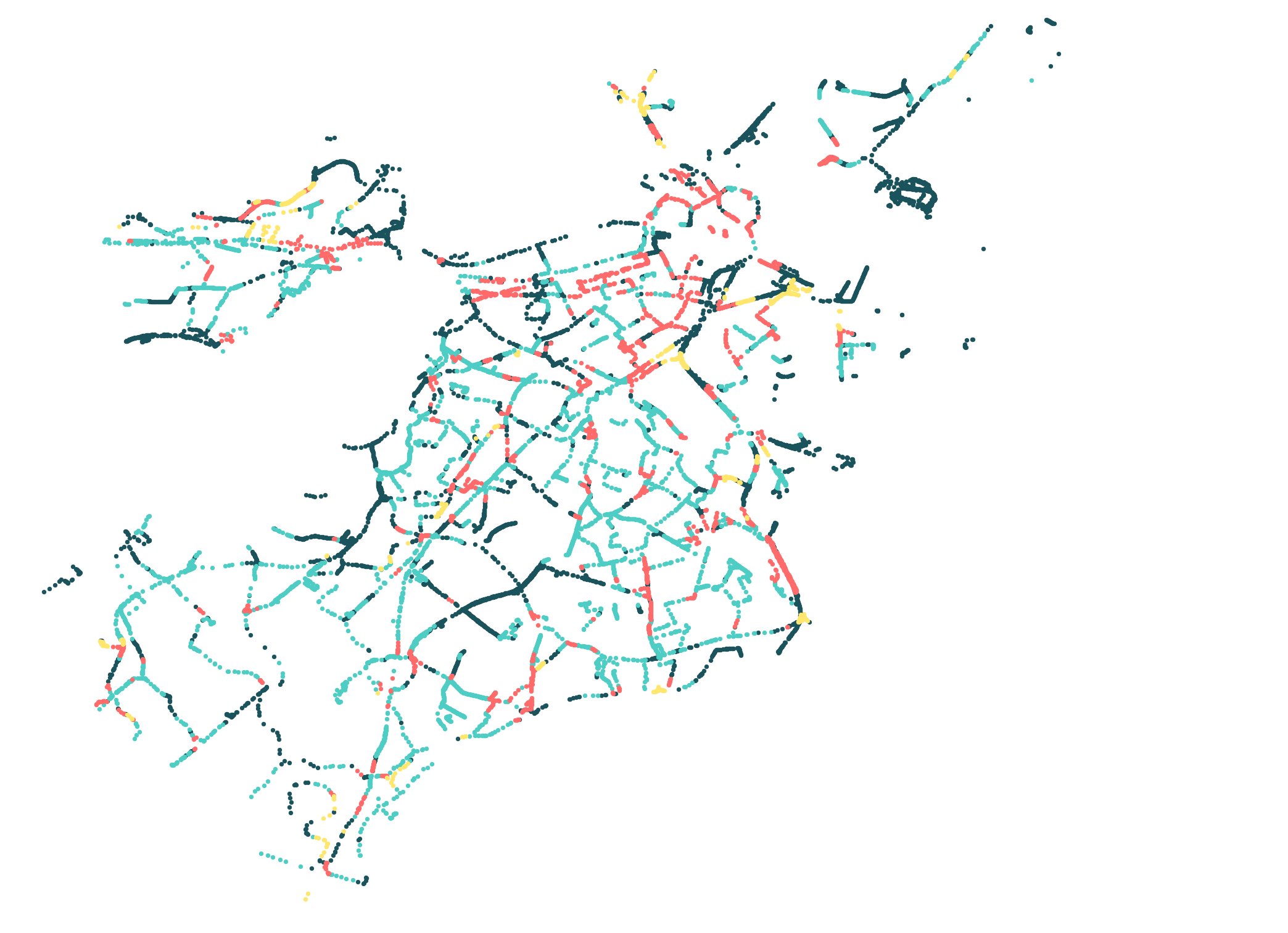} &
\includegraphics[width=.19\textwidth]{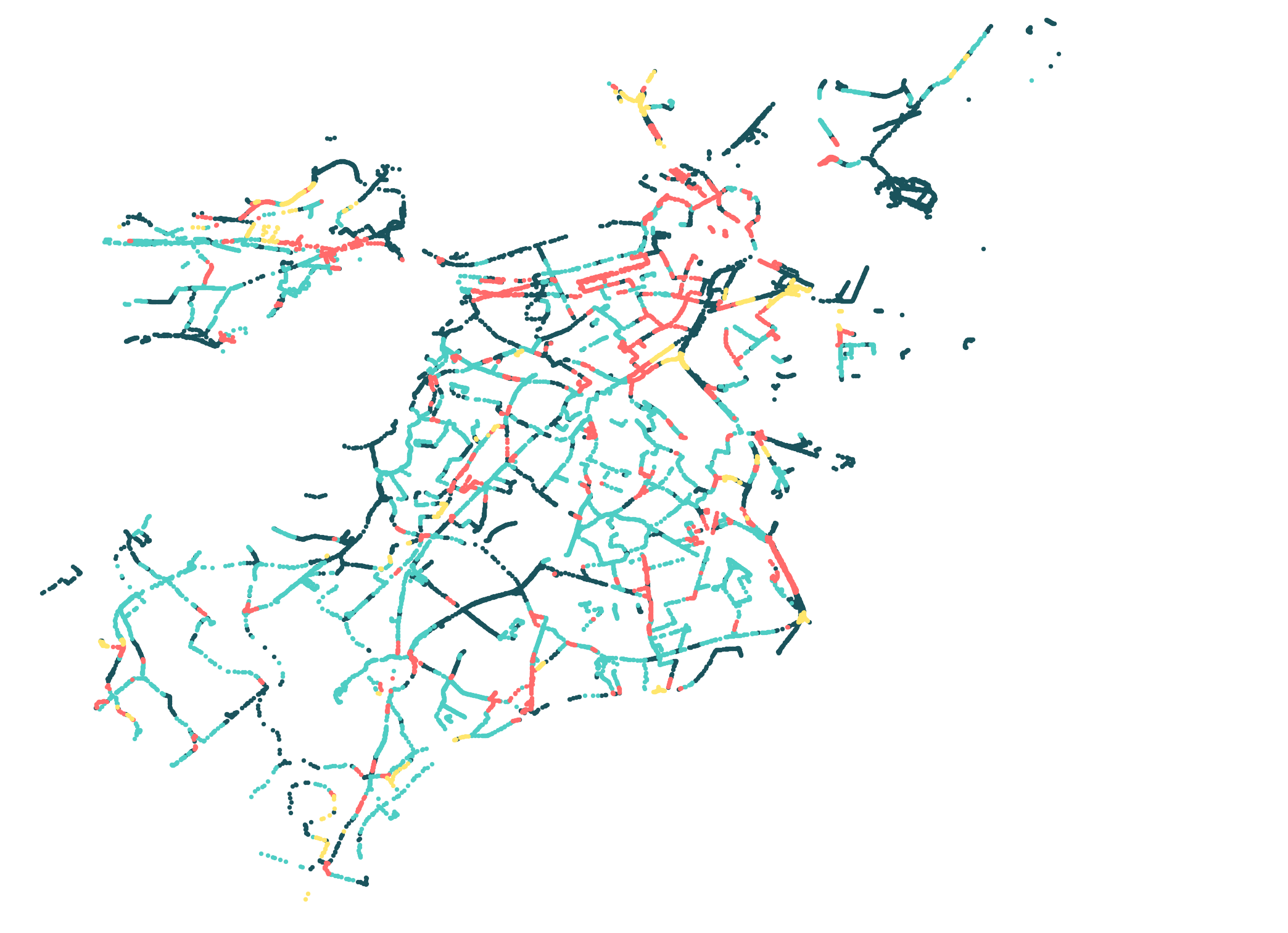} 
\\\midrule
\rotatebox{90}{BOS Results} & 
\includegraphics[width=.19\textwidth]{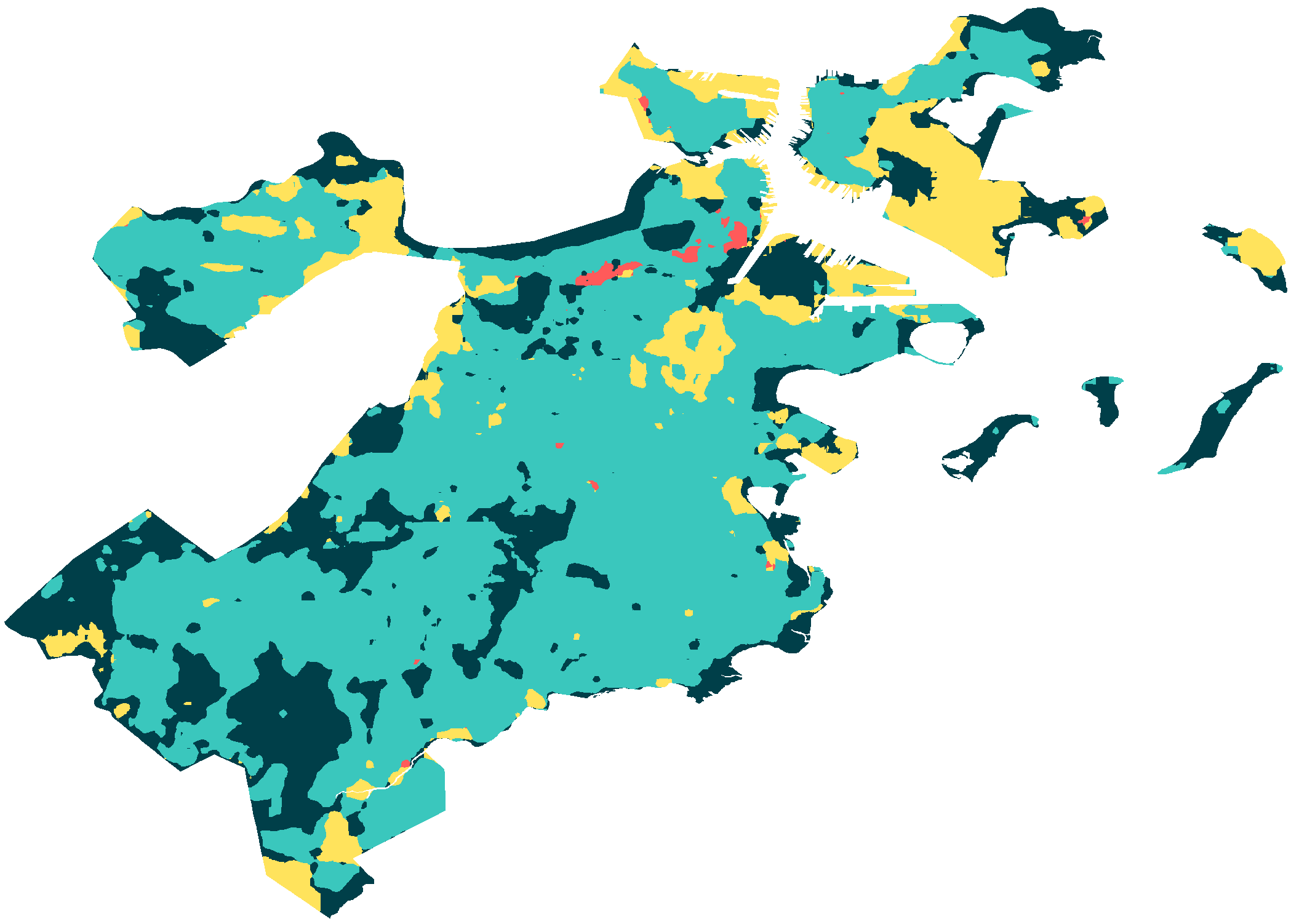} &
\includegraphics[width=.19\textwidth]{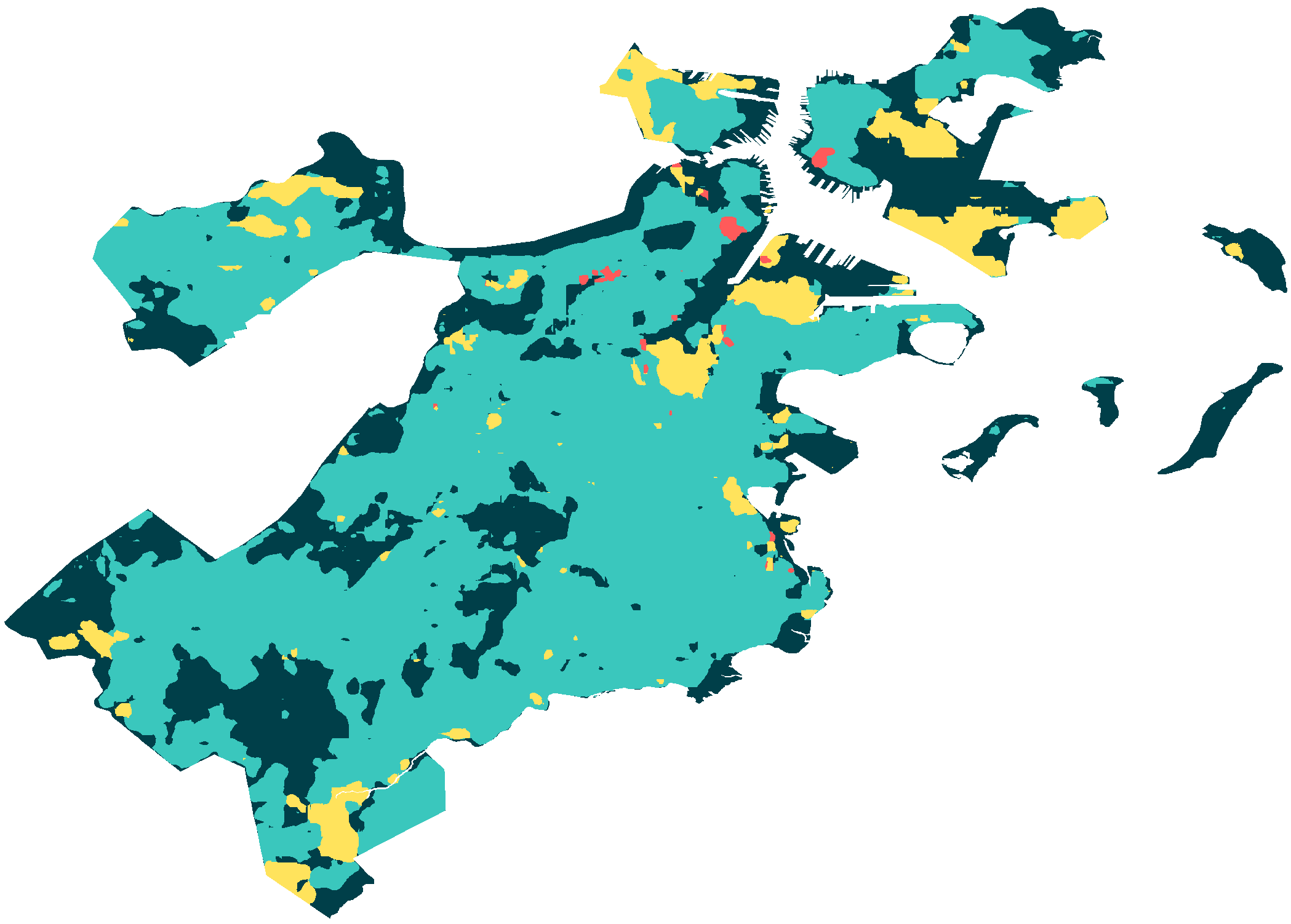} &
\includegraphics[width=.19\textwidth]{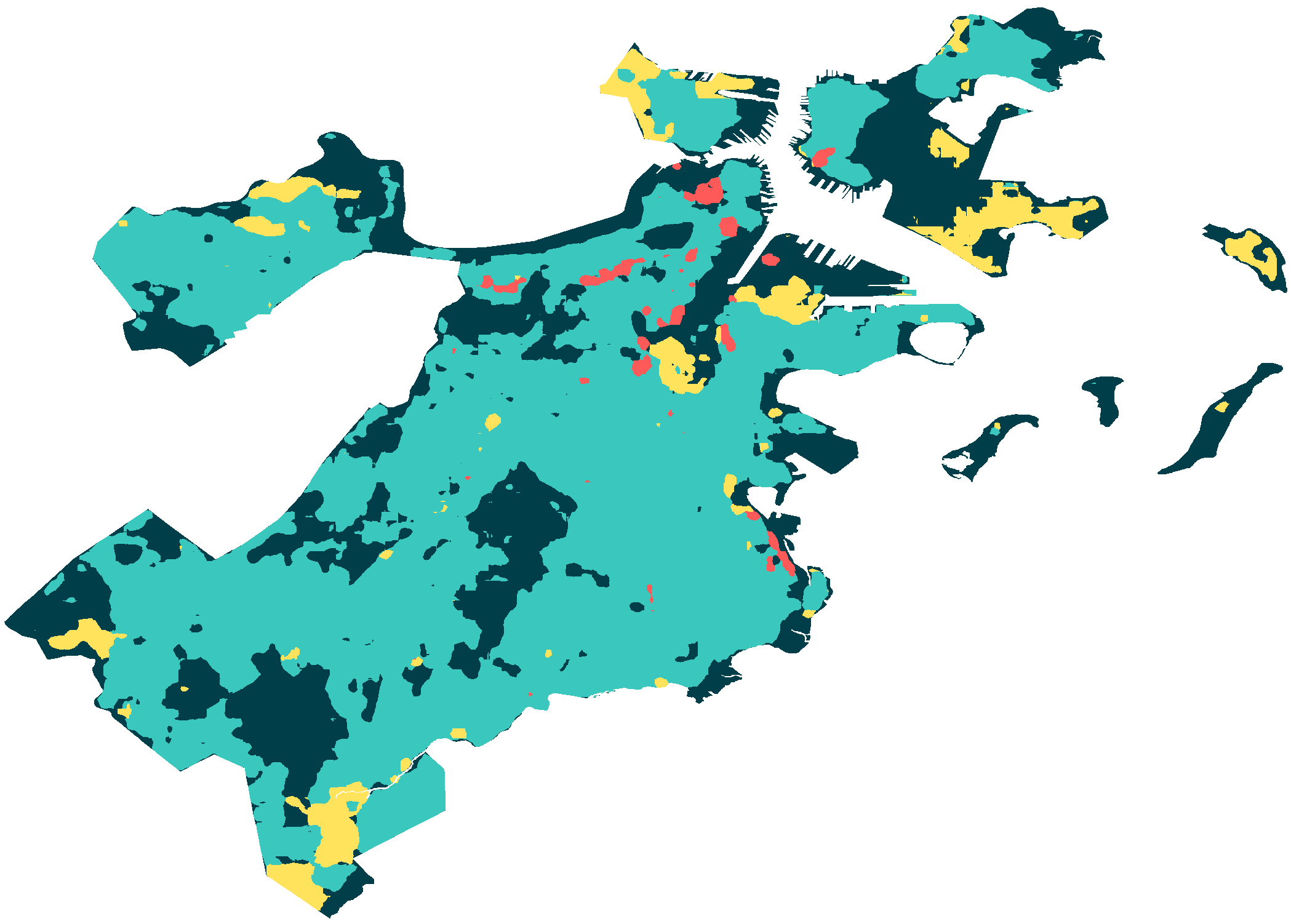} &
\includegraphics[width=.19\textwidth]{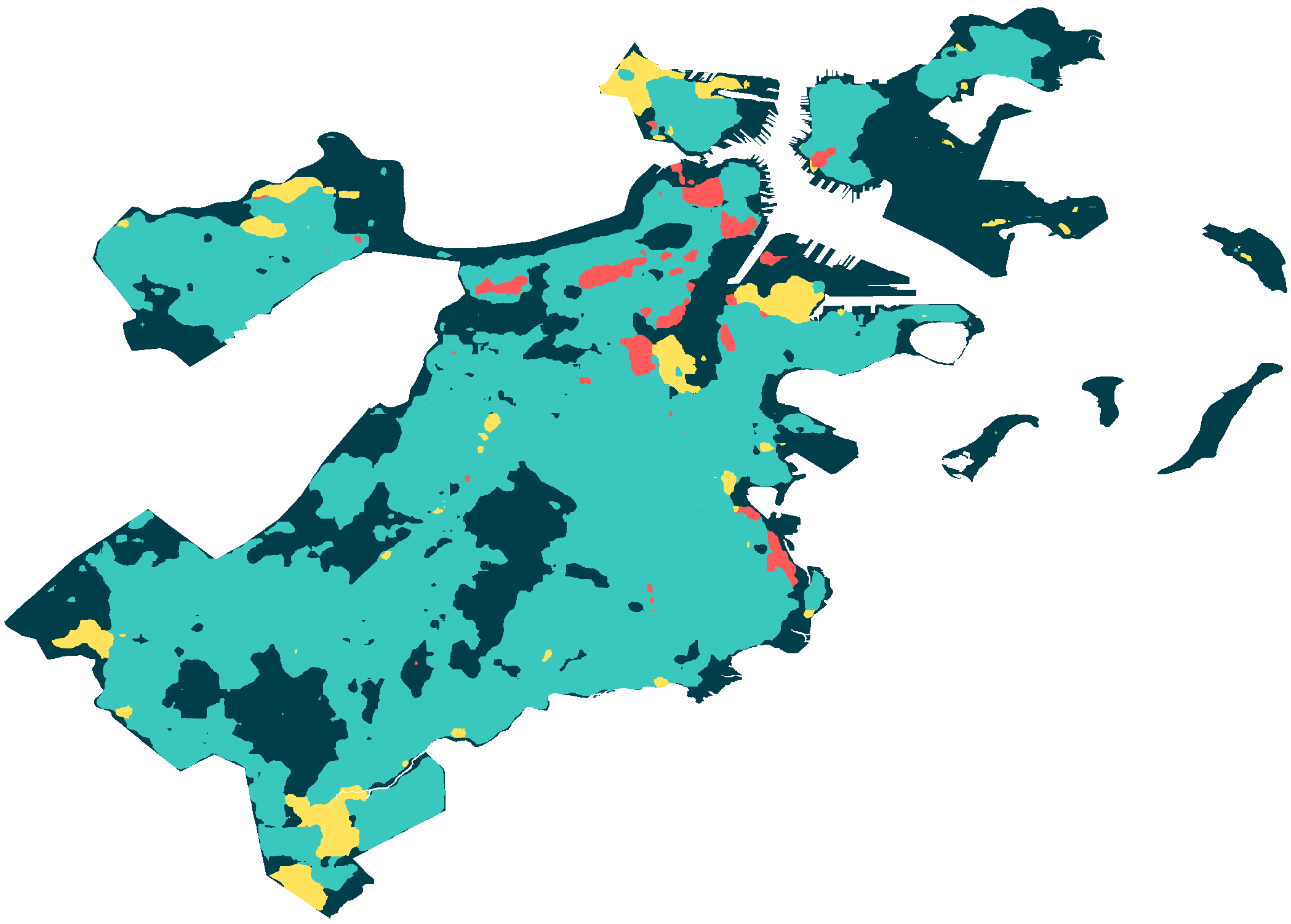} &
\includegraphics[width=.19\textwidth]{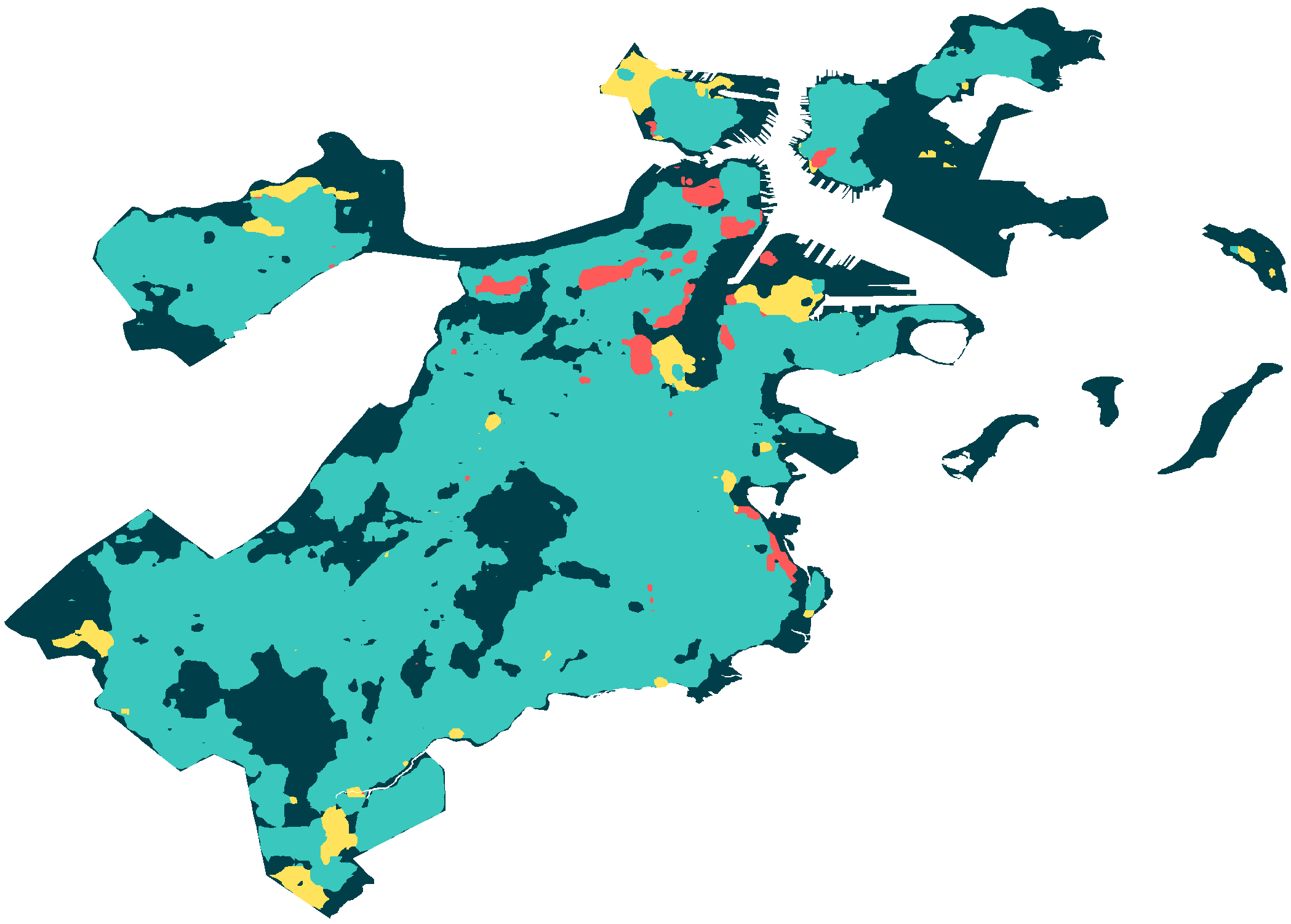} 
\\\bottomrule
\vspace{0.01pt}
\end{tabular}
\end{center} 
\addlegendimageintext{fill=others, area legend} Others
\hspace{3pt}
\addlegendimageintext{fill=residential, area legend} Residential
\hspace{3pt}
\addlegendimageintext{fill=commercial, area legend} Commercial
\hspace{3pt}
\addlegendimageintext{fill=industrial, area legend} Industrial
\caption{Results on zone segmentation \cite{feng2018urban} with varying brush strokes. Best viewed in color.}
\label{fig:urbanreduced}
\end{figure*}

\begin{figure*}[t]
\centering
% \scalebox{0.95}{
\begin{center}
\begin{tabular}{l@{\ }c@{\ }c@{\ }c@{\ }c@{\ }c@{\ }}
% \hline
 \toprule
\multicolumn{1}{c}{}
& \multicolumn{1}{c}{0\%}
& \multicolumn{1}{c}{20\%}
& \multicolumn{1}{c}{40\%}
& \multicolumn{1}{c}{60\%}
& \multicolumn{1}{c}{80\%}
\\\midrule
\rotatebox{90}{A05 Brush} & 
\includegraphics[width=.19\textwidth]{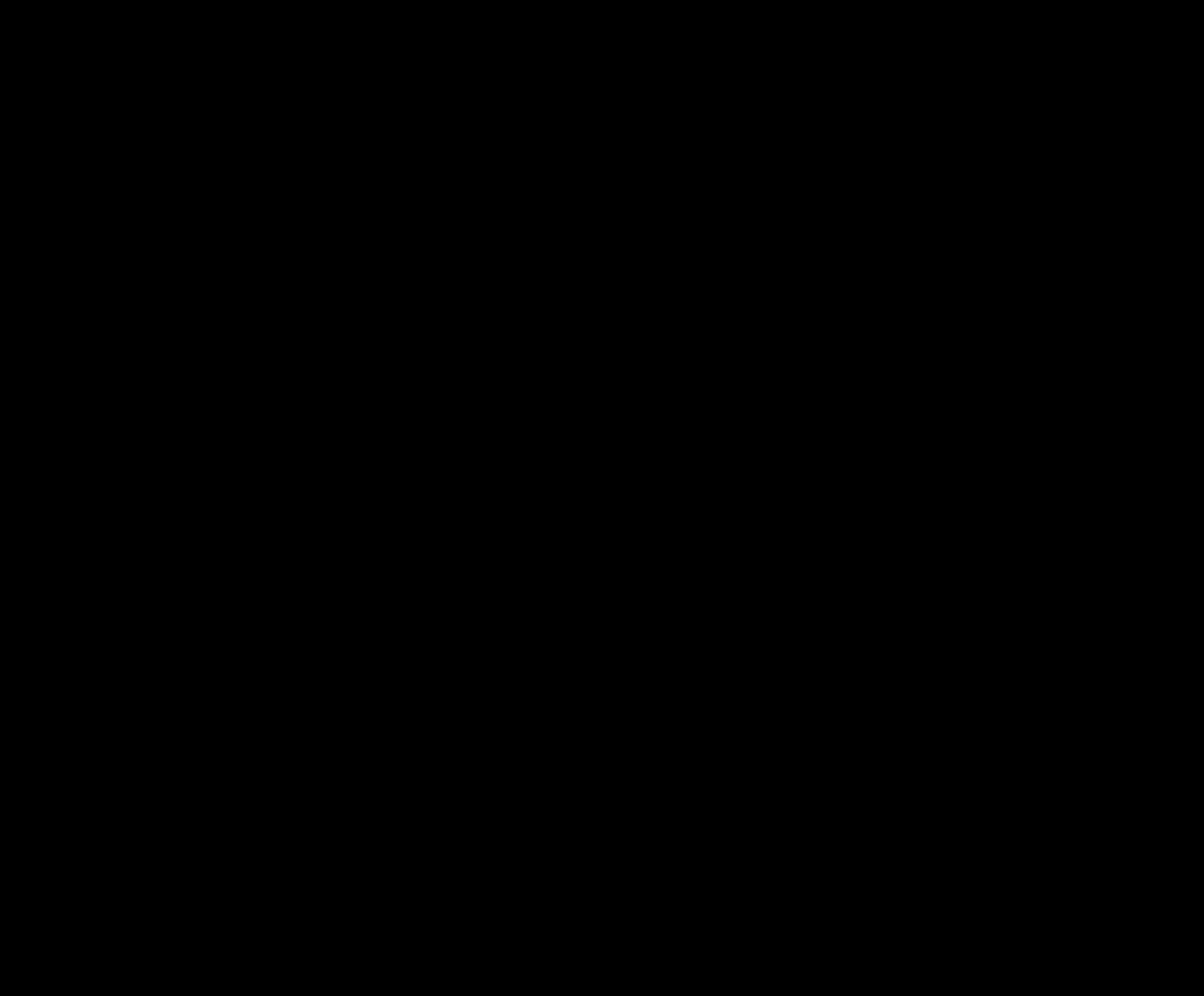} &
\includegraphics[width=.19\textwidth]{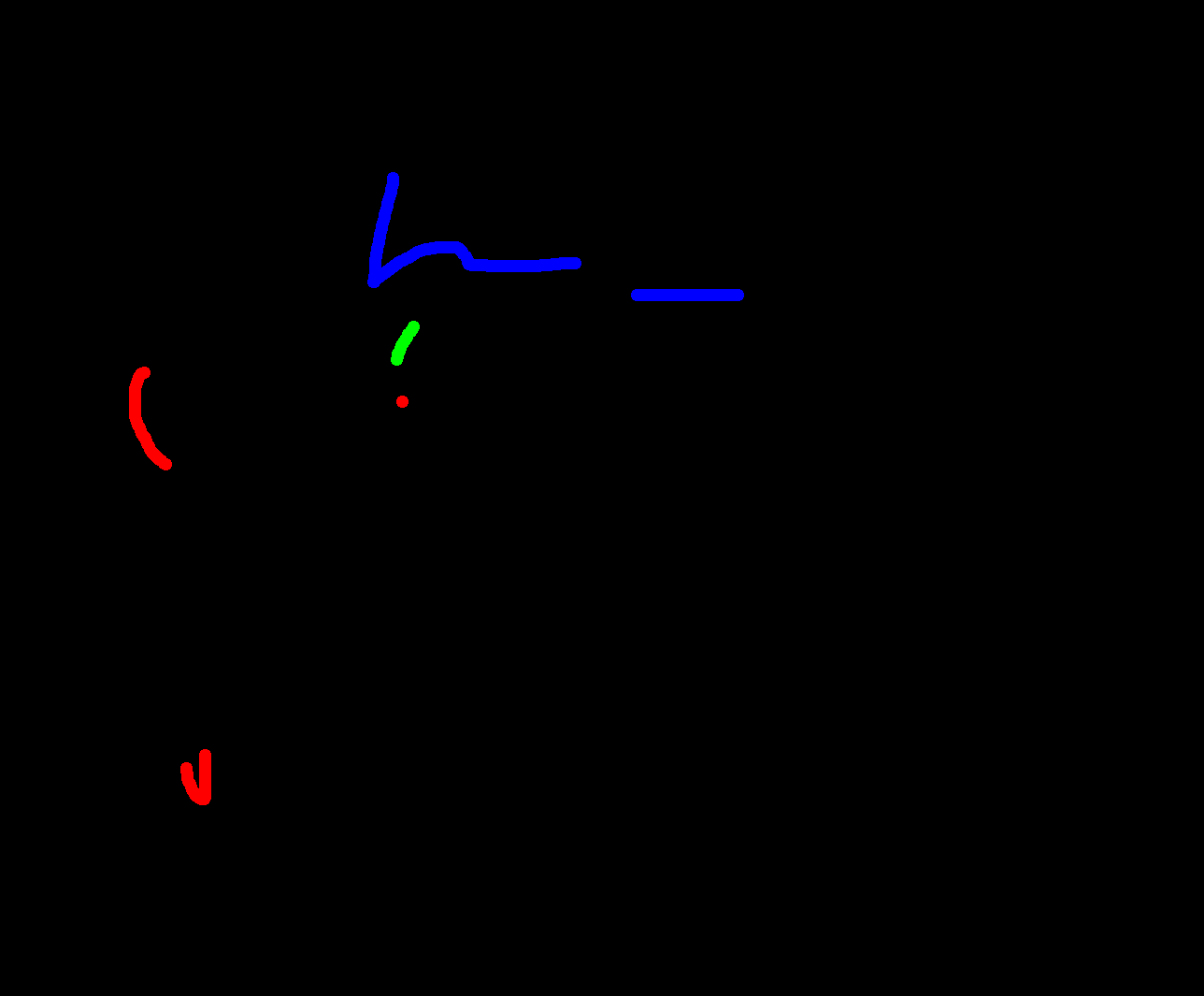} &
\includegraphics[width=.19\textwidth]{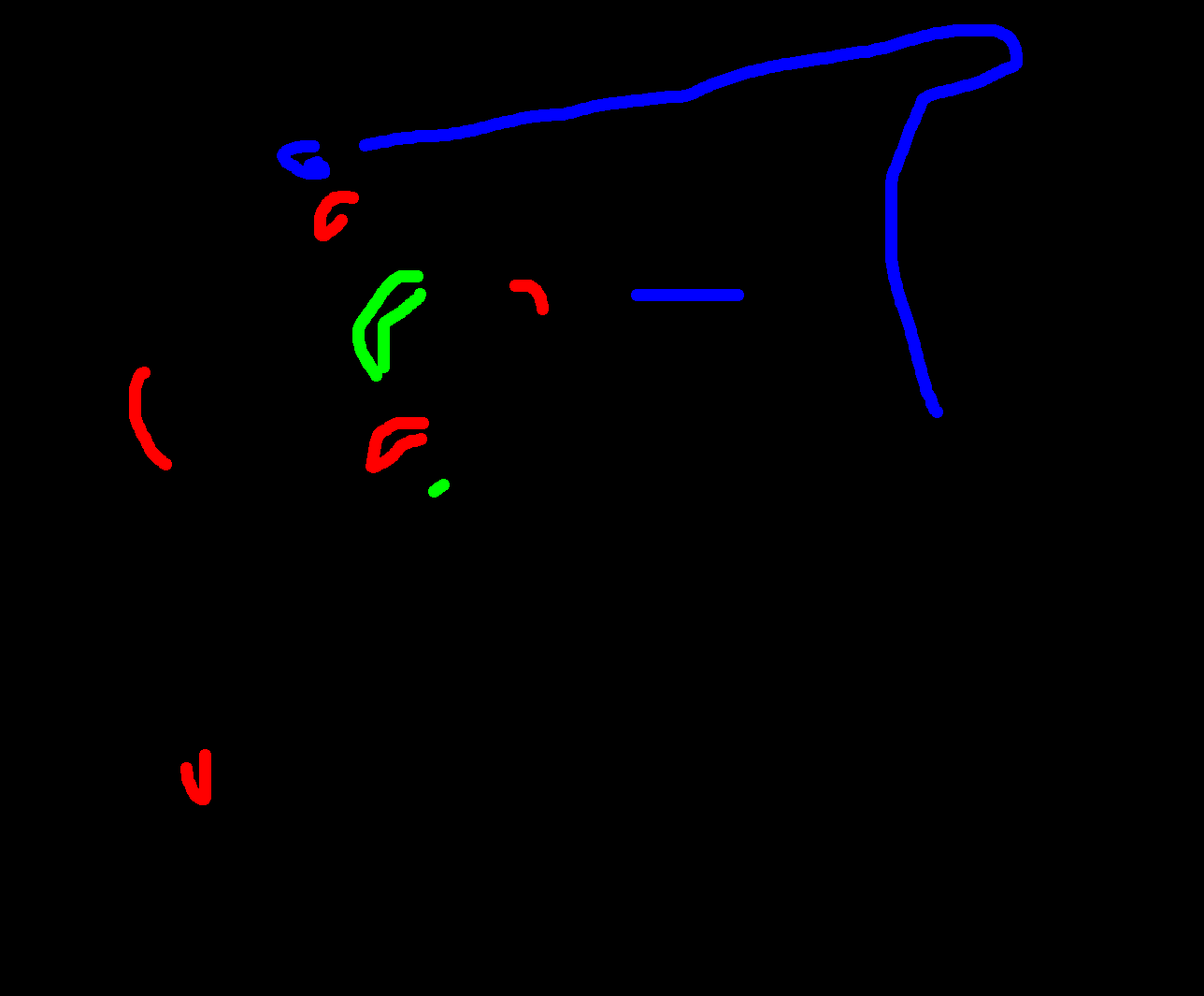} &
\includegraphics[width=.19\textwidth]{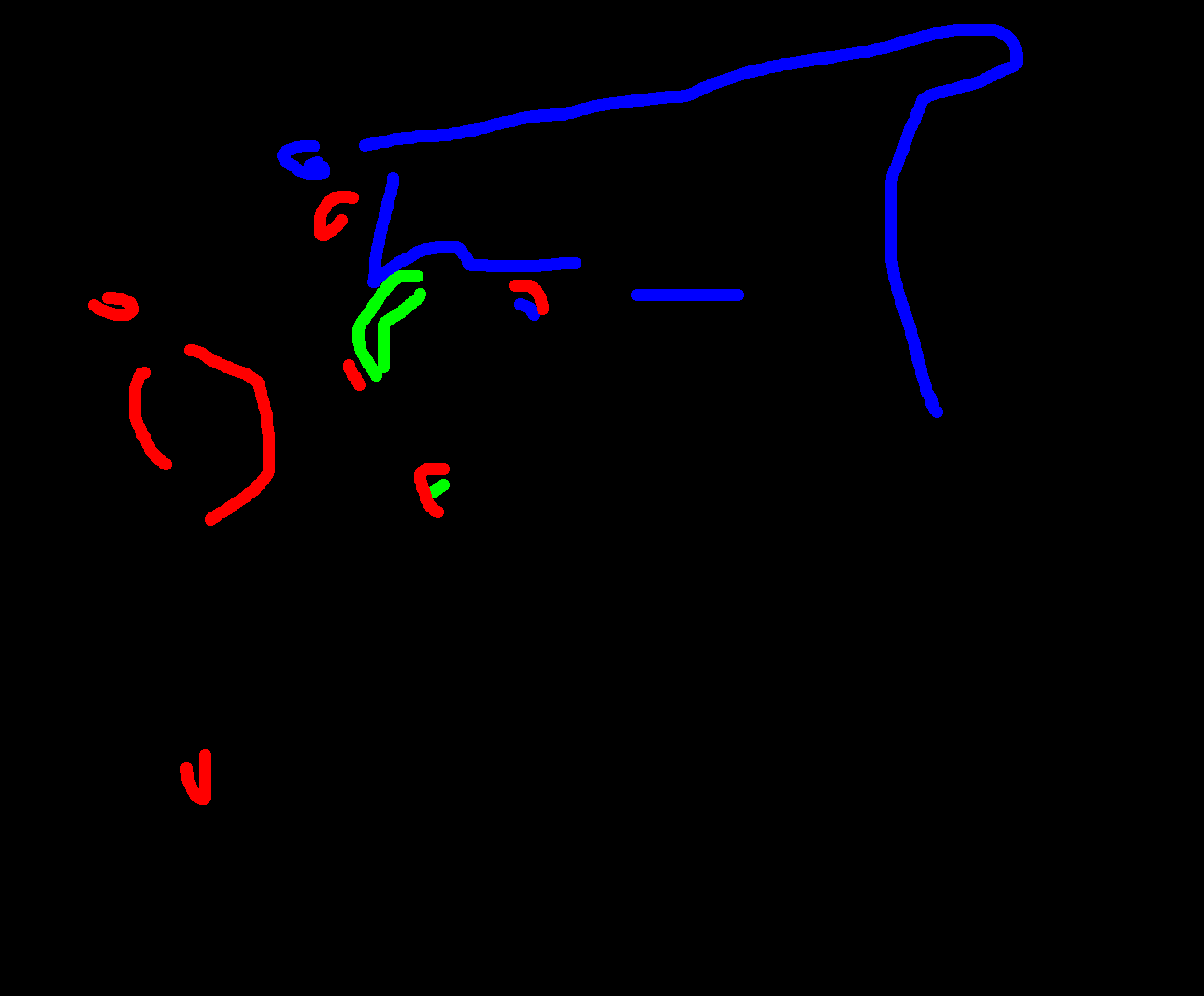} &
\includegraphics[width=.19\textwidth]{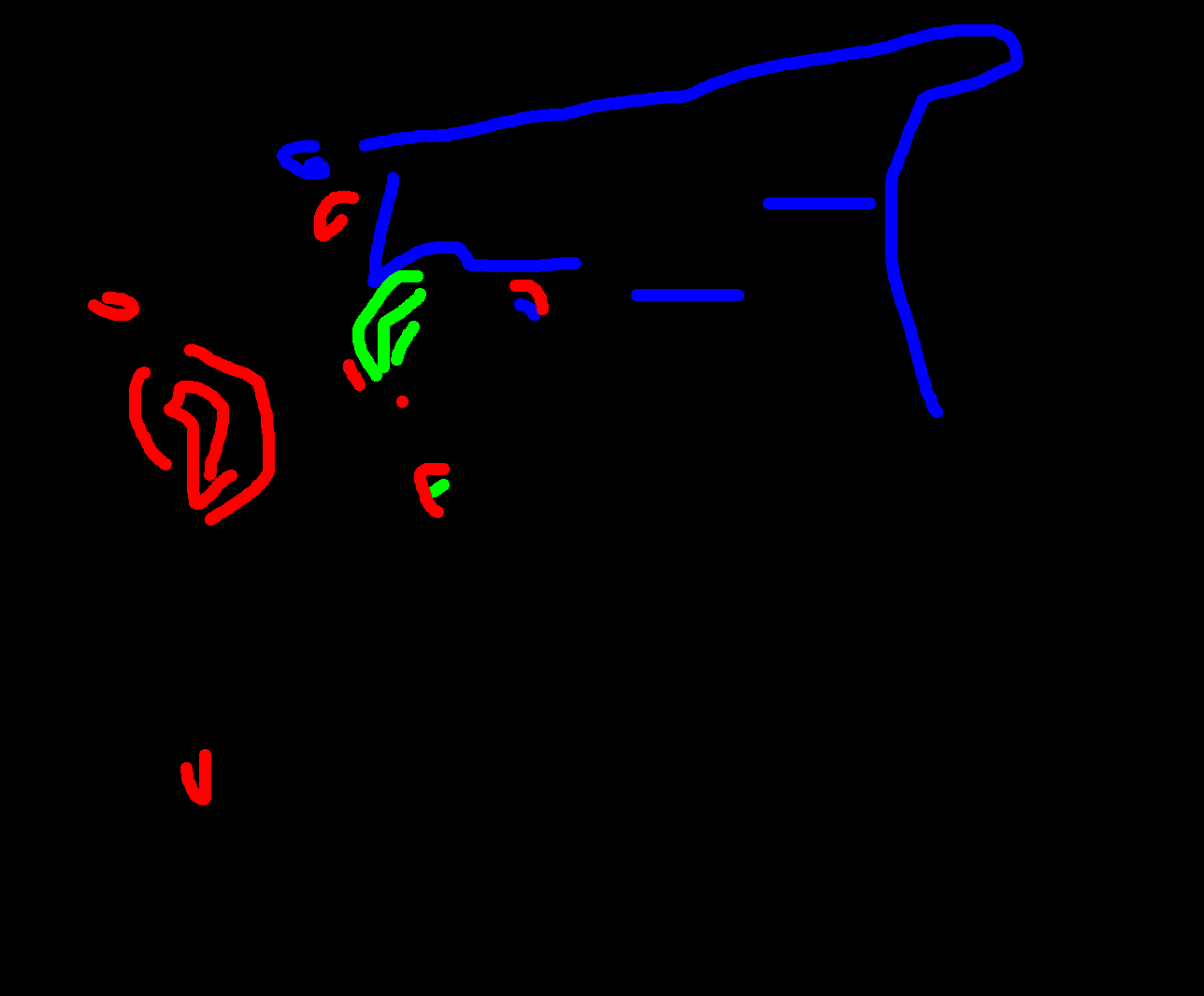}
\\\midrule
\rotatebox{90}{A05 Results} & 
\includegraphics[width=.19\textwidth]{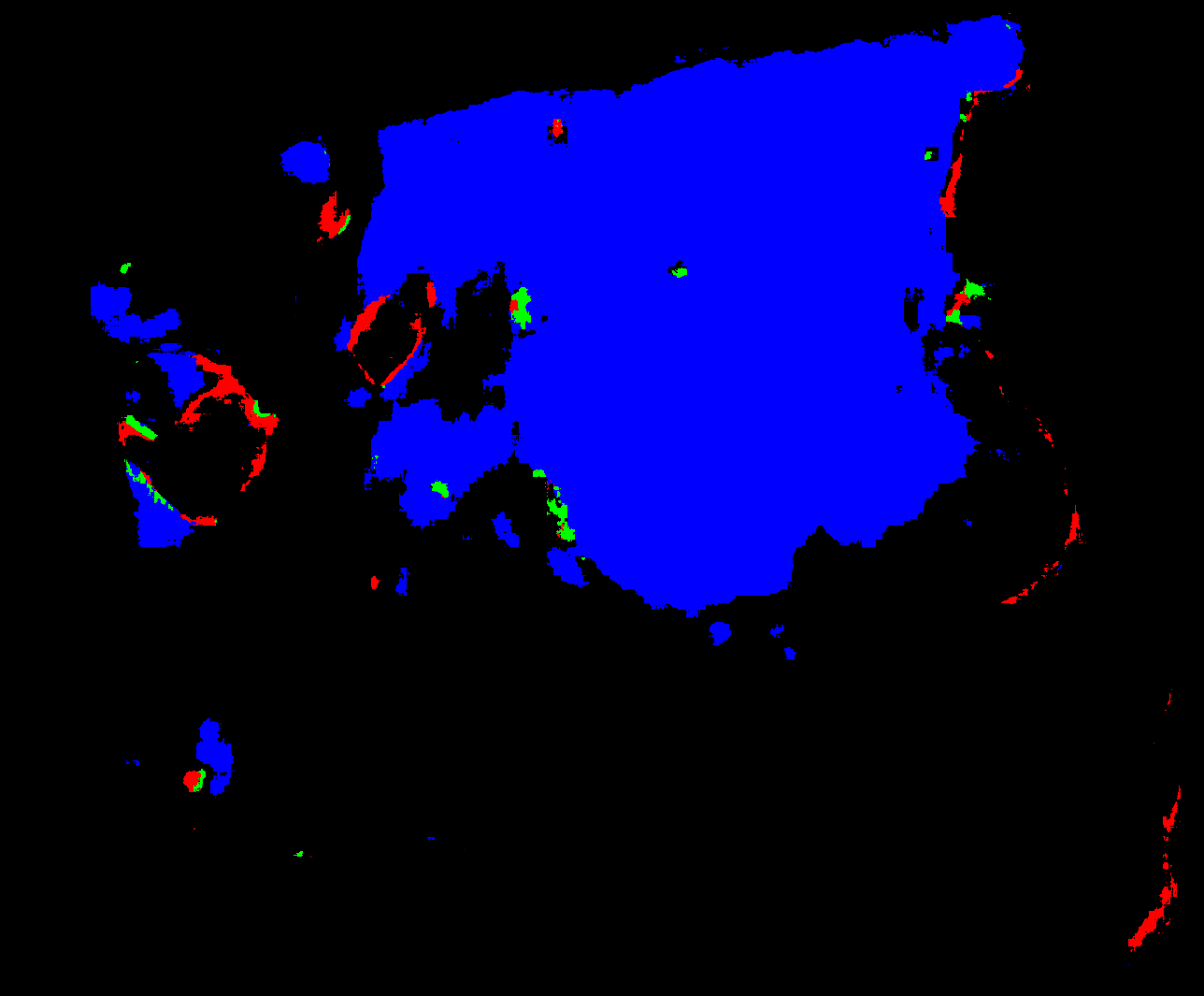} &
\includegraphics[width=.19\textwidth]{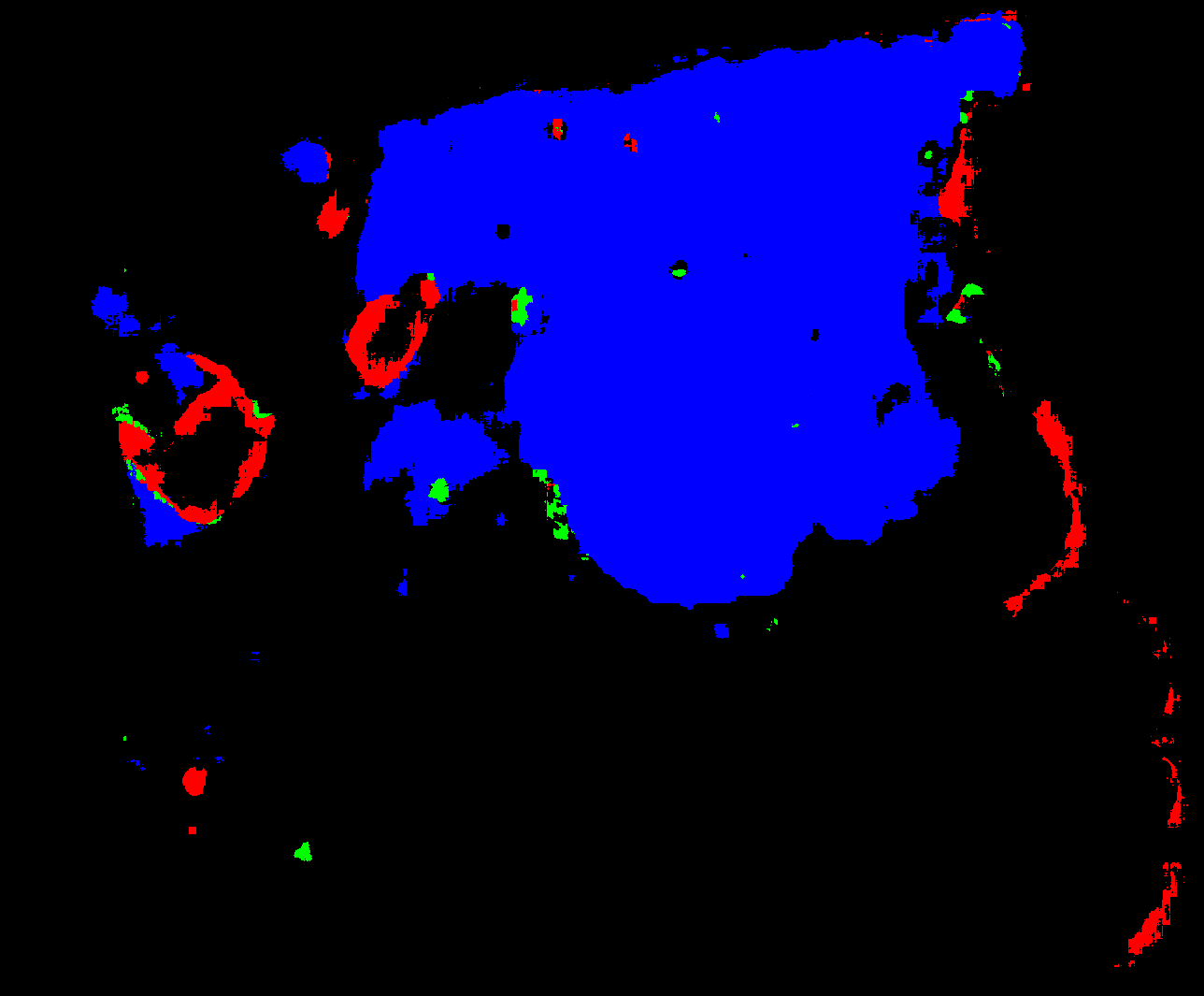} &
\includegraphics[width=.19\textwidth]{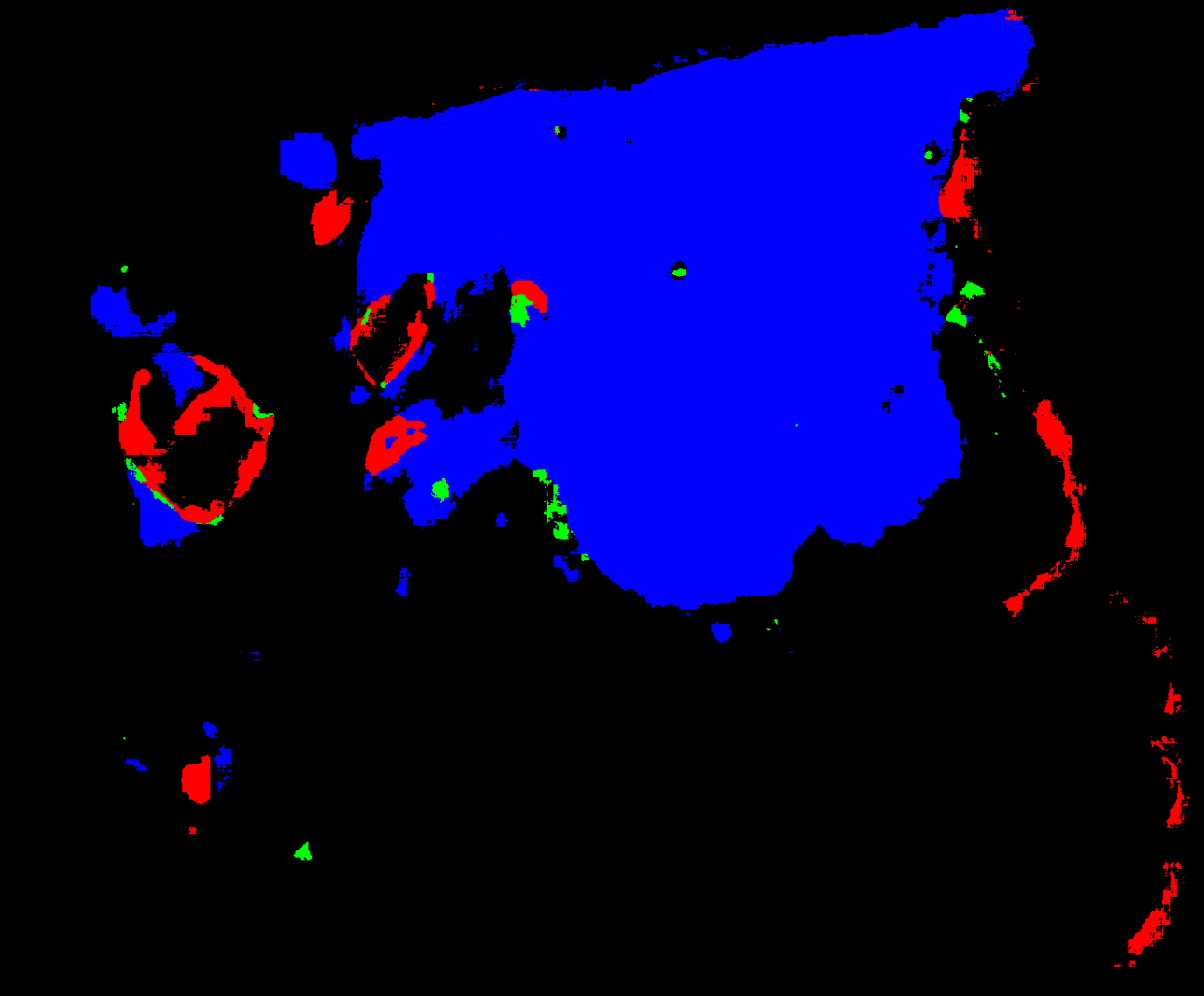} &
\includegraphics[width=.19\textwidth]{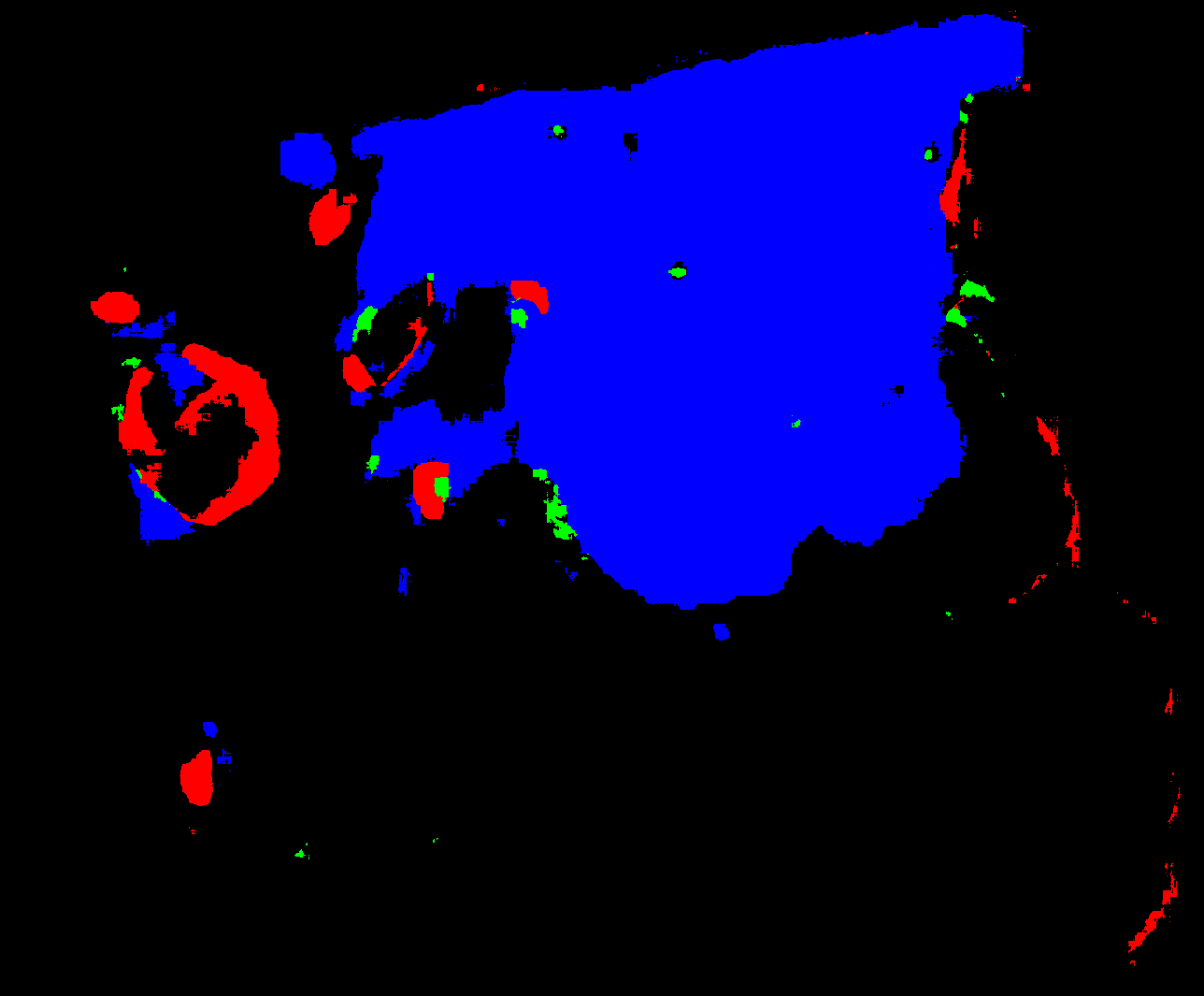} &
\includegraphics[width=.19\textwidth]{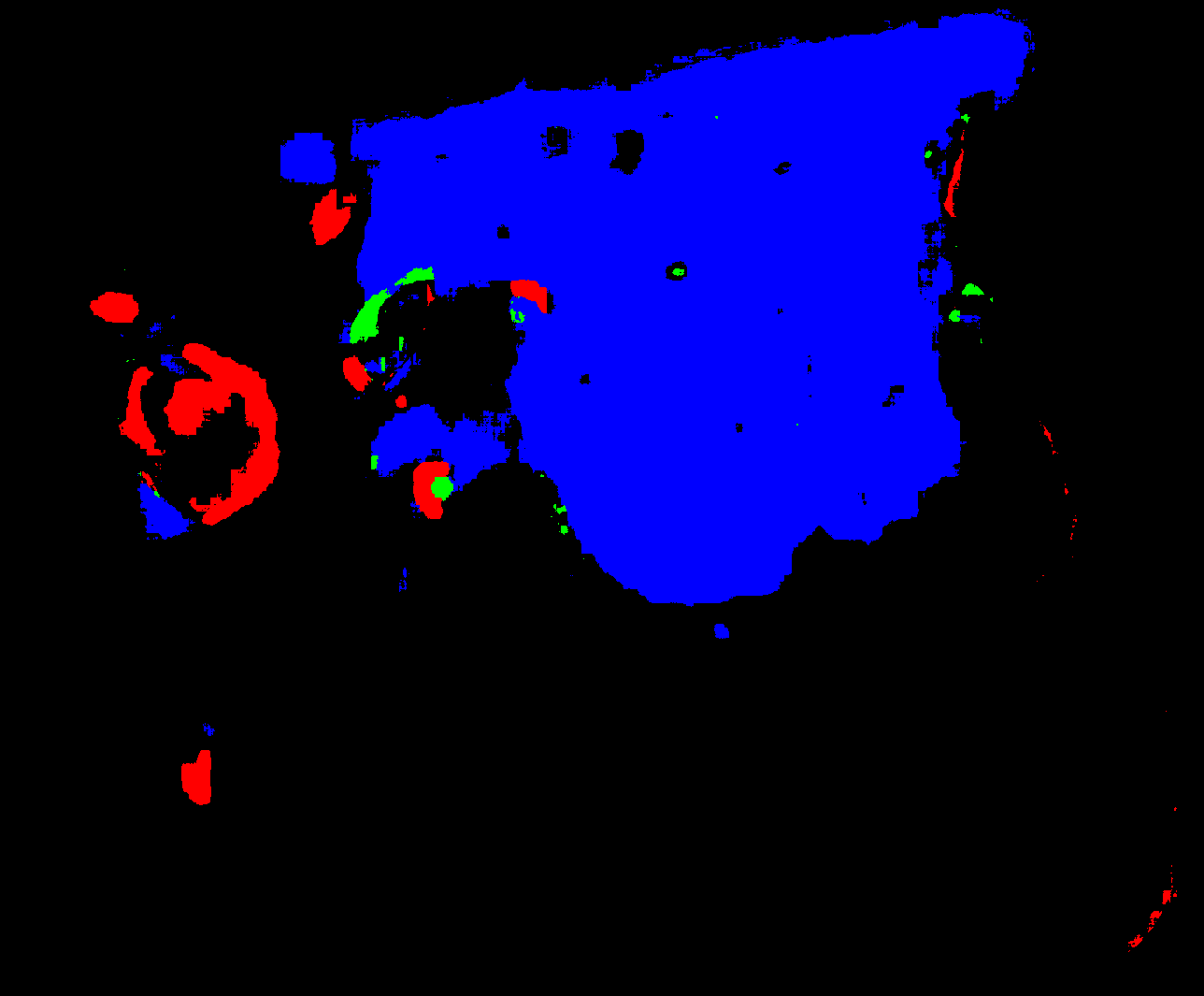}
\\\midrule
\rotatebox{90}{A10 Brush} & 
\includegraphics[width=.19\textwidth]{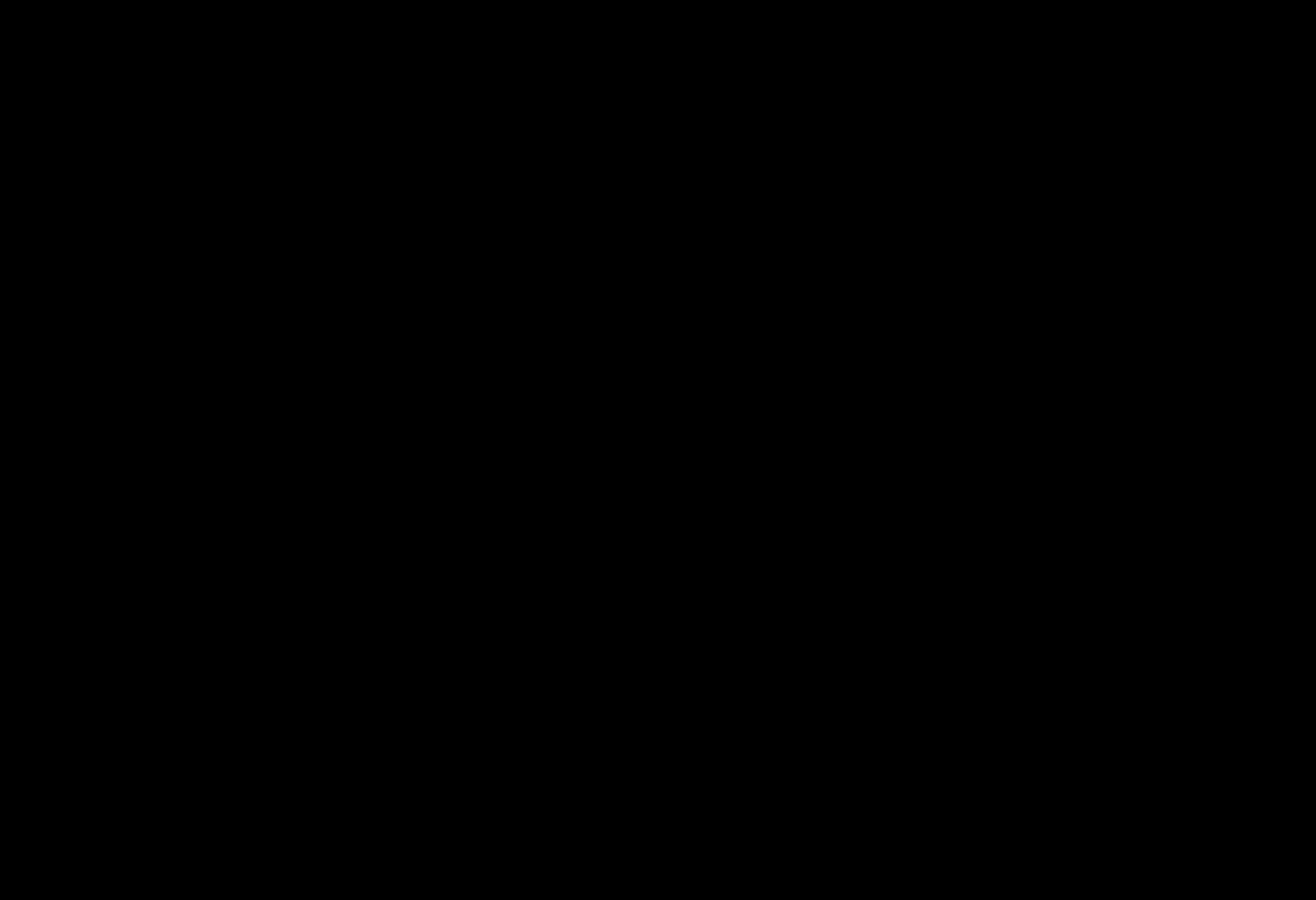} &
\includegraphics[width=.19\textwidth]{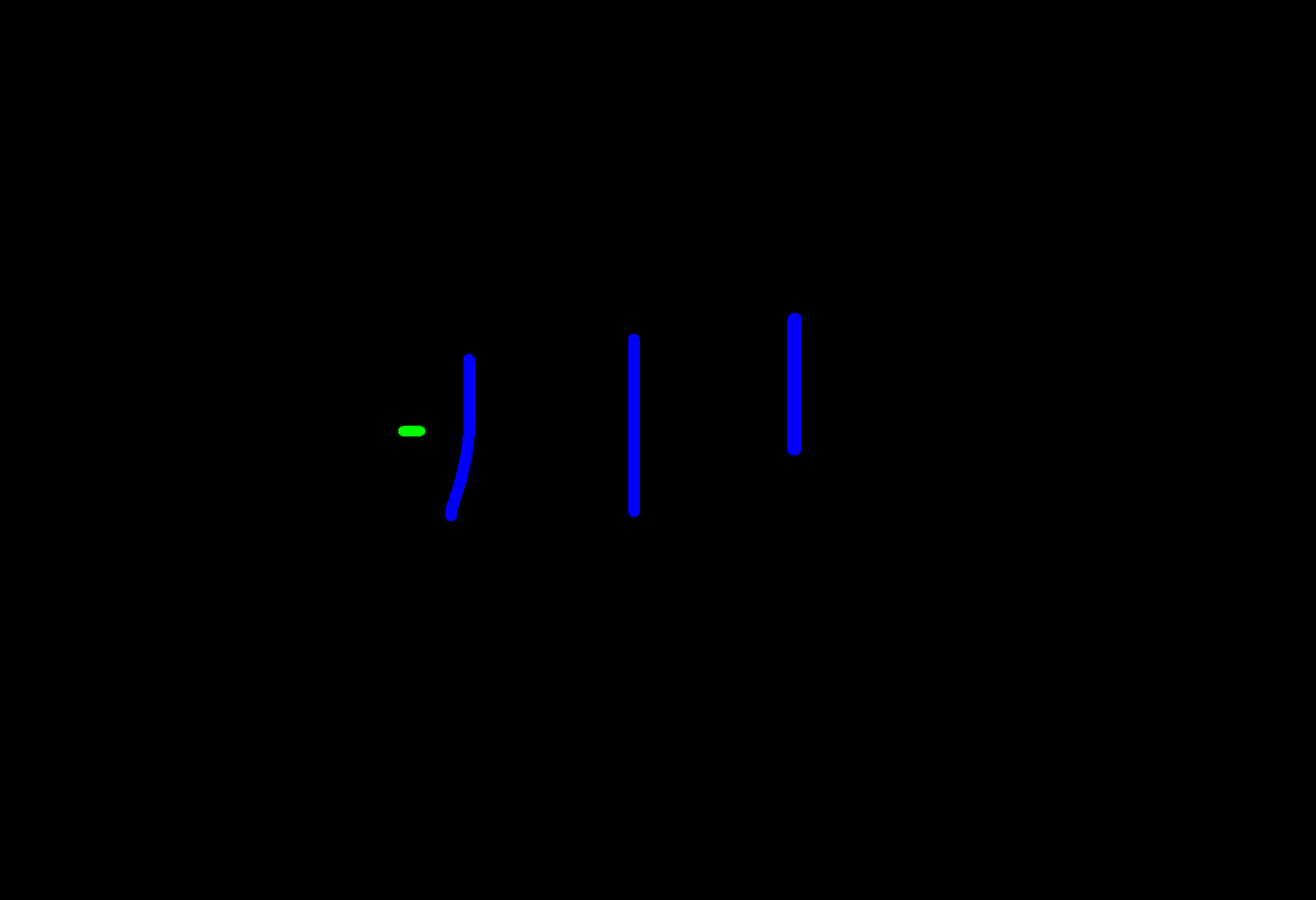} &
\includegraphics[width=.19\textwidth]{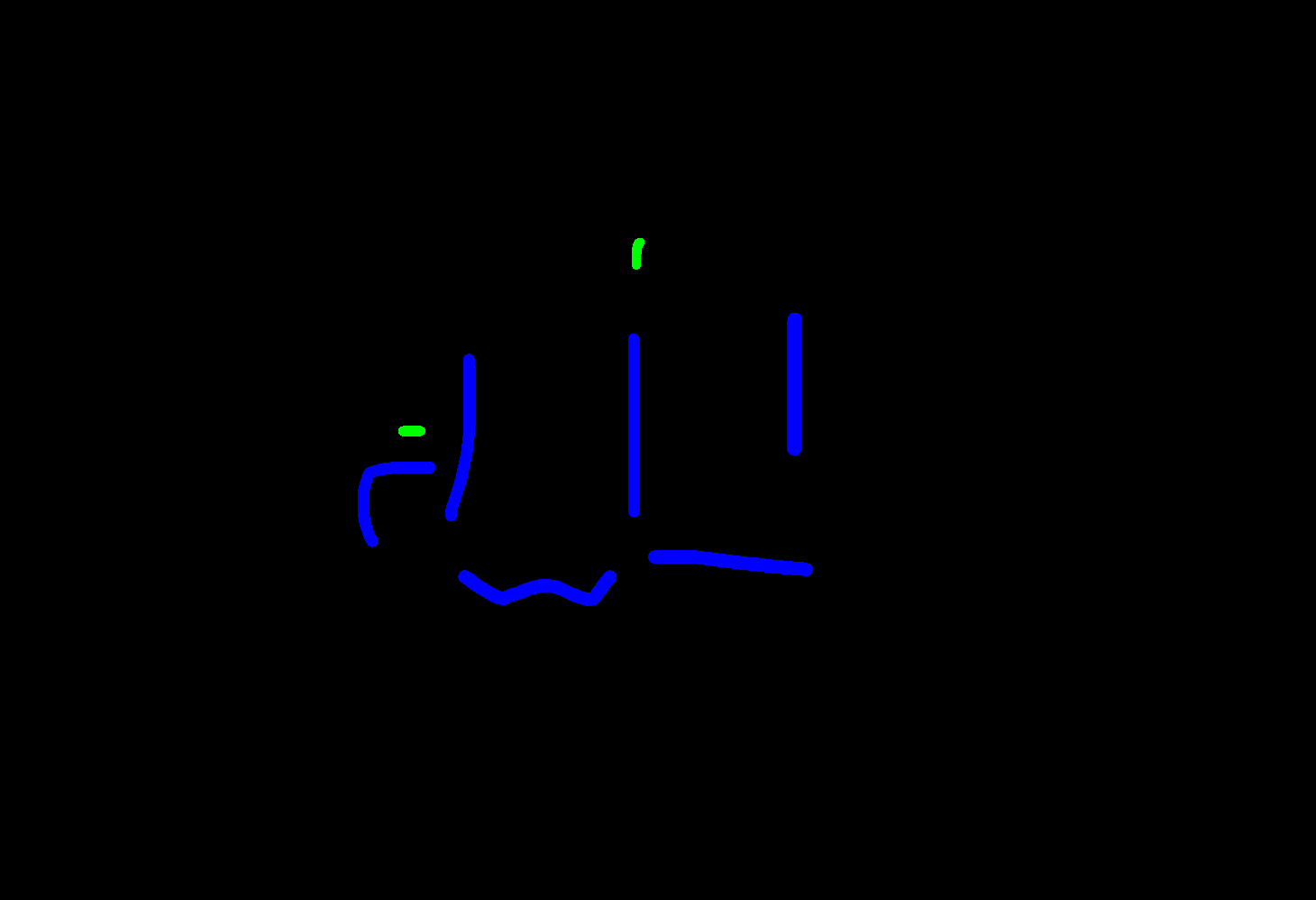} &
\includegraphics[width=.19\textwidth]{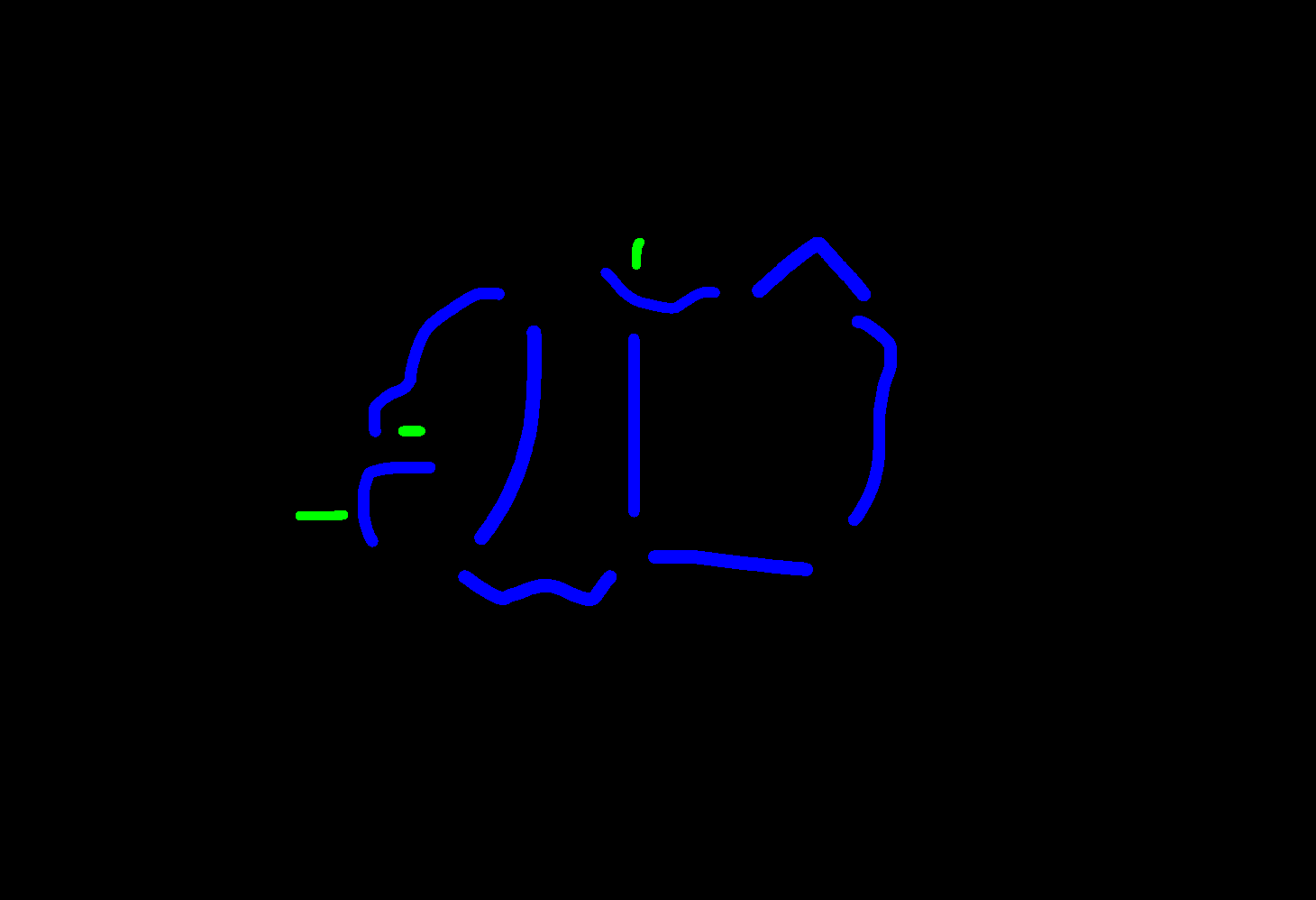} &
\includegraphics[width=.19\textwidth]{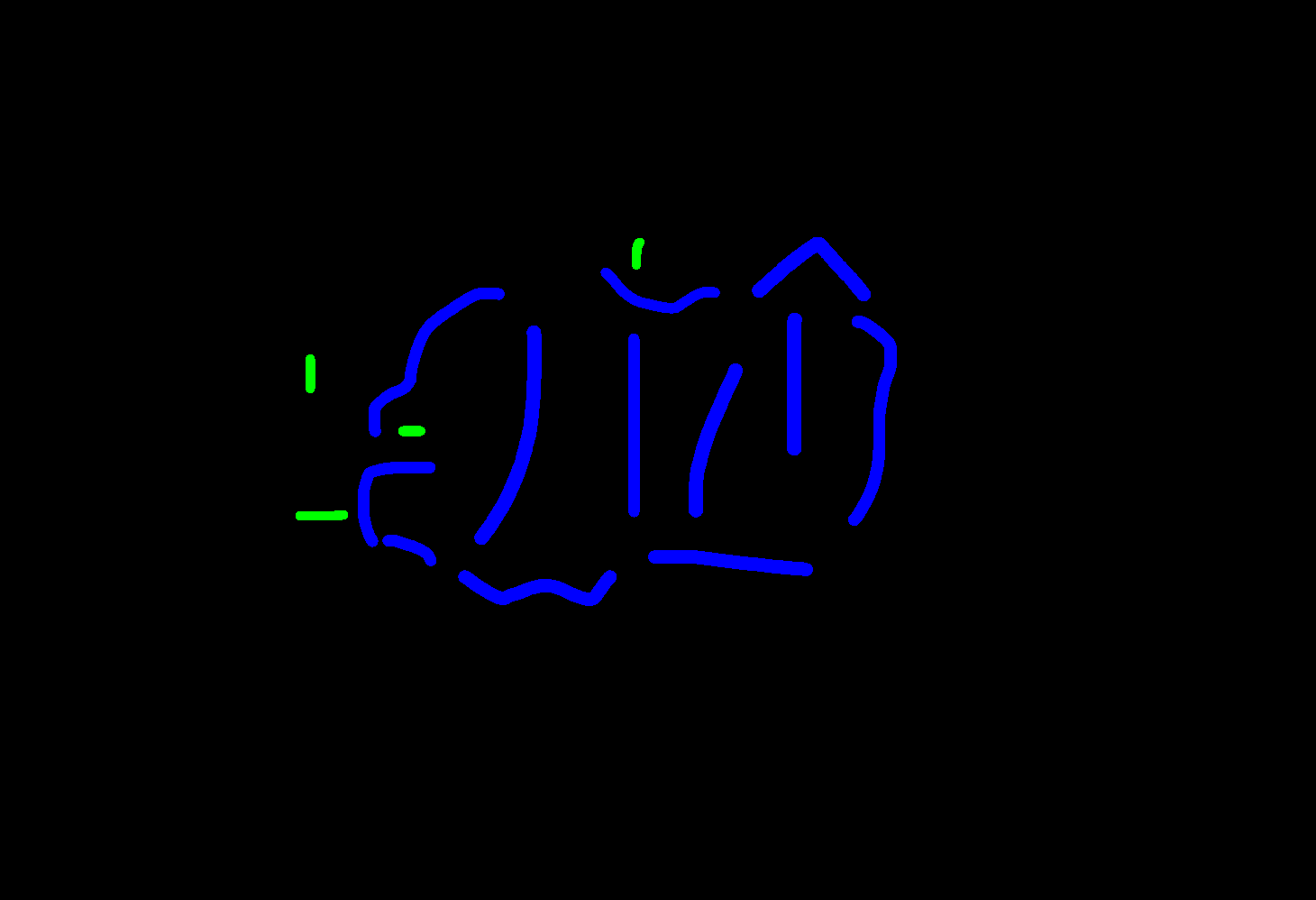}
\\\midrule
\rotatebox{90}{A10 Results} & 
\includegraphics[width=.19\textwidth]{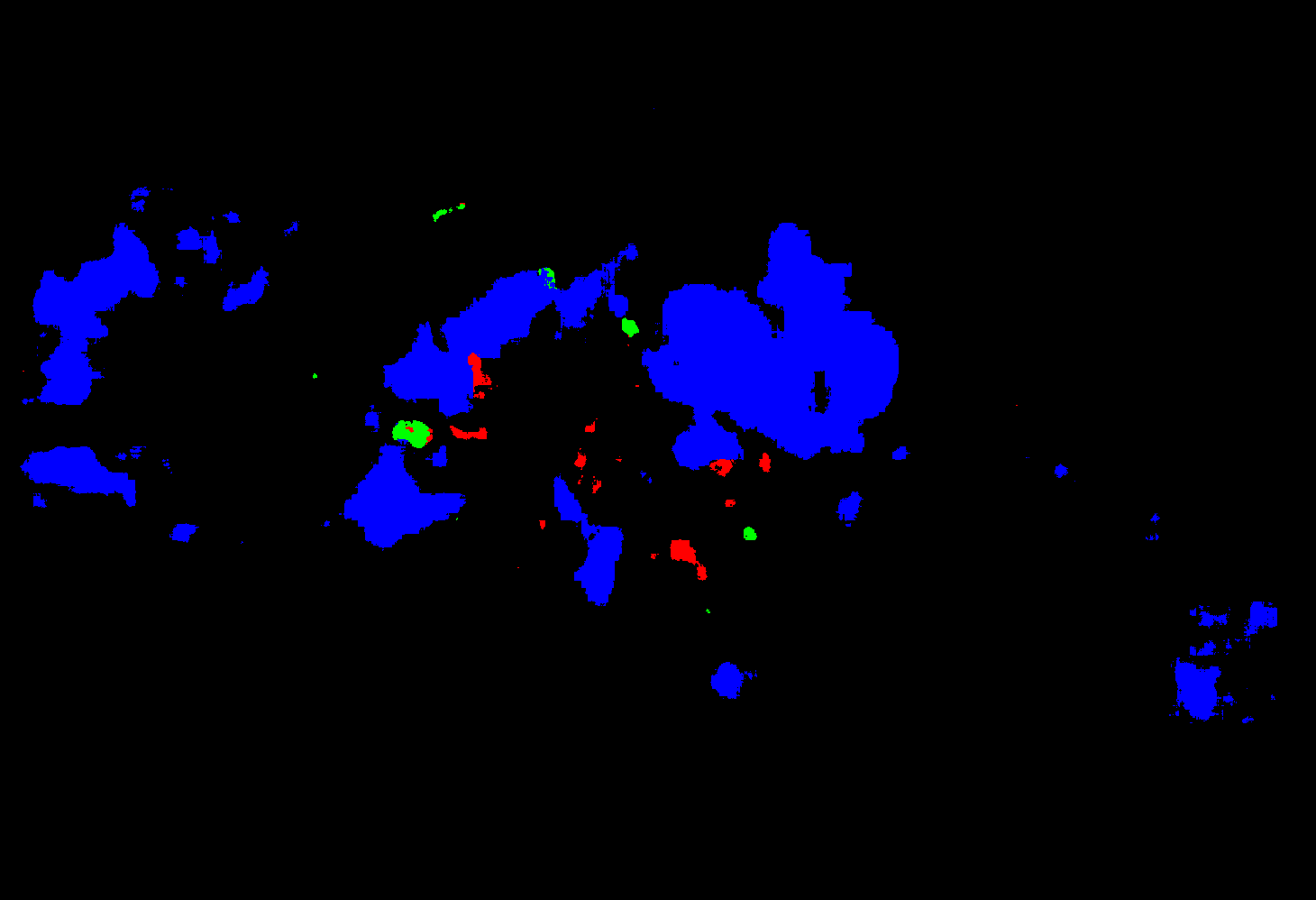} &
\includegraphics[width=.19\textwidth]{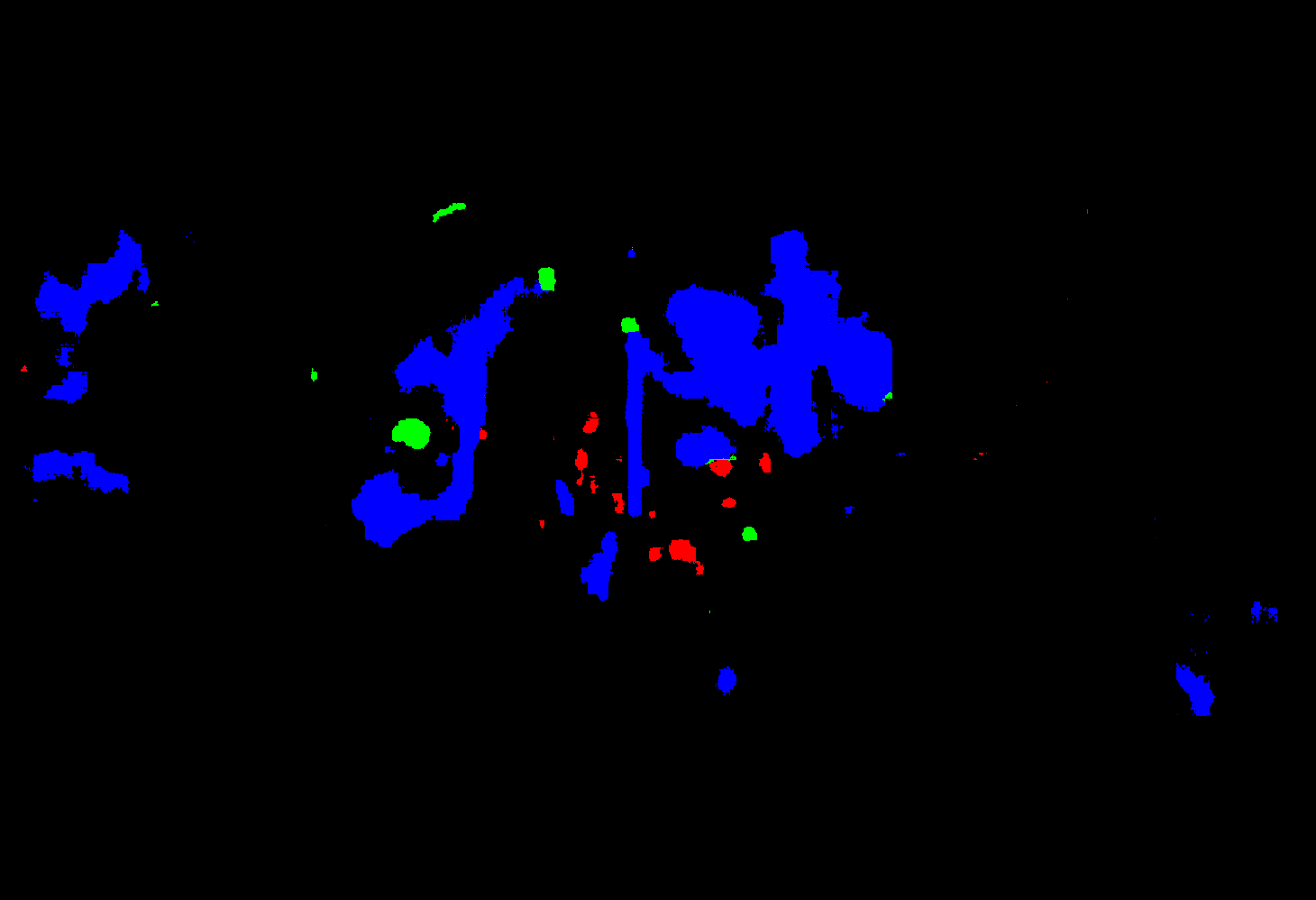} &
\includegraphics[width=.19\textwidth]{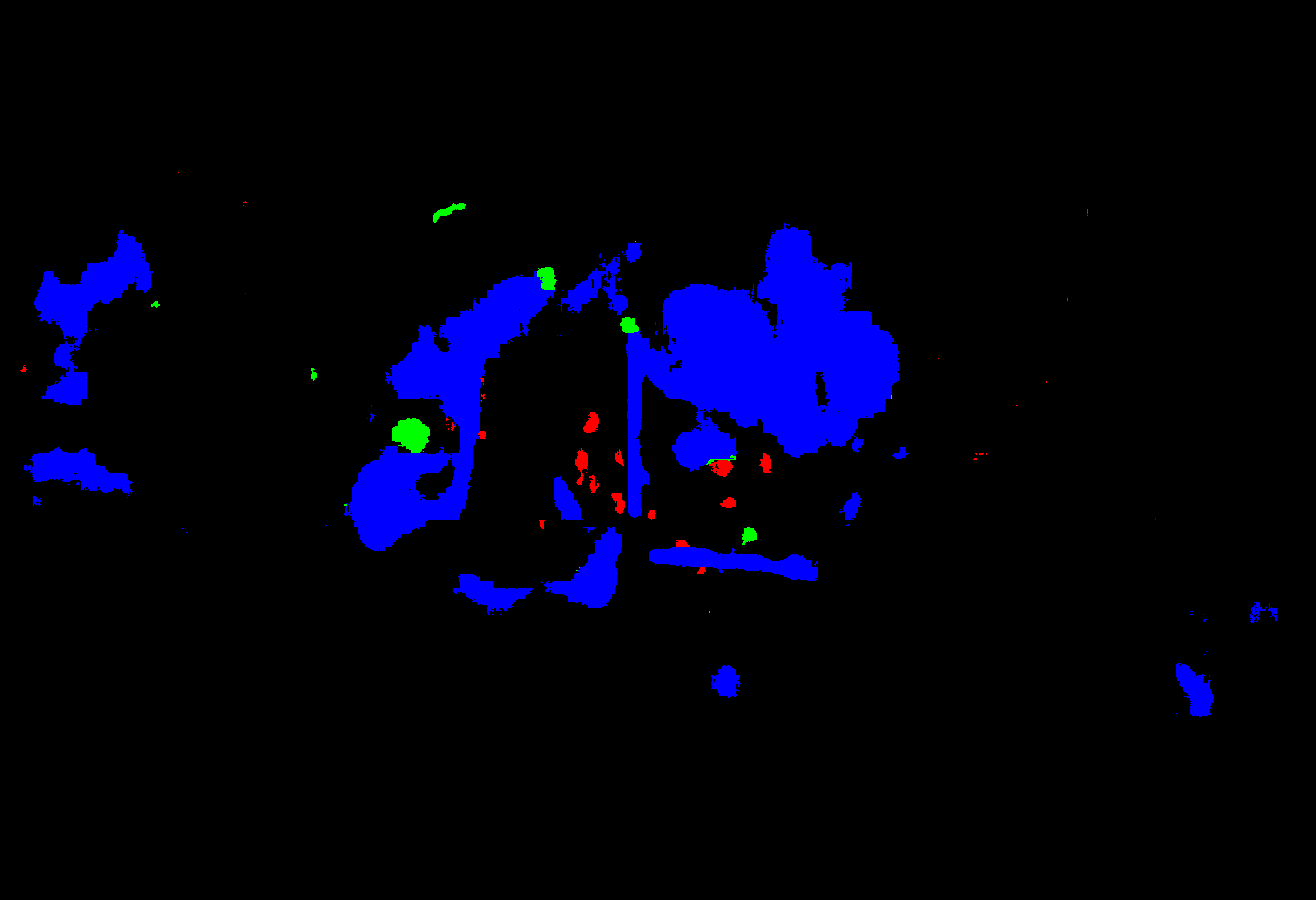} &
\includegraphics[width=.19\textwidth]{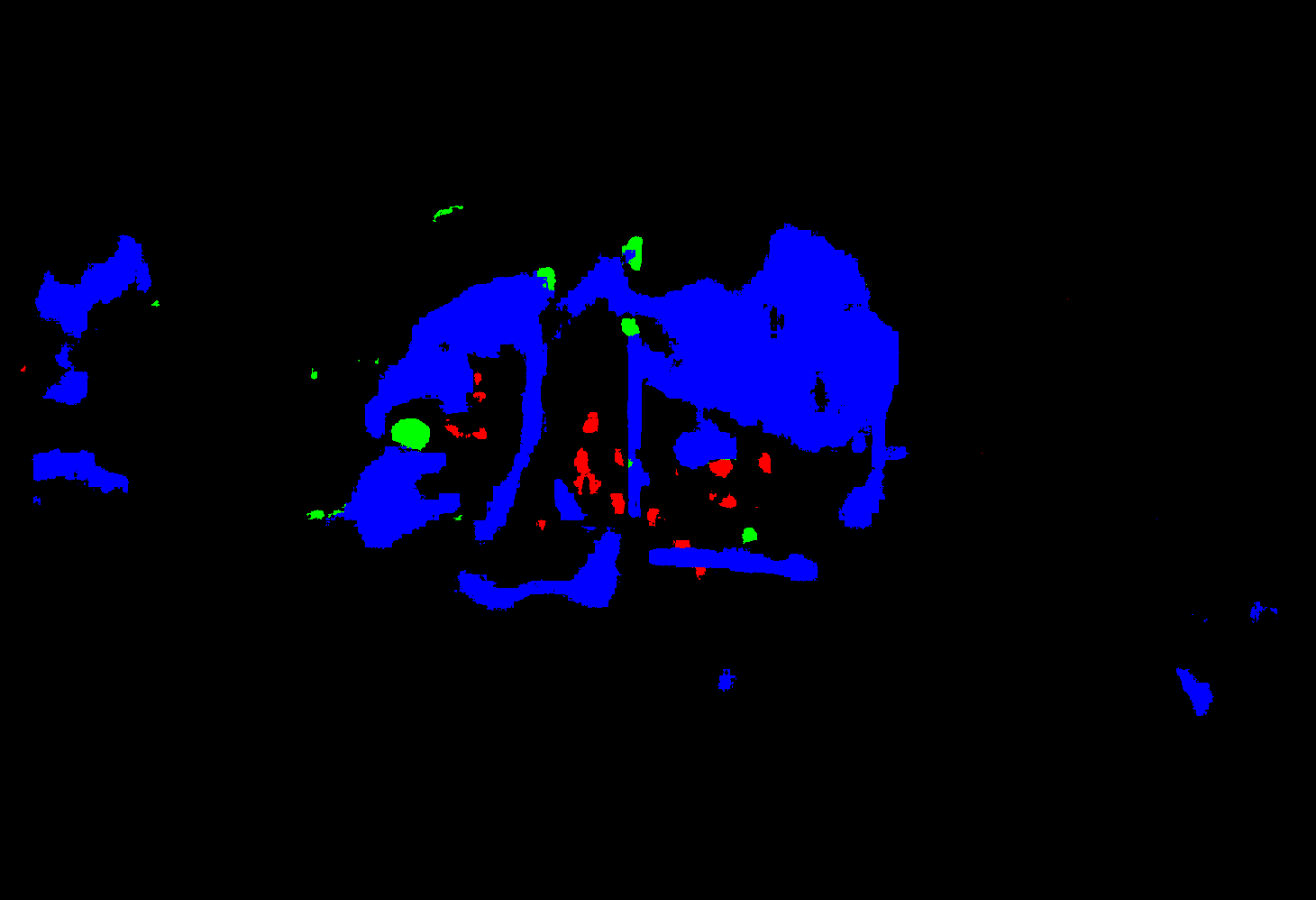} &
\includegraphics[width=.19\textwidth]{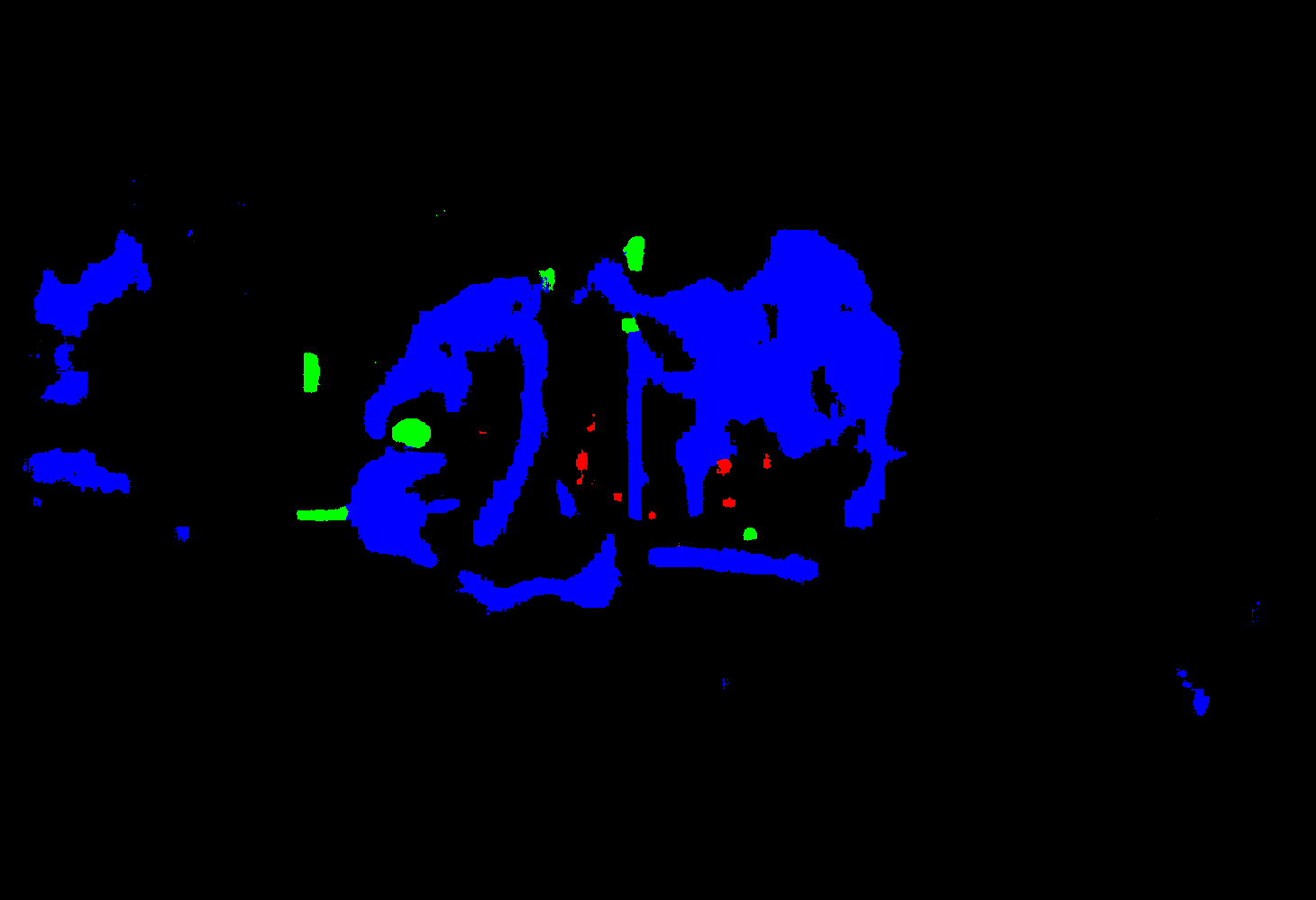}
\\\bottomrule
\vspace{0.01pt}
\end{tabular}
\end{center} 
\addlegendimageintext{fill=red, area legend} Benign
\hspace{3pt}
\addlegendimageintext{fill=blue, area legend} Invasive carcinoma
\hspace{3pt}
\addlegendimageintext{fill=green, area legend} In situ carcinoma
\caption{Results on BACH \cite{aresta2019bach} with varying brush strokes. Best viewed in color.}
\label{fig:bachreduced}
\end{figure*}

\begin{figure*}[t]
\centering
\begin{tabular}{l@{\ }c@{\ }c@{\ }c@{\ }c@{\ }c@{\ }}
 \toprule
\multicolumn{1}{c}{}
& \multicolumn{1}{c}{0\%}
& \multicolumn{1}{c}{20\%}
& \multicolumn{1}{c}{40\%}
& \multicolumn{1}{c}{60\%}
& \multicolumn{1}{c}{80\%}
\\\midrule
\rotatebox{90}{zh5 Brush} & 
 &
\includegraphics[width=.19\textwidth]{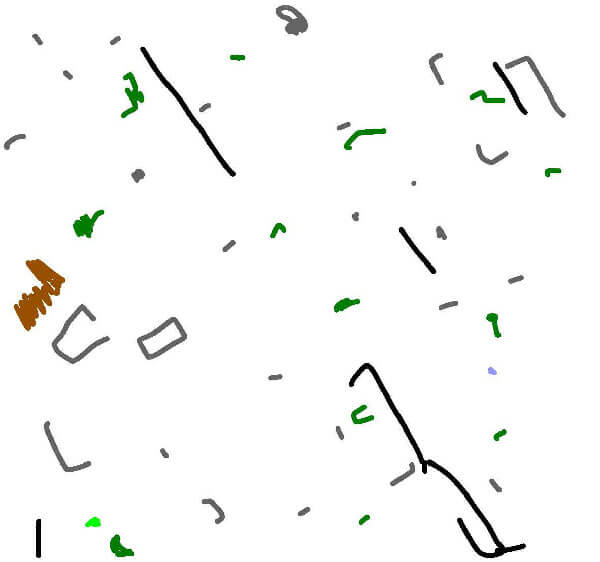} &
\includegraphics[width=.19\textwidth]{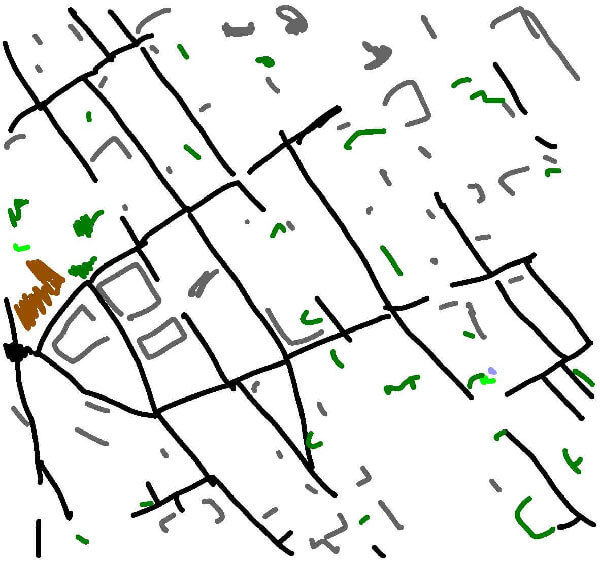} &
\includegraphics[width=.19\textwidth]{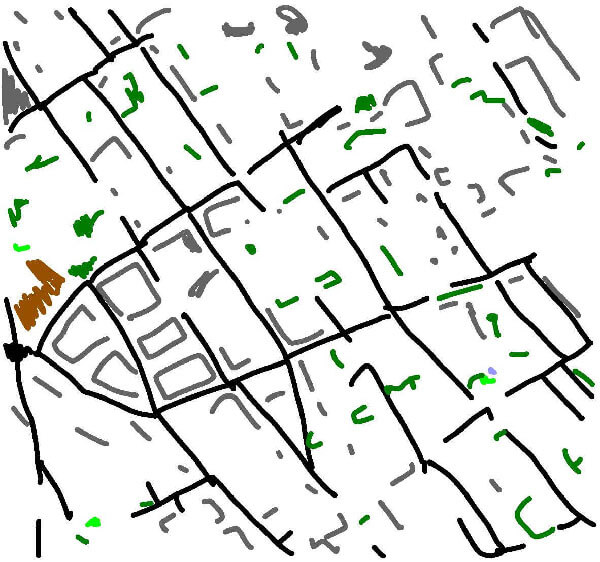} &
\includegraphics[width=.19\textwidth]{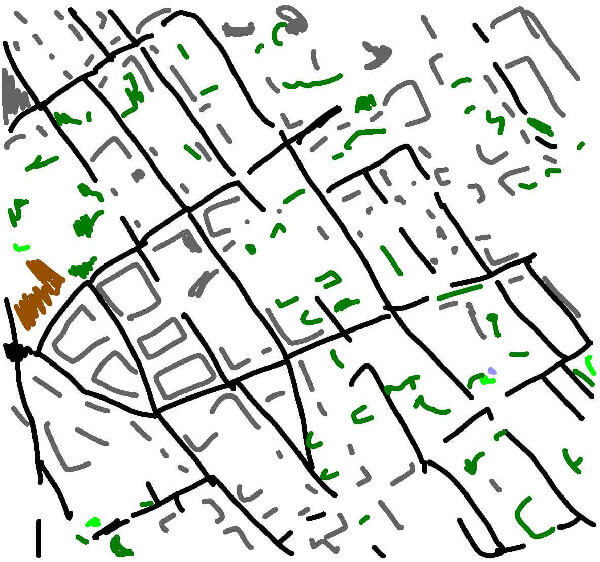} 
\\\midrule
\rotatebox{90}{zh5 Results} & 
\includegraphics[width=.19\textwidth]{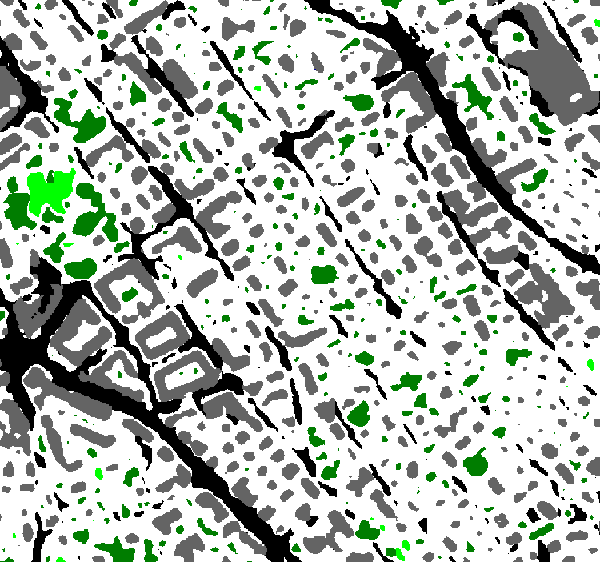} &
\includegraphics[width=.19\textwidth]{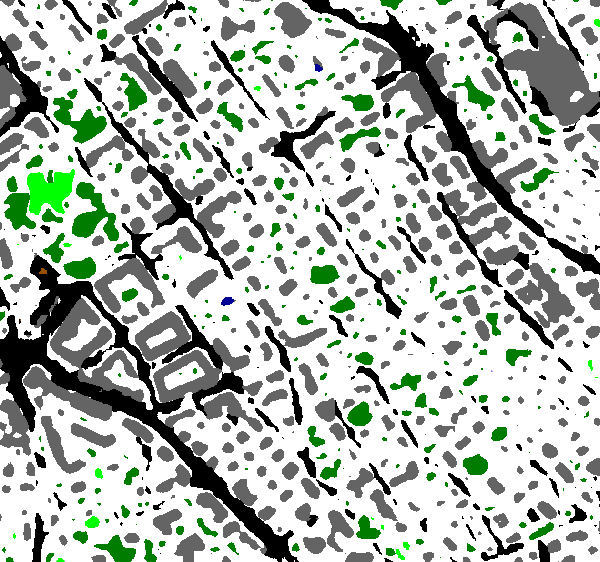} &
\includegraphics[width=.19\textwidth]{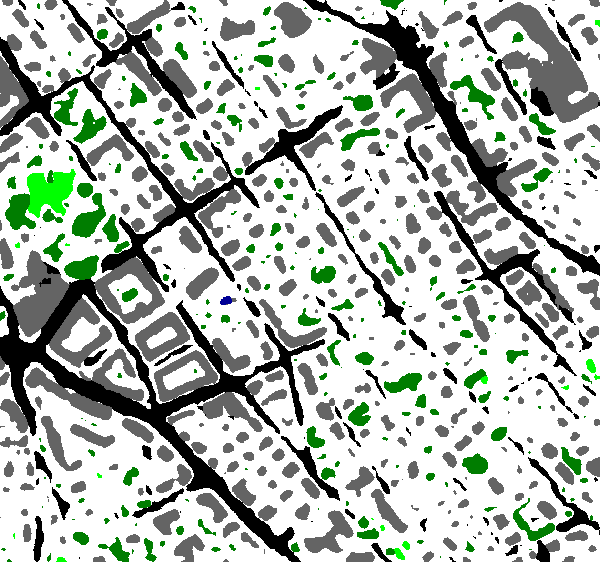} &
\includegraphics[width=.19\textwidth]{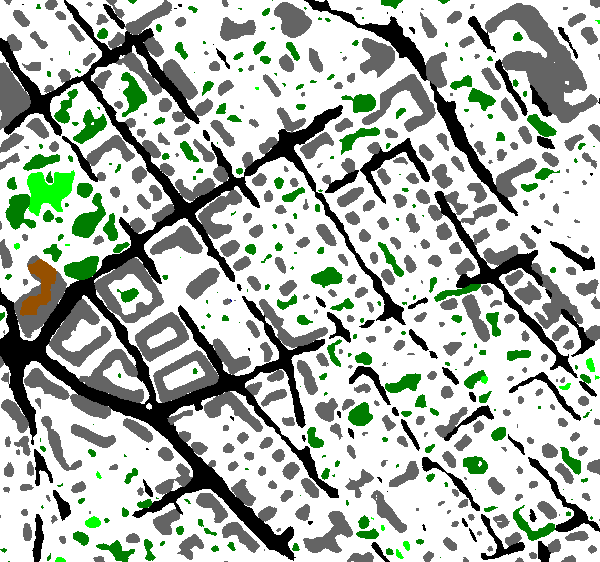} &
\includegraphics[width=.19\textwidth]{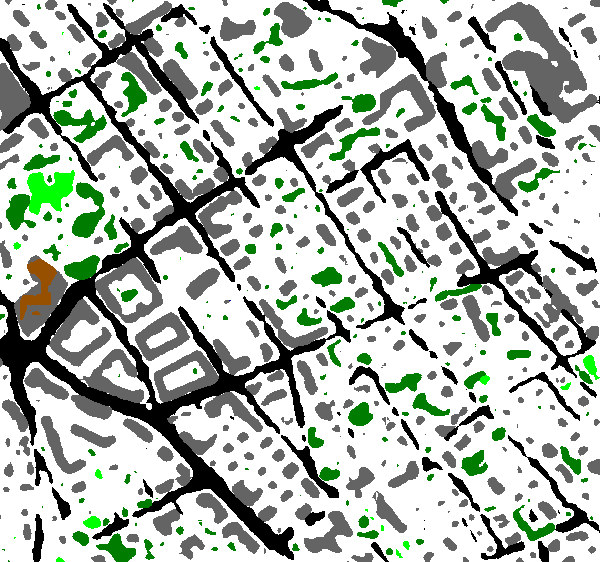} 
\\\midrule
\rotatebox{90}{zh7 Brush} & 
 &
\includegraphics[width=.19\textwidth]{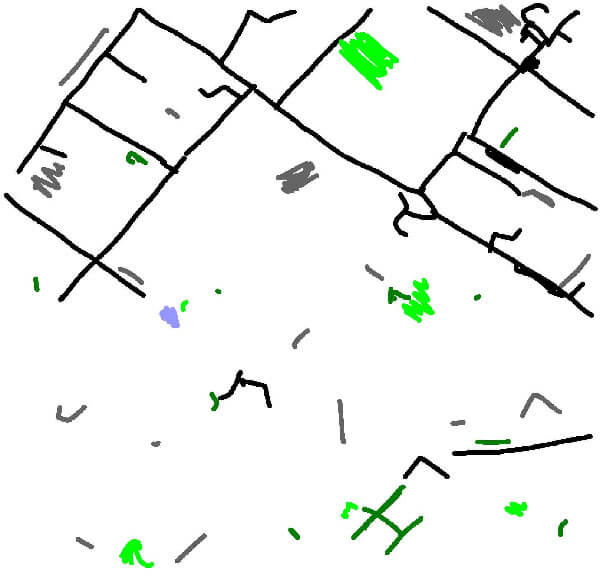} &
\includegraphics[width=.19\textwidth]{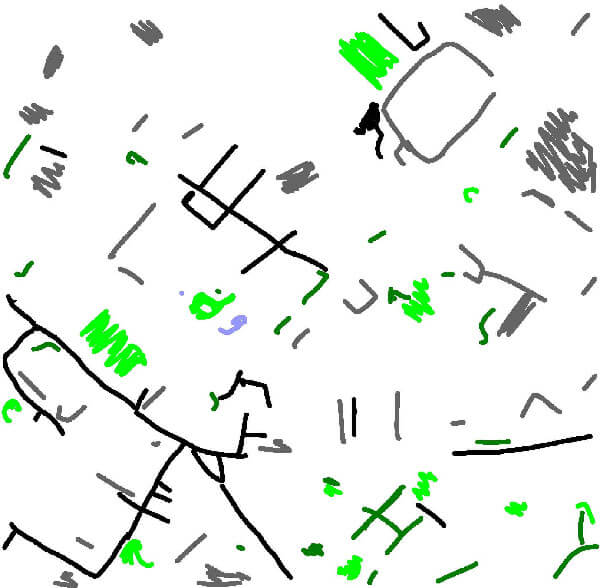} &
\includegraphics[width=.19\textwidth]{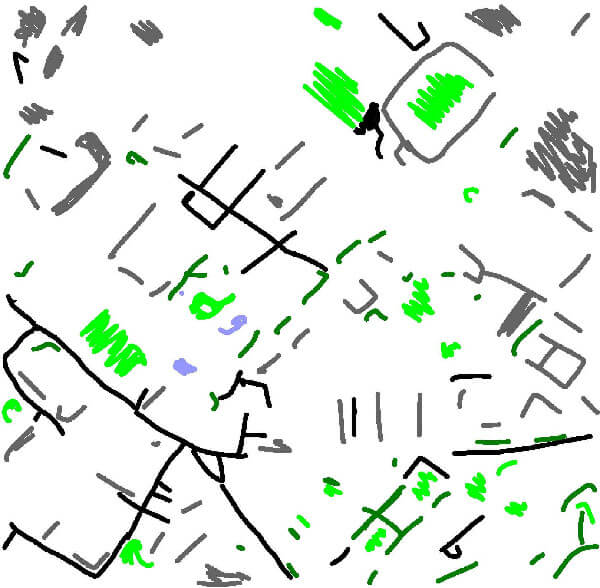} &
\includegraphics[width=.19\textwidth]{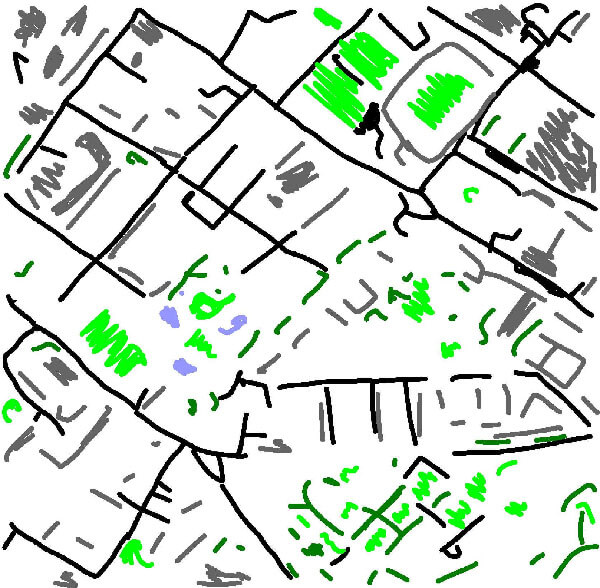} 
\\\midrule
\rotatebox{90}{zh7 Results} & 
\includegraphics[width=.19\textwidth]{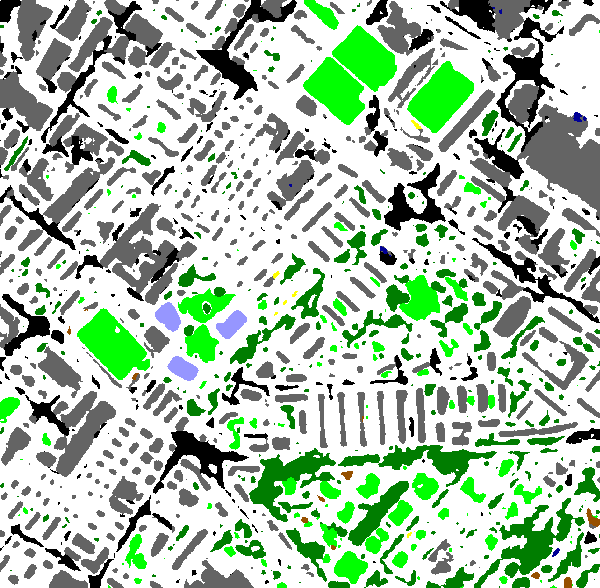} &
\includegraphics[width=.19\textwidth]{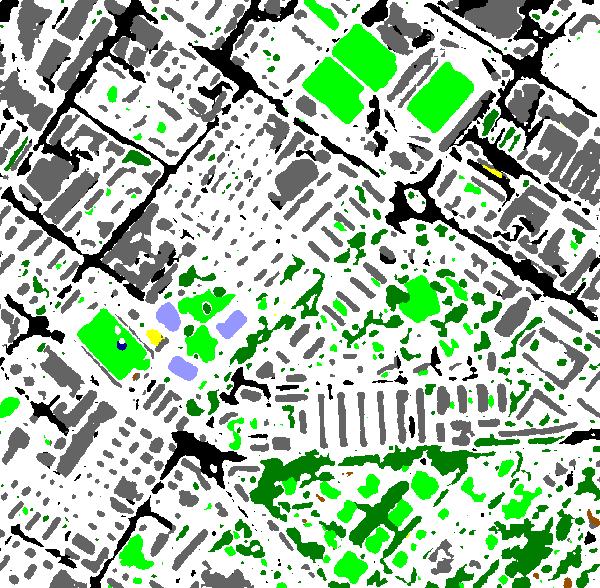} &
\includegraphics[width=.19\textwidth]{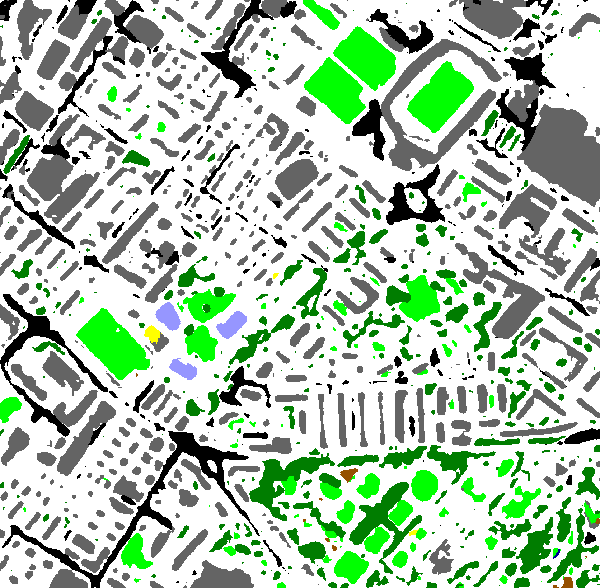} &
\includegraphics[width=.19\textwidth]{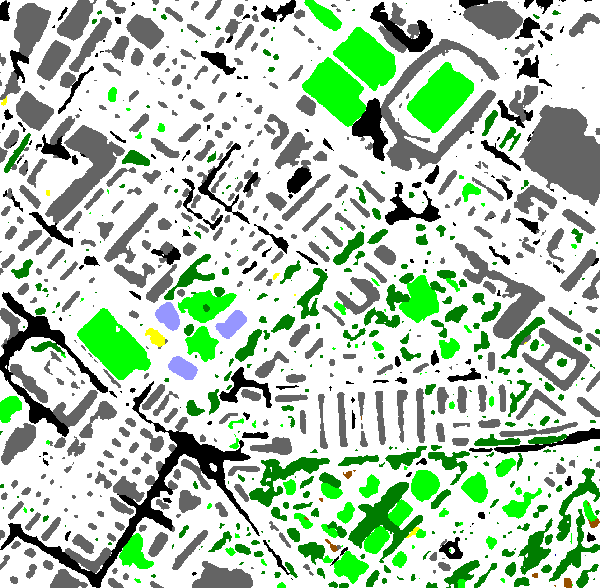} &
\includegraphics[width=.19\textwidth]{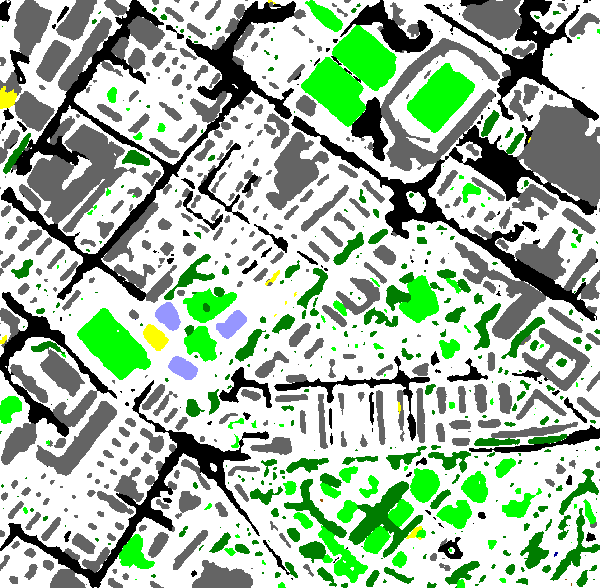} 
\\\midrule
\rotatebox{90}{zh8 Brush} & 
 &
\includegraphics[width=.19\textwidth]{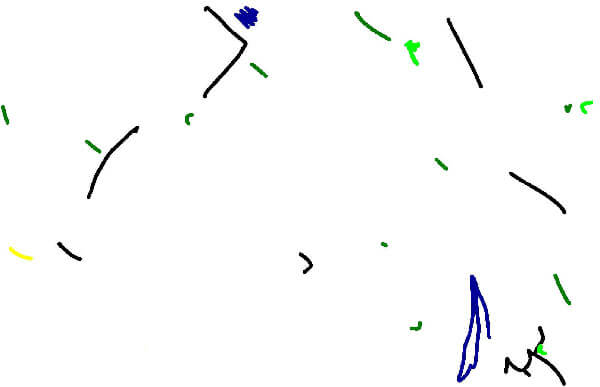} &
\includegraphics[width=.19\textwidth]{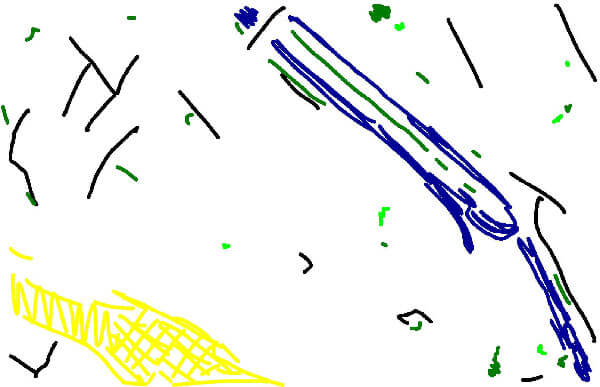} &
\includegraphics[width=.19\textwidth]{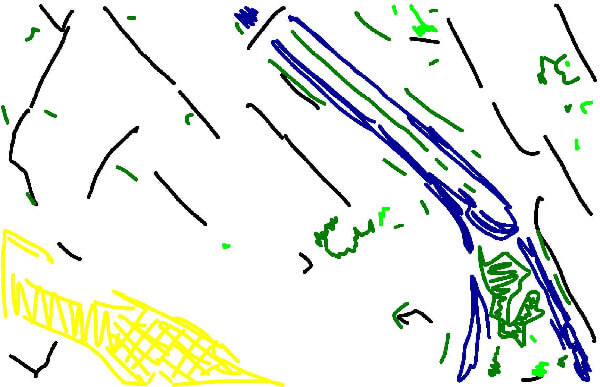} &
\includegraphics[width=.19\textwidth]{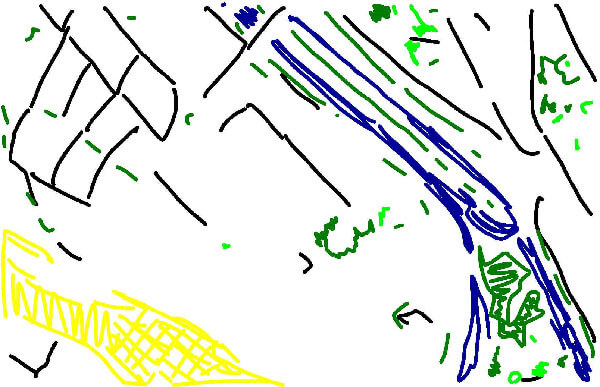} 
\\\midrule
\rotatebox{90}{zh8 Results} & 
\includegraphics[width=.19\textwidth]{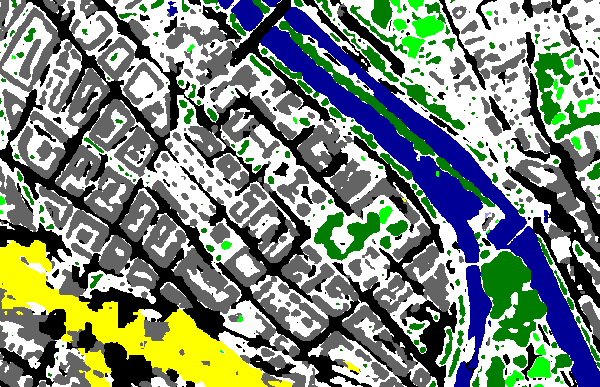} &
\includegraphics[width=.19\textwidth]{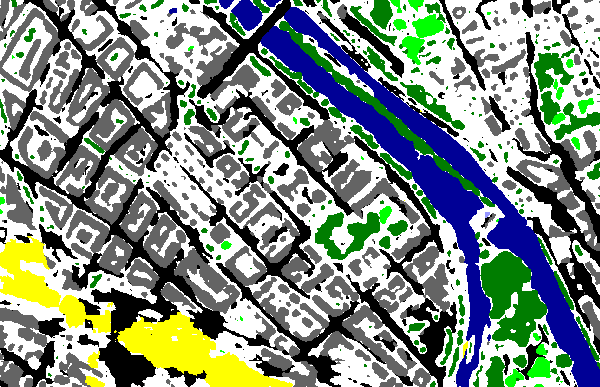} &
\includegraphics[width=.19\textwidth]{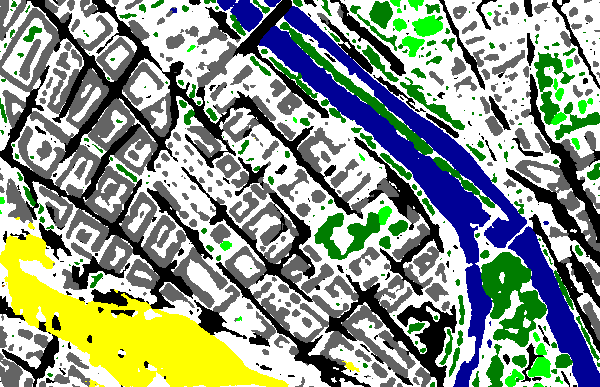} &
\includegraphics[width=.19\textwidth]{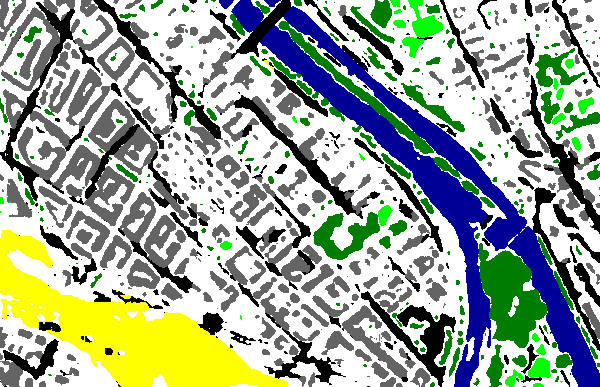} &
\includegraphics[width=.19\textwidth]{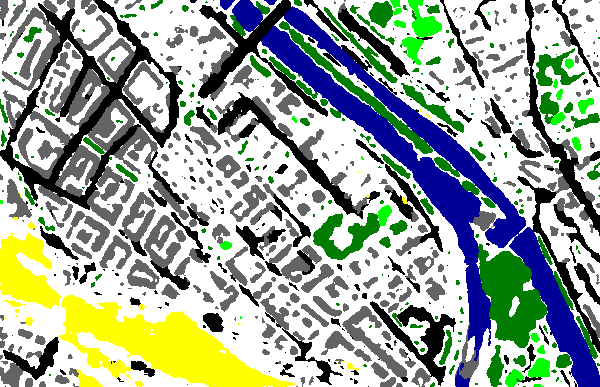} 
\vspace{0.01pt}
\end{tabular}
\addlegendimageintext{fill=black, area legend} Roads
\hspace{3pt}
\addlegendimageintext{fill=zurichgrey, area legend} Buildings
\hspace{3pt}
\addlegendimageintext{fill=zurichdarkgreen, area legend} Trees
\hspace{3pt}
\addlegendimageintext{fill=green, area legend} Grass
\hspace{3pt}
\addlegendimageintext{fill=zurichbrown, area legend} Bare Soil
\hspace{3pt}
\addlegendimageintext{fill=zurichdarkblue, area legend} Water
\hspace{3pt}
\addlegendimageintext{fill=zurichyellow, area legend} Railways
\hspace{3pt}
\addlegendimageintext{fill=zurichlightblue, area legend} Swimming Pools
\caption{Results on the Zurich Summer Dataset \cite{volpi2015semantic} with varying brush strokes. Best viewed in color.}
\label{fig:zurichreduced1}
\end{figure*}

\begin{figure*}[t]
\begin{tabular}{l@{\ }c@{\ }c@{\ }c@{\ }c@{\ }c@{\ }}
 \toprule
\multicolumn{1}{c}{}
& \multicolumn{1}{c}{0\%}
& \multicolumn{1}{c}{20\%}
& \multicolumn{1}{c}{40\%}
& \multicolumn{1}{c}{60\%}
& \multicolumn{1}{c}{80\%}
\\\midrule
\rotatebox{90}{zh11 Brush} &
 &
\includegraphics[width=.19\textwidth]{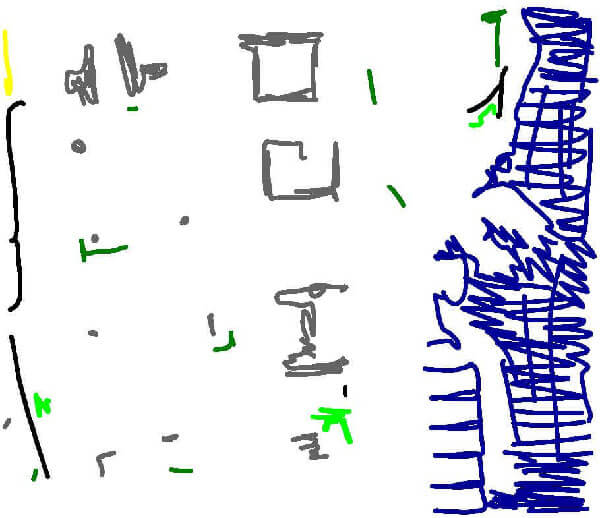} &
\includegraphics[width=.19\textwidth]{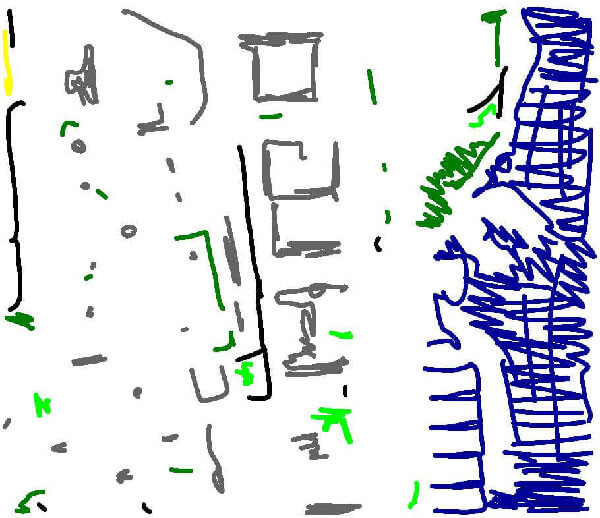} &
\includegraphics[width=.19\textwidth]{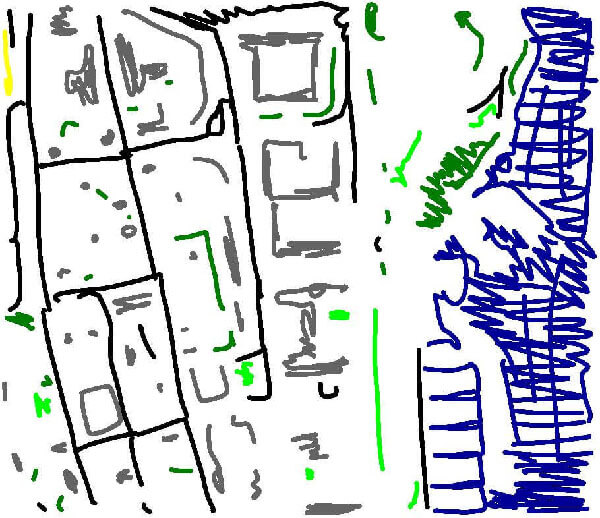} &
\includegraphics[width=.19\textwidth]{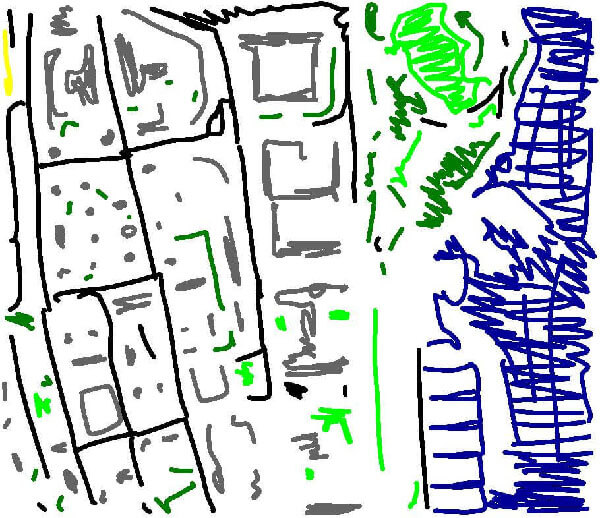} 
\\\midrule
\rotatebox{90}{zh11 Results} & 
\includegraphics[width=.19\textwidth]{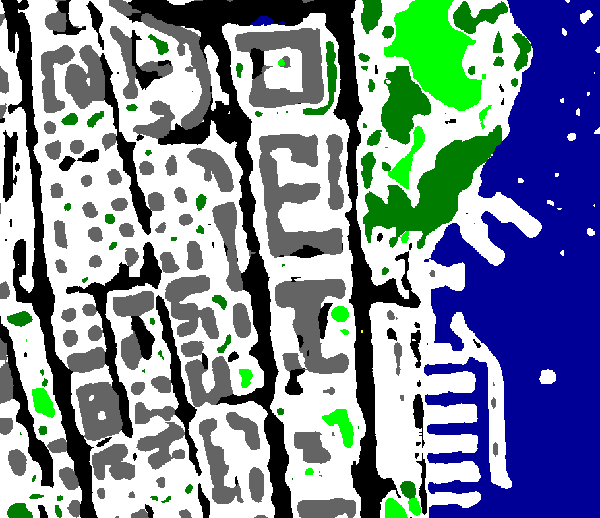} &
\includegraphics[width=.19\textwidth]{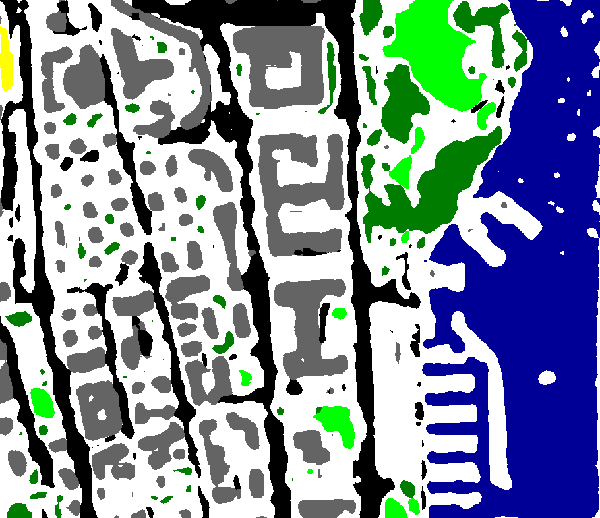} &
\includegraphics[width=.19\textwidth]{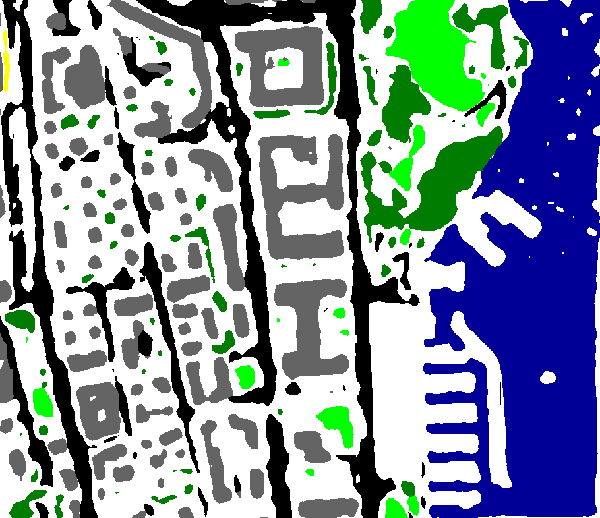} &
\includegraphics[width=.19\textwidth]{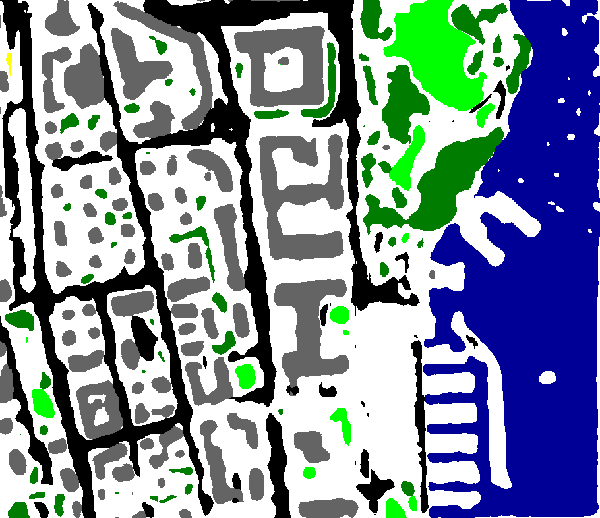} &
\includegraphics[width=.19\textwidth]{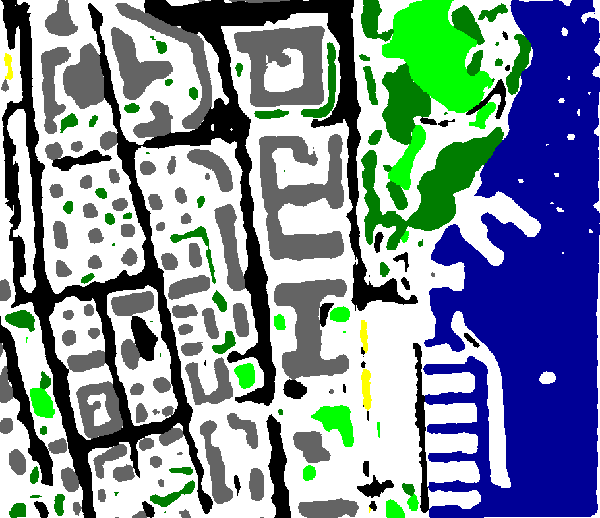} \\\midrule
\rotatebox{90}{zh18 Brush} & 
 &
\includegraphics[width=.19\textwidth]{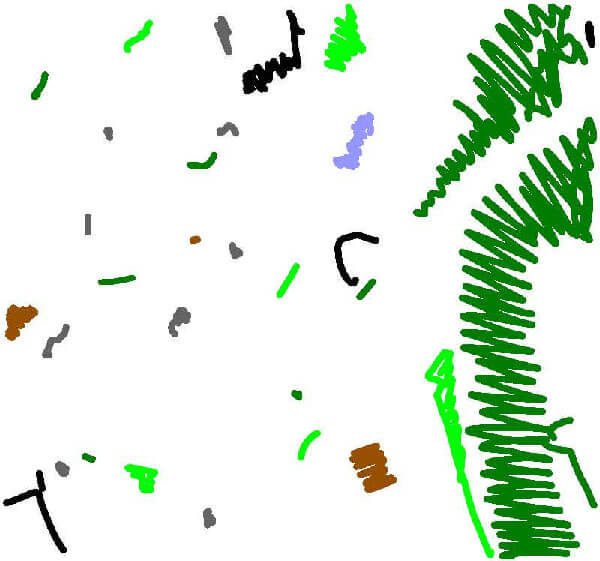} &
\includegraphics[width=.19\textwidth]{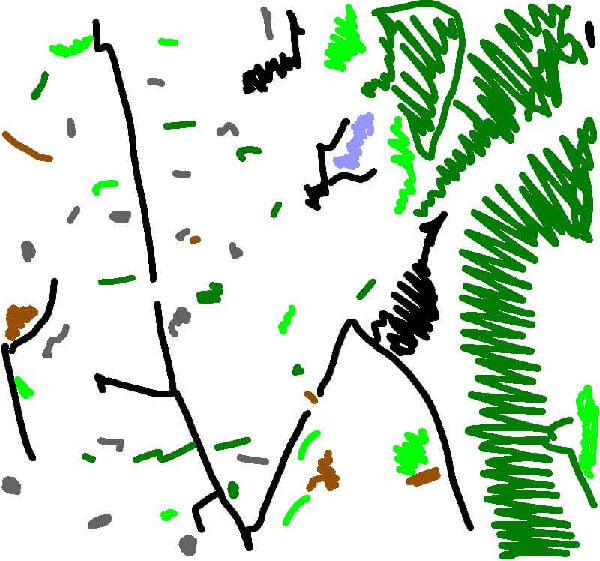} &
\includegraphics[width=.19\textwidth]{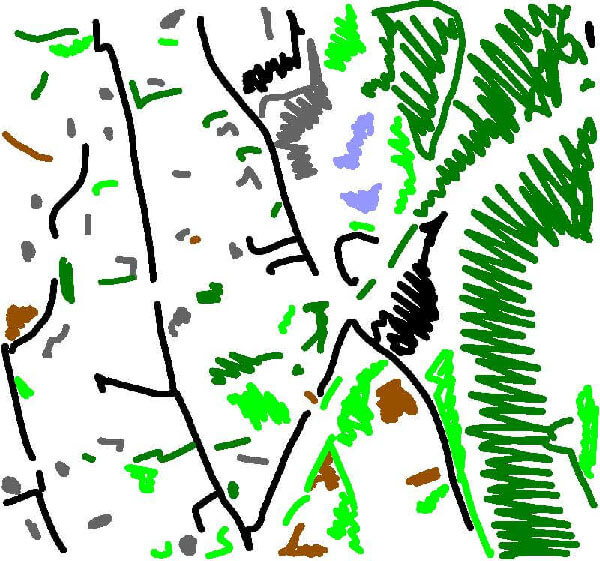} &
\includegraphics[width=.19\textwidth]{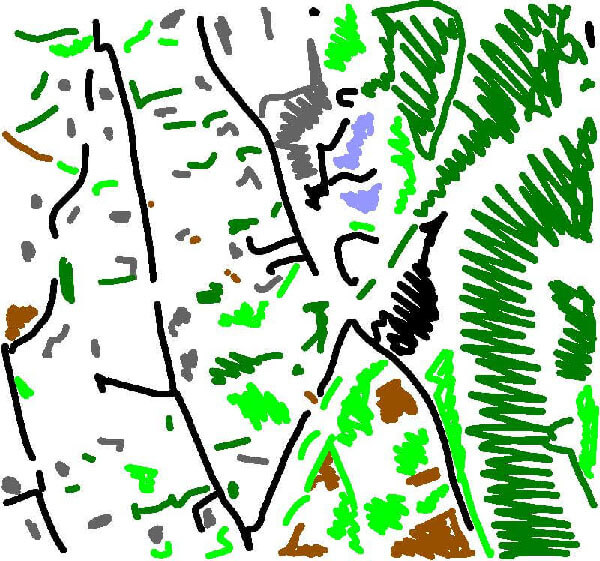}
\\\midrule
\rotatebox{90}{zh18 Results} & 
\includegraphics[width=.19\textwidth]{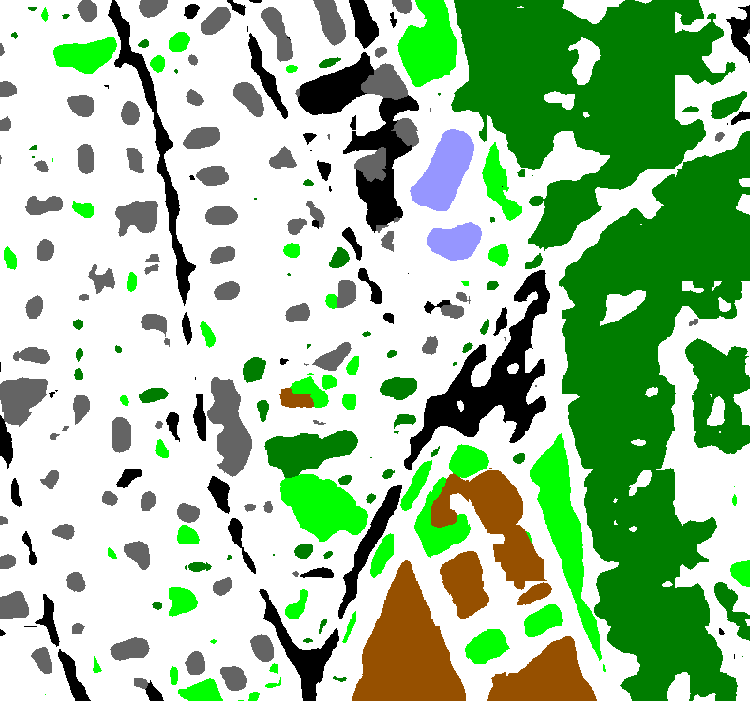} &
\includegraphics[width=.19\textwidth]{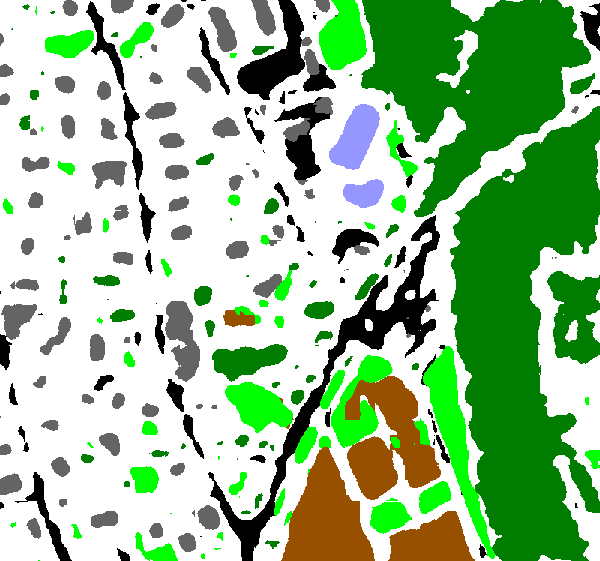} &
\includegraphics[width=.19\textwidth]{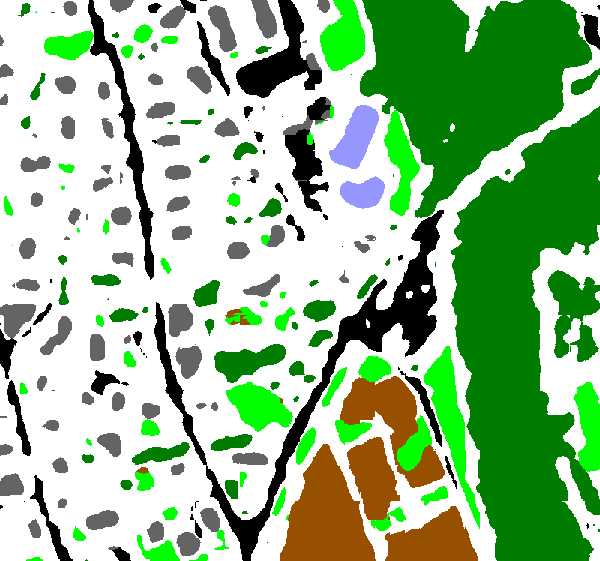} &
\includegraphics[width=.19\textwidth]{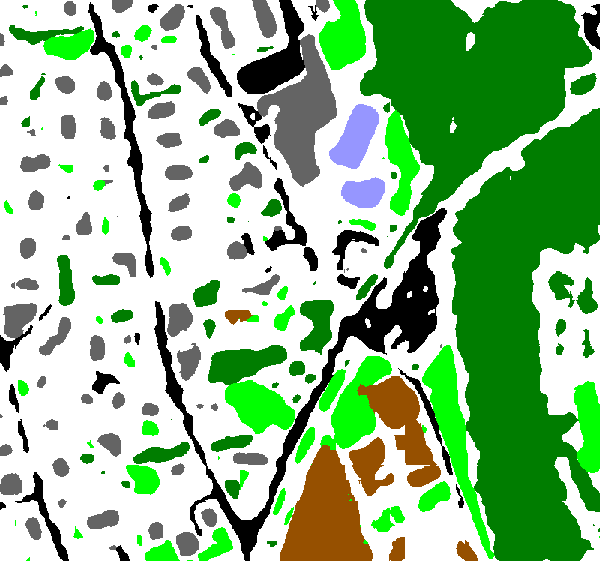} &
\includegraphics[width=.19\textwidth]{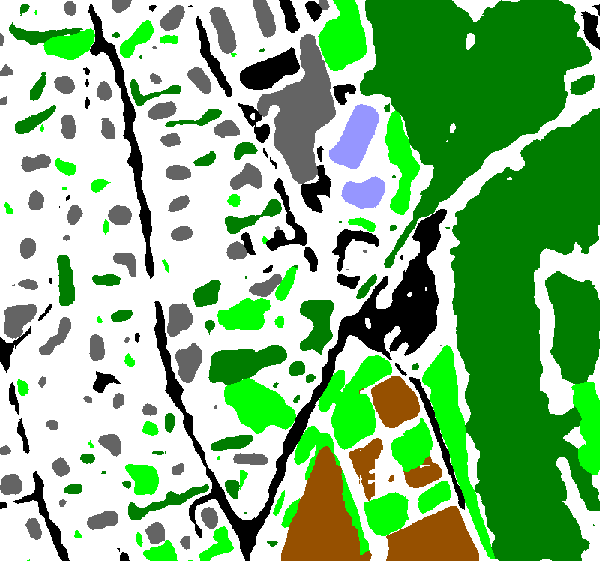} 
\\\bottomrule
\vspace{0.01pt}
\end{tabular}
\addlegendimageintext{fill=black, area legend} Roads
\hspace{3pt}
\addlegendimageintext{fill=zurichgrey, area legend} Buildings
\hspace{3pt}
\addlegendimageintext{fill=zurichdarkgreen, area legend} Trees
\hspace{3pt}
\addlegendimageintext{fill=green, area legend} Grass
\hspace{3pt}
\addlegendimageintext{fill=zurichbrown, area legend} Bare Soil
\hspace{3pt}
\addlegendimageintext{fill=zurichdarkblue, area legend} Water
\hspace{3pt}
\addlegendimageintext{fill=zurichyellow, area legend} Railways
\hspace{3pt}
\addlegendimageintext{fill=zurichlightblue, area legend} Swimming Pools
\caption{Results on the Zurich Summer Dataset \cite{volpi2015semantic} with varying brush strokes. Best viewed in color.}
\label{fig:zurichreduced2}
\end{figure*}

We present several qualitative results of varying the availability of side information on various datasets in Fig.~\ref{fig:urbanreduced}, Fig.~\ref{fig:bachreduced}, Fig.~\ref{fig:zurichreduced1}, and Fig.~\ref{fig:zurichreduced2}. We observe noticeable improvement of segmentation quality when side information is used. For instance, on the zoning dataset shown in Fig.~\ref{fig:urbanreduced}, many regions cannot be identified from satellite imagery. Without using geotagged photos, the baseline Deeplab-ResNet misclassifies the majority of \textit{commercial} regions as \textit{industrial} in SFO. As the amount of side information available increases, the segmentation quality is steadily improved. Similar trends are also found in NYC and BOS.

On the BACH dataset (see Fig.~\ref{fig:bachreduced}), an increased number of brush strokes help to overcome under-segmentation in contiguous regions. Rarer classes such as \textit{benign} in A05 slide and \textit{in situ carcinoma} in A10 slide are more consistently identified with the inclusion of brush strokes.

On the Zurich Summer dataset, as illustrated in Fig.~\ref{fig:zurichreduced1} and Fig.~\ref{fig:zurichreduced2}), the improvement is not as visually obvious as compared with the zoning dataset. This is likely due to the availability of high resolution imagery in the Zurich Summer dataset, which allows the model to make better baseline predictions without side information. However, the inclusion of side information via brush strokes also helps to correct errors made from the initial segmentation. For instance, in \textit{zh5} (see Fig.~\ref{fig:zurichreduced1}), side information helps to correctly identify the tiny \textit{Bare Soil} area. Similarly, in \textit{zh8} (see Fig.~\ref{fig:zurichreduced1}), our method is able to segment the \textit{Railway} class more accurately when provided with side information. We note that these classes are less presented in the dataset, which benefit the most when side information is included.

\subsection{SideInfNet with another CNN Backbone}

In our main paper, we experimented SideInfNet with VGG, the backbone used in the Unified model~\cite{workman2017unified}. In this section, we show in detail the performance of SideInfNet with VGG backbone on all the datasets. In addition, to prove the robustness of our proposed method of multi-modal data fusion over the Unified model~\cite{workman2017unified}, we provide results of SideInfNet and Unified model when the same baseline network is used. In particular, Workman et al.~\cite{workman2017unified} proposed a modified VGG-16 network to extract features from the overhead satellite images, in which feature maps were integrated at the seventh convolutional layer. We re-implemented the same architecture by fusing our constructed feature map at the same layer. In addition, we also re-implemented the Unified model with Deeplab-ResNet, our recommended backbone. We show the comparison results in Table~\ref{table:compareVGG}. As shown in experimental results, in general SideInfNet outperforms the Unified model \cite{workman2017unified} when the same baseline segmentation model is used, highlighting the advantages of our proposed method for multi-modal data fusion.

%In addition to the increase in performance, we also note that the proposed VGG-19 backbone is computationally more efficient than our original Deeplab-ResNet backbone due to the lack of multiscale processing. Hence, SideInfNet-VGG is significantly faster during inference compared to both the Unified and the original SideInfNet models, at the cost of a drop in accuracy compared to the original SideInfNet model.

\begin{table*}[t] % Placed here so it appears in the right page
\caption{Comparison of SideInfNet and Unified model~\cite{workman2017unified} on Deeplab-ResNet and VGG backbone.}
\centering
\centerline{(a) Zoning~\cite{feng2018urban}}
\begin{tabular}{@{}ccccccccc@{}}
\toprule
\multirow{2}{*}{} & \multicolumn{4}{c}{\textbf{mIOU}} & \multicolumn{4}{c}{\textbf{Pixel Accuracy}} \\ \cmidrule(l){2-9}
 & \textbf{BOS} & \textbf{NYC} & \textbf{SFO} & \textbf{Average} & \textbf{BOS} & \textbf{NYC} & \textbf{SFO} & \textbf{Average} \\
\cmidrule{1-9}
SideInfNet/DRN$^{\mathrm{*}}$ & 41.96\% & 39.59\% & 60.31\% & 47.29\% & 71.33\% & 71.08\% & 79.59\% & 74.00\% \\
SideInfNet/VGG & 41.94\% & 39.68\% & 56.73\% & 46.12\% & 68.28\% & 68.06\% & 75.95\% & 70.06\% \\
Unified~\cite{workman2017unified}/DRN$^{\mathrm{*}}$ & 37.61\% & 36.71\% & 57.31\% & 47.46\% & 66.87\% & 68.77\% & 77.96\% & 72.42\% \\
Unified~\cite{workman2017unified}/VGG & 40.51\% & 39.27\% & 55.36\% & 45.05\% & 67.91\% & 70.92\% & 75.92\% & 71.58\% \\
\bottomrule
\multicolumn{5}{l}{$^{\mathrm{*}}$ DRN: Deeplab-ResNet}\\
\end{tabular}

%\centerline{(a) Zoning dataset~\cite{feng2018urban}}
%\begin{tabular}{@{}ccccc@{}}
%\toprule
%\multirow{2}{*}{} & \multicolumn{4}{c}{\textbf{mIOU}} \\ \cmidrule(l){2-5}
% & \textbf{BOS} & \textbf{NYC} & \textbf{SFO} & \textbf{Average} \\
%\cline{1-5}
%SideInfNet (Deeplab-ResNet) & 41.96\% & 39.59\% & 60.31\% & 47.29\% \\
%SideInfNet (VGG) & 41.94\% & 39.68\% & 56.73\% & 46.12\% \\
%Unified model~\cite{workman2017unified} (Deeplab-ResNet) & 37.61\% & 36.71\% & 57.31\% & 47.46\% \\
%Unified model~\cite{workman2017unified} (VGG) & 40.51\% & 39.27\% & 55.36\% & 45.05\% \\
%\bottomrule \\
%\end{tabular}

%\centerline{(b) BACH dataset~\cite{aresta2019bach}}
%\begin{tabular}{@{}cc@{}}
%\toprule
%& \textbf{mIOU} \\ \cline{1-2}
%SideInfNet (Deeplab-ResNet) & 47.24\% \\
%SideInfNet (VGG) & 49.53\% \\
%Unified model~\cite{workman2017unified} (Deeplab-ResNet) & 34.66\% \\
%Unified model~\cite{workman2017unified} (VGG) & 29.37\% \\
%\bottomrule \\
%\end{tabular}

\centerline{(b) BACH~\cite{aresta2019bach}}
\begin{tabular}{@{}ccccccc@{}}
\toprule
\multirow{2}{*}{} & \multicolumn{3}{c}{\textbf{mIOU}} & \multicolumn{3}{c}{\textbf{Pixel Accuracy}} \\ \cmidrule(l){2-7}
 & \textbf{A05} & \textbf{A10} & \textbf{Average} & \textbf{A05} & \textbf{A10} & \textbf{Average} \\
\cmidrule{1-7}
SideInfNet/Deeplab-ResNet & 59.03\% & 35.45\% & 47.24\% & 89.68\% & 54.29\% & 71.99\% \\
SideInfNet/VGG & 66.34\% & 32.73\% & 49.53\% & 89.60\% & 46.50\% & 68.05\% \\
Unified\cite{workman2017unified}/Deeplab-ResNet & 47.94\% & 21.37\% & 34.66\% & 89.54\% & 40.42\% & 64.98\% \\
Unified\cite{workman2017unified}/VGG & 41.50\% & 17.23\% & 29.37\% & 91.38\% & 54.87\% & 73.12\% \\
\bottomrule \\
\end{tabular}

\centerline{(c) Zurich~\cite{volpi2015semantic}}
%\begin{tabular}{@{}cc@{}}
%\toprule
%& \textbf{mIOU} \\ \cline{1-2}
%SideInfNet (Deeplab-ResNet) & 58.31\% \\
%SideInfNet (VGG) & 49.73\% \\
%Unified model~\cite{workman2017unified} (Deeplab-ResNet) & 46.83\% \\
%Unified model~\cite{workman2017unified} (VGG) & 42.09\% \\
%\bottomrule

\begin{tabular}{@{}ccc@{}}
\toprule
& \textbf{mIOU} & \textbf{Pixel Accuracy} \\
\cmidrule{1-3}
SideInfNet/Deeplab-ResNet & 58.31\% & 78.97\% \\
SideInfNet/VGG & 49.73\% & 77.74\% \\
Unified\cite{workman2017unified}/Deeplab-ResNet & 46.83\% & 74.26\% \\
Unified\cite{workman2017unified}/VGG & 42.09\% & 68.20\% \\
\bottomrule
\end{tabular}
\label{table:compareVGG}
\end{table*}

\section{Computational Analysis}
\label{sec:computationalanalysis}

An additional advantage of our method is its computational efficiency, which comes into play with high density annotations. Specifically, the BACH dataset consists of very high resolution whole slide images, which is common in many medical datasets. Coupled with dense brush stroke annotations, this results in significant bottlenecks for prior works, e.g., the Unified model~\cite{workman2017unified}.

In order to evaluate the computational complexity quantitatively, we benchmark the inference speeds of the Deeplab-ResNet~\cite{chen2018deeplab}, Unified model~\cite{workman2017unified}, and our SideInfNet. As the HO-MRF model requires an additional post-processing step in the form of global normalization, we do not compare against it in this experiment. Evaluation results are averaged across the inference speeds over single patches (i.e., batch size of 1). However, in practice the method can be sped up with batch based processing. For instance, with a batch size of 64, SideInfNet averages 0.057s per patch on the BACH dataset.

The results are summarized in Table~\ref{table:speedbenchmark}. We observe that on datasets with smaller resolution images and sparser side information (e.g., the Zurich Summer dataset), the Unified model performs faster than SideInfNet. This is likely due to the multi-scale architecture of the Deeplab-ResNet, which increases the computational load as multiple images have to be processed. However, as we scale up to larger resolution images and denser side information, our method is much more efficient than the Unified model. In particular, on the BACH dataset which contains high resolution imagery and dense brush stroke annotations, we obtain approximately a 16 times speedup over the Unified model. This supports our hypothesis that on top of improved accuracy, SideInfNet is able to scale more efficiently to higher resolution images and denser side information.

%%%%%
\begin{table*}[t] % Placed here so it appears in the right page
\caption{Computational analysis performed on an NVIDIA Pascal Titan X GPU.}
\begin{center}
\begin{tabular}{ccccccc}
\toprule
\textbf{Approach} & \multicolumn{3}{c}{\textbf{Time (s)}} & \multicolumn{3}{c}{\textbf{GPU Memory (MB)}} \\
\cmidrule{2-7}
 & Zoning & BACH & Zurich Summer & Zoning & BACH & Zurich Summer \\
\cmidrule{1-7}
Deeplab-ResNet \cite{chen2018deeplab} & 0.047 & 0.101 & 0.048 & 779 & 821 & 781 \\
Unified model$^{\mathrm{*}}$ \cite{workman2017unified} & 0.034 & 2.003 & 0.062 & 739 & 1843 & 725 \\
SideInfNet & 0.105 & 0.121 & 0.139 & 783 & 857 & 785 \\
\bottomrule
\multicolumn{7}{l}{$^{\mathrm{*}}$ Our implementation.}
\end{tabular}
\label{table:speedbenchmark}
\end{center}
\end{table*}
%%%%%

\section{Additional Qualitative Evaluations}
\label{sec:qualitativeevalution}

\subsection{Qualitative Results on BACH dataset} 
\label{sec:bachresults}

\begin{figure*}[t]
% \centering
% \scalebox{0.95}{
\begin{center}
\begin{tabular}{l@{\ }l@{\ }c@{\ }c@{\ }c@{\ }c@{\ }c}
% \hline
 \toprule
\multicolumn{1}{c}{}
& \multicolumn{1}{c}{Brush Strokes}
& \multicolumn{1}{c}{Deeplab-ResNet \cite{chen2018deeplab}}
& \multicolumn{1}{c}{Unified \cite{workman2017unified}}
& \multicolumn{1}{c}{SideInfNet}
& \multicolumn{1}{c}{Groundtruth}
\\\midrule
\rotatebox{90}{A05} & 
\includegraphics[width=.19\textwidth]{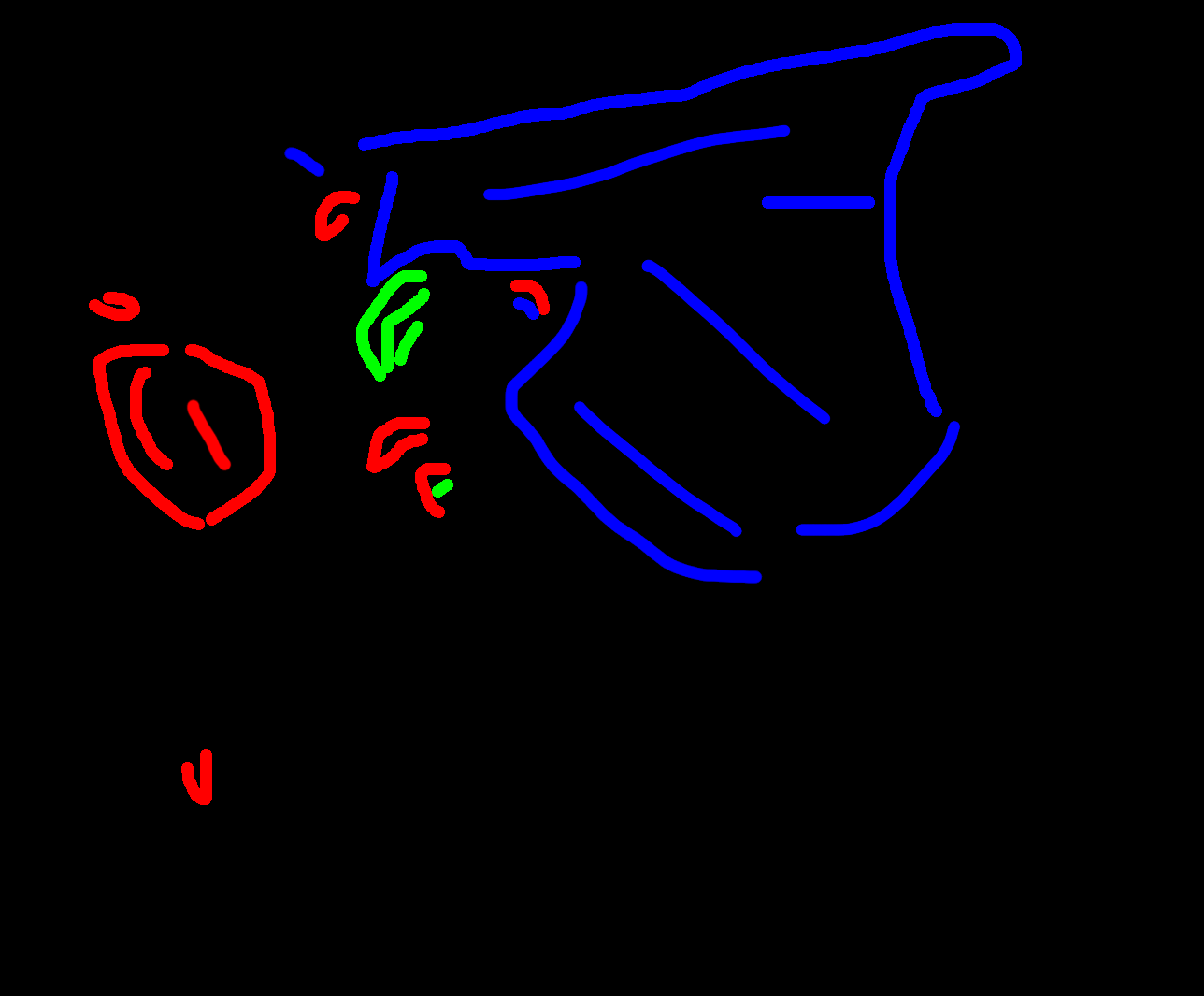} &
\includegraphics[width=.19\textwidth]{images/bach_inference/baseline/A05.png} &
\includegraphics[width=.19\textwidth]{images/bach_inference/unified/A05.png} &
\includegraphics[width=.19\textwidth]{images/bach_inference/adaptive/A05.png} &
\includegraphics[width=.19\textwidth]{images/bach_inference/gt/A05.png}
\\\midrule
\rotatebox{90}{A10} & 
\includegraphics[width=.19\textwidth]{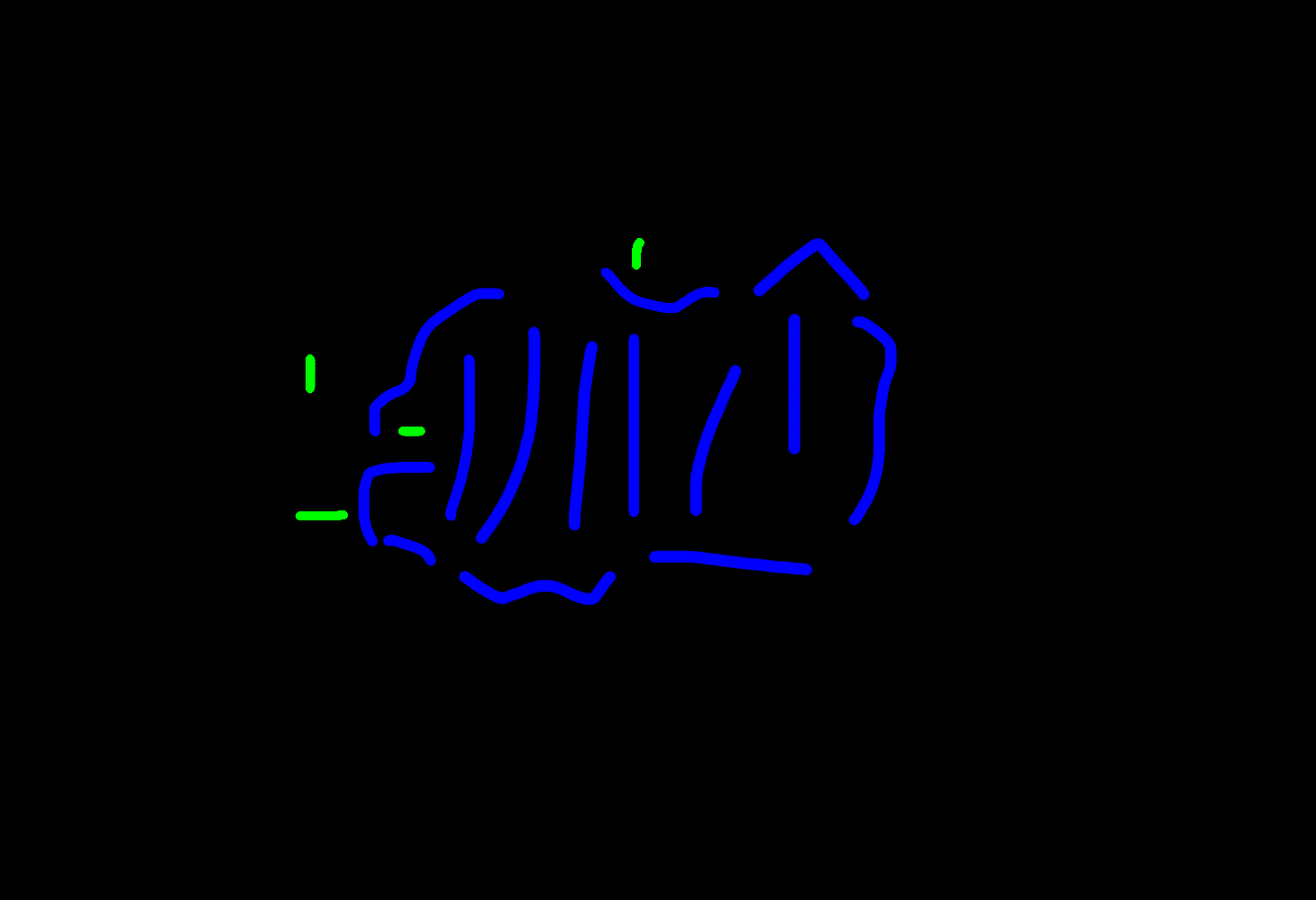} &
\includegraphics[width=.19\textwidth]{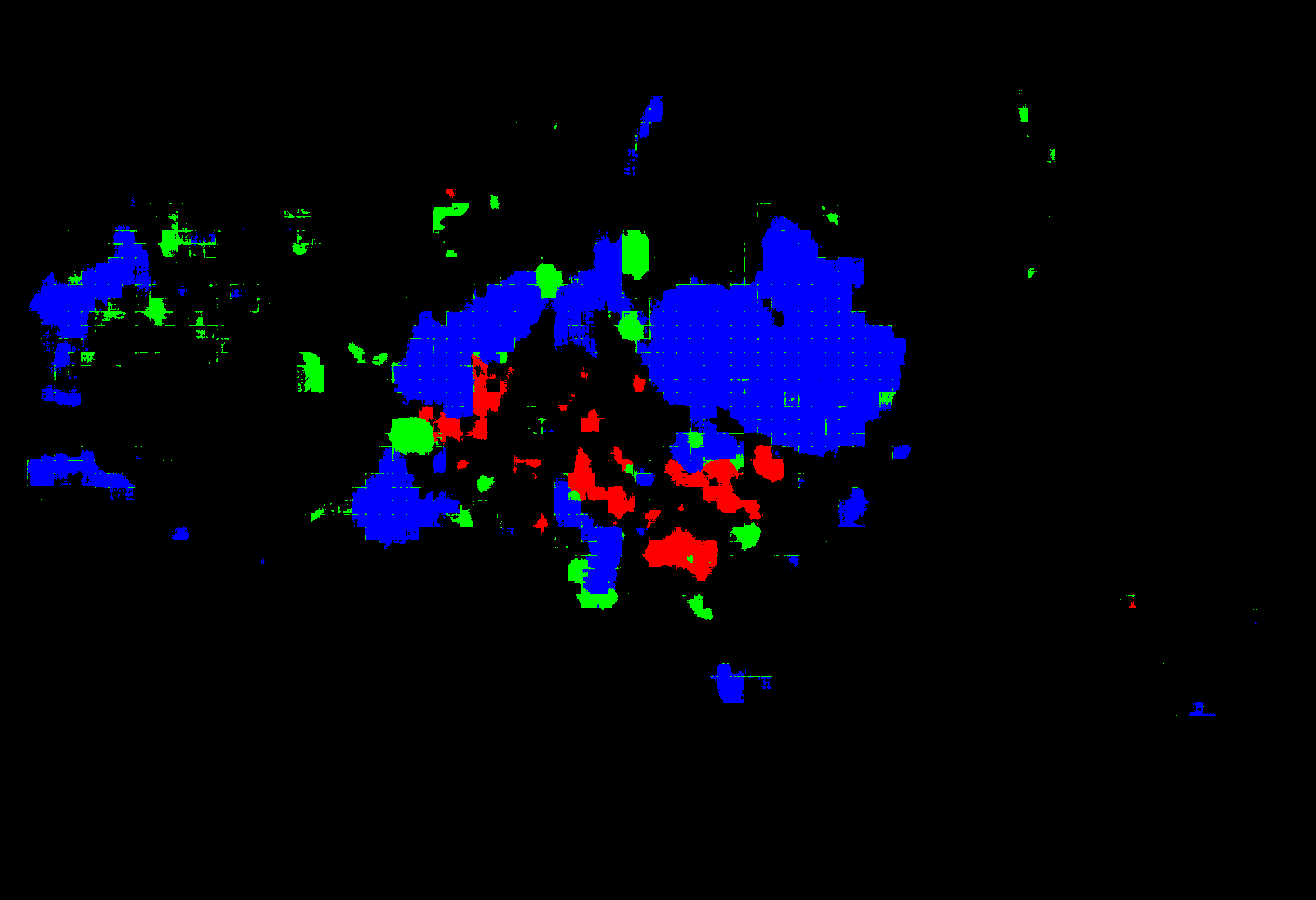} &
\includegraphics[width=.19\textwidth]{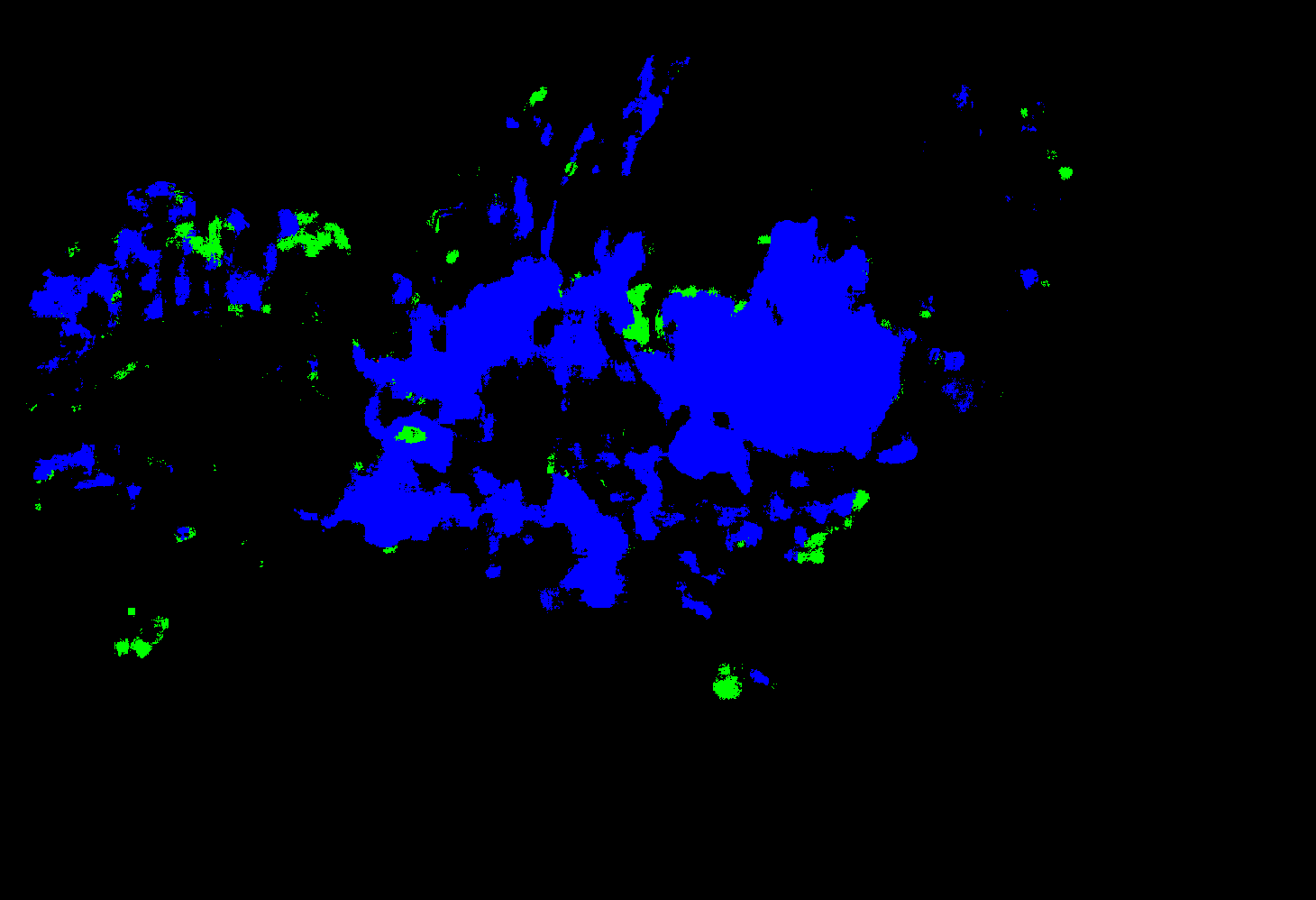} &
\includegraphics[width=.19\textwidth]{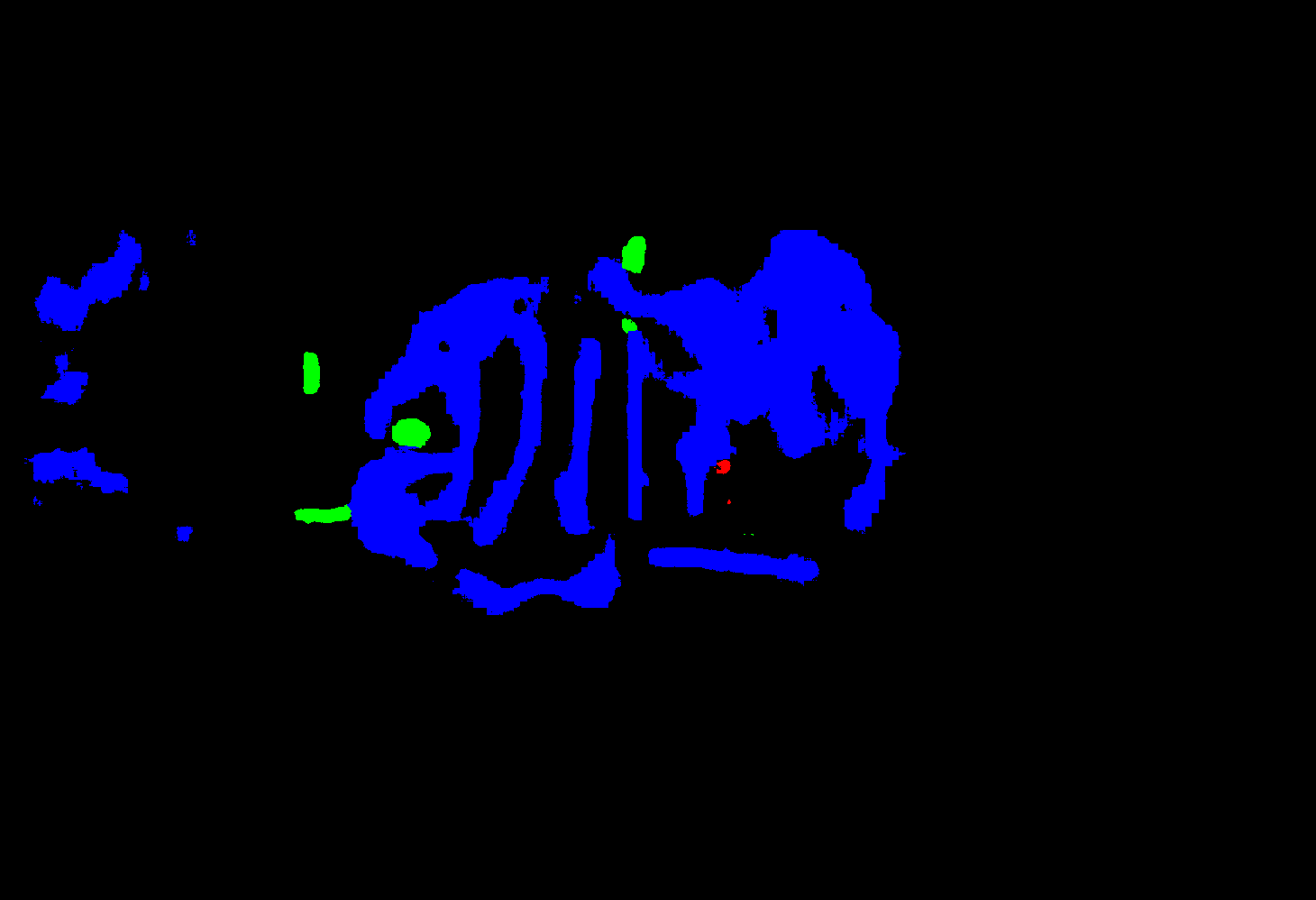} &
\includegraphics[width=.19\textwidth]{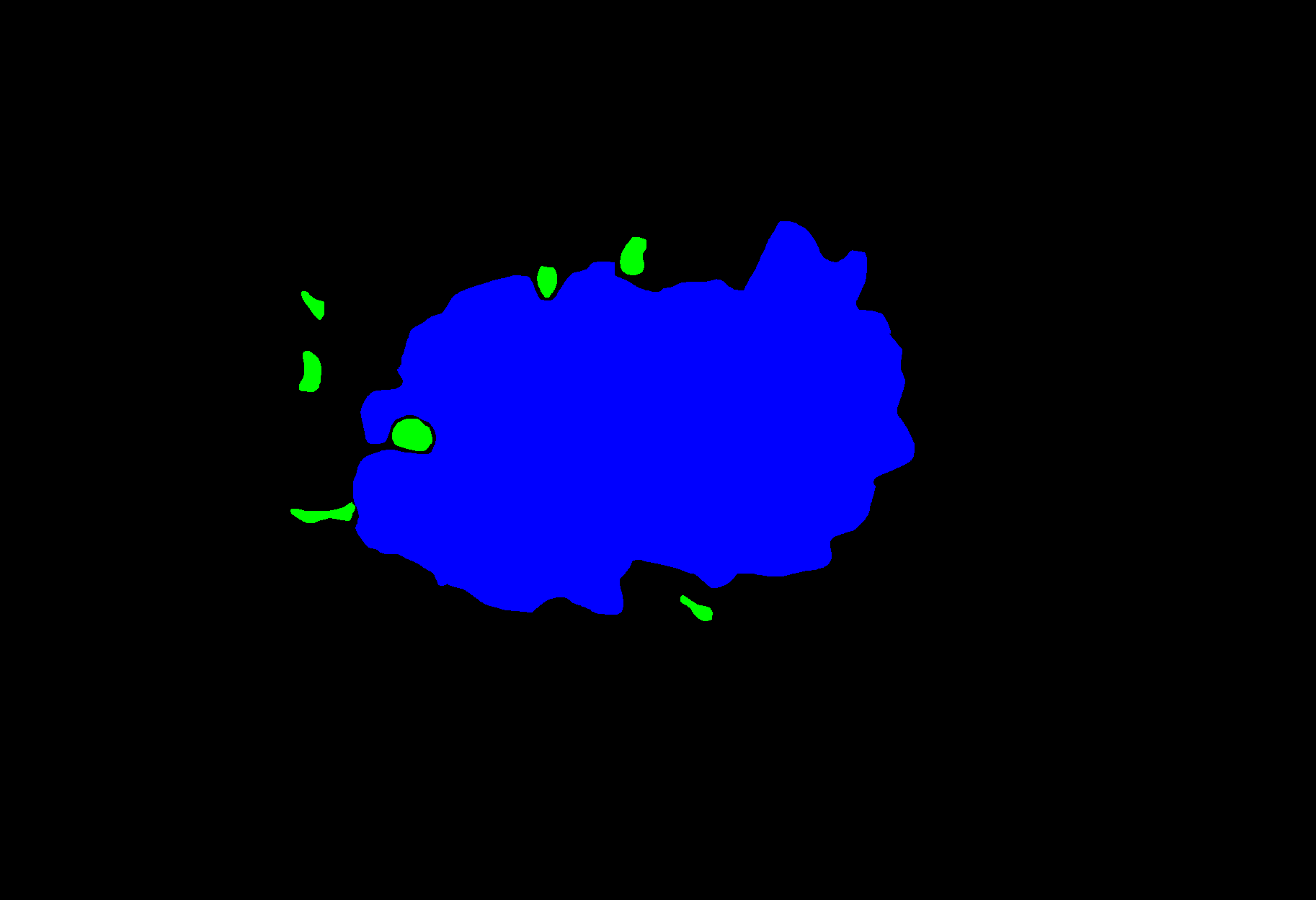}
\\\bottomrule
\vspace{0.01pt}
\end{tabular}
\addlegendimageintext{fill=red, area legend} Benign
\hspace{3pt}
\addlegendimageintext{fill=blue, area legend} Invasive carcinoma
\hspace{3pt}
\addlegendimageintext{fill=green, area legend} In situ carcinoma
\end{center} 
\caption{Qualitative results on the BACH dataset \cite{aresta2019bach}. Best viewed in color.}
\label{fig:bachinference}
\end{figure*}

Several qualitative results of our method on the BACH dataset are as shown in Fig.~\ref{fig:bachinference}. From the results presented, we observe that, compared with other methods, SideInfNet generally provides the highest quality results. The segmentation masks produced by SideInfNet are less noisy and sparse. In addition, compared with prior works, SideInfNet significantly produces less false positives.
 
%These predictions may require further expert intervention to resolve in a production environment, losing any efficiency gains from a semi-automatic inference process.

A common challenge for SideInfNet and Unified model is the spaces demarcated by brush strokes, leading to segmentation results that only contain shape outlines, such as the circular object in the A05 slide. A possible solution to this issue could be to perform global post-processing, e.g., by applying CRFs~\cite{krahenbuhl2011efficient} or HO-MRFs~\cite{feng2018urban}. However, these post-processing steps are computationally expensive and thus may not be feasible for high-resolution imagery data, e.g., the BACH images.

%In the case of BACH, this is likely not computationally feasible due to the high resolution imagery.

An alternative solution is applying manual post-processing. The refined results produced by SideInfNet allow these gaps to be easily filled in by users. These results suggest the viability of SideInfNet as a semi-automatic semantic segmentation tool.

\subsection{Qualitative Results on Zurich Summer Dataset} 
\label{sec:zurichresults}

\begin{figure*}[t]
% \centering
% \scalebox{0.95}{
\begin{center}
\begin{tabular}{l@{\ }l@{\ }l@{\ }c@{\ }c@{\ }c@{\ }c}
% \hline
 \toprule
\multicolumn{1}{c}{}
& \multicolumn{1}{c}{Brush Strokes}
& \multicolumn{1}{c}{Deeplab-ResNet \cite{chen2018deeplab}}
& \multicolumn{1}{c}{Unified \cite{workman2017unified}}
& \multicolumn{1}{c}{SideInfNet}
& \multicolumn{1}{c}{Groundtruth}
\\\midrule
\rotatebox{90}{zh5} & 
\includegraphics[width=.19\textwidth]{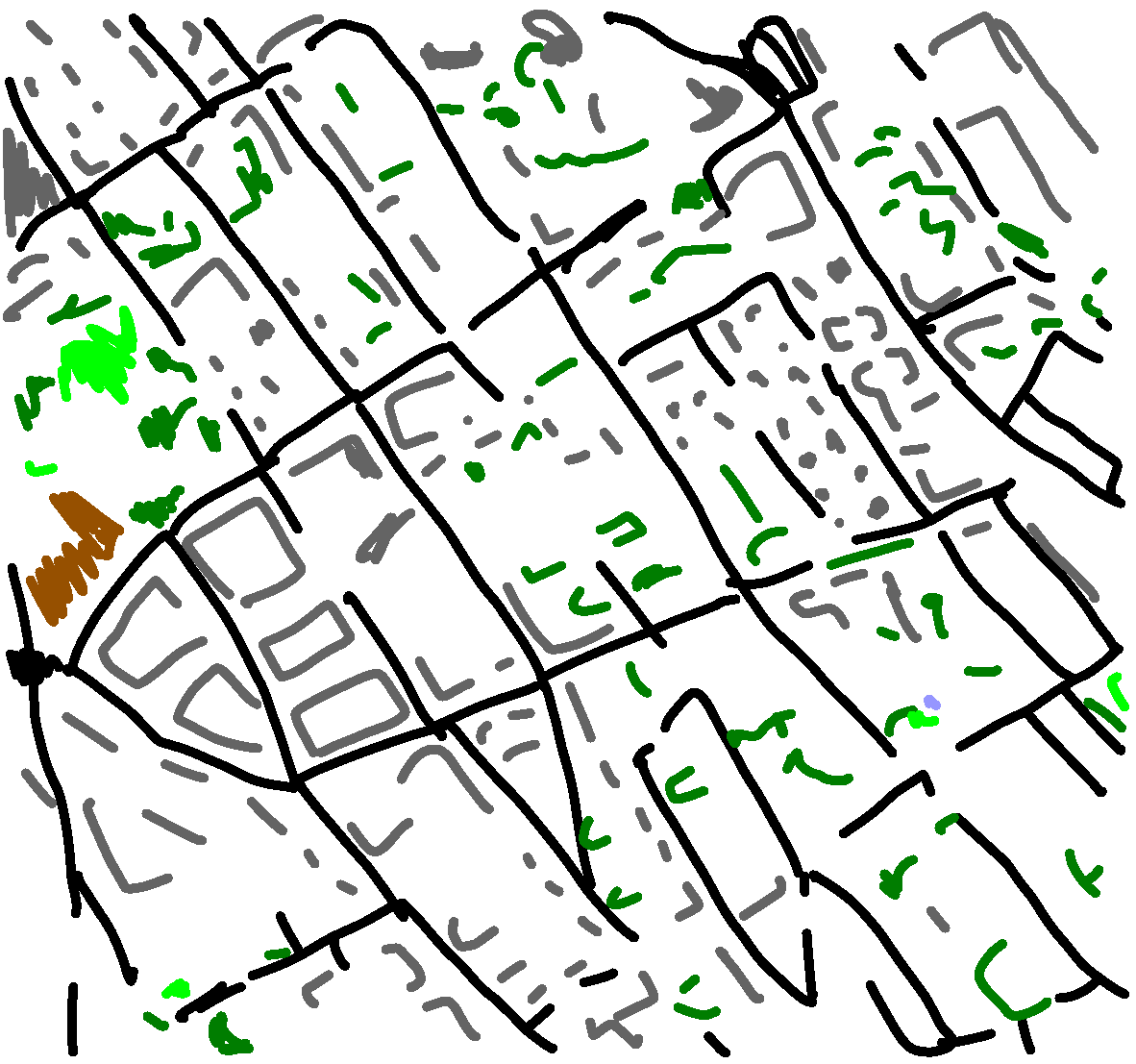} &
\includegraphics[width=.19\textwidth]{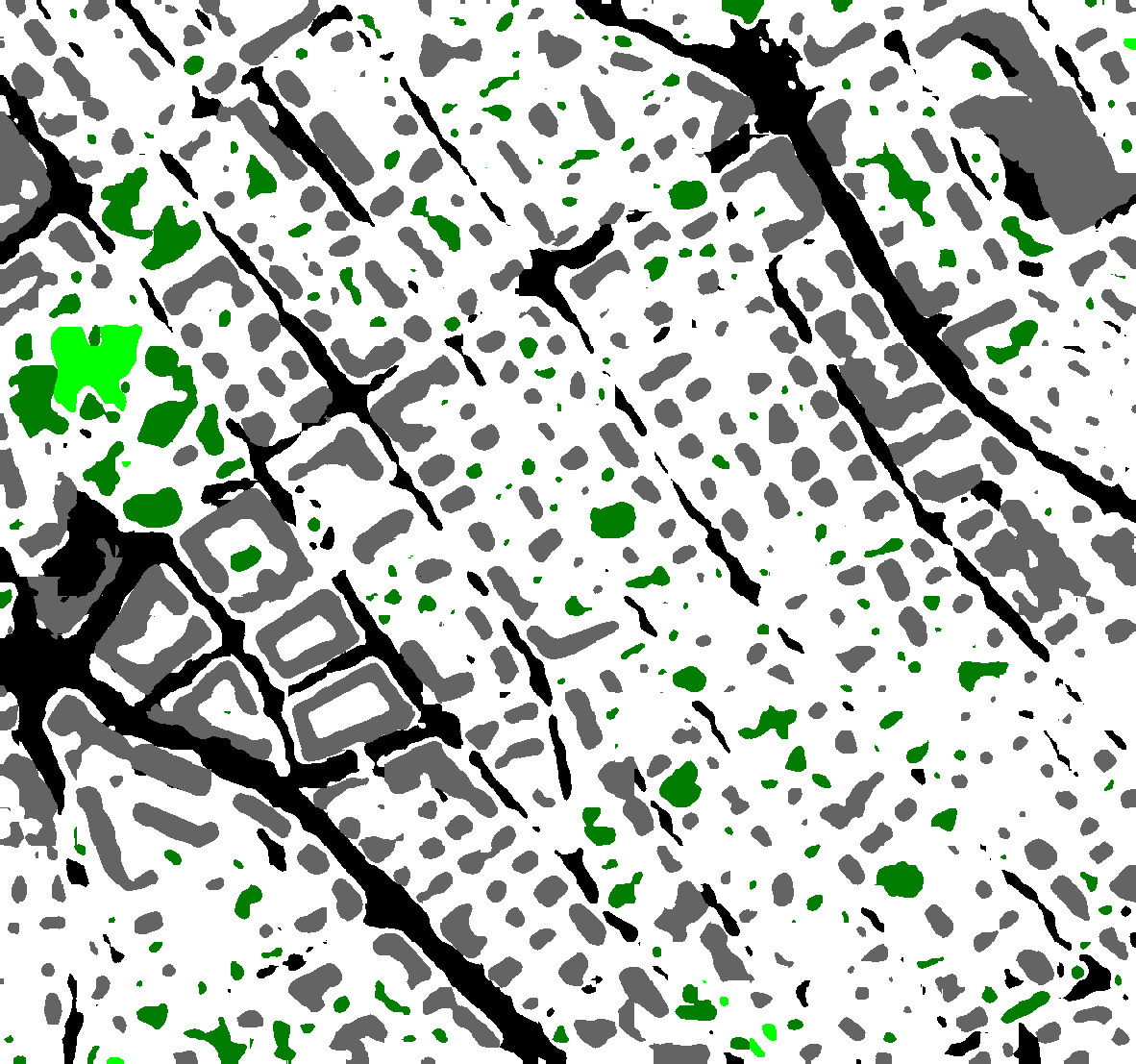} &
\includegraphics[width=.19\textwidth]{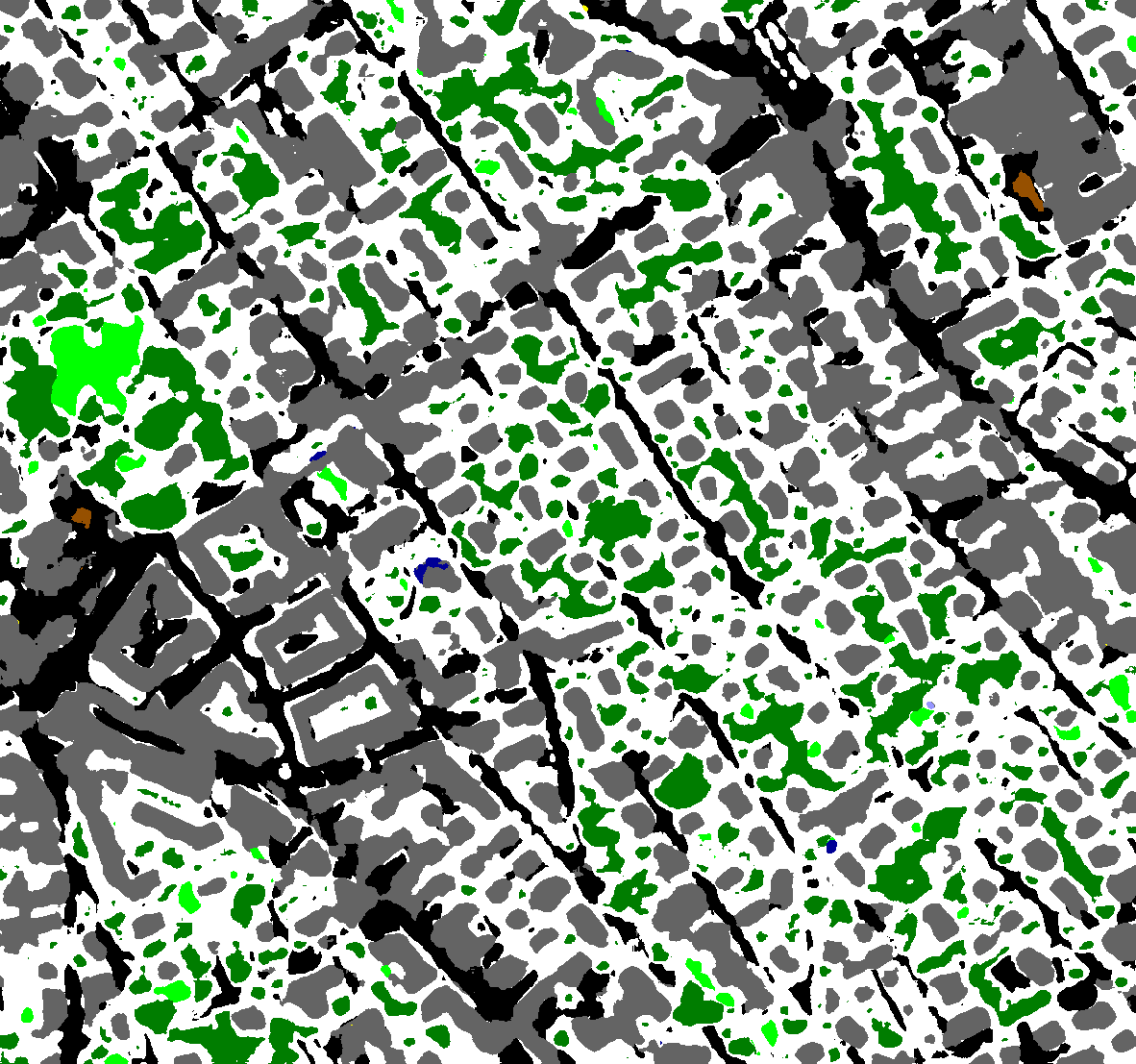} &
\includegraphics[width=.19\textwidth]{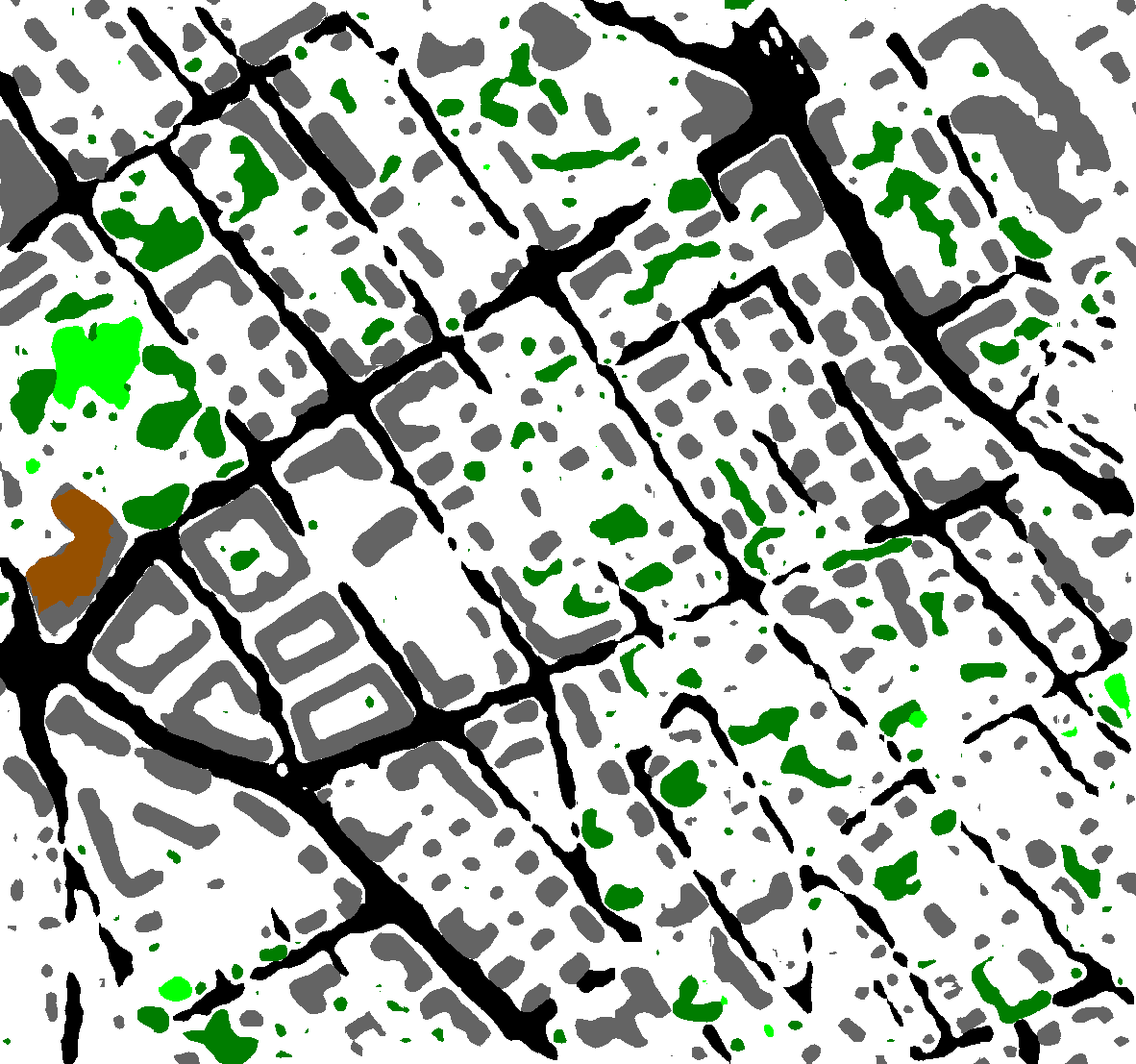} &
\includegraphics[width=.19\textwidth]{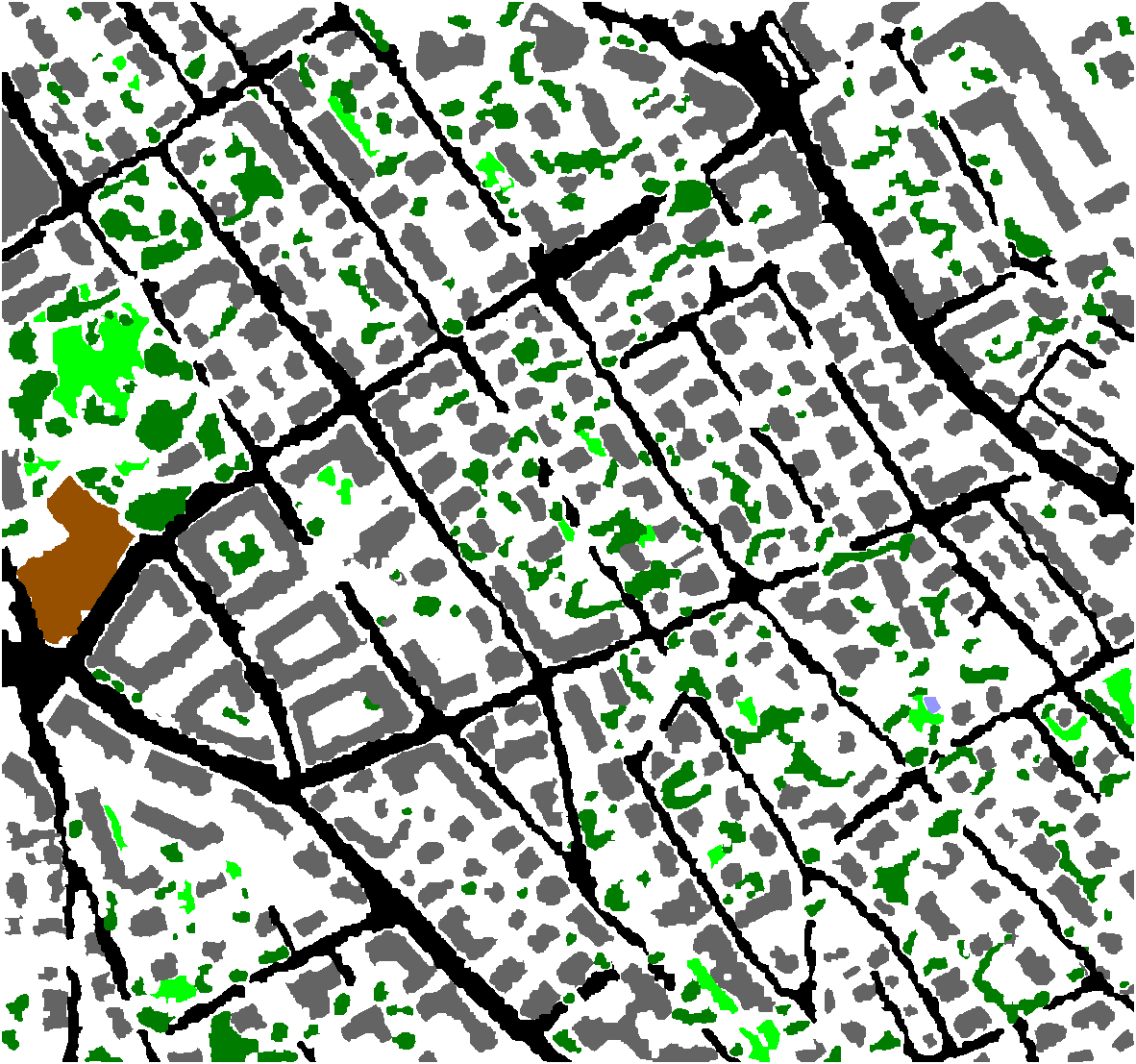}
\\\midrule
\rotatebox{90}{zh7} & 
\includegraphics[width=.19\textwidth]{images/zurich_inference/brush/zurich_brush_zh7.png} &
\includegraphics[width=.19\textwidth]{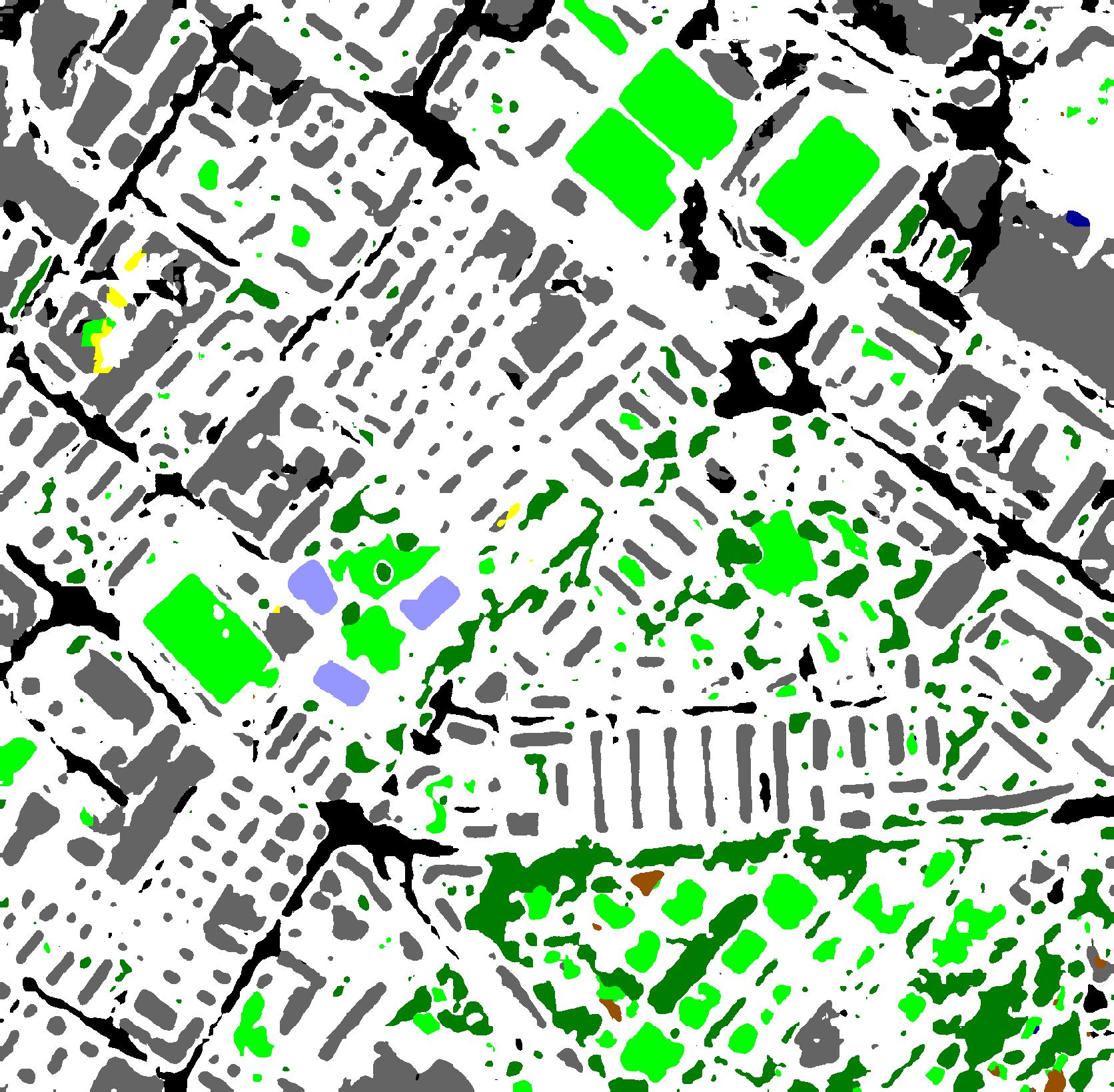} &
\includegraphics[width=.19\textwidth]{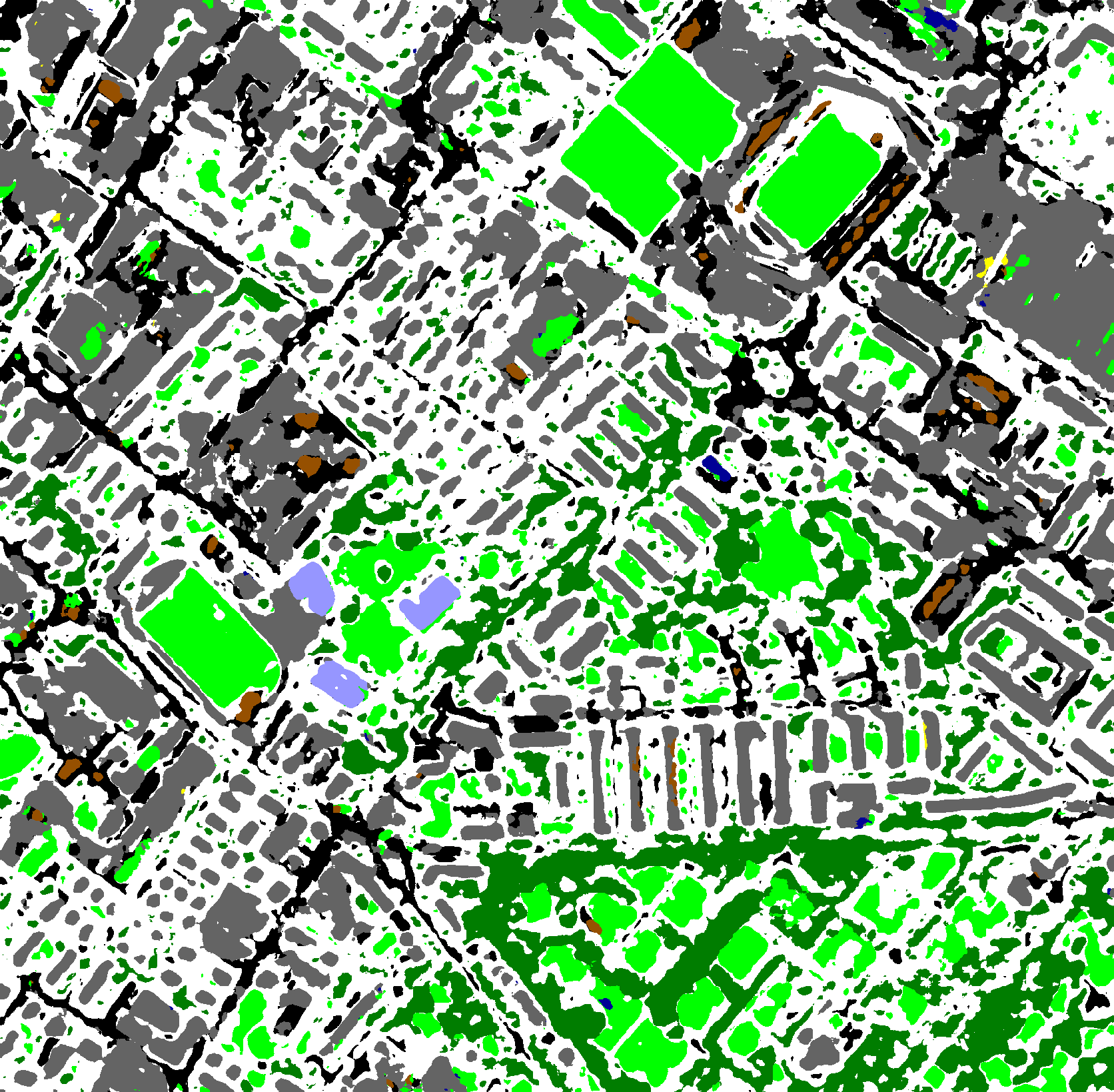} &
\includegraphics[width=.19\textwidth]{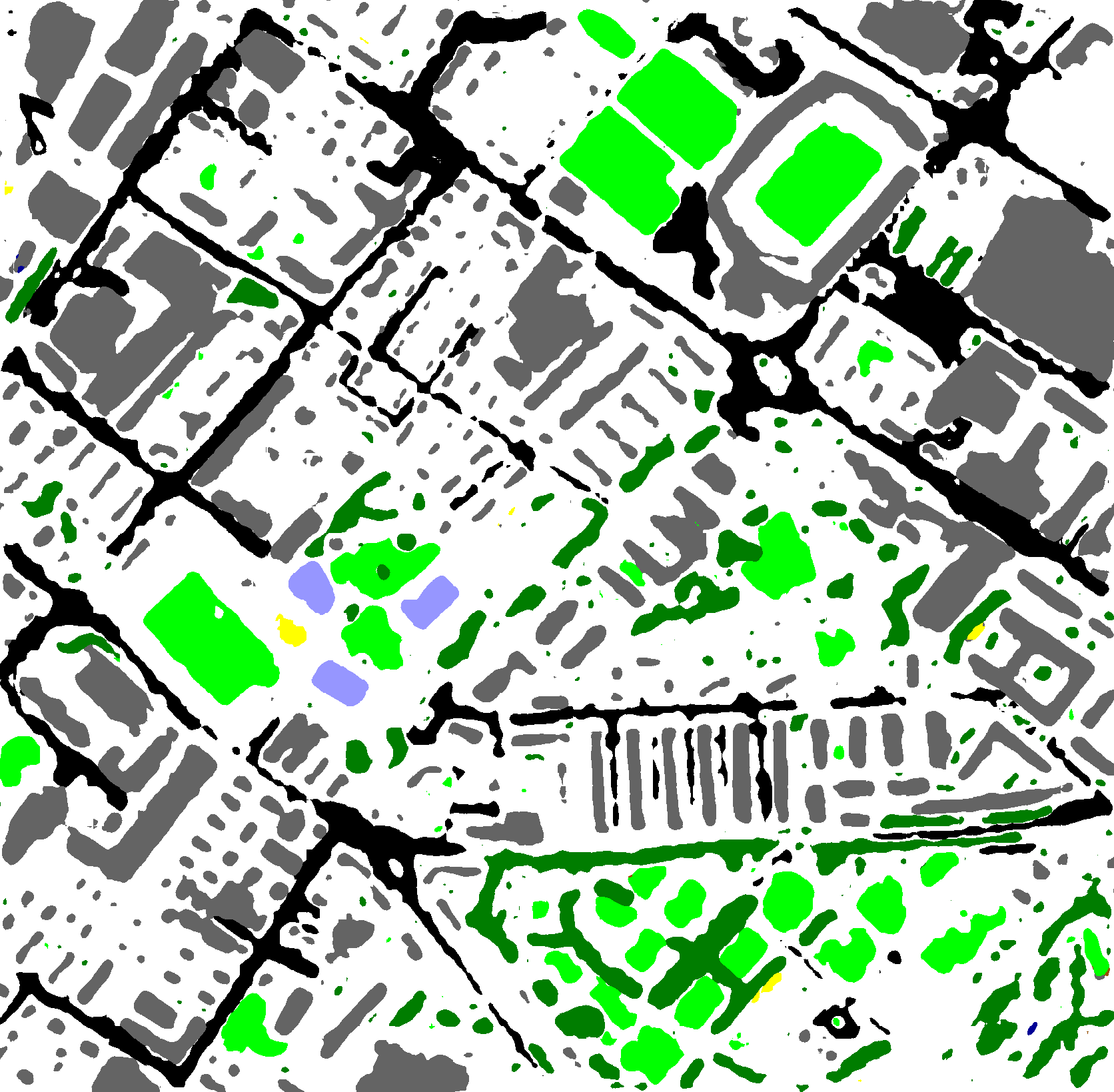} &
\includegraphics[width=.19\textwidth]{images/zurich_inference/gt/zurich_gt_zh7.png}
\\\midrule
\rotatebox{90}{zh8} & 
\includegraphics[width=.19\textwidth]{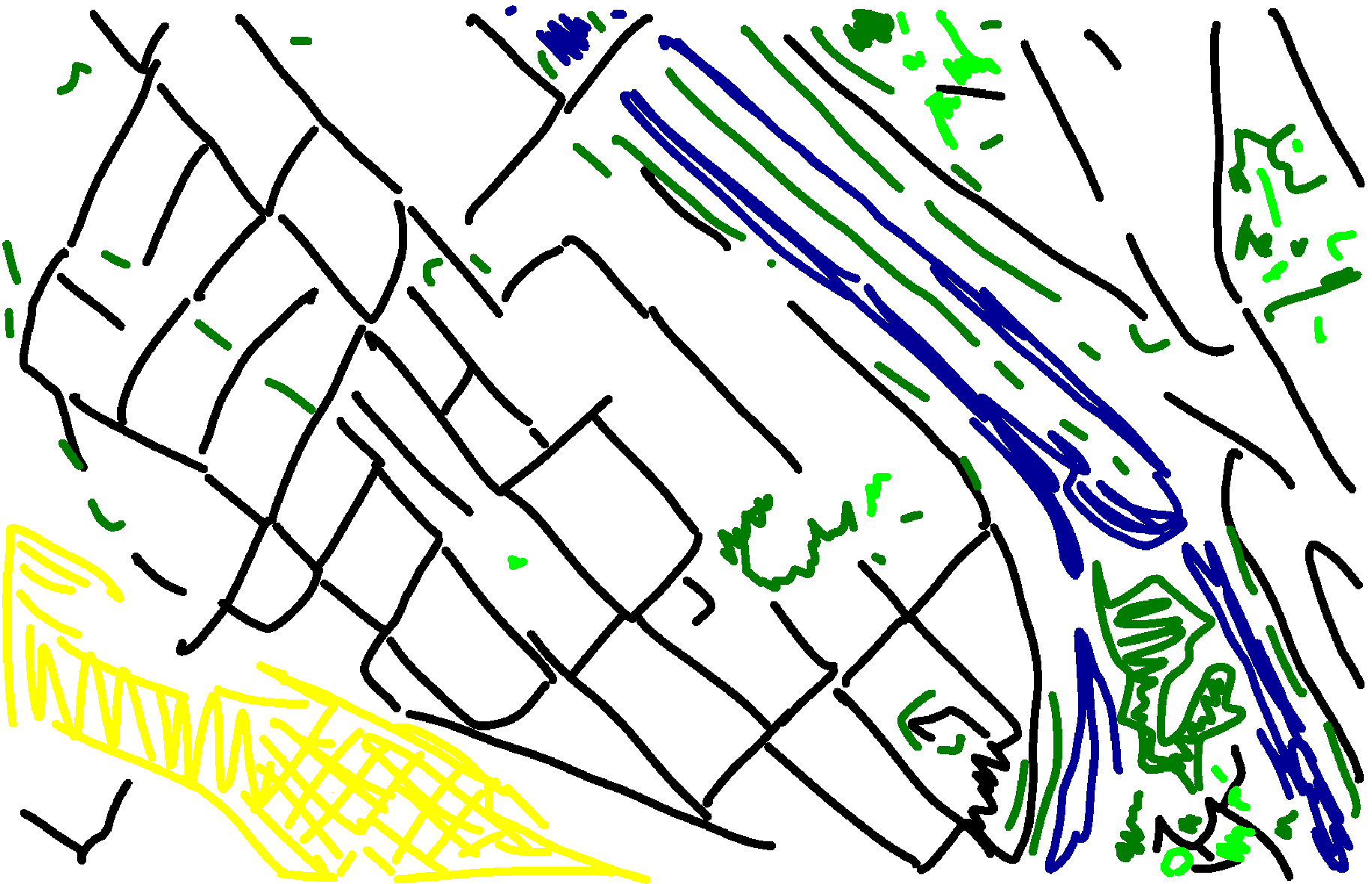} &
\includegraphics[width=.19\textwidth]{images/zurich_inference/baseline/zurich_baseline_zh8.png} &
\includegraphics[width=.19\textwidth]{images/zurich_inference/unified/zurich_unified_zh8.png} &
\includegraphics[width=.19\textwidth]{images/zurich_inference/adaptive/zurich_adaptive_zh8.png} &
\includegraphics[width=.19\textwidth]{images/zurich_inference/gt/zurich_gt_zh8.png}
\\\midrule
\rotatebox{90}{zh11} & 
\includegraphics[width=.19\textwidth]{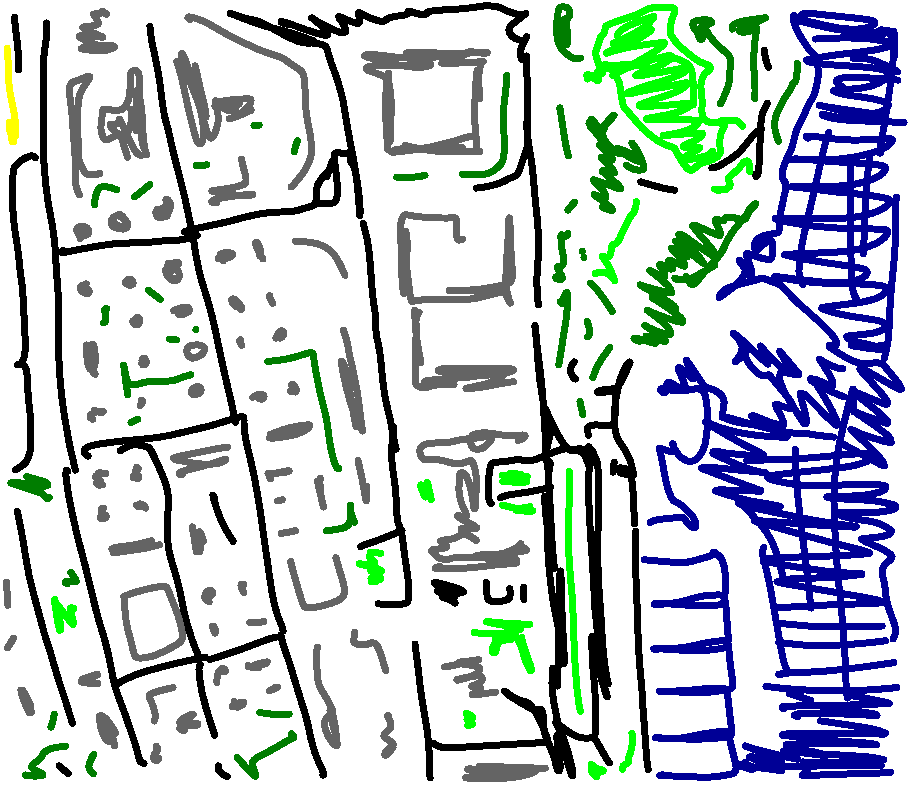} &
\includegraphics[width=.19\textwidth]{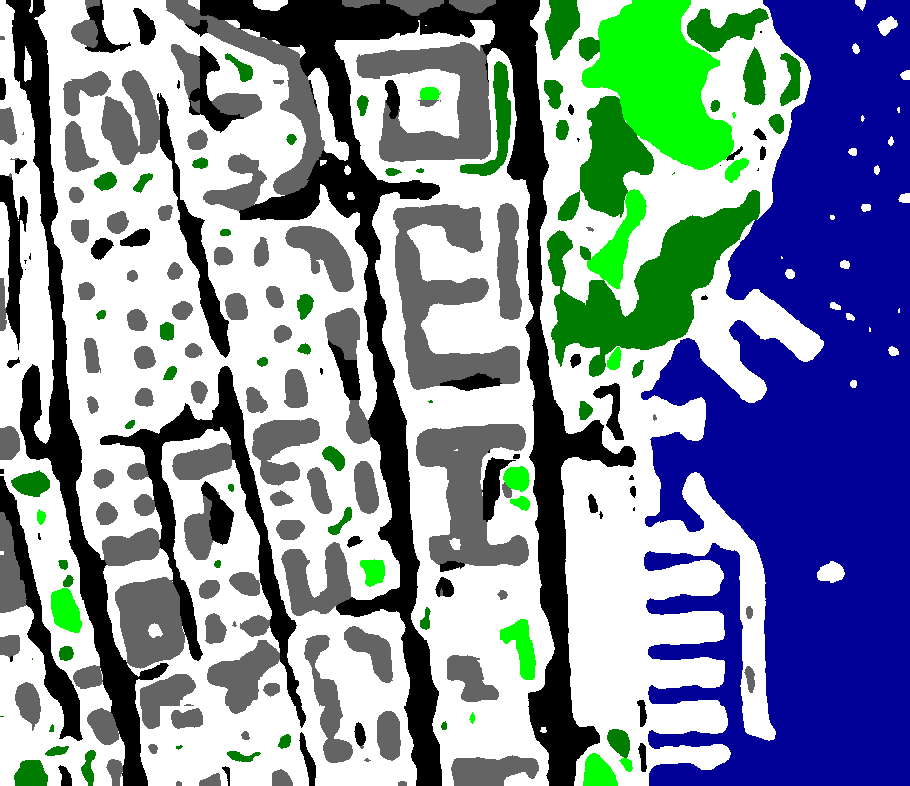} &
\includegraphics[width=.19\textwidth]{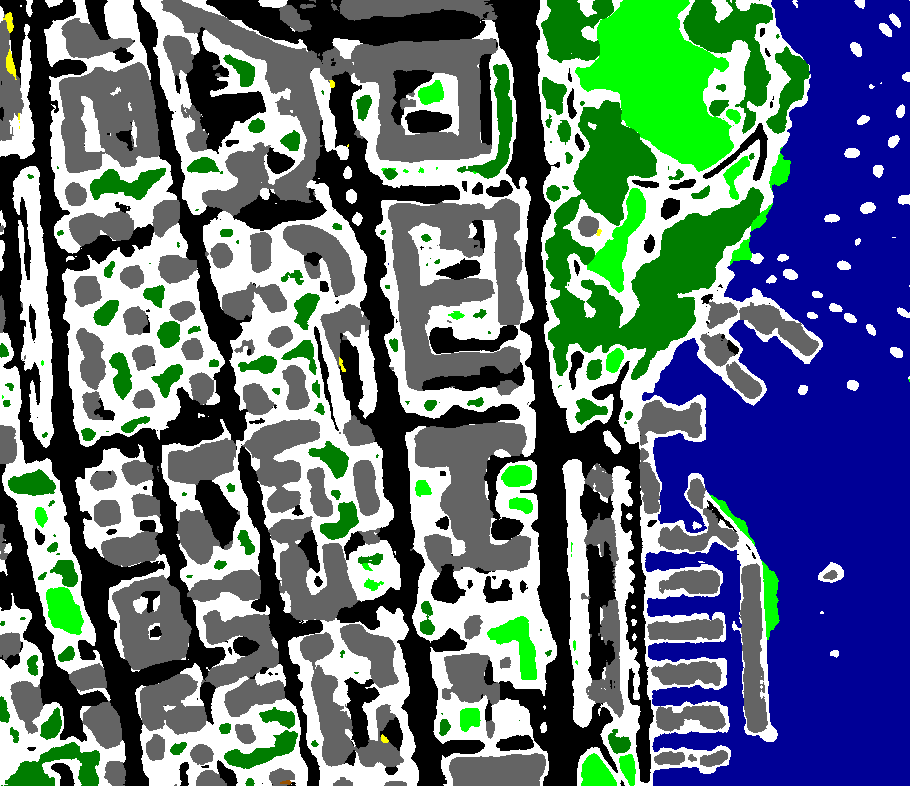} &
\includegraphics[width=.19\textwidth]{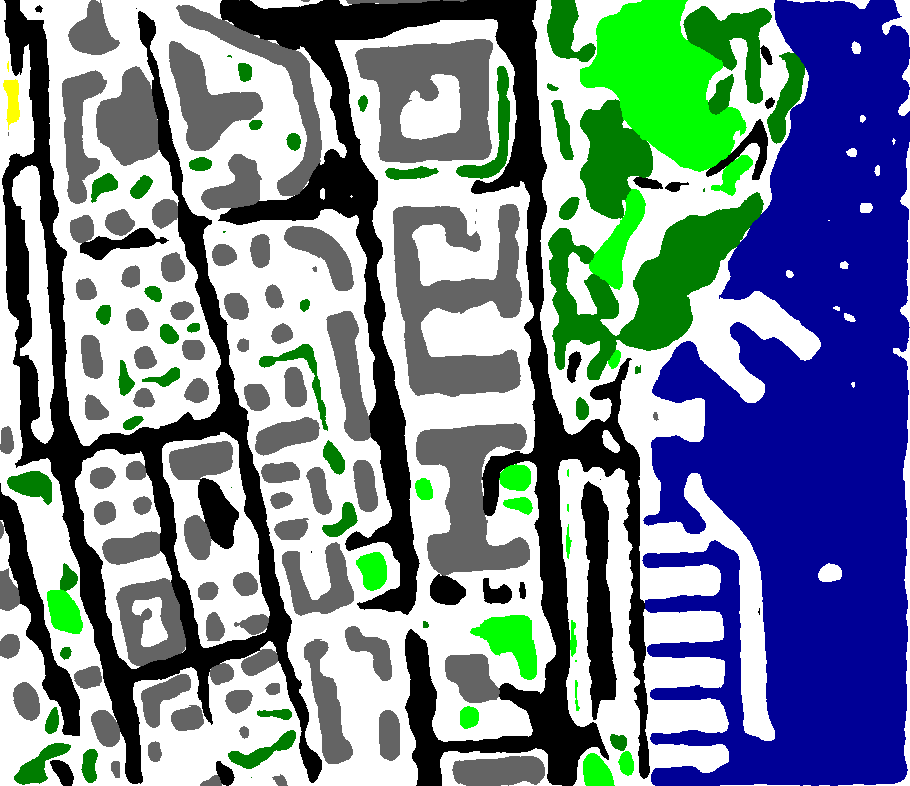} &
\includegraphics[width=.19\textwidth]{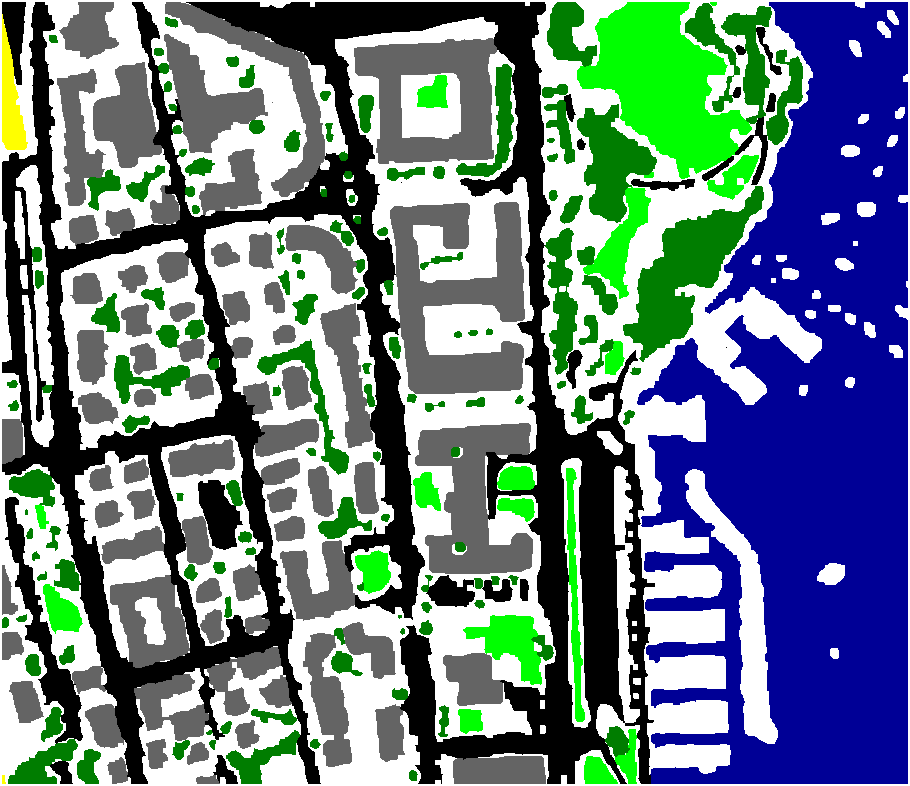}
\\\midrule
\rotatebox{90}{zh18} & 
\includegraphics[width=.19\textwidth]{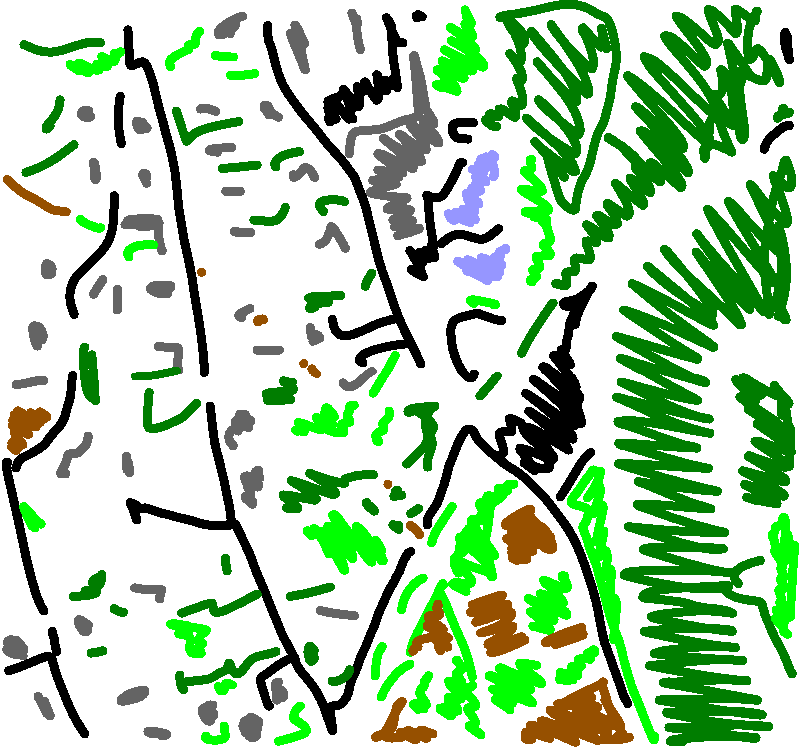} &
\includegraphics[width=.19\textwidth]{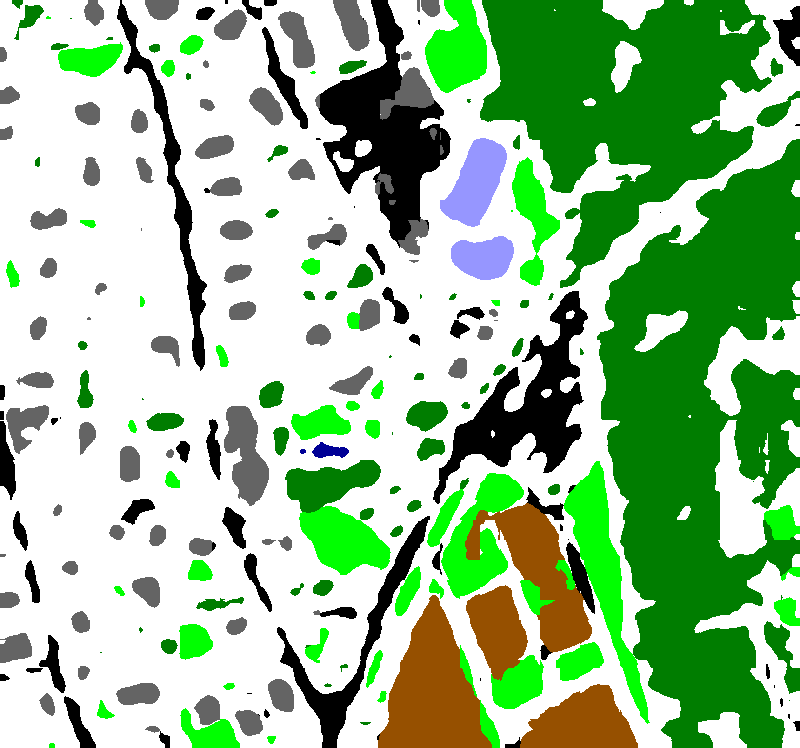} &
\includegraphics[width=.19\textwidth]{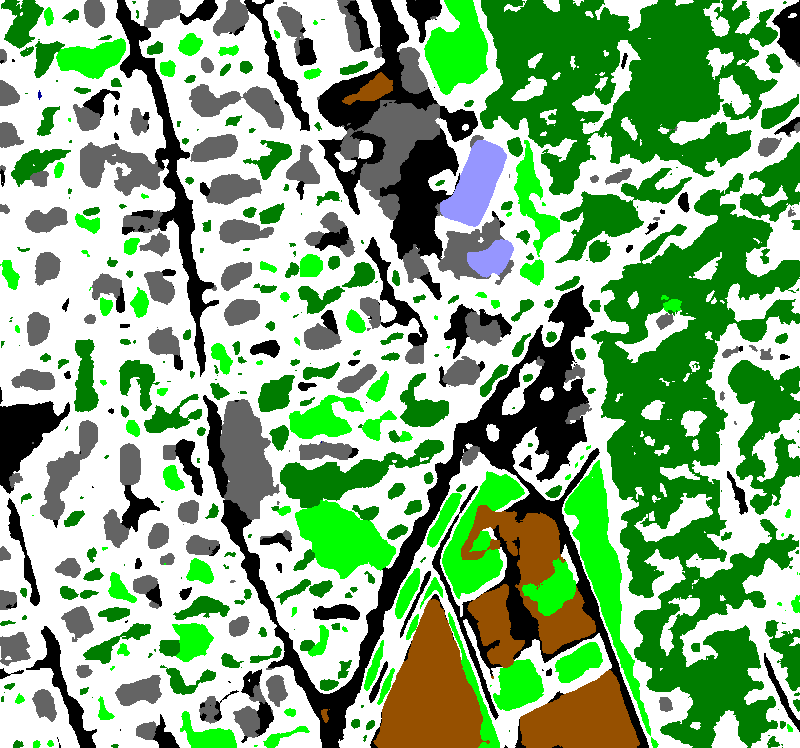} &
\includegraphics[width=.19\textwidth]{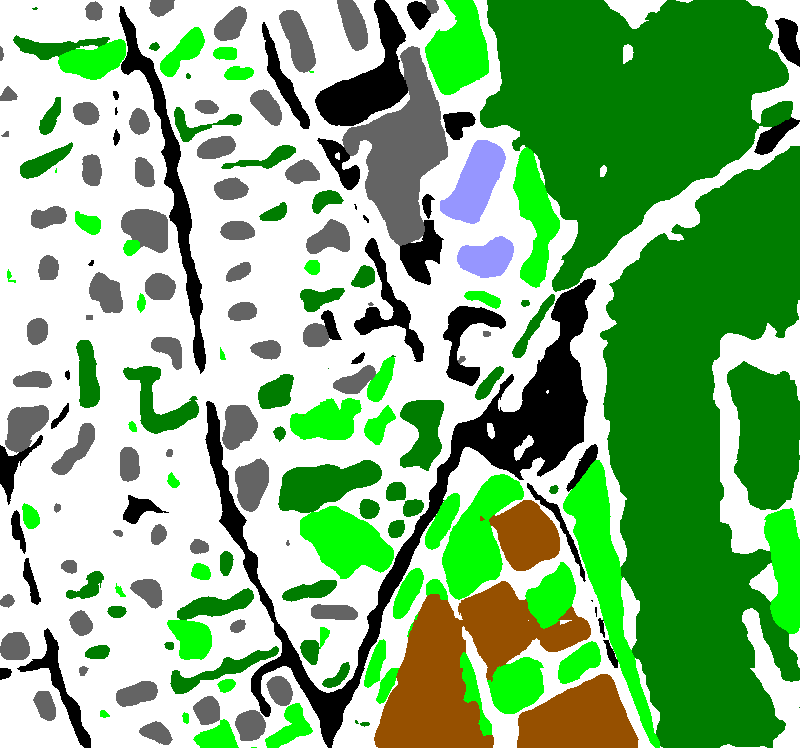} &
\includegraphics[width=.19\textwidth]{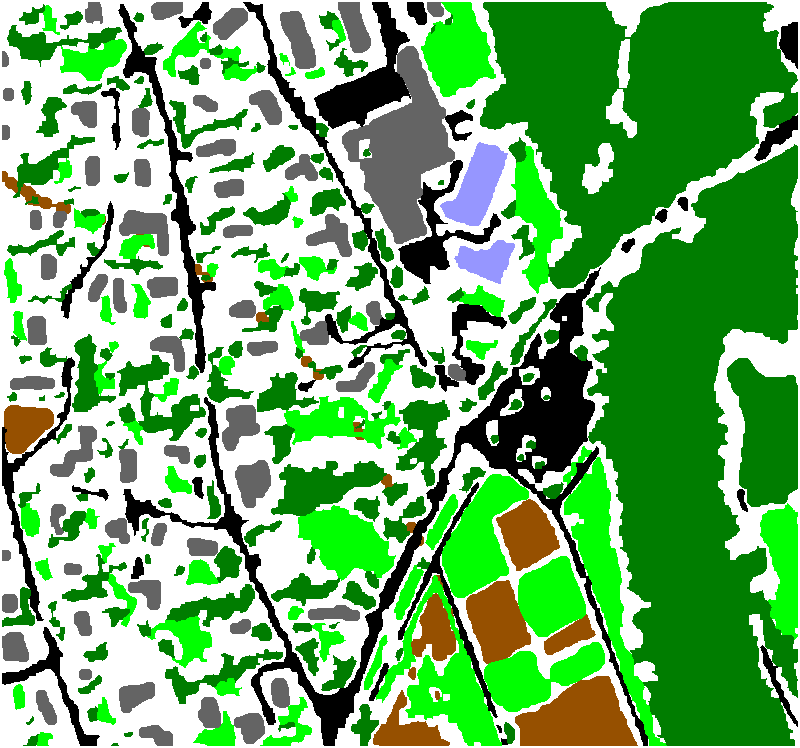}
\\\bottomrule
\vspace{0.01pt}
\end{tabular}
\addlegendimageintext{fill=black, area legend} Roads
\hspace{3pt}
\addlegendimageintext{fill=zurichgrey, area legend} Buildings
\hspace{3pt}
\addlegendimageintext{fill=zurichdarkgreen, area legend} Trees
\hspace{3pt}
\addlegendimageintext{fill=green, area legend} Grass
\hspace{3pt}
\addlegendimageintext{fill=zurichbrown, area legend} Bare Soil
\hspace{3pt}
\addlegendimageintext{fill=zurichdarkblue, area legend} Water
\hspace{3pt}
\addlegendimageintext{fill=zurichyellow, area legend} Railways
\hspace{3pt}
\addlegendimageintext{fill=zurichlightblue, area legend} Swimming Pools
\end{center}

\caption{Qualitative results on the Zurich Summer dataset \cite{volpi2015semantic}. Best viewed in color.}
\label{fig:zurichinference}
\end{figure*}

Our qualitative results on the Zurich Summer dataset are presented in Fig.~\ref{fig:zurichinference}. As shown in the results, SideInfNet is able to draw a balance between fully automatic inference (e.g., Deeplab-ResNet), and completely manual segmentation (e.g., by a human expert). Our method produces much more accurate segmentation results as compared to the Unified model. For instance, as shown in the docks at the bottom right area in the \textit{zh11} image, SideInfNet well distinguishes \textit{Background} (white) from \textit{Building} (gray). Docks are a relatively rare environmental feature, which make them difficult to be classified correctly. The Unified model, on the other hand, misclassifies this as \textit{Buildings}.

SideInfNet also produces higher quality results compared to other models. The Unified model generates more dilated segmentation masks, while the baseline Deeplab-ResNet produces sparser masks.

%This is a common theme with the results from BACH (Section \ref{sec:bachresults}).

SideInfNet is also able to accurately classifying smaller regions such as the \textit{Bare Soil} region in the \textit{zh5} image (see Fig.~\ref{fig:zurichinference}), which challenge other models. The segmentation results of SideInfNet on \textit{Railway} class in the \textit{zh8} image are also more coherent compared with other works.

% Placed here so it appears on the right page

\bibliographystyle{splncs04}
\bibliography{references}

\begin{thebibliography}{10}
\providecommand{\url}[1]{\texttt{#1}}
\providecommand{\urlprefix}{URL }
\providecommand{\doi}[1]{https://doi.org/#1}

\bibitem{mapillary}
AB, M.: Mapillary (2019), \url{https://www.mapillary.com}, Last accessed on
  2019-11-01

\bibitem{aresta2019bach}
Aresta, G., Ara{\'u}jo, T., Kwok, S., Chennamsetty, S.S., Safwan, M., Alex, V.,
  Marami, B., Prastawa, M., Chan, M., Donovan, M., et~al.: Bach: Grand
  challenge on breast cancer histology images. Medical image analysis  (2019)

\bibitem{caelles2017one}
Caelles, S., Maninis, K.K., Pont-Tuset, J., Leal-Taix{\'e}, L., Cremers, D.,
  Van~Gool, L.: One-shot video object segmentation. In: Proceedings of the IEEE
  conference on computer vision and pattern recognition. pp. 221--230 (2017)

\bibitem{chen2014semantic}
Chen, L.C., Papandreou, G., Kokkinos, I., Murphy, K., Yuille, A.L.: Semantic
  image segmentation with deep convolutional nets and fully connected {CRF}s.
  arXiv preprint arXiv:1412.7062  (2014)

\bibitem{chen2018deeplab}
Chen, L.C., Papandreou, G., Kokkinos, I., Murphy, K., Yuille, A.L.: Deeplab:
  Semantic image segmentation with deep convolutional nets, atrous convolution,
  and fully connected {CRF}s. IEEE transactions on pattern analysis and machine
  intelligence  \textbf{40}(4),  834--848 (2018)

\bibitem{bingmaps}
Corporation, M.: Bing maps tile system (2019),
  \url{https://msdn.microsoft.com/en-us/library/bb259689.aspx}, Last accessed
  on 2019-11-01

\bibitem{feng2018urban}
Feng, T., Truong, Q.T., Thanh~Nguyen, D., Yu~Koh, J., Yu, L.F., Binder, A.,
  Yeung, S.K.: Urban zoning using higher-order markov random fields on
  multi-view imagery data. In: Proceedings of the European Conference on
  Computer Vision (ECCV). pp. 614--630 (2018)

\bibitem{girshick2014rich}
Girshick, R., Donahue, J., Darrell, T., Malik, J.: Rich feature hierarchies for
  accurate object detection and semantic segmentation. In: Proceedings of the
  IEEE conference on computer vision and pattern recognition. pp. 580--587
  (2014)

\bibitem{goring2012semantic}
G{\"o}ring, C., Fr{\"o}hlich, B., Denzler, J.: Semantic segmentation using
  grabcut. In: VISAPP. pp. 597--602 (2012)

\bibitem{krahenbuhl2011efficient}
Kr{\"a}henb{\"u}hl, P., Koltun, V.: Efficient inference in fully connected
  {CRF}s with gaussian edge potentials. In: Advances in neural information
  processing systems. pp. 109--117 (2011)

\bibitem{li2018instance}
Li, S., Seybold, B., Vorobyov, A., Fathi, A., Huang, Q., Jay~Kuo, C.C.:
  Instance embedding transfer to unsupervised video object segmentation. In:
  Proceedings of the IEEE Conference on Computer Vision and Pattern
  Recognition. pp. 6526--6535 (2018)

\bibitem{lin2014microsoft}
Lin, T.Y., Maire, M., Belongie, S., Hays, J., Perona, P., Ramanan, D.,
  Doll{\'a}r, P., Zitnick, C.L.: Microsoft coco: Common objects in context. In:
  European conference on computer vision. pp. 740--755. Springer (2014)

\bibitem{long2015fully}
Long, J., Shelhamer, E., Darrell, T.: Fully convolutional networks for semantic
  segmentation. In: Proceedings of the IEEE conference on computer vision and
  pattern recognition. pp. 3431--3440 (2015)

\bibitem{mattyus2016hd}
M{\'a}ttyus, G., Wang, S., Fidler, S., Urtasun, R.: Hd maps: Fine-grained road
  segmentation by parsing ground and aerial images. In: Proceedings of the IEEE
  Conference on Computer Vision and Pattern Recognition. pp. 3611--3619 (2016)

\bibitem{papandreou2018personlab}
Papandreou, G., Zhu, T., Chen, L.C., Gidaris, S., Tompson, J., Murphy, K.:
  Personlab: Person pose estimation and instance segmentation with a bottom-up,
  part-based, geometric embedding model. In: Proceedings of the European
  Conference on Computer Vision (ECCV). pp. 269--286 (2018)

\bibitem{paszke2017automatic}
Paszke, A., Gross, S., Massa, F., Lerer, A., Bradbury, J., Chanan, G., Killeen,
  T., Lin, Z., Gimelshein, N., Antiga, L., Desmaison, A., Kopf, A., Yang, E.,
  DeVito, Z., Raison, M., Tejani, A., Chilamkurthy, S., Steiner, B., Fang, L.,
  Bai, J., Chintala, S.: Pytorch: An imperative style, high-performance deep
  learning library. In: Advances in Neural Information Processing Systems 32,
  pp. 8024--8035. Curran Associates, Inc. (2019),
  \url{http://papers.neurips.cc/paper/9015-pytorch-an-imperative-style-high-performance-deep-learning-library.pdf}

\bibitem{perazzi2017learning}
Perazzi, F., Khoreva, A., Benenson, R., Schiele, B., Sorkine-Hornung, A.:
  Learning video object segmentation from static images. In: Proceedings of the
  IEEE Conference on Computer Vision and Pattern Recognition. pp. 2663--2672
  (2017)

\bibitem{ren2015faster}
Ren, S., He, K., Girshick, R., Sun, J.: Faster r-cnn: Towards real-time object
  detection with region proposal networks. In: Advances in neural information
  processing systems. pp. 91--99 (2015)

\bibitem{rother2004grabcut}
Rother, C., Kolmogorov, V., Blake, A.: Grabcut: Interactive foreground
  extraction using iterated graph cuts. In: ACM transactions on graphics (TOG).
  vol.~23, pp. 309--314. ACM (2004)

\bibitem{shankar2015video}
Shankar~Nagaraja, N., Schmidt, F.R., Brox, T.: Video segmentation with just a
  few strokes. In: Proceedings of the IEEE International Conference on Computer
  Vision. pp. 3235--3243 (2015)

\bibitem{shazeer2017outrageously}
Shazeer, N., Mirhoseini, A., Maziarz, K., Davis, A., Le, Q., Hinton, G., Dean,
  J.: Outrageously large neural networks: The sparsely-gated mixture-of-experts
  layer. arXiv preprint arXiv:1701.06538  (2017)

\bibitem{tripathi2017pose2instance}
Tripathi, S., Collins, M., Brown, M., Belongie, S.: Pose2instance: Harnessing
  keypoints for person instance segmentation. arXiv preprint arXiv:1704.01152
  (2017)

\bibitem{veit2018convolutional}
Veit, A., Belongie, S.: Convolutional networks with adaptive inference graphs.
  In: Proceedings of the European Conference on Computer Vision (ECCV). pp.
  3--18 (2018)

\bibitem{volpi2015semantic}
Volpi, M., Ferrari, V.: Semantic segmentation of urban scenes by learning local
  class interactions. In: Proceedings of the IEEE Conference on Computer Vision
  and Pattern Recognition Workshops. pp.~1--9 (2015)

\bibitem{workman2017unified}
Workman, S., Zhai, M., Crandall, D.J., Jacobs, N.: A unified model for near and
  remote sensing. In: Proceedings of the IEEE International Conference on
  Computer Vision. pp. 2688--2697 (2017)

\bibitem{xu2017deep}
Xu, N., Price, B., Cohen, S., Yang, J., Huang, T.: Deep grabcut for object
  selection. arXiv preprint arXiv:1707.00243  (2017)

\bibitem{zhou2014learning}
Zhou, B., Lapedriza, A., Xiao, J., Torralba, A., Oliva, A.: Learning deep
  features for scene recognition using places database. In: Advances in neural
  information processing systems. pp. 487--495 (2014)

\end{thebibliography}
\end{document}